**VISVESVARAYA TECHNOLOGICAL UNIVERSITY**
**BELAGAVI, KARNATAKA, INDIA**

# DESIGN & IMPLEMENTATION OF ACCELERATORS FOR APPLICATION SPECIFIC NEURAL PROCESSING

Submitted in partial fulfilment of the requirements for the award of degree of

**DOCTOR OF PHILOSOPHY**

in

**Electronics and Communication Engineering**

By

**SHILPA MAYANNAVAR**
USN: 2KL16PEJ14

Under the Guidance of

**Dr. Uday V Wali**
Professor
KLE DR. MSS CET
Belagavi

**KLE DR M S SHESHGIRI COLLEGE OF ENGINEERING AND TECHNOLOGY,**
UDYAMBAG, BELAGAVI – 590008, KARNATAKA, INDIA

**September-2019**

# DECLARATION

I declare that the thesis entitled "**DESIGN & IMPLEMENTATION OF ACCELERATORS FOR APPLICATION SPECIFIC NEURAL PROCESSING**" is being submitted to **Visvesvaraya Technological University (VTU), Belagavi**, Karnataka, for partial fulfillment of requirements for the award of Doctor of Philosophy in Electronics and Communication Engineering by me, **Ms. Shilpa Mayannavar** bearing the **USN: 2KL16PEJ14**, registered in the Department of Electronics and Communication Engineering, KLE DR M S Sheshgiri College of Engineering and Technology, Belagavi – 590008. The work reported in this research is authentic and is an original work done by me during the period January 2016 to September 2019 and has not been submitted elsewhere for award of similar or other degree.

**Ms. Shilpa Mayannavar**

**(USN 2KL16PEJ14)**

**KLE DR M S SHESHGIRI COLLEGE OF ENGINEERING AND TECHNOLOGY**

**DEPARTMENT OF ELECTRONICS AND COMMUNICATION ENGINEERING**

## CERTIFICATE

Certified that **Ms. Shilpa Mayannavar** bearing **USN: 2KL16PEJ14** has completed the course work, comprehensive viva and all other academic requirements for the submission of the Ph.D thesis entitled "**DESIGN & IMPLEMENTATION OF ACCELERATORS FOR APPLICATION SPECIFIC NEURAL PROCESSING**" to Visvesvaraya Technological University, Belagavi during the year 2016-2019, for the award of **DOCTOR OF PHILOSOPHY** in **Electronics and Communication Engineering**. The thesis submitted does not contain any work, which has been carried out by others and submitted by the candidate herself for the award of any degree anywhere.

**Guide**

**Dr. Uday V Wali**

**Professor**
**KLE Dr. MSSCET**
**Belagavi**

**Head of the Department**

Head of Department
Dept. of Elect. & Commun. Engineering
KLE Dr. M. S. Sheshgiri College of
Engineering & Technology
Belagavi - 590 008

**Principal**

**Dr. Basavaraj G Katageri**

PRINCIPAL
KLE Dr. M. S. Sheshgiri
College of Engg. & Tech.
BELAGAVI.

## Acknowledgements

*This thesis and the research carried out in this course of work would not have been possible without the technical, professional and moral support from many people. I take this opportunity to thank each one of them for helping me complete this thesis.*

*Foremost, I would sincerely like to express my gratitude to my guide* ***Dr. Uday V Wali****, for his guidance and encouragement. I am also thankful to* ***Dr. Basavaraj Katageri****, Principal, KLE Dr. MSS CET, Belagavi,* ***Prof. S B Kulkarni****, Head, E&CE, KLE Dr. MSS CET, Belagavi, for their constant inspiration and support during my work.*

*I am thankful to* ***Dr. Rashmi Rachh****, Special Officer, VTU-Research Resource Centre, Belagavi, for her suggestions and support during my work. I am also thankful to* ***Doctoral Committee members*** *for their valuable suggestions and guidelines. I am grateful to* ***Dr M Sutaone****, CoE, Pune, for his support and encouragement throughout the work.*

*My sincere thanks to* ***C-Quad Research****, Desur I T Park, Belagavi,* ***VTU-RRC,*** *Belagavi and* ***KLE Dr. MSS CET****, Belagavi for all the opportunities and facilities provided. I am thankful to* ***BCWD, Govt. of Karnataka****, for providing me financial support through scholarship for three years of my research. My special thanks to* ***Ms. Deepa Pai, Ms. Prajakta Chandilkar*** *and* ***Mrs. Bharati Bhise*** *for their help during my work.*

*My heartfelt thanks to the source of my strength:* ***My beloved family****. An unconditional love & care, understanding, continuous support, encouragement, motivation and prayers of* ***my father Suresh K Mayannavar*** *and* ***my mother Sheela S Mayannavar****, has provided me a great strength and confidence throughout my journey. My love and gratitude for them can never be expressed in words. I dedicate this thesis to my parents. I extend my heartfelt thanks to* ***my brother Vishal Mayannavar*** *&* ***my sister Vinuta Mayannavar*** *for always being there for me with sweet supportive words.*

*Lastly, I thank all my friends and well wishers for their help & support.*

*With sincere thanks,*
***Shilpa Mayannavar***

# Table of Contents

# List of Figures

# List of Tables

# List of Symbols

| Symbol | Parameter name | Description |
|---|---|---|
| $\sigma$ | Standard deviation | A measure of amount of variation or spreading of a set of data values from an average (or expected value). Also referred as SD. |
| $\sigma^2$ | Variance | Square of a standard deviation. Measure of how far the data values are spread out from an expected value. |
| $\rho$ | Resonance control parameter | Also called as coverage control parameter or tuning parameter in Auto Resonance Networks. |
| T | Threshold | A pre defined value (usually above 90%). Acts as a condition for decision making in Auto Resonance Networks. |
| $b$ | Number base (binary, octal, decimal, etc. ) | To represent the number using different base in digital systems design. Examples of number base are binary (0 and 1), octal (0 to 7), decimal (0 to 9), etc. |
| C | Carry | An overflow of addition. |
| K | Gain control parameter | Used to adjust the peak value of output of a node in ARN. |
| M | Number of columns | Number of bits, octets, digits, etc. used to represent a number in base $b$. |
| N | Number of operands | It is also used to represent the number of inputs in a network. |
| P | Number of carry columns | Number of bits, octets, digits, etc. used to represent a carry in base b. |
| $\mathbb{R}$ | Range | Defines a set of values between minimum and maximum limit. |
| S | Column sum | Sum of a column. For example, in addition of two n-bit numbers, there would n column sums representing a sum of n-bits. |

| Symbol | Parameter name | Description |
|---|---|---|
| T | Translation parameter | Also called as scaling parameter. It is used to translate or scale the resonating curves of ARN nodes. |
| $x_m$ | Resonant input in ARN | The inputs values to which the new nodes are created. Also called as tuned values. |
| $x_c$ | Coverage points | The maximum and minimum values of x. If the applied input is within this range, then the node fires (triggers). |
| $y_m$ | Peak output | Peak value of output at $x = x_m$ |
| Z | Total sum | The result of adding two or more operands (including carry). |

# List of Abbreviations

| Abbreviation | Full Form and Description |
|---|---|
| AE | Auto Encoder |
| AFAU | Activation Function Acceleration Unit [2017 Wang] |
| AI | Artificial Intelligence |
| AGI | Artificial General Intelligence |
| ALU | Arithmetic and Logic Unit |
| ANN | Artificial Neural Network |
| API | Application Programming Interface ,<br>Application Peripheral Interface |
| ARN | Auto Resonance Network |
| ART | Adaptive Resonance Theory |
| ASIC | Application Specific Integrated Circuit |
| ASIP | Application Specific Instruction set Processor |
| AVX | Advanced Vector Instructions |
| B2B | Business to Business |
| B2C | Business to Customer |
| BM | Boltzmann Machine |
| BPN | Back Propagation Networks |
| BSL | Bit Select Logic |
| CCLU | Command Control Logic Unit |
| CEC | Constant Error Carousel |
| CIFAR | Canadian Institute For Advanced Research |
| CISC | Complex Instruction Set Computer |
| CNN | Convolutional Neural Network (ConvNet) |
| CNS | Central Nervous System |
| CPU | Central Processing Unit |
| CR | Cognitive Radio |
| CU | Control Unit |
| CUDA | Compute Unified Development Architecture (Obsolete, used as a pronoun now) |
| DL | Deep Learning |

| Abbreviation | Full Form and Description |
|---|---|
| DLAU | Deep Learning Accelerator Unit [2017 Wang] |
| DLNN | Deep Learning Neural Network(s), syn. DNN |
| DMA | Direct Memory Access |
| DNN | Deep Neural Network(s), syn. DLNN |
| DSP | Digital Signal Processor |
| FIFO | First In First Out |
| FMA | Fused Multiply and Accumulate |
| FPGA | Field Programmable Gate Array |
| FSM | Finite State Machine |
| GPGPU | General Purpose Graphic Processing Unit |
| GPU | Graphics Processing Unit |
| GAN | Generative Adversarial Network(s) |
| HBM | High Bandwidth Memory |
| HDL | Hardware Description Language |
| HTM | Hierarchical Temporal Memory |
| IMC | In-Memory Computing |
| IoT | Internet of Things |
| IP | Internet Protocol<br>Intellectual Property |
| ISA | Instruction Set Architecture |
| KNN | K-Nearest Neighbour (k-NN) |
| LSB | Least Significant Bit |
| LSTM | Long Short-Term Memory |
| LUT | Look-Up Table |
| MAC | Multiply and Accumulate |
| MIMD | Multiple Instruction Multiple Data |
| MIPS | Microprocessor without Interlocked Pipelined Stages,<br>Million Instructions Per Second |
| MKL | Math Kernel Library |
| ML | Machine Learning |
| MLP | Multi-Layer Perceptron |
| MNIST | Modified National Institute of Standards and Technology,<br>Hand written characters dataset from |

| Abbreviation | Full Form and Description |
|---|---|
| MP | Massive Parallelism |
| MSB | Most Significant Bit |
| NFU | Neural Functional Unit |
| NLP | Natural Language Processing |
| NN | Neural Network |
| NNP | Neural Network Processor (syn. NPU) |
| NPU | Neural Processing Unit (syn. NNP) |
| OFDM | Orthogonal Frequency Division Multiplexing |
| PDP | Parallel Distributed Processing |
| PE | Priority Encoder |
| PIM | Processing-In-Memory |
| PSAU | Part Sum Accumulation Unit [2017 Wang] |
| PWL | Piecewise Linear |
| RAM | Random Access Memory |
| RBF | Radial Basis Function |
| RBM | Restricted Boltzmann Machine |
| ReLU | Rectified Linear Unit |
| RF | Radio Frequency |
| RISC | Reduced Instruction Set Computer |
| RNN | Recurrent Neural Networks |
| RNA | Resistive Neural Acceleration |
| ROM | Read Only Memory |
| SDK | Software Development Kit |
| SDR | Software Defined Radio |
| SIMD | Single Instruction Multiple Data |
| SMA | Streamed Multiply Accumalate |
| SNN | Spiking Neural Network |
| SoC | System on Chip |
| SOI | Second Order Interpolation |
| SOM | Self Organizing Maps |
| SVHN | Street View House Numbers |
| SVM | Support Vector Machine |
| TFLOPS | Terra FLoating point Operations Per Second |

| Abbreviation | Full Form and Description |
|---|---|
| TMMU | Tile Matrix Multiplication Unit [2017 Wang] |
| TPU | Tensor Processing Unit |
| VALU | Vector Arithmetic and Logic Unit [2009 Farabet] |
| VGG | Visual Geometry Group |
| VLIW | Very Long Instruction Word |
| VLSI | Very Large Scale Integrated circuits |
| VNNI | Vector Neural Network Instructions |
| XAI | Explainable Artificial Intelligence |

# Abstract

Internet of Things (IoT) and Artificial Intelligence (AI) are quickly changing the computing eco system. Evolution of IoT has been slow, possibly due to lack of fine-grain standardization. On the other hand, developments in AI have been accelerating with a pace unprecedented in history of computing. Modern AI is largely driven by new structures in Artificial Neural Networks (ANN) like Convolutional Neural Network (CNN) and Long Short Term Memory (LSTM). Considering the types of neurons in mammalian Central Nervous System (CNS), it is easy to see that many new ANN structures need to evolve to attain the ultimate goal of Artificial General Intelligence (AGI). It is necessary to implement massively parallel architectures to realize modern AI systems, which in turn forces a complete overhaul of computing technology; everything ranging from processor design to end applications need to be reinvented.

Primary motivation for this work was the need to implement hardware accelerators for a newly proposed ANN structure called Auto Resonance Network (ARN) for robotic motion planning. ARN is an approximating feed-forward hierarchical and explainable network. It can be used in various AI applications but the application base was small. Therefore, the objective of the research was twofold: to develop a new application using ARN and to implement a hardware accelerator for ARN.

As per the suggestions given by the Doctoral Committee, an image recognition system using ARN has been implemented. An accuracy of around 94% was achieved with only 2 layers of ARN. The network also required a small training data set of about 500 images. Publicly available MNIST dataset was used for this experiment. All the coding was done in Python. Details and results of this work have been reported in an international conference in April 2019 and submitted to a journal.

During the present work, sigmoid based resonator has been generalized to improve controllability of the resonator. A relation between coverage of the neuron and half power point has been derived. A formal method of tuning of ARN neurons based on statistical properties of incoming data has been developed. This work helped in designing the necessary hardware modules for the accelerator.

Hardware implementation of non-linear activation functions involves resource intensive Taylor series expansion. Therefore, a module capable of piecewise-linear and second order approximation of non-linear functions was developed. It reduces the number of multiplications by a ratio of about 32:1 with less than 0.1% error, which can be implemented with 12 bit fractions. This is sufficient for ARN and other neural computations. It was also noted that such approximation will not affect the performance of ARN. Therefore, a new fixed point number format was defined to implement these approximation algorithms, resulting in faster implementation. These results were presented in an international conference in January 2018 and in a SCOPUS indexed journal in May 2019. Interestingly, such approximation often improves the performance of neural networks. As the demand for *computation at the edge* increases, instruction-level implementation of low precision computations will gain ground.

Massive parallelism seen in ANNs presents several challenges to CPU design. For a given functionality, e.g., multiplication, several copies of serial modules can be realized within the same area as a parallel module. Serial modules can also use serial data transfer over a single wire instead of a bus, improving the connection density. Advantage of using serial modules compared to parallel modules under area constraints has been discussed. These results, which can be useful in CPU design for AI systems, were presented at an international conference in December 2018.

One of the module often useful in ANNs is a multi-operand addition. One problem in its implementation is that the estimation of carry bits when the number of operands changes. A theorem to calculate exact number of carry bits required for a multi-operand addition has been presented in the thesis which alleviates this problem. A modular four operand serial adder has been implemented. The fast implementation of 4-operand parallel adder has also been implemented. An algorithm to reconfigure a set of such modules to implement larger multi-operand adders has been developed. The main advantage of the modular approach to multi-operand addition is the possibility of pipelined addition with low reconfiguration overhead. This results in overall increase in throughput for large number of additions, typically seen in several DNN configurations. This work was presented at international conference in April 2019. All the necessary hardware components for implementation of ARN and the neuron itself have been implemented. All the hardware modules are developed and verified using Verilog. All the code has been shared on github web portal.

**PhD Thesis**

# Design & Implementation of Accelerators
**for**
# Application Specific Neural Processing

**Shilpa Mayannavar**

**2KL16PEJ14**

**Visvesvaraya Technological University**

**Belagavi, Karnataka, India**

# Chapter 1

# Introduction

## 1.1 Preamble

Internet of Things (IoT), Machine Learning (ML) and Artificial Intelligence (AI) are considered as major development areas in electronics and computer science. These developments are creating strong ripple effects in various technologies like processor design, programming paradigms, power consumption and communication technologies, to name a few. However, the anticipated axes of development do not necessarily point in the same direction. For example, low level programming is necessary to achieve low latency in IoT devices. But the same devices could be using AI algorithms which require high level programming. Training AI systems requires large servers with AI specific accelerators. But quantization of neural computations and compaction of the network are necessary to run the same AI at the edge computer. This has brought in a technology divide that strongly needs to be crossed often and seamlessly.

Both IoT and AI bring disruptive social changes at a pace not seen at any time in history. These technologies will challenge the common knowledge hitherto shared among communities. Domains being influenced by IoT and AI include Defence, Cloud computing, B2B and B2C trading, stock market, elections, Government offices and procedures, teaching and learning process, human – machine interfaces, vehicular automation, surveys and maps, and many others. Integration of IoT and AI with arts and entertainment could possibly redefine social fabric itself. This is only the tip of the iceberg.

In 2015, when I started this work, ML was gaining ground in large corporations like Google who used it to filter spam from their email. Many on-line web stores developed recommendation systems to push marketable products to new consumer base. Map-reduce algorithms, distributed and non-sql databases (Hadoop, Mongo) were the tools used by Data Scientists to push the technological barriers. Machine learning algorithms extract data from various sources like text files, commercial transactions and social media etc. and apply statistical tools to the benefit of its users. It was easy to see that by integrating ML with IoT, new level of industrial automation could be reached. For example, ML allowed performance parameters to be extracted from a large number of equipment in the field rather than laboratory setup, which could be used to enhance the efficiency and reliability of machines. Electrical distribution systems could improve line regulation and stability. By integrating geographically scattered data, large dynamic systems like power systems could improve their stability. Internet communication would shift from personal and corporate information base to machine-to-machine communication. The number of machines that could be communicating would be at least two orders more than conventional computers. Therefore, IoT was seen as a technology game changer. It was called as the future of Internet.

The challenge was to redesign the communication system, aligning with the expected load from IoT devices. Given the large number of equipment manufacturers, many would like to put their own proprietary hardware on the internet while maintaining conformity with the communication standards. It was also necessary to support reconfigurable hardware to reduce cost of inventory. Streaming of machine data could overload the network resources unless some algorithms and methods were designed to extract intelligent and relevant data that could be pushed on the internet. Therefore, integration of ML and IoT was seen as an important step forward in achieving higher level of industrial productivity.

One of the issues related to dynamic reconfiguration of processor was the ability to stream an instruction set instead of a stored procedure. One example was the attempt made by ARM to implement Java byte code as basic instruction set in Jazelle series. ARM did not publish much literature on Jazelle but maintained that the processor can run Java byte code directly. Other issue was incorporating data aggregation, anomaly detection, data selection, compression and other similar operations at the instruction or hardware level.

By the end of 2015, AI and DL systems started gaining a lot of attention from industry and academia. That was the time when a computer game called AlphaGo was introduced by Google Deep Mind. It was able to beat the human European Go champion by 5 games to 0 [2016 Silver]. It was trained by combining both supervised and reinforcement learning, enabling to learn by playing with itself and be able to identify the position of the moves using 'value networks' and select the moves using 'policy networks'.

There were other developments in the field of AI as well. Image recognition and classification using Convolutional Neural Networks [2014 Jia] made it possible for computers to understand images, which was considered difficult for computers till then. Similar developments were also seen in natural language processing using Long Short Term Memory (LSTM) [1997 Hochreiter]. Developments in robotics were also dramatic (Boston dynamics). AI applications are growing since then and are being used for everyday applications. Therefore, focus of development shifted from IoT to AI.

When we look at the impact of AI and IoT on the processor design, it is easy to see the contrast: IoT requires a low power processor while AI requires extremely high performance multi-core processor. Processor designs could be incrementally improved to support IoT applications while AI required a complete revision of processor design strategies: A quad-core processor may suffice the IoT requirements but AI applications will require thousands of cores. Instruction Set Architecture (ISA) for IoT and AI is largely driven by divergent end applications. Bus design, which depends on how fast we need to move data between individual blocks, has to be different for IoT and AI applications because of the number of data shares in AI is several orders more than that in IoT devices. Therefore, introduction of IoT and AI will expand the need to build customized processors in years to come. The idea behind a customized System on Chip (cSoC) is derived from the fact that most of the processors can be built with a set of reusable building blocks and a customized controller, whose design depends on ISA, memory organization and the bus design.

Many companies are using their own processors to implement such hardware. For example, a special purpose processor for Software Defined Radio (SDR) and heuristic Cognitive Radio (CR) algorithms has been reported in [2013 Saha]. Use of Graphic Processor Units (GPUs) has shown significant performance advantages [2008 Lindholm]. Many new neural architectures are being proposed to address domain specific needs like robotic motion control [2016 Aparanji], image recognition [2014 Jia] etc. Introduction of open-source RISC-V ISA [2017 Ajayi] and the associated eco-system is expected to make it cost effective to produce vendor designed application specific processors.

Neural computing involves massively parallel operations which demand availability of special purpose processors with thousands of compute cores. To some extent, this need has been met by use of GPUs [2010 Nickolls], [2008 Che], [2011 Herrero], [2011 Keckler]. GPUs have been used in variety of applications including image processing, robotics, video and text processing and in many other intelligent use cases. Additional operations like scaling, rotation and transformations frequently used in GPUs are also useful in neural computations.

Hardware capable of vector, matrix and tensor representations of data can exploit massive parallelism of these architectures and provide a method to speed up the computations. However, several non-linear activation functions like sigmoid, hyperbolic tangent, ReLU and softmax are regularly required. Interconnections in neural network can be more varied than in case of GPU operations. Clearly, there are several features in neural computations that have no equivalent in graphics processing. Many research organizations like ARM [2018 Elliott], Google [2017 Jouppi], Intel [2017 Rao], Apple, Samsung [2019 Nagendra], Cadence Tensilica [2017 Cadence], Cambricon [2016 Liu], IBM [2015 Akopyan] etc. are developing hardware to implement neural architectures. A domain-specific ISA for neural accelerators, called Cambricon has been reported in [2016 Liu]. It has a load-store architecture that integrates scalar, vector, matrix, logical, data transfer and control instructions, designed after a comprehensive analysis of existing neural networks. Presently, Huawei's Kirin-970 [2019 Kirin] processor uses Cambricon IP for vector and matrix operations required to run AI software on mobile sets.

Intel's Nervana Neural Network Processor (NNP) is capable of performing tensor operations at processor level [2017 Rao]. Nervana NNP uses a fixed point number format named *flexpoint* [2017 Koster] that supports a large dynamic range using a shared exponent. Only mantissa is handled as a part of the op-code.

Neurons and synapses are two of the fundamental biological building blocks that make up the human brain. Crudely, they are analogous to components and interconnect of an electronic circuit: Neurons modulate the signal and synapses provide resistive interconnect. IBM's brain-like chip called *TrueNorth* [2015 Akopyan] has 4096 processor cores, each capable of emulating 256 neurons with 256 synapses each and therefore chip mimics one million human neurons and can perform 256 million synapses/operations at a time.

Functionality of these special purpose processors are specific to neural network architectures, e.g., Google TPU is best suited for TensorFlow implementations, TrueNorth from IBM is suited for Multi Layer Perceptron (MLP) implementations, etc.

Therefore, it is not difficult to foresee a situation where many companies would like to design their own processors to support Deep Learning systems.

While special purpose chips are being developed for Deep Neural Network (DNN) applications, traditional processors are implementing extended instructions to support such applications. Notably, Intel’s Advanced Vector Instructions (AVX), broadcast instructions, streaming SIMD (Single Instruction Multiple Data) instructions have reduced the performance gaps between CPU and GPUs to some extent. Of special interest are the Fused Multiply and Accumulate (FMA) and FMA Packed Single Precision instructions [2014 Intel AVX]. These instructions reduce the need to transfer data between memory and registers during repeated computation of multiply-add cycles used in many DNN architectures. DynamIQ is a new technology for ARM Cortex. It has dedicated processor instructions for AI and ML with faster and more complex data processing. It is reported that *dynamIQ* performs AI computations with 50x increased performance compared to Cortex-A73 systems [2017 Wathan]. It supports flexible computation with up to 8 cores on a single cluster with SoC, where each core can have different performance and power characteristics.

Need for large number of computing cores is a hallmark of neural networks. As the number of processing elements becomes large, sharing memory and the required bandwidth to transfer information becomes a bottleneck. Another characteristic of neural processing is the sufficiency of low precision arithmetic. Most of the neural computations are meaningful only in a limited range of values, dictated by the activation functions used to implement these computations. Therefore, computation by approximation could result in faster neural networks with low hardware overhead.

Further developments in processor design will be guided by the demands posed by IoT and AI systems. An understanding of these technologies is therefore necessary for research in this direction.

Our group has been working with a new type of neural network called Auto Resonance Network (ARN) [2016 Aparanji]. Initially, this network was proposed for use in robotic path planning and time series prediction but it is generic enough to be used in several other applications. In this thesis, use of these networks for image recognition and classification has been demonstrated. I have also investigated issues related to implementation of these networks in hardware.

## 1.2 Motivation

Developments in IoT and AI will have an impact on the design of processors, instruction set architecture, memory organization, bus design, on-chip communication, etc. all of which need to be explored. Support for IoT includes communication and peripheral interconnect. However, AI requires fundamental changes to design of the computing eco-system itself. Therefore, these issues need to be understood and addressed. Along the progression of AI, new types of neural networks have been introduced and many more will follow. Newer ANN structures used in AI will expand to a larger application base, exerting pressure on innovation. Therefore, primary motivation for this research has been the change in computing environment imposed by the advancements in AI. Its effects on processor and accelerator design and application design have been studied in this research.

As a case, Auto Resonance Network (ARN) has a potential to be used in diverse applications ranging from image recognition to predictive analytics. ARN form a class of explainable neural networks which may be compacted enough to run on edge computers. They can be easily scaled because adding new nodes will not affect the existing network. They can be organized into multi-layered deep hierarchical networks with assured stability. Primary features of AI systems like massive parallelism, low precision and use of non-linear functions are all present in ARN. Therefore, while I have focused on this type of neural network, work reported here can be useful without loss of generality of application in other types of neural networks.

## 1.3 Objective

Broadly, the objective of this research was twofold:

(1) To develop and test an image recognition application using a hierarchical Auto Resonance Network (ARN) and

(2) To implement a set of hardware accelerators for neural networks in general and ARN in specific.

At a detailed level, the objectives could be stated as follows:

a) To enlist problems in implementation of deep learning systems and their effect on processor architecture and design.

b) To develop image recognition system using hierarchical ARN with an accuracy of recognition exceeding 85% using a publicly available MNIST hand written data set.

c) To define and use a low precision number format to improve the computational performance without significantly reducing accuracy of the network (5%).

d) To design and verify necessary modules to implement an ARN neuronal model in hardware.

Most of the time would be required to address (b) and (d).

## 1.4 Research Methodology

As per the objectives of the research, the problem was divided into two parts, viz., design of hierarchical ARN for image recognition and implementation of hardware modules for ARN neuron.

After a quick overview, it was decided to use Modified National Institute of Standards and Technology (MNIST) database consisting of hand written numerals. This data set has been used by several research groups to validate image recognition and classification algorithms. Python programming language was to be used to implement software as it is frequently used in AI applications. Methods of tuning the ARN were to be worked out. Learning characteristics of ARN were to be studied and documented. Alternate methods to improve performance of ARN were to be suggested if found. Though Python provides large number of libraries to support most of the ML & AI applications, I had to write my own libraries to perform the operations using ARN. This helped me to identify the hardware modules required to accelerate the neural computations. These issues became the basis for implementing hardware modules.

Activation functions require calculation of exponential functions and hence there was a need to develop fast approximators with low error rate. The next was the use of serial hardware multiplier instead of parallel hardware multiplier as the area overhead of parallel multiplication would reduce the throughput. It was also noted that multi-operand addition is a frequent requirement in DNN hardware. The necessary theoretical studies have been carried out and method to implement modular adders was designed. All the units were later integrated in to a single hardware accelerator. Hardware simulation was done using Verilog HDL and Xilinx-ISE simulator (version 9.2i).

Design of multi-operand adder was taken after implementing the resonators and activation functions. One of the theoretical constraints was to decide the number of carry bits, for which necessary proofs were derived. All the modules were integrated and the performance of the accelerator was evaluated.

The Gantt chart in Figure 1.1 shows the overview of work done during this research. In the initial part, I studied the requirements for a processor eco-system including development frameworks, languages and compilers, etc. A prototype of framework for development of customized processors was presented in a conference. However, as the focus shifted to AI, I spent a lot of time to understand the neural architectures and their applications including image identification and classification, robotic path simulations, natural language processing, time series prediction and other related topics.

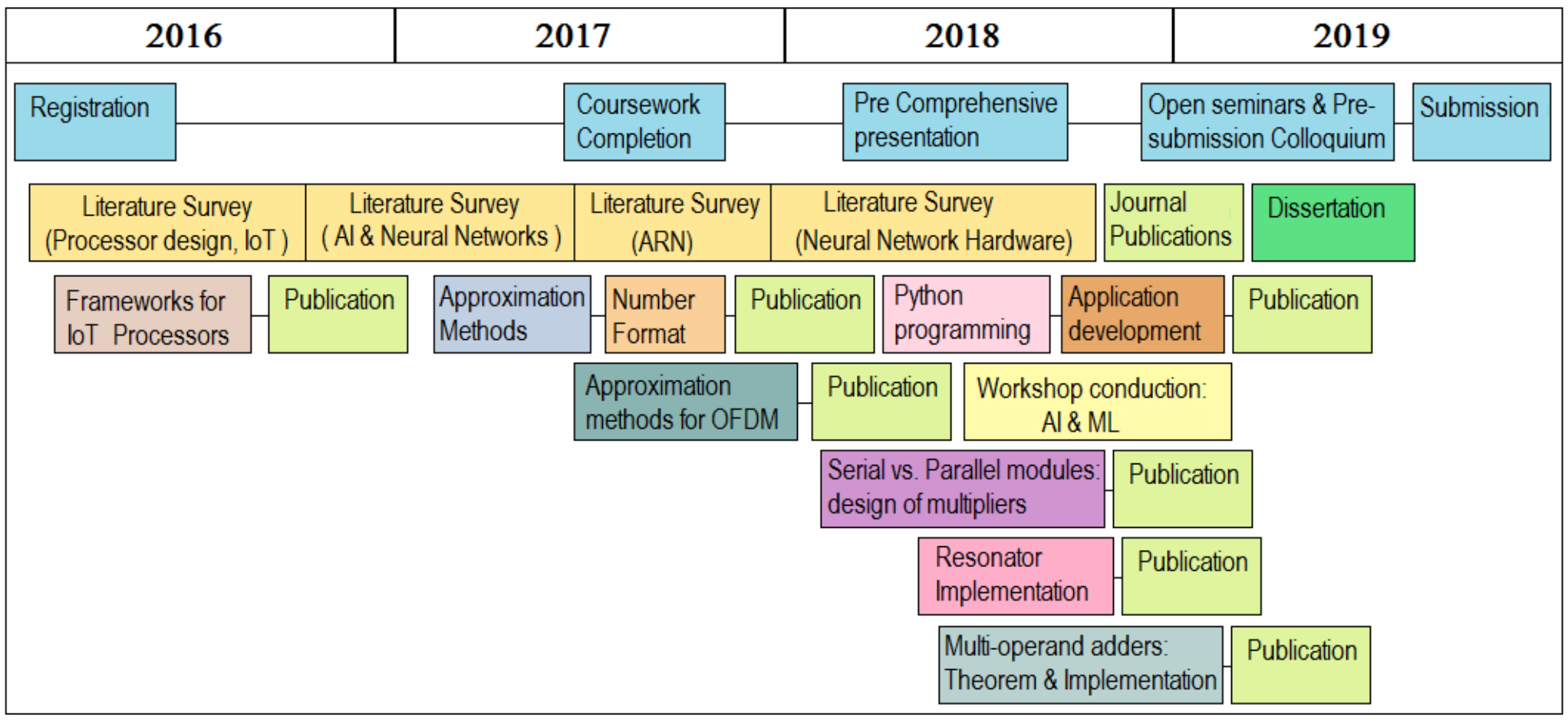

**Figure 1.1 Gantt Chart**

A list indicating the steps involved is given below.

1. Extensions to Auto Resonance Network (ARN) basic model
   - 1.1 Theory of ARN
   - 1.2 Generalization of ARN equations and their interpretations
   - 1.3 Charting of learning and control curves
   - 1.4 Selection of ARN Model

2. Image Recognition Using ARN

   (Use MNIST data set of hand written English numbers)
   - 2.1 Create Data Sets, separation into training and test sets
   - 2.2 Selection of image partitioning method to reduce computational load and implement some kind of feature extraction
   - 2.3 Initial design of a 2 layer Hierarchical ARN for image processing

2.4 Design of similarity functions (image hash)

2.5 Coding in Python

2.6 Training the model with gradually increasing number of samples

2.7 Testing the accuracy of recognition

2.8 Improving the network as necessary

3. Implementation of Hardware Modules

(Use Bottom – Up approach, build all necessary modules, use Verilog)

3.1 Identify necessary basic building blocks

3.2 Implement and test individual blocks

3.3 Implementing a command – control unit

3.4 Overcome any theoretical bottlenecks and redesign when necessary

3.5 Integrate the modules to realize ARN node

3.6 Testing of the module

## 1.5 Organization of thesis

The organization of this thesis reflects the progression of this research. However, some parts of the published work which were not relevant to the contents of this thesis has been omitted, notably the initial work on framework for IoT Processor [2016 P2] and use of approximation techniques in OFDM [2018 P8].

Chapter 2 covers a large body of published literature in the field of AI and DNN. The chapter also introduces Auto Resonance Network, which is extensively used in this work. Effect of AI and DNN on processor designs and hardware accelerators has been covered. Research on AI is still evolving and often represents work in progress and hence some literature is picked up from web sites and technical forums.

Chapter 3 describes some of the important critical design issues of neural network hardware. These observations form the basis of my present and future work. Therefore, this chapter is partially a literature survey and partially a work done in this research.

Chapter 4 presents the basis of Auto Resonance Network and my extensions to the theory. The chapter then continues with use of ARN for image identification and classification. In specific, use of MNIST database of handwritten digits in this research has been elaborated. All the necessary theoretical background, mathematical equations and implementation details are discussed in this chapter. Results of training and testing the ARN for MNIST data set are also presented.

Chapter 5 discusses some solutions to the design issues mentioned in chapter 3. Performance comparison of area constrained parallel and serial modules in massively parallel environment is presented in this chapter. Design and implementation of components for a neural network accelerator, e.g., approximations to activation functions, resonators for ARN, fixed point number format used for these hardware modules are discussed. Simulation results are also included in the chapter.

Chapter 6 describes the design and implementation of multi-operand adders for use in neural processing. The multi-operand adders are systematically studied. Estimation of the upper bound on the size of carry and its proof are provided. Use of theorem to implement multi-operand adders is discussed. The need for re-configurability in such design is also briefly discussed.

Chapter 7 presents the integration of all the hardware modules discussed in chapter 5 and 6, to implement a ARN neuron. Similar concept has been extended to implement MLP neuron (perceptron). Simulation results are also included.

Chapter 8 presents conclusion, with an overview of work done, important observations and interpretation of the results presented in earlier chapters. The scope for future work is also discussed in this chapter.

The work reported in this thesis is published in International Conferences, International Journals and Book chapters as listed in Publications. All cited literature is included in References. Literature used during this work but not cited is listed in Additional References. Appendices contain software developed as a part of this work. The pseudo code for image recognition using ARN is given in Appendix 1. The actual code is available on github web portal. The hardware modules and their implementation details are given in Appendix 2. Justification for using ARN is explained in Appendix 3.

# Chapter 2

# Literature Survey

Initial part of this chapter contains a quick introduction to related topics like biological and artificial neural networks, artificial intelligence etc. The reason for their brief introduction is to provide a better context for current work. Basis of the reported work has been Auto Resonance Network (ARN), which has potential applications in several areas of Artificial Intelligence (AI). Specific objective of this work was to design a hardware accelerator for image recognition using hierarchical ARN. Hence, this literature survey covers topics related to ARN, image recognition using Deep Neural Networks (DNN), and hardware designs for DNN. Later part of this chapter contains brief description of recently introduced processors for DNNs. Due to the proprietary nature of the design, implementation details are not available and hence not included.

Many AI processor architectures are specific to particular types of applications, e.g., Google TPU for TensorFlow library, IBM TrueNorth for MLP, etc. However, most of the DNN architectures use gradient descent algorithms to find local or global minima on a multi-dimensional learning surface. Therefore, many intrinsic operations like activation functions used in DNNs are same in all architectures. Massive parallelism of DNN architectures can be easily implemented using vector and matrix arithmetic. It is more efficient to perform repeated operations using SIMD architectures and systolic arrays. DNN accelerators are also affected by the type and size of data expected to be handled. DNNs are best implemented as dataflow machines rather than procedure flow. Resource rich processors need to be reconfigured to achieve better throughput. Such reconfiguration overhead is expected to be small compared to the speed advantage of reconfigured hardware. Some of the conceptual issues to be considered while designing a neural hardware are briefly discussed in chapter 3.

Instruction sets have to incorporate specific provisions for DNN computations. These include SIMD vector and matrix operations, calculation of exponents and activation functions at hardware level, multi-operand addition and multiplication, hardware reconfiguration, etc. Instruction sets of such processors will support vector and matrix operations. Streamed data and hardware reconfiguration instructions will also be part of the ISA. This is clearly different than the issues addressed by RISC and CISC computers used today in everyday life.

## 2.1 Learning in Biological Systems

Understanding the mechanisms of learning in biological systems has been a topic of interest since early days of written history. However, after the advent of computing, the idea of building an artificial neural system could become a reality. Neural systems of biological systems consist of various kinds of neurons connected in a large network. Largely, there are three types of neurons, viz., sensory neurons, motor neurons and inter-neurons that connect two types of layers of neurons. Not all neurons have input and outputs; some have inputs, some have outputs but most of them have both. In the following discussion and elsewhere in literature, the term *Neuronal* refers to description of a single neuron while the term *Neural* refers to a class or collection of neurons, such as a nerve bundle connected to a muscle.

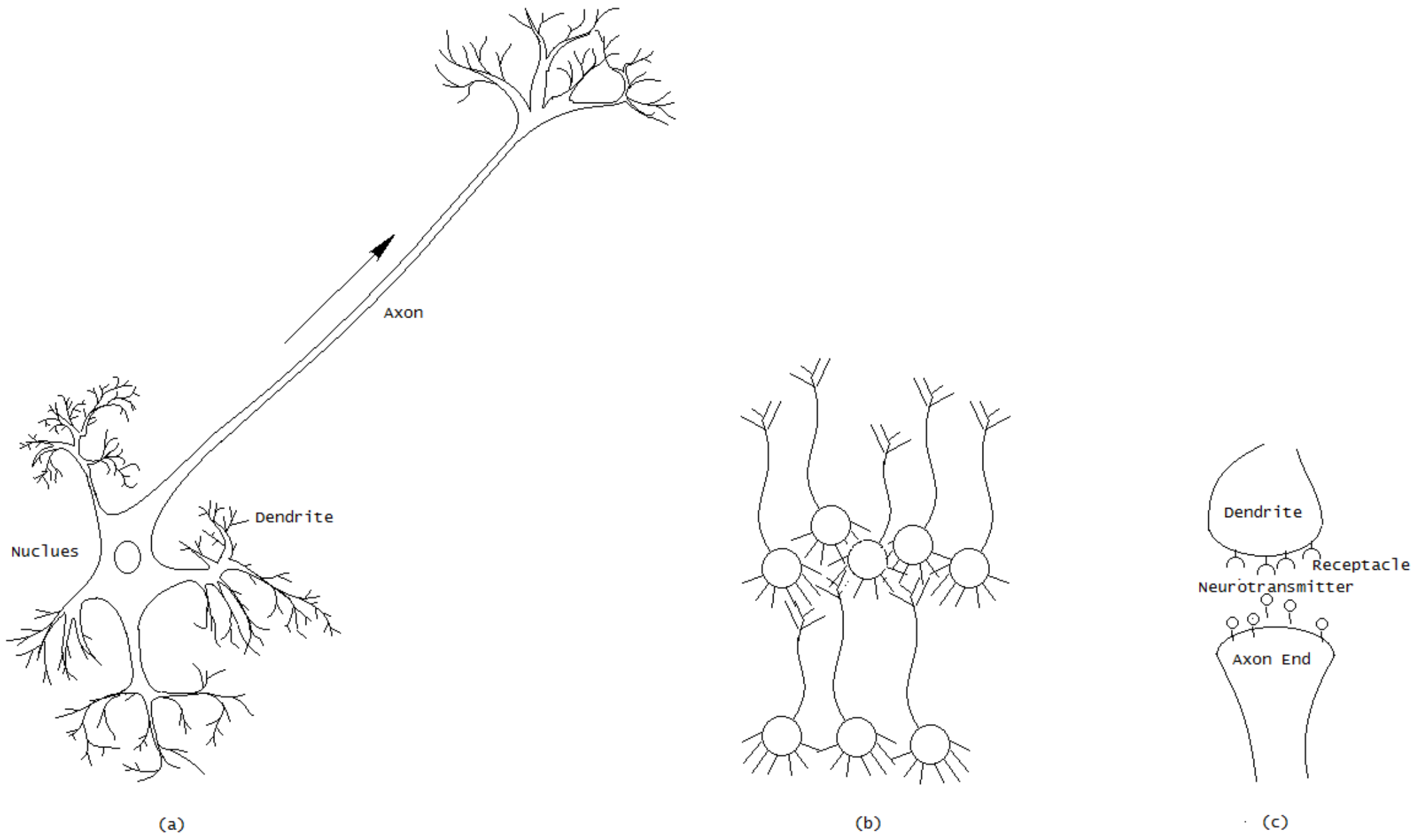

**Figure 2.1 (a) Biological Neuronal structure (b) Layers of neurons (c) Synapse**

Structure of a typical neuron is given in Figure 2.1(a). Connection between sets of neurons is shown in Figure 2.1(b). The connection between two neurons is called the synapse (Figure 2.1(c)). Neurons exchange ions to modulate the potential at their membranes through molecular gates. Each neuron may have thousands of synaptic connections with neurons in other layers. The strength of the synaptic connection is proportional to the contribution of an input neuron to output of the node. A neuron receives a set of inputs from the environment or from other neurons. Inputs are largely electro-chemical in nature. The dendrites carry a decaying electrical excitation to the areas inside the nerve cell. Normally, a neuron is in a polarized state (resting membrane potential ~ -70mV). Synaptic input depolarizes it. Small inputs received via small synaptic activity (graded potentials) alter the potential of the membrane but it is only when there is sufficient accumulation of synaptic input, the neuron is triggered, i.e., reaches a peak potential (action potential ~ +30mV) which pushes the neurotransmitters at the end of the axon to the exterior. Dendrites of connecting neuron will depolarize, effectively, carrying a signal from first neuron to the next. Triggering also causes the neuron to repolarise to resting potential or reach a temporary hyperpolarized state. A neuron can receive both depolarizing and/or hyperpolarizing inputs simultaneously. The action potential, whenever reached, has the same value, independent of the synaptic input.

Strength of a synapse indicates the number of molecular gateways available at a connection. These gateways constitute both short term and long term memory. Hebbian learning suggests that the strength of synaptic connection between two neurons improves with repeated use. In unconstrained form, this leads to unstable learning systems because the strength has to increase with use and there is no scope for reducing the strength of the connection. If we observe that action potential of a neuron does not exceed a fixed value (~ +30mV), it becomes apparent that there is fixed upper bound on the value to be attained by a Hebbian connection. Therefore, Hebbian learning should be seen only as a preliminary description of the underlying mechanism and not as a linear relation between usage and connection strength. It may also be noted that activation potential can be reached even when partial input is present. So, a model that uses constrained Hebbian learning should be able to perform recognition in presence of partial data. Constrained Hebbian learning is an intrinsic feature of the equations used in ARN. Hence, ARN always achieves a stable learning state. Further details on Constrained Hebbian Learning are presented in chapter 4.

Human Central Nervous System (CNS) has thousands of types of neurons. Overall functioning of these neurons is similar but their actual function depends on various factors like their structure, position in the network, connectivity, dendrite density, length of axon, neurotransmitters and inhibitors used, chemical receptors and gateways, type of input and output, timing of potentials, and a plethora of other factors. It is therefore important to explore various neural architectures to move towards realization of anything closer to Artificial General Intelligence (AGI). Some of the architectures that can achieve AGI are Multi-Layer Perceptron (MLP) [2016 Tang], Spiking Neural Network (SNN) [2007 Fujii], [2019 Bouvier], Long Short-Term Memory (LSTM) [1997 Hochreiter], Brunovsky systems [2006 Liu] using Radial Bias Function (RBF) etc. ARN explored in this thesis is also one such architecture capable of adapting to varying demands of AGI. Some of these are discussed in next section.

It is worth noting that nature has provided multiple solutions to intelligent behaviour. Each such solution has some specific advantage in a given environment. Dormant features may prove useful as the environmental conditions change. As the environment of digital systems continues to evolve, classical as well as newer architectures have to be reviewed in the newer context.

## 2.2 Artificial Neural Networks

Recent developments in several fields like image recognition and natural language processing are due to development of new ANN architectures. However, the idea of using brain like structures to realize intelligent machines is not new. We will have a look at several of the classical and modern ANN structures in this section.

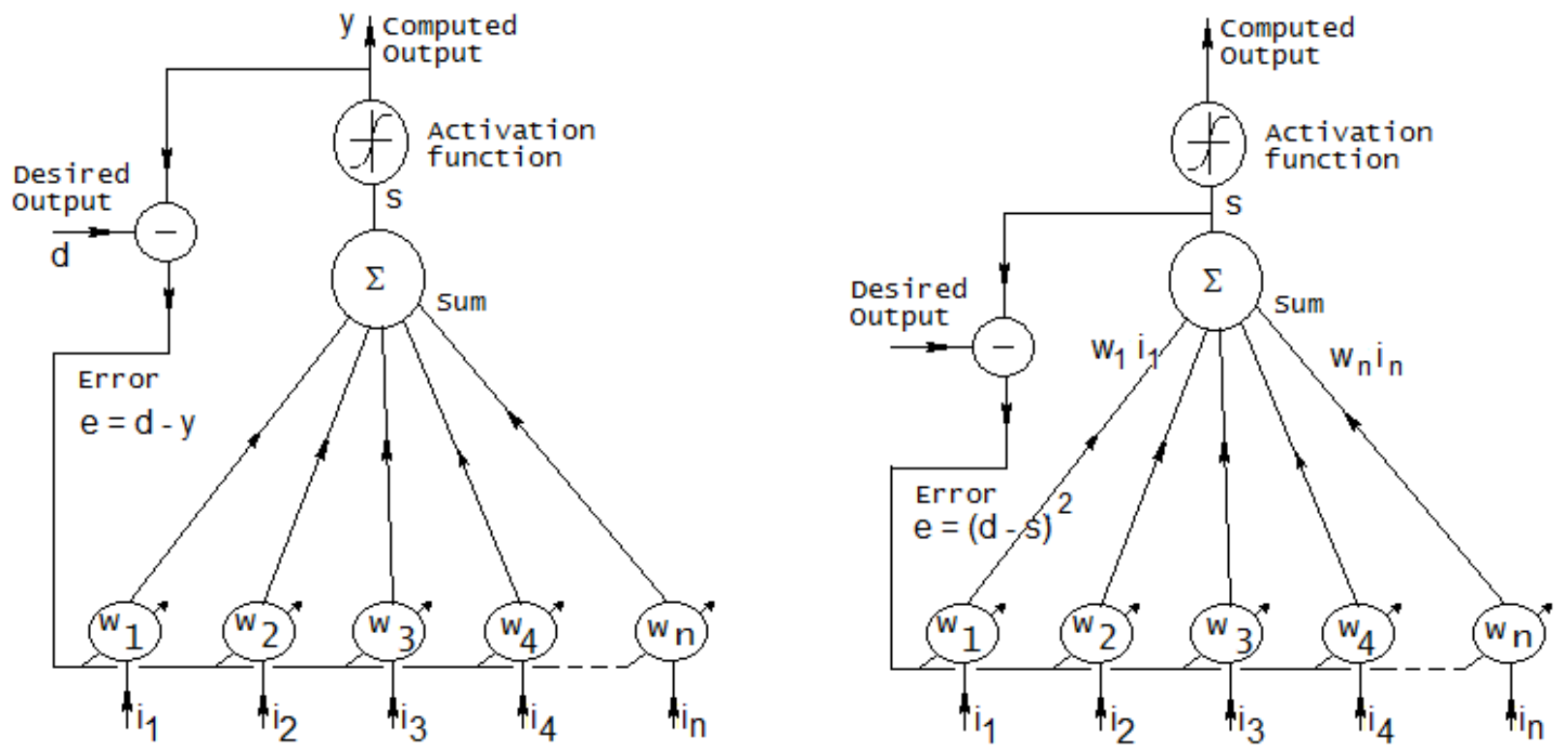

(a) (b)

**Figure 2.2 Early Artificial Neuronal models (a) Perceptron (b) Adaline**

### 2.2.1 Classical ANN structures and Back Propagation

Roots of modern ANNs can be traced to the early 1960s. Some of the early models of artificial neural networks are Adaline [1960 Widrow] and Perceptron [1962 Rosenblatt], [1969 Minsky]. It was demonstrated that learning in Adaline follows a gradient descent, mathematically proving that learning can take place in an electronic circuit. Perceptron was implemented in analog hardware using variable resistors to scale the input.

Figure 2.2 shows the structure of perceptron and adaline neurons. They differ in the way error is computed. The error is a first order difference in Perceptron while it is the second order difference in Adaline. The weights in Perceptron are updated as

$$w_i(t+1) = w_i(t) + \eta\,(y-d)x_i \quad (2.1)$$

where $\eta$ is the learning rate. Adaline on the other hand, takes the square of the difference between the desired output and the computed output (before activation). These weights are updated as

$$w_i(t+1) = w_i(t) + \eta\,(y-d)^2 x_i \quad (2.2)$$

Note that the term $(y-d)$ is independent of the input node index. Therefore, both equations (2.1) and (2.2) indicate that the correction is proportional to the product of an error term and contribution of individual input. These networks demonstrated that neuronal functionality can be approximated in hardware. However, developments in ANNs took a downturn when it was shown that simple functions like XOR which are not linearly separable cannot be implemented using perceptron [1969 Minsky]. Their proof was considered as an indication that ANNs were not a viable solution for complex real life problems.

Neural systems are complex and therefore a single neuron or a small set of neurons may not represent performance of a neural system. Therefore, a simple 2-input 1-output system considered for implementation of XOR logic is hardly a model to represent the complexity of biological neural system.

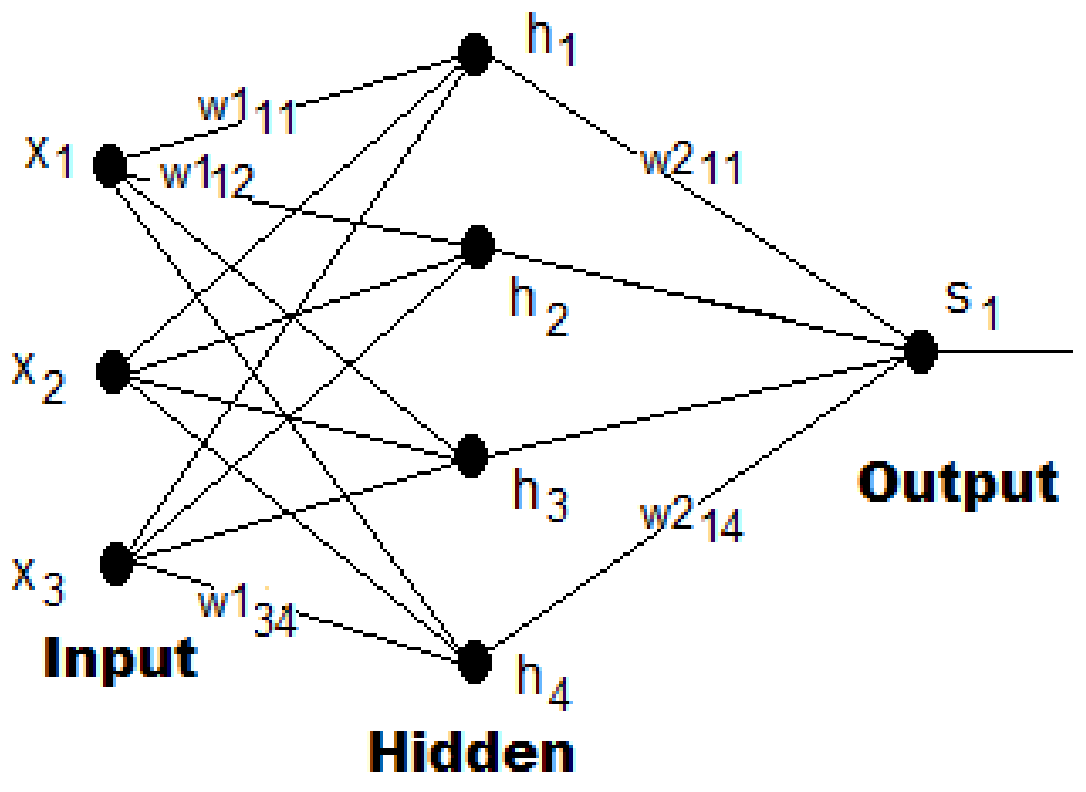

**Figure 2.3 3-layer neural network model for Back-propagation**

Basis of learning in neural networks like Percepton and Adaline happens by adjusting the weights in response to the difference between the observed and expected output. In a single layer, these quantities are completely specified. However, when new layer is introduced between the output layer and the input layer, the expected output from the middle (hidden) layer could not be estimated directly. There could be multiple sets of values that could be reached by the neurons in the hidden layer (Figure 2.3). This problem was addressed by the introduction of Back-propagation algorithm by Rumelhart et al. [1985 Rumelhart]. Work done by the Parallel Distributed Processing (PDP) group in southern California changed the way neural networks would be used. Back propagation algorithm consists of two step computation: Forward error computation and backward weight updates. In the forward calculation, existing weights were used to compute the network output. Then the difference between the desired and computed output is taken as error. Considering that error E at the output node is a function of the weight w, and there are as many inputs to the output nodes as the number of hidden nodes, it is possible to express a change to weight as a partial derivate

$$\Delta w_i = \eta \frac{\partial E}{\partial w_i} \tag{2.3}$$

The contribution of individual weight to the error cannot be known exactly. However it is possible to estimate how the error would change if the weight is changed by a small amount. It is possible to compute the desired output at the hidden neurons, which in turn may be used to calculate weights before the hidden layer. Back propagation algorithm computes the change in weights as

$$\Delta w_i = -\eta E w_i \varphi_i \tag{2.4}$$

where $\varphi$ is the value of the derivative of the activation function at the current output. Desired output at the nodes of hidden layer can be estimated once the change in weights is known. Similar procedure can be applied to weights connecting input neurons to hidden neurons.

### 2.2.2 Other Classical ANN Structures

The work at PDP group was based on a set of ideas which was referred to as ‘framework’ of neural computing, which formed the basis of back propagation algorithm. However, there were other alternate models of neural networks that were considered ‘outside the framework’ that were feasible but failed to attract enough attention. These models are still relevant in the context of ‘deep learning’, and hence discussed here.

All the models presented till this point use a bottom up approach: simulating the behaviour of a neuron in the hope that a collection of them will represent the whole of brain. One of the strong proponents of the idea that brain is more than just a collection of neurons and hence ANNs should be designed using a top-down approach (brain to neuron) was Grossberg [1987 Grossberg]. Some of his earlier work also showed that (short term) memory based systems can provide solutions to AI problems [1973 Grossberg]. Carpenter and Grossberg [2009 Carpenter] described several variants of their top-down ANN architecture called as Adaptive Resonance Theory (ART). Overall operation of ART is given in Figure 2.4(a). This network performs both analysis and synthesis. It introduces many concepts of biological learning to ANN architecture, viz., scale and orientation invariance, long and short term memory, attention system, clearing the recognition engine, etc. Essentially, the architecture moved the ANN design from a combinational to sequential system. ART networks memorize historic inputs to the network and compare them with incoming input. Several transformations are applied on the input and compared with existing templates. These transformations ensure that partial match can also result in recognition of input pattern.

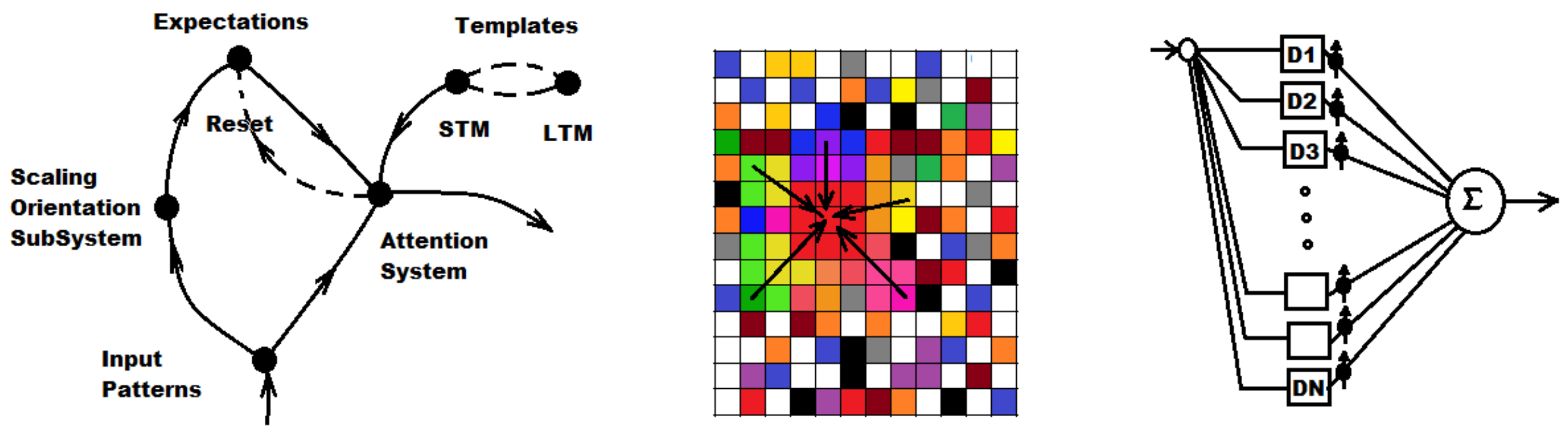

(a) (b) (c)

**Figure 2.4 Neural Systems (a) Adaptive Resonance Theory (b) Self-Organizing Maps and (c) Time Domain Neural Network**

Another interesting architecture called Self Organizing Maps (SOM) was presented by Kohonen [1990 Kohonen], see Figure 2.4(b). Here the neurons form an ordered sequence or a surface. Each neuron has a fixed location, described by Euclidean coordinates. Neurons are initialized to random values. All neurons receive all input values. Neuron with best matching stored value (or a hash) becomes the winner, if it is above a threshold. The winner forces neighbouring nodes to adjust their weights such that their output approaches that of the winner.

$$w(t+1) = w(t) + h_{ci}(t)\eta(t)(I(t) - w(t)) \tag{2.5}$$

where I(t) is the input vector and $h_{ci}(t)$ is a scalar kernel function of the form

$$h_{ci} = e^{(-\|r_i - r_c\|^2/\sigma^2)} \quad (2.6)$$

where $r$ is a Eucledian distance measure between the winner node $c$ and $i$-th node, $\sigma$ is a function of time. This allows the network to harden over period of time. As inputs are applied, the network forms regions of recognition. The network allows variations in input to be handled easily.

Both ART and SOM have a complexity of the order of ($O^n$) which implied that they could be used only in small systems, which was opposite of their aspirations. Capabilities of hardware available at that time were very limited and hence they did not attract much attention.

Applications that require temporal sequences as input also have been handled by introducing delay chains in every input line. Outputs from these delay chains could then be input to a regular ANN such as Perceptron or Back-propagation network. Several such architectures that convert a temporal order to a spatial order have been suggested. For example, Waibel et al. [1989 Waibel] presented a time delay neural network for speech recognition. A delay chain synthesizes the required input set (sequence) for a Perceptron like neuron. Figure 2.4(c) shows an overview of these networks. It is possible to use various learning algorithms on these networks. Wang and Arbib [1990 Wang] suggested a neural network capable of learning complex temporal sequences with short term periodicity. They claimed to introduce two new concepts in neural modelling, viz., short term memory and sequence conversion. Other researchers, especially Grossberg, had introduced these concepts in ART and developed working models by then.

### 2.2.3 Stability Plasticity Dilemma

One of the problems in understanding how the biological neuron works was the stability – plasticity dilemma pointed out by Malsburg [1987 Malsburg]. It is known that neural network in biological species stops growing after early childhood. So, it was not clear how the network acquires new learning. If the new information continues to occupy the same neurons which till that time stored older information, then the old information has to be lost to some extent to allow the new information to override the stored information. Introducing new nodes altered the behaviour of the complete network. This was clearly a problem. ART networks overcome this problem by allowing a clear mechanism to add new memory without losing old memory.

More than a decade after stability – plasticity dilemma was discussed by Malsburg, Doetsch and Buylla reported that certain types of cells called astrocytes in the mammalian brain are actually neural stem cells [1999 Doetsch]. Existence of such cells and their connection with dendritic growth and repair was also known. But the knowledge that they can create new neurons was completely new. This supports Grossberg's ART like architectures where new nodes (neurons) can be added without affecting the performance of existing network. ARN, used in this research also supports such addition of neurons to a mature network without affecting the existing neural network.

Many researchers of that time put effort in proving that neural nets can perform Boolean operations (like XOR) and implement Finite State Machines (FSM). The trend persisted for many long years even after the back propagation and other usable ANNs were reported in literature. For example, Siegelmann and Sontag used a recursive neural net to show that a generalized Turing machine can be implemented using ANN with finite number of neurons [1991 Siegelmann]. However, real life situations like Natural Language Processing (NLP) and image recognition and classification require machines more complex than what can be implemented with FSM. Implementing such super-Turing machines required a much finer understanding of neuronal dynamics. Research suggested that the temporal order or sequence of input/output should also be used as a parameter to model the neuron. As the neuronal activation and re-polarization are electro-chemical in nature and that molecular transport requires finite time, inclusion of temporal order in neuronal modelling gained significance.

### 2.2.4 Temporal Aspects of ANN

The stability – plasticity dilemma and the super-Turing machine required some new directions in ANN development. Thorpe [1990 Thorpe] took a step forward in this direction and suggested that the neuronal response depends on the interval between two spikes of input, essentially saying that a temporal coding mechanism is active in (vision related) neural networks. Thorpe gives elaborate examples on how and why spike interval allows fast image processing in biological vision systems. Important aspect of this discovery was that it could add a large degree of freedom, if not an unbound degree of freedom, to the representation of real world data. These networks implement a fading input model at the dendrites and a lossy integral at the neuronal output. Output of a neuron may be represented as

$$u(t) = \sum f_i(t)\, \delta_i\, (t) + s \quad (2.7)$$

where $\delta(t)$ indicates a spike, filtered by $f(t)$. The filter function performs a delayed decay of the spike. A small bias $s$ is also added where needed. This output may be integrated to achieve a low pass filter like behaviour, followed by thresholding, i.e., the neuron fires on persistent activation. The integration can also result in a slow reset of firing neuron. Recently, spiking neural network (Figure 2.5) has been used as basic structure in design of Intel's 'Loihi' neuromorphic processor [2018 Davies]. Details of this processor are discussed in section 2.5.3.

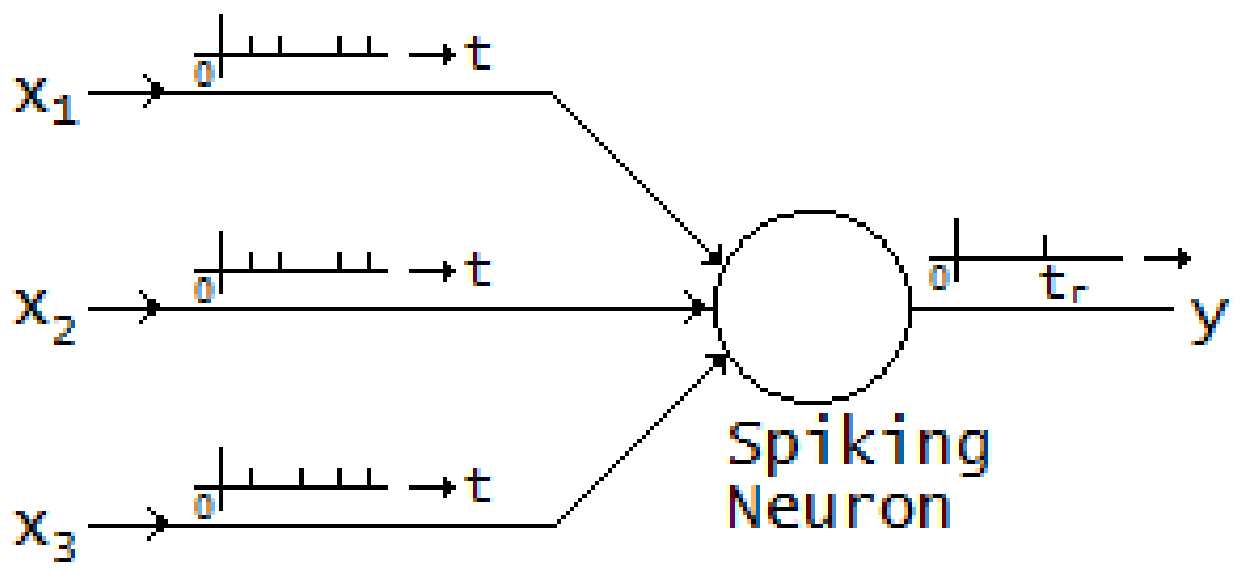

**Figure 2.5 Spiking Neural Network**

*Pathnet*, a temporal neural network architecture using ARN as basic module was suggested by Aparanji et al. [2018 Aparanji] for robotic motion planning. The paper shows how path optimization can occur in a time delayed neural network when polarization of neurons is influenced by the strength and arrival time of input. The path here refers to a series of transitions through the solution space (work area). The response of the neurons in pathnet take four potential levels labelled as *resting*, *threshold*, *activation* and *maximum* as shown in Figure 2.6. When the sum of scaled inputs reaches a threshold, the neuron fires indicating a recognition event. This firing is one time, after which the neuron goes to resting level. However, if the value reaches above the threshold to a level marked *activation* the neuron holds the axon potential for some time, before it starts decaying. This allows new paths to be explored. The hold time depends on the strength of activation. Note that the activation potential will not exceed a predefined maximum. ARN model used in Pathnet is an approximation system. The system can create chordal paths to reduce path length.

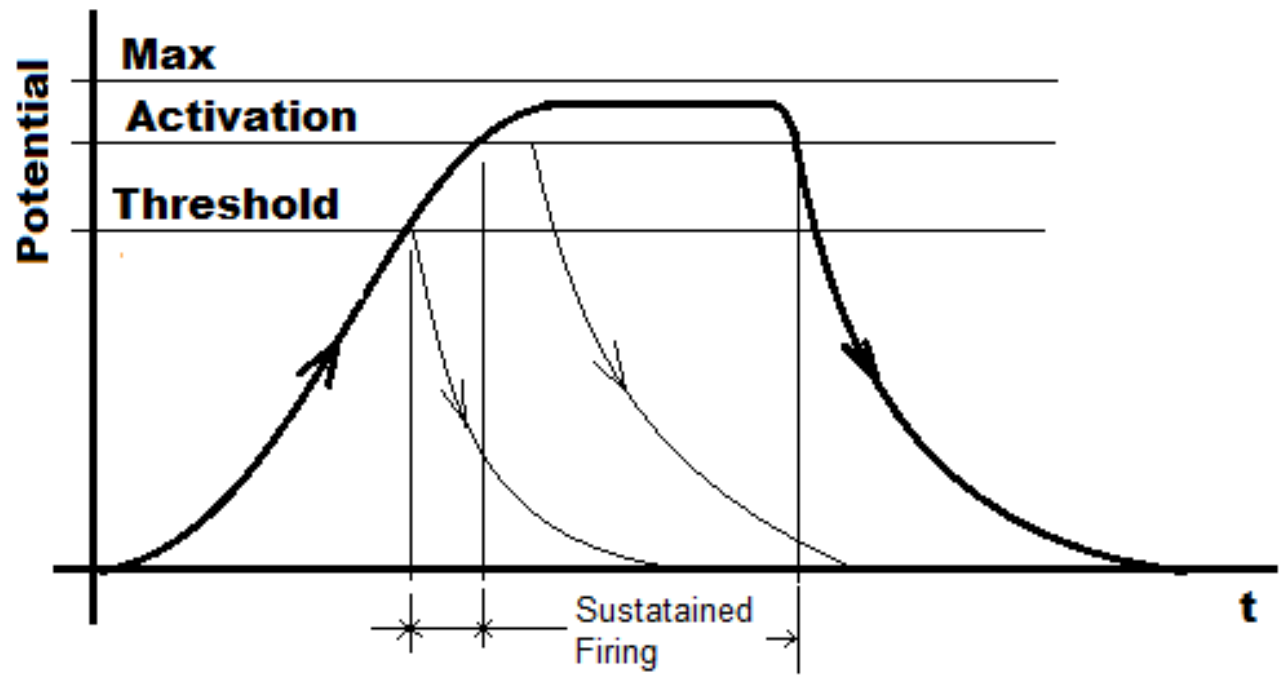

**Figure 2.6 Neuronal activation in Pathnet, from [2018 Aparanji]**

### 2.2.5 Modern Artificial Neural Networks

One of the early ANN structures to incorporate time domain effects was the Recurrent Neural Network (RNN) shown in Figure 2.7(a). These are similar to back propagation networks but some or all of the nodes receive input from previous outputs. This allows the network to respond to previous and present input. The network will be useful if older inputs can be made to influence present output. However, as the activation functions like sigmoid tend to push the computed output towards saturation quickly, this problem is documented as the problem of vanishing gradients.

Use of ANN for complex systems started from the days of back propagation network. For example, Fukushima et al. presented an ANN based system for image recognition as early as 1980 [1980 Fukushima]. However, it was only in recent years, Deep Learning (DL) systems, which evolved from ANNs, have been very successful in solving such complex problems. Some of the popular DL approaches that changed the computing scenario were Convolutional Neural Network (CNN) [2012 Krizhevsky], [1989 Waibel], Long Short-Term Memory (LSTM) [1997 Hochreiter] and Generative Adversarial Networks (GANs) [2014 Goodfellow] etc.

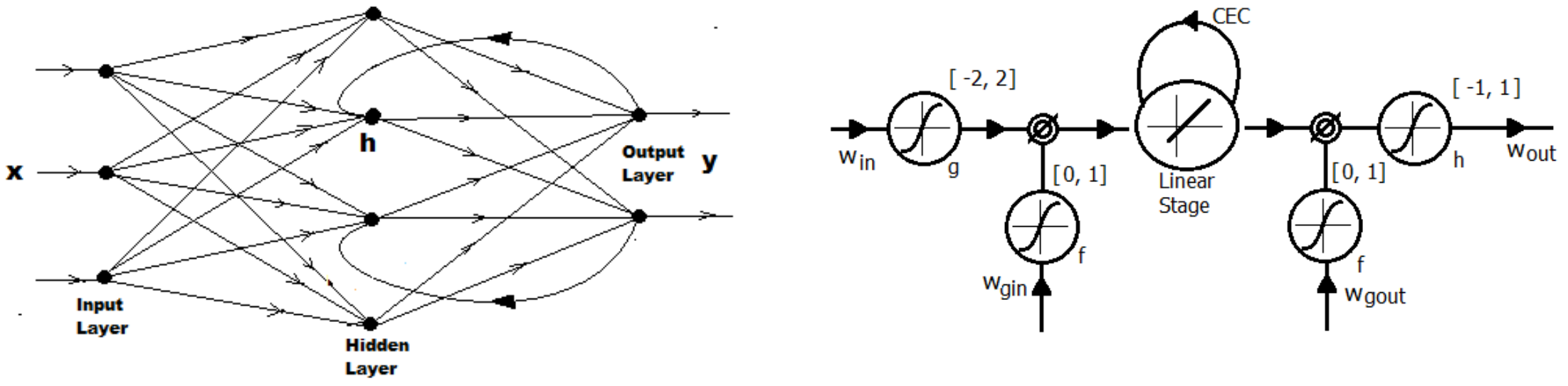

(a) (b)

**Figure 2.7 (a) Recurrent Neural Network and (b) Long Short-Term Memory**

Typically, recurrent neural networks can use few time steps in their design [2014 Schmidhuber]. In order to overcome this problem, Hochreiter and Schmidhber developed a somewhat more complex LSTM network [1997 Hochreiter]. The basic idea of LSTM was to keep the terms unchanged till required (see Figure 2.7(b)). A structure called Constant Error Carousel (CEC) with unit function as activation, allowed the stored term to be retained through thousands of time steps. This in turn allowed long time dependencies to be used as part of the output computation. To further facilitate the computation, signal gating was extensively used. LSTM also used at least three types of activation functions that gave it a good control over the recurrent loops as well as forward computation. LSTM has been very successful in time series analysis and prediction.

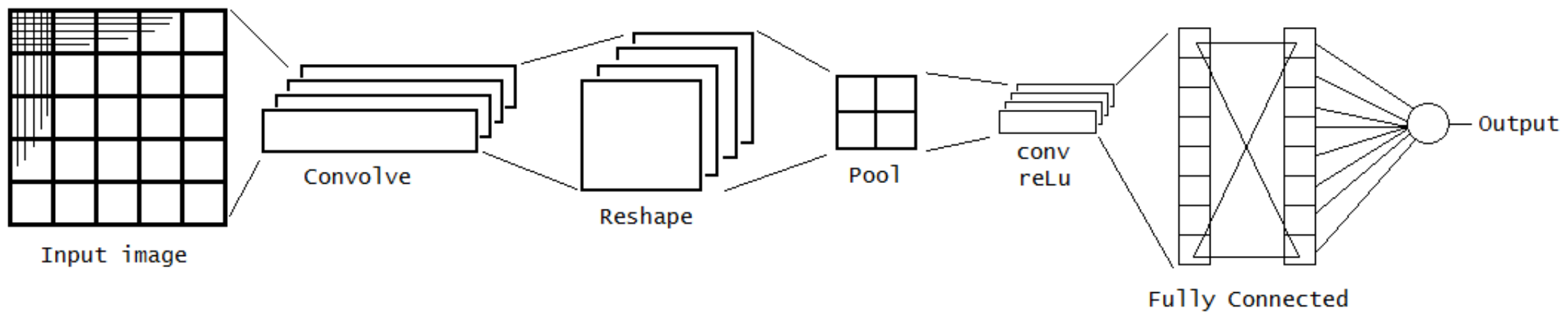

**Figure 2.8 Convolutional Neural Network**

Convolutional Neural Networks (CNNs) have been used successfully for image classification. CNN is not one specific structure like Perceptron and Adaline but an assembly of several components. Overall flow of computations is shown in Figure 2.8. The input image is scanned in small sized segments. These segments may or may not overlap. These segments are used to train a network that identifies features. These set of extracted features are rearranged to generate a set of vectors, in a process referred to as convolution. The output of convolution layer is input to a reshape layer which changes the set of vectors into matrices of a different size. Some of the operations that can be performed on the reshaped matrix are convolution, applying activation functions like ReLU, Pooling, etc. ReLU removes negative terms while the Pooling operation combines several cells in various ways. Max-pooling for example, picks up the maximum pixel value from a set of four or nine pixels. These operations are repeated several times. Near the end of this processing is a fully connected network which combines the generated values into one or more recognition values or labels. CNN is a complex network and hence computationally intensive. However, the results have been nothing less than spectacular.

As any other technology, deep learning systems have scope for improvements. Some of the leading researchers including Geoffrey Hinton [2017 Sabour] point out some such issues. For example, pooling layers that perform some approximation and image scaling incur loss of features. CNNs serialize the image and therefore loose spatial relations among the image features. Further improvements to CNN are being suggested by other groups as well. For example, replacing a convolutional layer with perturbation layer to reduce the size of the learning sample set has been reported in [2018 Juefei-Xu]. The authors of this paper acknowledge that using millions of images to train a digital neural system is an achievement in itself. However, the authors show that new input images can be perturbed with a set of pre-defined random additive noise masks, achieving accuracy comparable to larger input image set. It is reported that the method is more efficient than using a convolution layer. Simonyan and Zisserman [2015 Simonyan] report how the accuracy of a model can be improved by increasing the depth of a neural network. In general, it is now recognized that the depth (number of layers) of neural network increases the quality of recognition. Addition of a higher layer amounts to introducing a higher level of abstraction of the data. Therefore, adding layers improves overall quality of recognition. However, attention needs to be given to loss of gradient in long chains of neural layers. In absence of gradient, the networks get stuck to local minima, which in turn disengage the network from any further learning.

Another important issue in development of ANNs is the density of node usage. Having large number of rarely used nodes will increase computational cost without contributing much to the overall throughput. In that sense, most of the ANNs used in DL are sparse. Therefore, it should be possible to increase the overall performance by increasing the overall density of the network. However neural networks do not clearly demarcate the neurons in the path of recognition or their dependencies. So, pruning the networks becomes difficult. Developing explainable neural architectures can alleviate this bottleneck.

Many DNNs use stochastic gradient search algorithms, wherein every neuron corrects its weights on every input such that overall system reduces the error in recognition. It is not clear if biological systems also perform stochastic gradient search. Further evolution in ANN has to see several new types of neural architectures evolving, generalizing as well as specializing in the functionality of a neuronal type. For example, Hinton's capsule networks for image recognition are an important milestone in image recognition [2017 Sabour]. Use of Spiking Neural Networks (SNN) suggested by Robert Fujii et al. are an excellent example of such network [2007 Fujii]. Very interestingly, it has been pointed out by the designers of ‘Loihi’ processor at Intel [2018 Davies] that complex problem solving ability of neural networks without use of linear algebra is possible and likely to be faster and more efficient in terms of recognition and power consumption.

## 2.3 Auto Resonance Networks

Auto Resonance Network (ARN) proposed by Aparanji et al. [2016 Aparanji] is a resonance based approximator that can be tuned by the statistical parameters of local data available at the neuron. We have extended this work and provided more generalized equations and better learning algorithm. Full details are available in chapter 4. Please note that the terms node and neuron are synonymously used in the thesis.

Intelligence in ARN is due to its ability to change the locus and quality of resonance based on input and to some extent history of input data. The output of ARN is real value but constrained to a designated small upper bound which permits use of limited accuracy in computation. The output can be controlled by a small number of parameters whose values are derived from the input data. At the input layers, when neurons match a specific range of values to which it is tuned, the neuron fires. Such a neuron is said to be in resonance. Every node (neuron) in ARN represents a feature or a sequence of features. Multiple layers of ARN nodes can be used in a sequence or hierarchy, each higher layer representing a higher abstraction in recognition of input. Each node in a layer responds to a specific combination of inputs, independent of the other nodes in the same layer. Labelling of the feature recognized by ARN allows separate nodes to point to a single feature, allowing a group of nodes to cover any convex or concave set of input.

Perturbation techniques are used by ARN to reinforce the training and reduce the size of training set. Perturbed nodes are similar to biological mutations. Such nodes can be created by perturbation of input, output and the system parameters. Mechanisms to distinguish between real and perturbed nodes are described in [2017 Aparanji]. As the perturbed nodes can respond to similar input, we can say that they can be generated by applying an affine transform on existing node parameters. These properties of ARN are useful in building image recognition and classification systems. Increasing the number of training samples will increase the accuracy of recognition. But perturbation increases the effective number of samples available to a network and hence improves overall learning efficiency of ARN. Therefore, often small sets of input are sufficient to train ARN.

ARN can be modified to suit an application at hand. For example, the type of perturbation used in robotic path planning [2017 Aparanji] is very different than that used in image recognition [2019 P7]. It is also possible to implement various network parameters with a given type of application. Some of these possibilities include the choice of resonator, spatial spread of neural input, neuronal density, selection of threshold and resonance control parameter (tuning or coverage parameter) $\rho$ (rho), perturbation parameters, number of layers, etc. It is also possible to inter-spread resonators, constrained Hebbian nodes [2017 Aparanji] and logical decision making layers to improve overall quality of recognition.

Multi-layer ARN is a generic data classifier capable of classifying non-linear input data. ARN allocates specific paths to be associated with a recognition event and establish a priority on recognized paths. Therefore, it is easy to find how and why ARN performed a particular task. Paths that are not significant contributors to performance of the system can be truncated from the network. This allows compaction of the network. Temporal behaviour of ARN and its application has also been discussed in literature. Use of ARN for image recognition is reported in [2019 P3], [2019 P7] and in chapter 4 of this thesis.

## 2.4 Impact of DNNs on Processor Design

Generic use of neural nets as computing devices was discussed by Siegelmann and Sontag [1991 Siegelmann]. The authors show that if a problem can be solved by a computer then the same is also possible by a neural net. They also noted that computations in brain rely on different strategies than a conventional computing. Architecture of neuro-microprocessor called SPERT was designed using moderate precision-arithmetic for Back-Propagation Networks (BPN) [1993 Wawrzynek]. The processor is capable of performing $10^{11}$ connections per second. Most of the modern neural hardware uses a digital control circuit which can also perform logic operations.

A massively parallel SIMD computer architecture called MM32k, capable of implementing highly parallel neural network architectures is reported in [1994 Golver].

It has 32768 bit serial processing elements connected by a 64x64 crossbar switch network. It acts as a coprocessor to accelerate vector operations. The most commonly used operation in neural network classifier i.e., multi-dimensional nearest-neighbour match is included in this design. Three different examples of neural networks viz., Radial Basis Functions (RBF), Kohonen's Self-Organizing Maps (SOM) and neocognitron were implemented using this design.

An analog processor architecture for non-parametric classification of data has parallel implementation of analog VLSI processor [1994 Verleysen]. It involves the operations like addition, multiplication, distance computation and nonlinear functions.

Omondi et al. described two ways of implementing an arithmetic unit for neural network processors [2000 Omondi]. The first one is to have a combination of basic neural networks capable of performing weighted sum, activation and weight assignment operations. The second one is to have a special arithmetic unit capable of performing operations specific to the neural network and its applications. The author also argues that conventional processors can be used for neural computations. However, at the time of this publication, the modern neural networks like CNN were not well known.

CNP is an FPGA (Spartan and Vertex) based processor for Convolutional Networks (ConvNets) [2009 Farabet]. A software compiler is used to generate the instructions from a trained ConvNet. The architecture of CNP contains Control Unit (CU), a Parallel/Pipelined Vector Arithmetic and Logic Unit (VALU), an I/O control unit, and a memory interface. The operations of VALU are controlled by CU, which is based on PowerPC architecture. The VALU is designed to perform the instructions specific to ConvNet namely 2D convolution, spatial pooling/ subsampling, point-wise non-linear activation functions, square root, division etc. A dedicated hardware arbiter is designed to provide the common memory interface. At a time, 8 ports are allowed to read/write from/to the memory. This simultaneous access to the memory is achieved using FIFO strategy with a limited time slice for each port. All the basic operations required by a ConvNet are implemented at hardware level and provided as macro instructions. 2D convolution is implemented using shift and dot product operations and the *tanh* function is implemented using a piecewise linear approximation. This architecture is used to implement a face detection system.

Modern trends in computing indicate a paradigm shift driven by DNNs and AI [2014 Schmidhuber], [2015 LeCun]. DNNs require massively parallel processing capability and GPUs offer such capability at low cost. However, GPUs were primarily designed to handle display tasks which are different than ANN tasks. Some functions like non-linear activation functions are not a part of GPU designs. Noise tolerance of artificial neural network systems implies that numerical accuracy is not very critical to successful implementation of ANN hardware. However, as the number of computations is very large, speed of computation is critical to success. It is to be noted that companies like NVidia are developing GPUs primarily targeting DL applications.

In neural architectures, most of the time is spent in data movement rather than the computations. The data movement dominates energy consumption and therefore, some research on in-memory computing (IMC) and low precision arithmetic [2017 Sze] is necessary. The challenge of memory bottleneck in neural computing systems is addressed by using a ReRAM based memory architecture, which is configured as Neural Network accelerator or a regular memory as required [2016 Chi]. Numerical precision required by DLNN varies with the type of the neural network and a type of layer within a neural network. Therefore, a precision based approach may be useful to improve the energy consumption. The hardware accelerator called Strips (STR) uses a bit-serial computation to improve the energy consumption and performance of neural computing [2017 Judd].

The impact of low precision multiplication on the training error is reported in [2015 Courbariaux]. The authors have trained Maxout neural network with three different number formats namely fixed point, floating point and dynamic fixed point. The network was trained for three different data set viz., MNIST, CIFAR-10 (Canadian Institute For Advanced Research) and SVHN (Street View House Numbers). The authors have reported that it is possible to train and run the networks by reducing the precision of multipliers. Further, author also suggests the use of dynamic fixed point number format for training deep neural networks.

Recently, Intel has announced its numerical format called flexpoint, designed for deep learning systems [2017 Koster], [2018 Rodriguez]. It is reported that, performance of a network with 16-bit flexpoint closely matches with that of 32-bit floating point. Many researchers are focusing on the numerical representation [2018 Ortiz], [2018 Hill], [2018 Johnson] because the performance depends on the speed of computation rather than precision and accuracy of calculation. Reports suggest that it is possible to use 16-bit multipliers for training and 8-bit multipliers or fewer for inference and achieve the same performance as compared to 32-bit multipliers [2018 Rodriguez]. Intel's AVX 512 instructions for vector processing also use 16-bit multiplications.

It is interesting to note that our work, presented in a conference in 2018 [2018 P9], shows why bit serial computation is faster than parallel computation in a massively parallel environment. The paper highlights the need to use single wire asynchronous communication over bus oriented communication.

Recently, Intel released a white paper, which reviews the importance of In-Memory Computing (IMC) in today's big and fast data applications [2017 Intel]. These RAM are placed across the Processing Elements (PEs) so that they can perform the computations from local memory. It is further reported that data consistency and scalability can also be achieved with the use of IMC.

A neural accelerator called RAPIDNN [2019 Imani] maps all the functionalities like activation function, pooling, multiplication and addition inside the memory block. It consists of DNN composer and accelerator: composer converts the neural operation to a table to be stored in the accelerator memory and accelerator performs in-memory computing using those stored values. Accelerator contains a crossbar memory to store the input data processed by a neural network and a resistive neural acceleration (RNA) memory for processing the data. Each RNA has multiple memory banks to perform parallel computations and the results are stored in FIFO buffer. One RNA block is designed to compute the output of one neuron (including multiplication, activation and pooling).

Popularity of DL computations has influenced commercial processors also, e.g., Intel AVX-512 extensions added four new instructions called Vector Neural Network Instructions (VNNI) to ease CNN implementations [2014 Intel AVX]. These new instructions support 'multiply and accumulate' operations that can be concatenated to implement multi-operand operations.

Some of the challenges in the design of multi-operand adders for CNNs and the problems faced during the implementation are discussed in [2018 Abdelouahab]. A multi-bit adder using approximate computing methods is reported in [2018 Tajasob]. An optimized multi-operand design based on Wallace tree reduction (series of half and full adders) is presented in [2019 Farrukh]. It is proposed as a replacement for typical two operand tree structure. A group of 8 or 16 inputs is given to Wallace tree and the final results are added using a traditional two input adders. The implementation of 16×4 (4 inputs of 16-bits each) and 16×8 and 16×16 adders is presented. An algorithm for multiple operand addition and multiplication using bit-partitioning is proposed in early 1973 [1973 Singh]. Design of multi-operand binary adder using two methods viz., a tree of 2-input 1-output adders and a tree of 3-input 2-output carry save adders is discussed in [1978 Atkins].

To summarize, the following points may be noted:

a) Low precision computation is both necessary and sufficient. Fixed point arithmetic is preferred.

b) Number of computations can be very large. Hence the computation has to be fast but power consumption should be low.

c) Serial connectivity is useful. Dividing the input to smaller sets and pipelining the transfers is also useful.

d) In-memory computations can reduce data transfer load.

e) SIMD and Streaming instructions can increase the throughput.

f) Hardware accelerators for regular maths functions are to be implemented.

g) Multi-operand addition is difficult to implement but necessary for faster computation.

## 2.5 Contemporary Processors for DNN implementation

As the complexity of DL networks is growing, the computational power required to implement such networks is increasing. Graphics Processing Units (GPUs) have been taking some of this burden [2016 NVIDIA]. Generally, a neural processor will contain a large number of compute-cores in a massively parallel configuration. Instruction set of such processors should allow these compute-cores to be reconfigured or provide facilities to transfer data between these cores. However, as the demand on DL/AI is increasing, processors/accelerators designs for neural networks are being developed. Several DL/AI specific hardware designs like Cambricon [2016 Liu], IBM TrueNorth [2015 Akopyan], Intel [2017 Rao] and Google TPU [2017 Jouppi] have been reported in recent past.

Some of the recent processors for DL/AI are discussed below.

### 2.5.1 NVIDIA CUDA GPUs

NVIDIA developed CUDA as a parallel computing platform and programming interface to provide access to the parallel compute engines and memory inside their general purpose GPUs. It provides kernel-level support for hardware initialization, configuration and a device-level API for developers. C programming language can be easily used to develop applications with CUDA. However it can also be integrated with other programming languages such as Java, python, etc. It has shared memory, which can be used by the threads for communication. Integer and bitwise operations are supported and the compiled code can be directly run on GPU. These GPUs are credited largely for the developments in modern AI. At a time when no massively parallel and user programmable CPUs available, CUDA GPGPUs provided an opportunity for the industry to evolve. Top of the line NVidia K80 dual GPU card deliver 2.9 double precision tera flops and 8.7 single precision tera flops with a peak 480 GB/sec bandwidth.

### 2.5.2 Cambricon IP for ANNs

Cambricon for Neural Networks [2016 Liu] proposed by Cambricon Technologies was possibly one of the first among ISAs designed for DL applications. It uses a 64 bit load-store architecture that offers data level parallelism, customized vector/matrix instructions and on-chip scratch pad memory. The silicon IP contains 64 32-bit general purpose registers. It supports complex instructions like computation of element-wise exponential of vector which are not covered by general purpose linear algebra libraries but frequently used in DL applications. The instructions for ANNs require variable-length access to vector/matrix data and therefore, fixed-width register set is no longer effective to access the data from memory. For this reason, Cambricon replaces fixed-width register set with on-chip scratch pad memory to support flexible-width data accesses. Scratch pad memory is visible to programmers and compilers. DL models such as CNN, RNN, SOM, SVM, k-NN, k-Means, decision tree and other algorithms are supported. Cambricon ISA is used in Huawei Kirin 970 and higher processors for mobile computing.

### 2.5.3 Intel processors for AI

Intel's Nervana Neural Network Processor (NNP) [2017 Rao] is specifically designed to provide the flexibility required by DL systems. It is built with new software managed memory architecture that supports the data movement instructions such as matrix multiplication, convolution and etc. This design comes with high speed on-chip and off-chip interconnects to enable bi-directional data transfer across multiple chips. Numerical computations on each chip are constrained by power and memory bandwidth and not by resources. NNP follows its own low-precision 16-bit numeric format called *flexpoint* [2017 Koster] to achieve increased performance while simultaneously decreasing power consumption. Flexpoint has a common exponent which is used for integer values in a tensor and it is adjusted dynamically so as to minimize the overflow and maximize the range. This format has been validated by training AlexNet, deep residual networks and generative adversarial networks (GANs). Area and power requirements are reduced by use of flexpoint compared to floating point. Majority of computations in deep learning can be performed using fixed point operations, as common exponent is shared among all the tensor elements.

Modelling of spiking neural networks in silicon is reported by Intel [2018 Davies]. The neuromorphic chip called *Loihi* is a 60-$mm^2$ chip fabricated in Intel's 14-nm process. The spiking neural network is implemented by incorporating time as an explicit dependency in their computations. The architecture of Loihi has 128 neuromorphic cores embedded with three x86 processor cores, 33MB of SRAM, 16MB of synaptic memory and off-chip communication interfaces. Each neuromorphic core implements 1024 spiking neural units (130,000 neurons and 130 million synapses). This provides a total of 2.1 million unique synaptic variables per $mm^2$, which is three times higher than that of IBM's TrueNorth [2015 Akopyan].

Intel has an open source library called MKL-DNN (Math Kernel Library for DNN), for deep learning applications. It includes the necessary building blocks required for neural implementations. This library is optimized for Intel processors. Compute intensive operations such as convolution, inner product, pooling, activation functions etc. are supported.

### 2.5.4 IBM TrueNorth

TrueNorth is a brain inspired, neurosynaptic, 70mW reconfigurable silicon chip with 5.4 billion transistors fabricated in Samsung's 28nm LPP process technology [2015 Akopyan]. It has 4096 neural cores arranged in 64x64 array, capable of implementing one million neurons and 256 million synapses. Each core represents a neural network with 256 neurons. Each of these neurons has 256 synapses. Recently, IBM has delivered the largest neurosynaptic computer with 64 million neurons (64 TrueNorth chips); however the details have not been disclosed online [2019 DeBole].

TrueNorth is configurable to support wide range of applications, specially targeting sensory information processing, machine learning, and cognitive computing applications using synaptic operations [2016 Tsai]. A novel ecosystem for implementing neural networks based on TrueNorth is reported in [2016 Sawada]. This ecosystem includes hardware platforms, training algorithms, software languages, tools, libraries and all that is required to implement neural network applications.

### 2.5.5 Google TPU

Recently, Google has introduced a custom, domain specific hardware called Tensor Processing Unit (TPU) to accelerate the inference phase of neural networks. TPU has 256x256 Matrix Multiply Unit, which can perform 8-bit multiply and additions on both signed and unsigned integers, which is 25 times as many MACs and 3.5 times as much on-chip memory as the NVIDIA K80 GPU. TPU ISA is similar to CISC with 10-20 average clock cycles per instruction. It uses 4-stage pipelines for matrix multiply instruction. Floating point numbers are transformed to 8-bit integers using quantization technique. It is reported that 8-bit integer multiplies have 6x improvement in area and energy, while 8-bit integer additions have an improvement of 13x in energy and 38x in area when compared with IEEE 754 16-bit floating point. This performance improvement is achieved by reusing the weights across independent examples during training and inference. There is on-chip weight FIFO, which reads from off-chip 8GiB weight memory. A programmable DMA controller is used to transfer the data to or from CPU. The applications of TPU can be seen in several Google products like photos, Gmail, voice assistant, search assistant, translate and etc.

### 2.5.6 ARM DynamIQ

*DynamIQ* for ARM Cortex is designed to meet the challenges and opportunities in AI machines. It has dedicated instructions for AI & ML with faster and more complex data processing. It is reported that dynamIQ performs AI computations with 50x increased performance compared with Cortex-A73 systems [2017 Wathan]. It supports up to 8 compute cores on a single cluster with SoC, where each core can have different performance and power characteristics. ARM claims that both high performance big CPU's and high efficiency little CPU's can be built using this architecture with shared coherent memory. More computation on CPU leaves less information on cloud, resulting in to a more secure AI experience required by *computing at the edge*.

### 2.5.7 Videantis v-MP6000UDX

A new DSP based deep learning and vision processor named v-MP6000UDX has been introduced by a German vendor named Videantis [2018 Videantis]. The architecture is scalable up to 256 cores and has instructions specially optimized to run ConvNets to address low cost to high performance applications. It is capable of performing 64 Multiply and ACcumulate (MAC) instructions per cycle per core, giving a a total of 16384 MACs per cycle. It has its own multi-channel DMA engine for data transfer to/from on-chip and off-chip memories. VLIW/SIMD media processors are combined with a number of stream processors to make a heterogeneous multi-core architecture. The company also announced a new tool called v-CNNDesigner to design and deploy neural networks using Tensorflow, PyTorch, Caffe and other popular frameworks. The tool is capable of converting FP32/FP16 to INT8/INT16 formats for optimizing the calculations.

### 2.5.8 DaDianNao for ML

A Chinese chip called *DaDianNao* is another interesting accelerator for neural networks. It can perform low-energy execution of large CNNs and DNNs, beyond a single-GPU performance [2014 Chen], [2017 Tao]. It provides a throughput of 452 Giga operations per second on a 3.02mm$^2$ 485mW chip. The authors call the chip as a *machine learning super computer*. The chip implements four essential components of CNN, viz., convolution, normalization, pooling and classifiers. It is capable of performing the basic operations of neural networks such as synaptic weight multiplications and additions. Several activation functions like sigmoid, tanh, ReLU, argmax, etc. are implemented at instruction level. The chip design incorporates many new architectural details, e.g., the chip has no main memory and all storage is at or close to the neuron. It is possible to build arrays of processors that can handle larger neural network. It contains buffers for input neurons (NBin), output neurons (NBout) and synaptic weights (SB), connected to a computational unit called Neural Functional Unit (NFU) and control logic. The chip design is based on a single node NFU, which performs the computations of classifier layers (multiplication, addition, sigmoid), convolutional layers and pooling layers. 16-bit fixed-point arithmetic operators are used and the outputs are truncated to 16 bits. It is possible to design versions with several NFUs on a chip. Overall performance of the chip is reported to be around 4 times that of a comparable GPU.

## 2.6 Data transfer challenge and re-configurability

Deep learning processor architectures are characterized by large number of low precision parallel operations using distributed memory. Artificial neural networks are designed to behave like biological neuronal bundles. Each of these neurons in the bundle performs a non-linear transformation on input data and communicates the results to hundreds to thousands of other neurons. Each neuron performs scaling, summation and a non-linear transformation on the input data supplied by the synapses to the cell. As the data is pushed through the network hierarchy, the processor should be able to reconfigure the underlying modules into structures best suited for such operations. This calls for a dynamically changing processor configuration.

Traditional processors are designed to consume more time in computation compared to the time taken for data movements. However, DL networks are characterized by large number of interconnections between the neurons. This increases bandwidth required for data transfer. As neural hardware becomes more capable, the number of synapses is bound to increase, directly increasing the time required to communicate among various neurons, over limited transfer lines. Accessing a central memory, to read and write from compute nodes (representing neurons) and distributing the output to nodes on the next layer, etc. can present several bottlenecks and hence design challenges. The problem becomes more severe as several layers of neural hardware compete to get data from a common pool of memory.

To overcome this bottleneck, DaDianNao chip does not use heap memory at all [2017 Tao]; all the memory is located at the compute cores called Neural Function Unit (NFU), reducing the external memory transfer. Use of programmer accessible local scratch pad memory at NFUs will also reduce the demands on memory bandwidth. The authors mention that it is impossible to meet the memory bandwidth on-chip. Therefore, the accelerator is implemented as a multi-chip module using electrical and optical interconnect.

Another Deep Learning Accelerator Unit (DLAU) for large scale DL networks is reported in [2017 Wang]. The methods used in the accelerator minimize the memory transfer and reuse compute units to implement large neural networks. A tiling method is used to divide the massive input data into smaller subsets, which are then put in a FIFO buffer to be executed in a timely manner. The operations like matrix multiplication and activation functions are supported by DLAU. Three architectures namely TMMU (tiled matrix multiplication unit), PSAU (part sum accumulation unit) and AFAU (activation function acceleration unit) are used to compute over streamed data. TMMU performs the multiplication of a tile obtained from input buffer with the corresponding weights obtained from the memory. The PSAU is responsible for accumulating the partial sums generated by the TMMU and send the final sum to AFAU which will then compute the activation function using piecewise linear interpolation. The architecture is tested with MNIST dataset and the training of a network is done using MATLAB. The trained values are applied to DLAU memory to compute 32 hardware neurons with 32 weights per clock cycle.

Eyeriss is an energy efficient accelerator for deep neural networks [2017 Chen]. Reducing the data movement is the main design key of this accelerator as it has a large contribution to achieve energy efficiency. This is in turn achieved by reusing the data locally, than accessing it from the memory all the time. A new compute scheme called *Row Stationary dataflow* is used to support the parallel computation while minimizing the data movement from both on-chip and off-chip memories. Further, a reconfigurable hardware is used to minimize the data movement cost by reusing the data and to support different network size/shape. In addition to these, data statistics (such as ignoring the input values with ‘0’) have also been used to reduce the energy consumption. It contains 168 processing elements arranged in a four-level memory hierarchy. The performance of this architecture is tested with AlexNet, VGG-16 and it has also been integrated with Caffe.

Another solution to overcome the data transfer overhead by reconfiguration is reported in [2016 Schabel]. The algorithms used in hierarchical ANNs change according to the requirements posed by the end application and the desired performance. Sparsity of ANNs such as Hierarchical Temporal Memory (HTM) [2011 Hawkins] and Sparsey [2010 Rinkus] poses another level of difficulty. According to the authors, these networks can switch into learn mode for the new data set without completely retraining the network. Due to this capability, the networks are memory intensive and can benefit from accelerator with memory co-design. The custom processor can achieve high parallelism using a configurable SIMD architecture with processing in memory (PIM) as a solution for memory bottleneck issue of neural computing. Fused instructions and operational blocks are introduced in execution pipelines to reduce data transfers. Each processing element (PE) has one 16-wide SIMD engine with its own register file and a scratch pad memory. Several PEs are connected to a common shared memory. Custom instructions are allowed to be added to SIMD engine by a PE. This architecture includes implementation of Sparsey networks having 93,312 neurons in three layers [2010 Rinkus]. Data-level parallelism has been achieved at the neuron level (each neuron has thousands of synaptic connections), while the instruction and thread-level parallelism has been achieved at PE level.

The number of parallel computations to be performed can be estimated by looking at how typical CNN systems are designed. For example, AlexNet [2014 Krizhevsky] uses more than 650 thousand neurons. The size of neural network may vary depending on the learning type, end application and other performance requirements. Therefore, neural hardware should be reconfigurable to suit the different neural network suite. In general, it is necessary to move towards compute intensive SIMD and MIMD streaming architectures that will reconfigure the hardware to suit the configuring instruction set or end application.

Multi-operand adders are required at the output of a neuron to add the outputs from a previous layer. As the number of inputs in ANNs is very large and dynamic, there is a need to design reconfigurable multi-operand adders. Some of the design issues to be considered while designing multi-operand adders and the algorithm to implement larger adders is discussed in chapter 6 and presented in International conference [2018 P9].

## 2.7 Frameworks and Software Tools for DNN Development

Several DNN development frameworks are available to provide a platform for designing, training and validating deep neural networks. For example, Eclipse DL4J Java libraries, MXNet from Apache (2019), Theano [2010 Bergstra] from Univ. of Montreal (2007), Caffe [2014 Jia] from Univ. of California at Berkeley (Dec 2013), TensorFlow [2015 Abadi] from Google (Nov 2015), PyTorch by Facebook (Oct 2016), Intel's Neon etc. are popular. Most of these platforms implement low level code in C++ and provide Python API for easy implementation. Most of them can run on CPU or use NVIDIA CUDA GPU for faster operations.

There are several Software Development Kits (SDKs) available to design and deploy deep neural networks using the above mentioned frameworks. NVIDIA cuDNN (CUDA Deep Neural Network) is a GPU accelerated library which provides the support for DNN operations such as convolution, pooling, activation, normalization and etc. Intel's SDK supports few frameworks viz., Tensorflow, MXNet and Caffe for the applications such as image recognition, object detection & localization, speech recognition, language translation, recommended systems and etc.

## 2.8 References Analytics

The graph of number of publications referred vs. published year is shown in Figure 2.9. It is an indicative of how the research is evolved.

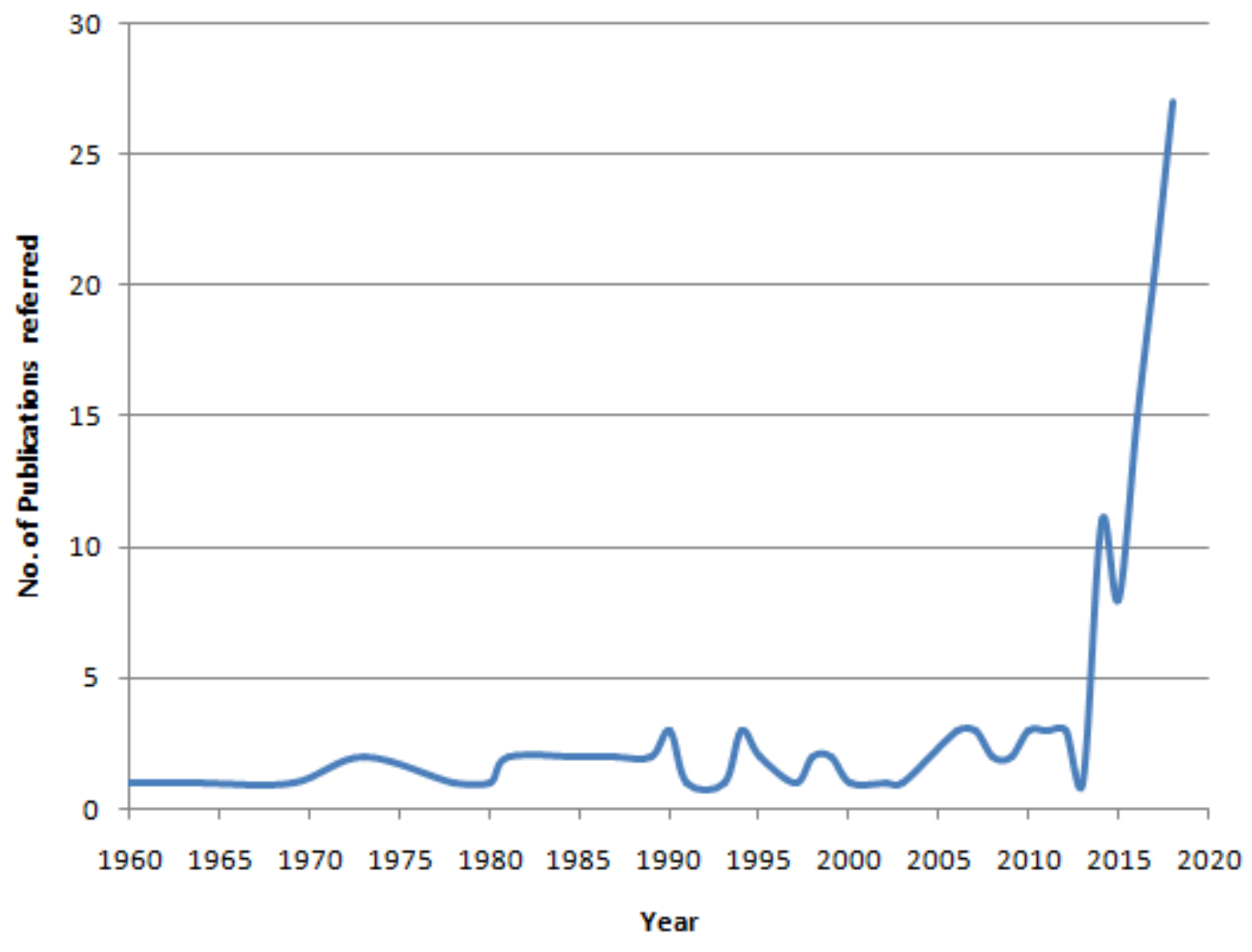

**Figure 2.9 Reference analytics**

This literature survey has helped me understand different neural architectures. As a result, it was possible to identify common operations required by neural networks and the complexity involved in their implementation. For example, while I was working on the 16-bit number format for ARN, Intel just released their new number format. It was an encouraging factor for me to continue the work. During later stages of this research, I could see many groups focused on the low precision arithmetic for neural computations.

Developments in IoT and AI will have an impact on the design of processors, instruction set architecture, memory organization, bus design, on-chip communication, etc., all of which need to be explored because all this is driven by a fundamental change in the complexity to be handled by new processors. Moving from a scalar design to a vector or matrix design can be seen as a continued evolution but the massive parallelism of ANNs throws new challenges that need complete rethinking of design strategies. In recent years several new types of neural networks have been introduced to support specific applications and many will follow. Biological systems use thousands of types of neurons to achieve highly efficient sensory and motor control. Therefore, it is imperative that we need to explore many more ANN architectures on our quest to reach anything close to Artificial General Intelligence (AGI). This will require both evolutionary and exploratory approaches to find new architectures.

Some of the contemporary issues to be considered for the design of processors/accelerators for neural networks are presented in the next chapter.

# Chapter 3

# Contemporary Issues in Processor Design

Processor design is going through revolutionary changes imposed by two trends in technology, viz., Internet of Things and Artificial Intelligence. Both technologies have come out of long held dormancy and growing at an exponential rate. Interestingly, much of these developments owe their success to advances in hardware technology and networking. New processor designs have to address the needs of these technologies to penetrate the market. Developments in Field Programmable Gate Arrays (FPGAs) with analog and radio components on chip have enabled a new market for customized processor designs.

Internet of Things, when viewed in isolation, is an extension of the existing technologies. Main features to look for in a chip for IoT are Fast CPU with large memory, DSP on chip, analog IO ports, system specific hardware like sensors and drivers, RF on chip, battery management, data encryption, etc. Typical applications of IoT involve data logging and device control. They could be used in everything everywhere, viz., industrial and building automation, safety, vehicular automation and tracking, agriculture and animal husbandry, forest canopy management, home appliances and hand tools, etc. Microcontrollers like CC3200 from Texas Instruments and i.MX-RT600 from NXP Semiconductors incorporates many such features [2019 NXP]. Integration of web services has provided a stable interface so that products can become vendor neutral, without which, the growth in IoT was stunted.

IoT scenario is undergoing the next level of evolution in the last couple of years. Processors are becoming increasingly capable of implementing complex compute intensive DNNs on chip. This opens whole new technological capability, generally called *AI at the edge*. Till now, AI services were accessed from the cloud. However, AI at the edge allows real time object recognition on a single processor. This is enabling technologies like driverless vehicles from Tesla and new generation robots like Atlas from Boston dynamics that can perform local AI based decisions. Object recognition on edge devices can be implemented with new AI enabled processors like Videantis MP6000UDX [2018 Videantis]. DaDianNao chip implements CNN related functions on chip, allowing vision based application development [2017 Tao]. This chip is designed with a focus on single neuron architecture with local memory. Eyeriss implements a special data flow architecture called Stationary Row Dataflow [2017 Chen].

Apart from the popular CNN and LSTM, there are many other DL algorithms and structures, e.g., Recurrent Neural Networks (RNNs), Auto-encoders (AE), Generative Adversarial Networks (GAN), Restricted Boltzmann Machines (RBM), Radial Basis Functions (RBF), word2vec, etc. which address specific domains. Many more AI techniques and DNN architectures will be developed in next few years. While general purpose GPUs and TPUs can address the server market, there is a need to study processor level implementations of these algorithms to be able to implement AI at the edge. The following sections discuss some of the key issues in implementing these processors.

## 3.1 Instruction Set

Instruction sets of Turing machines in both Princeton and Harvard architectures essentially implement *instruction flow* design [1994 Hazewinkel]. On the other hand, neural networks are best implemented as *data-flow* machines [2014 Graves]. Current GPU implementations use MIMD or streamed SIMD architectures. Streamed processors are data-flow machines with implicit or programmed computational chain. Streamed-data instructions persist over long data sets, performing the same operation over the entire data set [2006 DeHon]. This saves the time required to perform fetch-decode steps to be performed at every unit of data. Similarly, systolic arrays and data flow architectures can be used to perform Fused Multiply and Accumulate (FMA) instructions [1981 Kung]. It is possible that a fixed set of instructions are run on a data stream. Therefore special hardware and instructions for gating the data over programmed pipelines will be necessary. Further, computational chains need to be implemented without intervening memory calls, to be computationally efficient. Otherwise, the memory-register data transfer overhead will dominate the CPU time.

Instruction sets for DNNs support scalar, vector and tensor operations. Introduction of newer instruction sets like Intel AVX [2014 Intel AVX] are an example. Many operations of DNNs can be implemented using linear algebra, but implementing non-linear activation functions can be a challenge task. Using Taylor series expansion for such functions is computationally expensive. Neural computations can tolerate small errors in computation and hence use of low precision arithmetic is often sufficient.

Therefore, study of approximation methods for implementation of activation function is necessary for efficient hardware implementation of non-linear functions.

Updating synaptic weights during training is an essential part of learning. If memory has to be accessed at every step, data transfer can choke the processor performance. *Computation-in-memory* provides some answers but fragmented memory reduces the memory density (see DaDianNao, Section 2.5.8). On the other hand, mapping heap memory to individual cores can become complex. Several other candidates for hardware implementation of instructions may exist, which need to be explored with reference to the type of network being implemented.

## 3.2 Training & Run-time Environments

In a conventional processor, an algorithm works the same way during development (debug) and run time environments. However, the scenario is very different when handling DNNs: They require computationally intensive training, which has no equivalent in conventional algorithms. Training DNNs is a slow process as millions of weights have to be computed and adjusted to reduce the overall error in classification or recognition. On the other hand, run time environment of DNNs is largely feed forward and hence fast. Building processors with additional hardware to facilitate training will make the chip area-inefficient and hence expensive. Tuning the hardware to runtime will deteriorate already slow training time. Implementing a trained network has different set of constraints from those present during training. Therefore, the task would be to make the processor efficient during training as well as during runtime. If we consider training as a process that actually builds the "program", an opportunity exists in optimizing the built network to perform better during run time. This process will lead to compaction of the trained network. Further, compaction of trained DNNs can be easy if the contribution of individual nodes in functioning of the network can be easily identified. However, such identification is not possible in many DNNs, making the compaction difficult. Therefore, there is a need to explore explainable AI (XAI) and evolve methods to compact trained DNNs.

## 3.3 On-Chip Compute Core Count

Conventional processors have a single computing core per chip. More modern processor chips may have multiple cores, e.g., Quad-cores, Octa-cores etc. These architectures are excellent for running multi-threaded applications. But, the number of parallel threads possible in DNNs can run into millions. Some of the contemporary special purpose processors and GPUs have thousands of compute cores. For example, NVIDIA Tesla V100 GPU [2017 NVIDIA] contains 84 streaming multiprocessors, each with 64 32-bit floating point units, 64 integer units, 8 tensor units, etc. bringing the throughput to 125 TFLOPS (Tera Floating Point Operations per Second). Videantis MP6000UDX can perform 64 MAC operations per core per clock [2018 Videantis]. A chip module can contain up to 256 cores, giving a total of 16K MAC operations per clock. This is only a beginning of the ramp expected ahead. With such a large number of compute cores on chip, many design challenges can be anticipated, e.g., power and clock distribution, global memory, on-chip networking, etc. Some of these issues are discussed in literature architectures [2014 Szegedy], [2014 Lin], [2015 He]. But many situations will arise over period of time as the field usage improves.

## 3.4 Efficiency in Massively Parallel Operations

Operations like multiplication can be performed serially or in parallel. Parallel implementations will run several times faster than serial implementations, but at the cost of silicon area. Given the area of a parallel unit, it is possible to implement several serial units in the same area. As the number of parallel units increases, the number of serial units increases faster. Therefore, a threshold/cross-over of *area vs. speed* will exist. When the number of parallel units is small as in case of conventional processors; the advantage will be in favour of parallel units. However, as the number of parallel units exceeds a threshold; throughput of combined serial units will exceed that of parallel units. Such situations will be present in implementation of processors/accelerators for DNNs.

Multiplication and addition are the most frequently used operations in neural computations. Among these, multiplication is considered as the most complicated operation as it takes several clock cycles to complete. Considering Multi Layer Perceptron, there could be 10-20M multiplications per learning step. Thousands of multipliers would be required to implement such large number of operations. Many researchers are working on the design of multipliers for neural network hardware [2012 Lotric], [2016 Mrazek]. Lotric et al. have designed approximate multiplier for use in feed forward networks. The exact or high precision multipliers require large resources and consume more power and time. Use of approximate multipliers will increase the energy efficiency, often without any compromise in performance of DNNs.

The performance of both serial and parallel multipliers in massively parallel environment has been studied during this research (see section 5.1). It has been observed that serial multipliers will perform better in terms of speed and silicon area when the area ratio is larger than execution time ratio.

## 3.5 Re-configurability

As noted earlier, computational demands during training and runtime are different. Does re-configurability of hardware holds the key to training-runtime dilemma? The type of neural network to be used depends very strongly on the type of input, e.g., CNN for image classification and LSTM for audio [2014 Sak]. If the processor is designed with a specific DNN in mind, it may become inefficient for other DNNs. Sometimes, it may be acceptable to allow some inefficiency in favour of flexibility provided by re-configurability. So there is a need to estimate common needs of DNNs and define a maximal but functionally complete implementation that allows easy re-configurability.

## 3.6 Memory Bottleneck and Computation in Memory

This issue has been discussed in section 2.6, where how existing chips have overcome this problem is presented. Some additional views are presented here. As the number of compute cores in DNN hardware tends to be large, the need to transfer bulk data between heap memory, local memory and registers get complex. The number of busses that can be realized on a chip will become a limiting factor. Small number of busses means larger memory latency. Large number of busses will not only occupy space on chip but partition the chip area into isolated sections. We have to make two observations here:

a) Linear Memory organization is inefficient for ANN implementations and
b) Use of traditional bus oriented architecture becomes a bottleneck.

Currently, NVIDIA Tesla V100 with a matrix type of organization achieves a transfer rate of 900GB/s to 82 streaming multiprocessor units using 4096-bit HBM2 memory interface. There is a need to rethink of memory design as well as transfer mode. A good hierarchical memory organization may use high speed serial transfer between heap and cores but use short buses within complex compute cores: Much like the hierarchical memory, the data transfer also needs to be hierarchical.

## 3.7 Numerical Precision and Accuracy of Computation

Most of the ANNs will use saturating activation functions. That means the output of a neuron is always bound to an asymptotic limit. Derivate of the activation functions approaches zero when the output is approaching asymptotic limits. It is significant only in a limited range of the number scale. Multiplication or addition has very little effect on the output of a node once it reaches the asymptotic limit. Therefore, the range of input values where the output will significantly change is limited to a small range of values. The output itself is bound by the activation function.

Additionally, neural networks should produce useful output even in presence of noisy environments or input. In fact noise tolerance is a hallmark of neural networks. That is one of the reasons ANNs produce correct results even when the applied input is partial or has no precedence. These conditions imply that precision of input or of computation should have a small bearing on the decision taken by a neural network. The accuracy of digital systems comes at high cost of precision but is not necessary to achieve correctness of output using neural computations. Many of the modern DNN processors use a low precision arithmetic to achieve speed, e.g., Intel uses a format called flexpoint [2018 Rodriguez], eyeriss uses 8-bit and 16-bit fixed point [2017 Chen].

During implementation of ARN, we used piece-wise linear and second order approximations to compute the resonance curve. It was clear that the resulting reduction in numerical precision did not deteriorate the recognition performance. Therefore we have been using a low precision number format for use in ARN [2019 P1] [2019 P5]. Use of low precision arithmetic reduces the run time in simulations. It also improves area-speed performance in hardware implementations.

## 3.8 Multi-Operand Instructions

Most of the DNNs use sum of product computation to generate neuronal outputs. The number of inputs (synapses) to these nodes can be of the order of hundreds. For example, first convolution layer in Alexnet [2012 Krizhevsky] uses nodes with 363 synapses. TrueNorth analog AI chip from IBM [2015 Akopyan] supports 256 synapses per neuron. Each such computational node has to perform as many multiplications per neuron followed by an addition. Therefore, performing two operand operations on such computation would be very inefficient in terms of throughput; most of the time would be spent moving the data between partial sum storage elements. Therefore, use of multi-operand operations would be preferable. Systolic arrays could be used to implement successive multiplication and addition. Modern instruction sets with operations like "Fused Multiply and Accumulate Single Precision" were designed to address this problem. Streamed SIMD architecture may also be used to compute the sum of multiple products operation.

## 3.9 Explainable Neural Networks

One of the bottlenecks to improving the performance of deep neural networks is the ambiguity in explaining the decision the network took. Largely DNNs are seen as a black box because of the large number of inputs and synapses connecting to individual nodes in the network. This problem is generally called the *Binding Problem* and was highlighted by Malsburg [1999 Malsburg] as the inability to assign a state or node of a network to how and why a network took a specific decision. More modern networks like CNN work on parts of the input and merge the components in successive layers of neurons. But the data association is not recorded or remembered explicitly. Therefore, it becomes difficult to express a causal relation between the input and output [2017 Gunning] [2018 Vaughan]. Newer DNNs will store such information about relations and causes as an associated part of the network, similar to the semantic net of the classical AI networks. Some DNNs like ARN have better opportunities over other networks.

### Summary

I have discussed some of the architectural issues of processors for AI but not considered physical aspects of chip like power and clock distribution, as they are completely at a different level of technology. It is interesting to note that the number of synapses of a neuron in human brain is of the order of 100,000 and that in mice is about 45,000. While the current range of processors for DNNs has an amazing complexity, we are at least two to three orders behind the natural intelligence. The contemporary issues discussed in this chapter and the design of accelerators to address some of these issues are published in SCOPUS indexed International Journal [2019 P1].

# Chapter 4

# Image Recognition using Auto Resonance Network

This chapter presents a new application of a multi-layer neural model called *Auto Resonance Network* (ARN). These networks appear to be capable of addressing problems in varied fields but have not been sufficiently explored. In this chapter, we explore the capabilities of ARN for image recognition.

Auto Resonance Networks (ARN) proposed by Aparanji et al. [2016 Aparanji] are different that other DNNs. Therefore, an overview of the model is made here.

ARN is an approximation network. Input and output values are real numbers but limited to a small range, typically zero to one. It does not perform explicit stochastic or algorithmic gradient descent but it performs a likelihood search. It performs recognition optimization solely based on local information. At higher layers local information represents higher level of abstraction, which in turn leads to global optimization. ARN was originally used for graph based neural networks and therefore neurons are sometimes called *Nodes*. Every neuron has a specific *coverage* of input values within which it resonates. Resonance allows the neuron to be noise tolerant and tunable. Resonating neuron can fire (connect to next neuron) for a short period of time depending on the degree of resonance. There are two distinct learning mechanisms in an ARN. Primary learning occurs when a neuron is assigned to the network with a specific locus of resonance. This mode of learning can correspond to registering a set of input values or an associative or a causal relation between two or more neurons. Second mode of learning is caused by morphing the node coverage or panning of locus in response to variations in input. This second mode of learning has been referred to as Tuning of a neuron. Tuning can be explicit in case of supervised learning or implied by the input data as in case of reinforcement learning. In addition to this, nodes may be added to the network layer by perturbation of existing nodes, which reduces the number of samples required to train the network.

Resonance in an ARN neuron is somewhat different than traditional view of resonance, in the way output is generated. In a classical resonator like RLC tank, the peak can reach high values depending on the value of R. Resonance in ARN neuron has a finite and fixed upper bound. Higher selectivity (or quality) of resonance simply means lower spread of the node coverage. The peak value is always the same. Spread of resonance is the coverage of the neuron. It is interesting to note that the activation value of a biological neuron is always fixed. It is only how many times or how long that value is retained that decides the quality of neuronal activation (or firing). This has two fold advantages. Firstly a finite upper bound avoids stability problems in Hebbian-like systems that learn by association. Secondly, several layers of neurons may be connected in series allowing long strands of patterns to be identified.

The locus of resonance is unique to every neuron. Therefore, activation of a neuron is associated with a specific input pattern or causal relation. Therefore, an activation path(s) corresponding to any specific input - output relation can be traced, interpreted and explained. Therefore, it is easy to teach ARN. In fact, teaching can be systematically planned to achieve a goal. It is also easy to prioritize the nodes and paths, which may be used to improve network density by eliminating outlier nodes. This property will make ARN a suitable candidate for *AI at edge*.

Implementation of an image classification and identification system using multi-layer ARN is discussed in the following section. Recognition accuracy of 94% has been achieved for MNIST data set with only two layers of neurons and using as few as 50 to 500 samples per numeral. ARN based solution presented here can achieve performance comparable to CNN but using very shallow networks, hinting at their usefulness in computing at the edge.

## 4.1 Biological basis for ARN

Initial development of ARN by Aparanji et al. [2016 Aparanji] was based on a biological pull-relax mechanisms used to control muscle movement. Current models of ARN are significantly evolved versions of the old analogy. A typical biological neuron has several dendrites that receive inputs from sensor neurons or intra neurons in its vicinity (see section 2.1). The strength of the dendritic connection presents a scaled version of the input to the neuron. Such inputs accumulate in the soma of the neuron and shift the polarization (internal state of excitation). When the polarization at the base of the axon exceeds a threshold, the neuron fires output (reaches activation potential) along the axon to the other neurons. Membrane potential of a firing neuron is constant irrespective of the strength of input but the frequency of firing can vary with strength of neuronal activity. Further, physical transport mechanisms put additional limits on conduction of activation. Therefore the output may be represented as multiple copies of activation rather than an increased activation potential. The soma of the neuron performs a saturating non-linear integration of scaled inputs. As variations in inputs can still result in the same level of excitation, there is a certain noise tolerance built into this neuronal excitation. Further, connection strength of individual dendrites varies with repeated use, providing more gateways to transfer the polarization. Therefore the neuronal dynamics controls learning by the neural network which is the basis of Hebb's description of neuronal learning.

ARN nodes have many similarities to biological neurons: ARN nodes are approximation devices. The incoming data is scaled and summed to produce a resonance, similar to activation of a biological neuron. There is always a known locus of input that pushes the node to resonance. They tune to the incoming data, similar to the description of Hebb's law. They have saturating outputs similar to activation of a biological neuron. Typical threshold of an ARN is above 90%, which is similar to the constant activation potential of neurons. Multi-layer ARN is a feed forward network, though a feedback loop can always be added. ARN as in pathnet [2017 Aparanji] has nodes with delay characteristics that can establish temporal and causal relations among input. It is also to be noted that addition of new nodes to any layer does not affect the coverage or locus of resonance of existing nodes. This feature allows ARN to build an explainable network.

## 4.2 Constrained Hebbian Learning

Hebb's law states that the strength of a neural connection improves as a particular input is applied more frequently or reduces if not used. The rule is somewhat partial as it does not clearly state how the output of a neuron is limited to a saturation level. Most of the current interpretations of Hebbian rule use it without any constraint on the strength of the connection [1999 Yegna]. This leads to the classical weakness of Hebbian learning, i.e., continued use of a connection will scale the input so much that the node becomes unstable. In a biological system, this never happens as the physical limits on transport will limit the strength of connection and hence the Hebbian rule must be subjected to an upper limit. ARN implements a constrained Hebbian learning mechanism and remains stable for reasons explained in next sections.

## 4.3 Resonator

Consider a simple resonator like

$$y = x * (k - x) \tag{4.1}$$

with a resonance (peak) value of $y_m = k^2/4$, occurring at $x_m = k/2$, where k is a gain control parameter. Both $y_m$ and $x_m$ are independent of input x. Therefore, this peak value remains unaffected by translation, scaling or other monotonic transformations of x like sigmoid or hyperbolic tangent. This is an important feature as the output is bound irrespective of how x is scaled by the strength of connection. It is possible to shift the point of resonance by translating the input to required value by translation or scaling.

For translation, $X \Leftarrow x - t$, peak occurs at

$$x_m = (k/2) + t \tag{4.2}$$

For scaled input, $X \Leftarrow xt$, peak occurs at

$$x_m = k/(2\,t) \tag{4.3}$$

In either case, peak value is bound by

$$y_m = k^2/4 \tag{4.4}$$

Though k may be used for gain control, using fixed value per layer or network yields stable systems. We have used k = 1 giving a peak value of ¼. Thus output may be multiplied by 4 to yield a peak value of 1. It is also possible to use k = 2 for peak value of 1.

Another interesting input transformation is the sigmoid:

$$X = \frac{1}{1+e^{-\rho(x-x_m)}} \tag{4.5}$$

where ρ, called the resonance control parameter, is a positive real number and $x_m$ is the resonant input. For now, we will assume k = 1. Note that $(1 - X)$ has form similar to eqn. (4.5) and is given by

$$(1 - X) = \frac{1}{1+e^{\rho(x-x_m)}} \tag{4.6}$$

Substituting (4.5) and (4.6) in (4.1), we get

$$y = X\,(1 - X) = \frac{1}{(1+e^{-\rho(x-x_m)})}\,\frac{1}{(1+e^{\rho(x-x_m)})} \tag{4.7}$$

Note that the response y is symmetric about $X_m$. The output of a resonator expressed by eqn. (4.7) attains a peak value of $y_m = 1/4$ at $X_m = 1/2$ as expected. Incidentally, eqn. (4.7) is also the derivative of eqn. (4.5) which has a well known bell shaped curve.

Figure 4.1 shows some node outputs for $k = 1$. Figure 4.1(a) and Figure 4.1(b) show shifting resonance by input translation and input scaling given by eqn. (4.2) and (4.3) respectively. Figure 4.1(c) shows the output curves for sigmoid transform for different values of $x_m$. Resonance of each shifted sigmoid in Figure 4.1(c) can be altered by using resonance control parameter ρ, as shown in Figure 4.1(d). These equations allow stable Hebbian like learning to be incorporated into ANNs without the associated instability problems. It must be emphasized that our usage of the term Hebbian learning is only a holistic description of the ARN's learning algorithm. Actual learning algorithm may be a variation of Hebbian learning, suitable for specific use case.

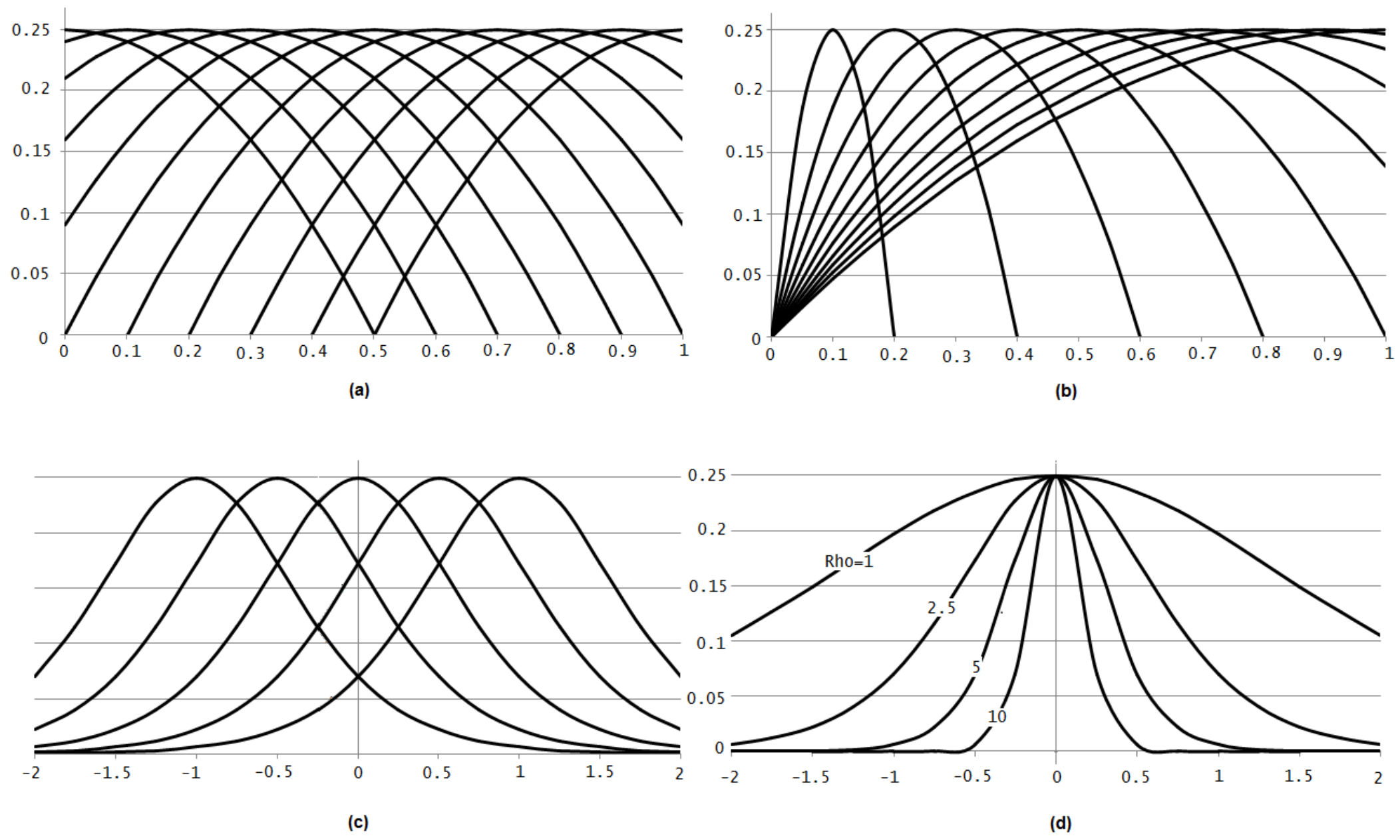

**Figure 4.1 Resonating Curves of ARN nodes (k = 1): a) Shifted Resonators, b) Scaled Resonators, c) Shifted, scaled Sigmoid Resonator (ρ=2.5) d) Tuning of Sigmoid Resonator**

## 4.4 Selecting the Resonance Parameter

From eqn. (4.2) and eqn. (4.3), the location and peak value of resonance are controlled by k while the location is also controlled by translation parameter t. Generally the input translation is more complex than those implied in these equations. In such cases, the effect of k on the resonance should be studied explicitly. For example, eqn. (4.5) transforms real input x to X. Corresponding ARN in eqn. (4.7) has a peak of ¼ occurring at $X_m = k/2 = 1/2$ or equivalently $x = x_m$. In order to scale the peak to 1, we may use $k = 2$. However, $X_m$ shifts to 1 and X does not cross 1 for any x. Therefore, the resonator can only implement left side of the resonance function. This suggests that choice of k must ensure that the reverse transformation $x = f^{-1}(X)$ will yield x on either side of $X_m = k/2$, unless it is desirable otherwise. It is not always necessary that the transformation is symmetric about $X_m$ but must have values spread on either side. For example, Figure 4.2 shows the effect of transformation given in eqn. (4.5). As k shifts, $X_m$ and $x_m$ also shift. There may be other parameters like ρ in eqn. (4.7) that are specific to the chosen transformation which also need to be checked for validity of the input range.

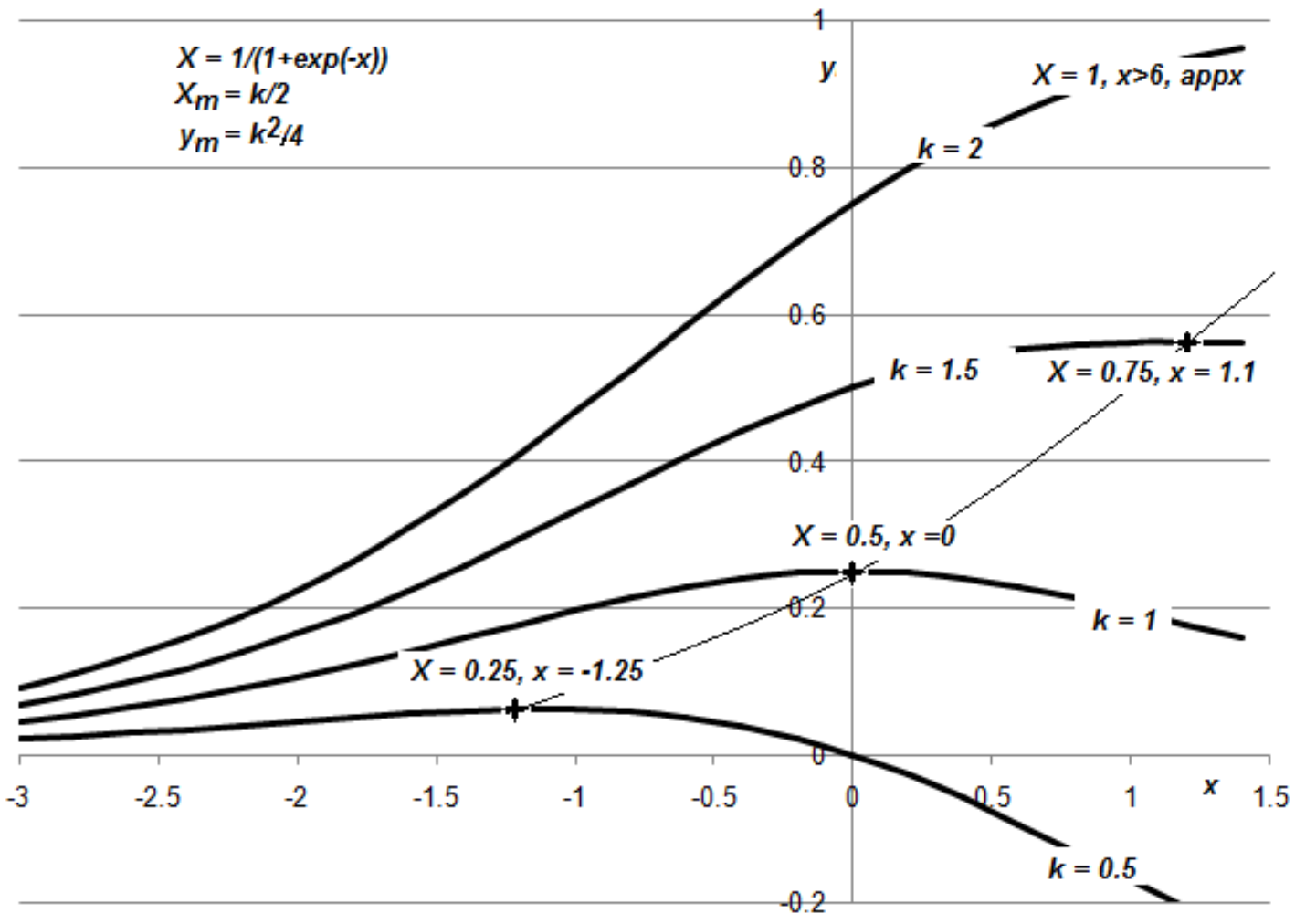

**Figure 4.2 Dependence of resonance on k for eqn. (4.5)**

## 4.5 Threshold, Coverage and Label of a Node

Some basic operations of the nodes are described here. Their actual implementation will become clear after further discussions.

A node is said to be triggered if the computed output is above a pre-selected threshold T. Input points in the neighbourhood of the peak will cause the node to be triggered. The set of all input values to which the node is triggered is called the 'coverage' of the node. As mentioned earlier, node learns by adjusting the coverage in response to the input. No two nodes will have same coverage. However, coverage of two or more nodes may overlap partially, which sometimes can cause several nodes to be triggered for some specific inputs. If only one node is triggered, it means that the network has clearly identified the input and hence can classify it. If the input is covered by more than one node, one of the nodes may adjust its coverage to produce stronger output. If no node produces an output above threshold T, then a new node is created.

Clearly, the number of nodes in ARN varies depending on the type of inputs, training sequence, resonance control parameter (ρ) and threshold (T).

The output of a node is a constrained real value, which may be input to the nodes in higher layers. Nodes may also be labelled with a class which reflects the type of primary input. Many nodes may be tagged with the same label. Therefore, a label may indicate an arbitrary collection of nodes. This allows input space corresponding to a label to be arbitrarily set: continuous or disconnected, convex or concave, linear or non-linear, etc. This feature of ARN layers gives them an ability to learn from any input set.

## 4.6 The Aggregator

A node in ARN can have several inputs, each with its own resonator and a distinct coverage. Assuming that suitable transformation has been applied, output of the node with N inputs is given by, say

$$y = \frac{4}{Nk^2} \sum_{i=1}^{N} X_i \ (k - X_i) \qquad (4.8)$$

with a peak value of 1. If this normalized output is above a threshold T, the node is said to be triggered and it identifies the input as belonging to a specific label (or class).

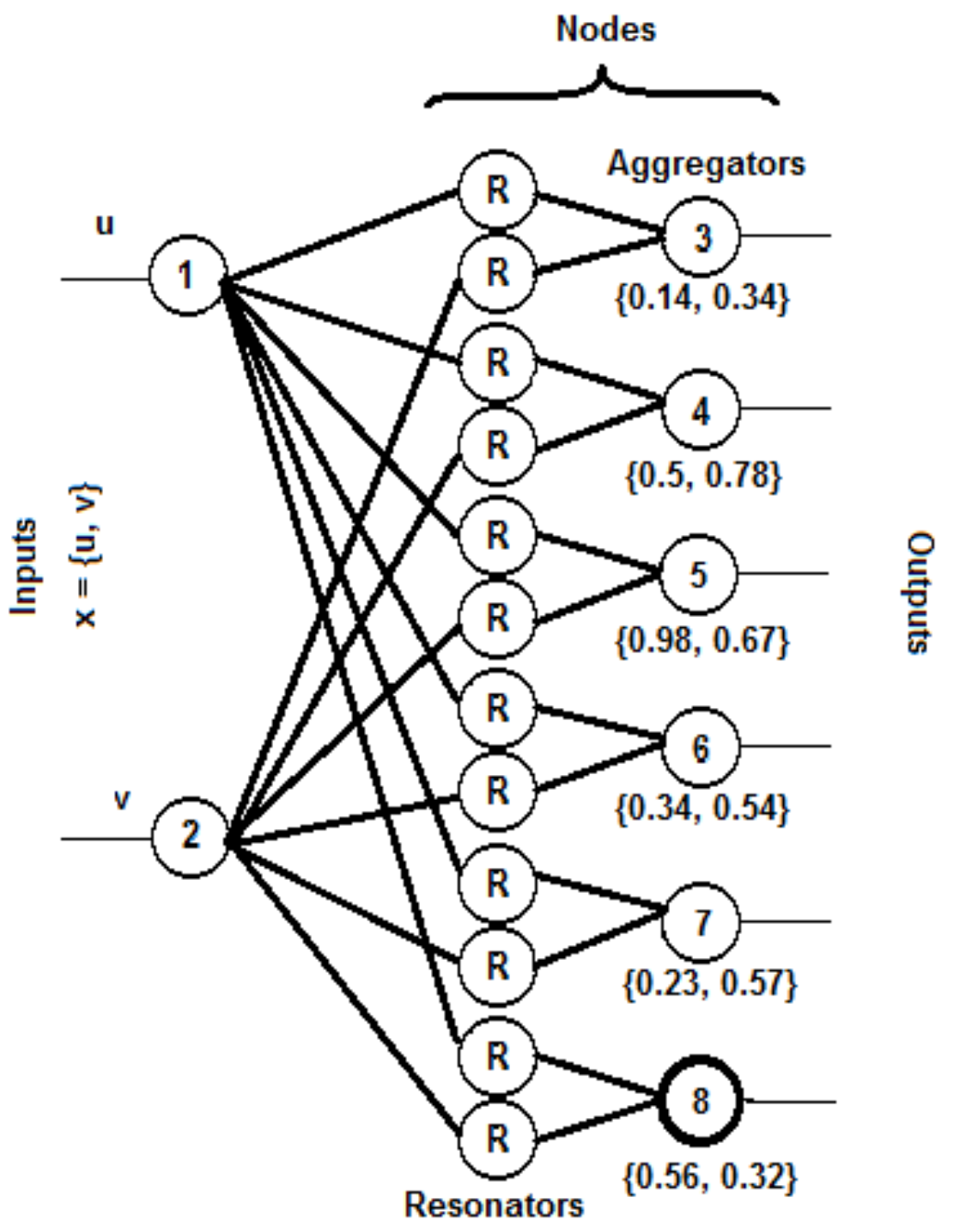

**Figure 4.3 A layer of ARN structure**

**Table 4.1 Resonant values of nodes created for network in Figure 4.3**

| Input values: x= {u, v} | Output Node label |
|---|---|
| {0.14, 0.34} | 3 |
| {0.5, 0.78} | 4 |
| {0.98, 0.67} | 5 |
| {0.34, 0.54} | 6 |
| {0.23, 0.57} | 7 |
| {0.56, 0.32} | 8 |

As an illustration, consider Figure 4.3. On initialization, only inputs 1 and 2, representing the open synaptic connections from the primary sensor or outputs of other layers are present. The ARN layer is empty on initialization. Nodes get appended to the layer as the inputs are applied. The resonant inputs to which the nodes are tuned are shown in Table 4.1. To understand how the layer grows, consider the first input {0.14, 0.34} applied to ARN shown in Figure 4.3. There will be no output from the network as there are no output nodes. This causes node 3 to be created as first node in the layer. The second input {0.5, 0.78} is outside the coverage of node 3 and hence the network does not produce any output again. Therefore, another node, node-4, is created to resonate at this input and appended to the layer. The process continues on arrival of unmatched input. An input of {0.45, 0.29} is within the coverage of node 8 and hence, it is triggered. Figure 4.4 shows the scenario graphically. The winner node is highlighted with green colour. The two and three dimensional views of coverage of each output node are shown in Figure 4.4 (a) and Figure 4.4 (b) respectively.

The rectangles in Figure 4.4(a) represent the coverage of each output node. There are six rectangles corresponding to six output nodes of ARN shown in Figure 4.3. Each node is tuned to one unique input and has its own coverage. For example, the rectangle in green colour representing the node 8 has the coverage area bound by the range $0.42 < y < 0.68$ and $0.25 < x < 0.39$. The test point {0.45, 0.29} is within the coverage area of node 8 and therefore it is triggered. The location of test input shown as red dot in Figure 4.4(a) and in Figure 4.4(b).

Tuning of a node changes coverage of the node. Increasing tuning reduces the range, while decreasing tuning will result in wider coverage (see Figure 4.4(c)). It may also be noted that increasing the coverage lifts the surface up and hence has a higher output value.

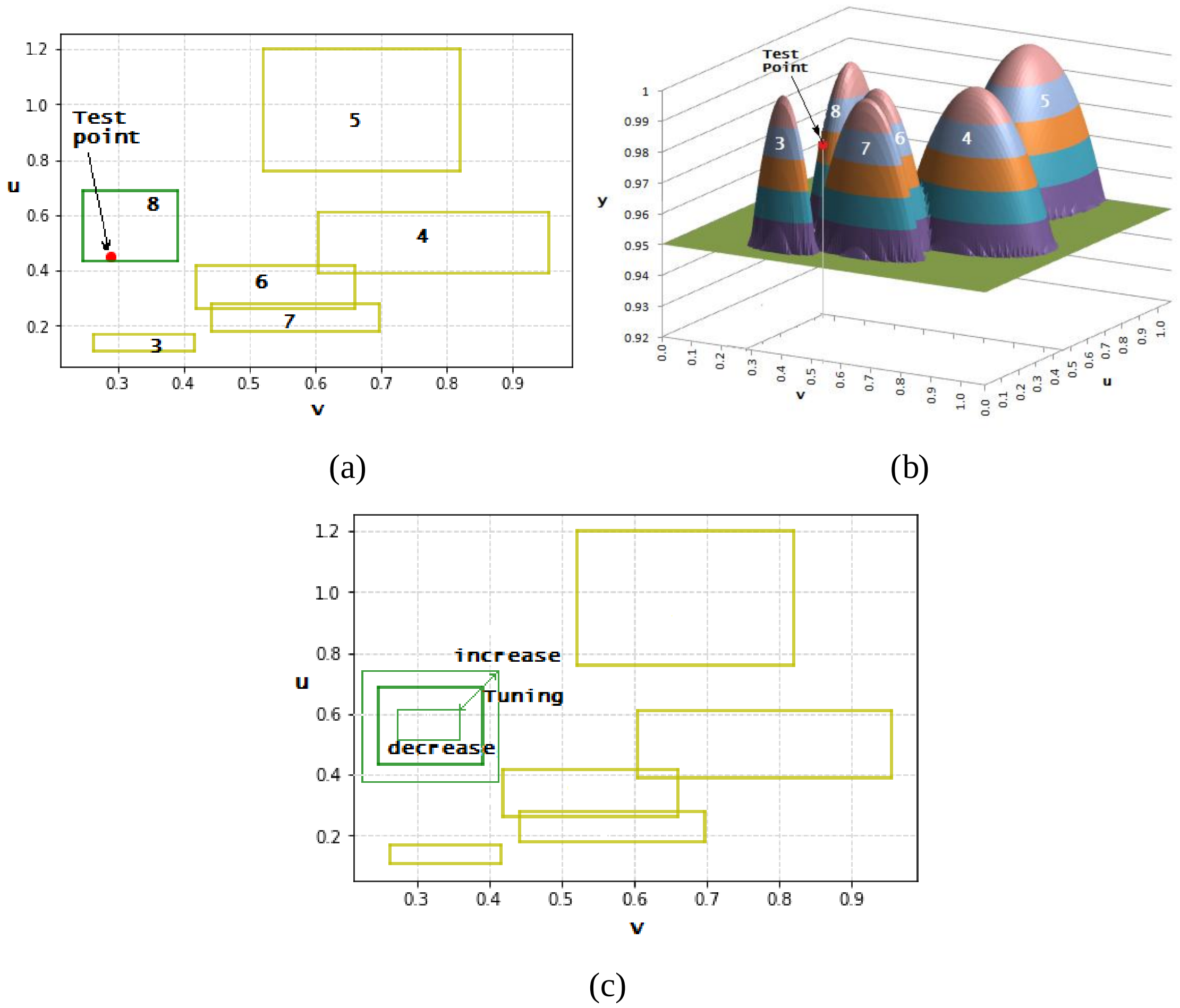

**Figure 4.4 (a) 2-D view of coverage and tuning, (b) 3-D view, (c) Effect of tuning on coverage**

## 4.7 Input Range, Coverage and Tuning

When a new node is created, default coverage is assigned to the node. Under certain transforms, coverage depends on point of resonance, as in Figure 4.4. In this figure, nodes closer to origin have a smaller coverage compared to the points that are near maximum value. Statistical parameters of the input data may be used to set the resonance control parameter and selection threshold of ARN nodes. Typical procedure to calculate initial coverage for a node is given below, for a peak value of ¼ (or k = 1). The method may be suitably modified for other transformations.

One reasonably good threshold point is half power point, as shown here. Assuming k = 1 in eqn. (4.1) yields a peak value of 1/4 for eqn. (4.4), i.e.,

$$T = \text{half power point} = \sqrt{\frac{0.25^2}{2}} = 0.176 \qquad (4.9)$$

Assuming the output of a node to be equal to threshold (T), we can equate the eqn. (4.1) to T and write

$$X(1 - X) = T = 0.176 \tag{4.10}$$

We will use scaled input model of Figure 4.1(b), i.e., X = xt as an example. By solving for x we get two points $x_c$ indicating bounds of coverage points:

$$x_c = \frac{1 \pm \sqrt{(1-4T)}}{2t} \tag{4.11}$$

Both $x_c$ points are symmetrically placed around the resonant point, which is given by eqn. (4.3) to be $x_m = 1/(2t)$. This is also evident from Figure 4.1(b). Note that as t shifts the locus of resonance, the coverage changes (e.g., see Figure 4.4(c)). For sigmoid transform the coverage is independent of locus of resonance (given by t) but controlled only by the resonance control parameter ρ. From eqn. (4.7), assuming $x_m = 0$, we can write

$$0.176 = \frac{1}{(1+e^{-\rho x})} \frac{1}{(1+e^{\rho x})} \tag{4.12}$$

Simplifying, and substituting $\frac{(e^{\rho x} + e^{-\rho x})}{2} = \cosh(\rho x)$, we get

$$x_c = \frac{\pm \cosh^{-1}(1.8409)}{\rho} = \frac{\pm 1.2198}{\rho} \tag{4.13}$$

A graph showing this relation between resonance control parameter and coverage, i.e., ρ vs. $x_c$, is shown in Figure 4.5. Range between the two $x_c$ is the coverage.

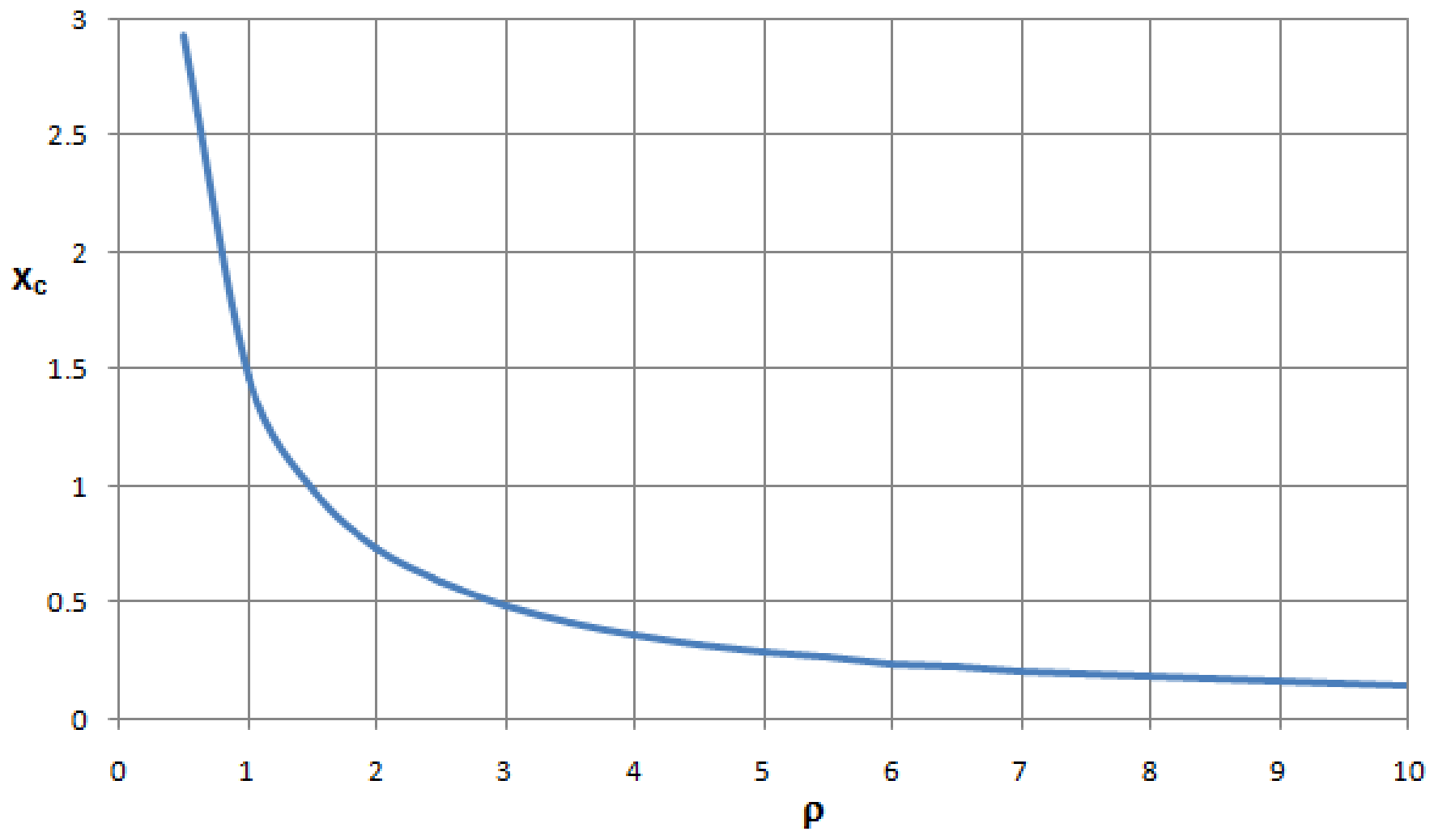

**Figure 4.5 $x_c$ as a function of resonance control parameter, ρ from eqn. (4.13)**

The resonance control parameters depend on the transform used. Reduced coverage indicates focused learning. It is also helpful in resolving ambiguities in learning: In case of multiple winners, one of the nodes can relax control parameters to increase the coverage and enhance the output.

## 4.8 Data Variance and Node Coverage (ρ)

As coverage is a data dependant parameter, we may try to relate it to the distribution of values arriving at a node. Assuming that the incoming data has a Gaussian distribution, we can derive a relation between the statistical variance of the data and coverage of ARN node. The Gaussian distribution is given by eqn. (4.14).

$$y = \frac{1}{\sqrt{2\pi\sigma^2}} e^{\frac{-(x-x_m)^2}{2\sigma^2}} \tag{4.14}$$

where, $x_m$ indicates resonant input and $\sigma^2$ is variance. At $x = x_m$, we get the maximum (peak) value of y as

$$y_{peak} = \frac{1}{\sqrt{2\pi\sigma^2}} \tag{4.15}$$

Value of Gaussian distribution function at half power point can be written as

$$y_c = \sqrt{\frac{(y_{peak})^2}{2}} = \frac{1}{2\sqrt{\pi\sigma^2}} \tag{4.16}$$

Equating equations (4.14) and (4.16) and substituting $x_m = 0$, which is similar to the condition for eqn. (4.13), we get,

$$x_c = \pm 0.8325\sigma \tag{4.17}$$

In eqn. (4.17), $x_c$ represents the range of inputs (coverage of a node) and σ represents standard deviation. From equations (4.13) and (4.17) we can get the relation between ρ and σ:

$$\rho = \frac{\pm 1.4652}{\sigma} \tag{4.18}$$

When the mean and standard deviation are known, we can compute the corresponding values of $x_c$ and ρ. This in turn yields the coverage of the node. Graph in Figure 4.6 describes the coverage of node in range 0: 1 at $x_m = 0.5$. For other values of $x_m$ the coverage in terms of standard deviation can be expressed as

$$x_c = x_m \pm \alpha\sigma \tag{4.19}$$

where α indicates a scaling factor, similar to the constant in eqn. (4.17).

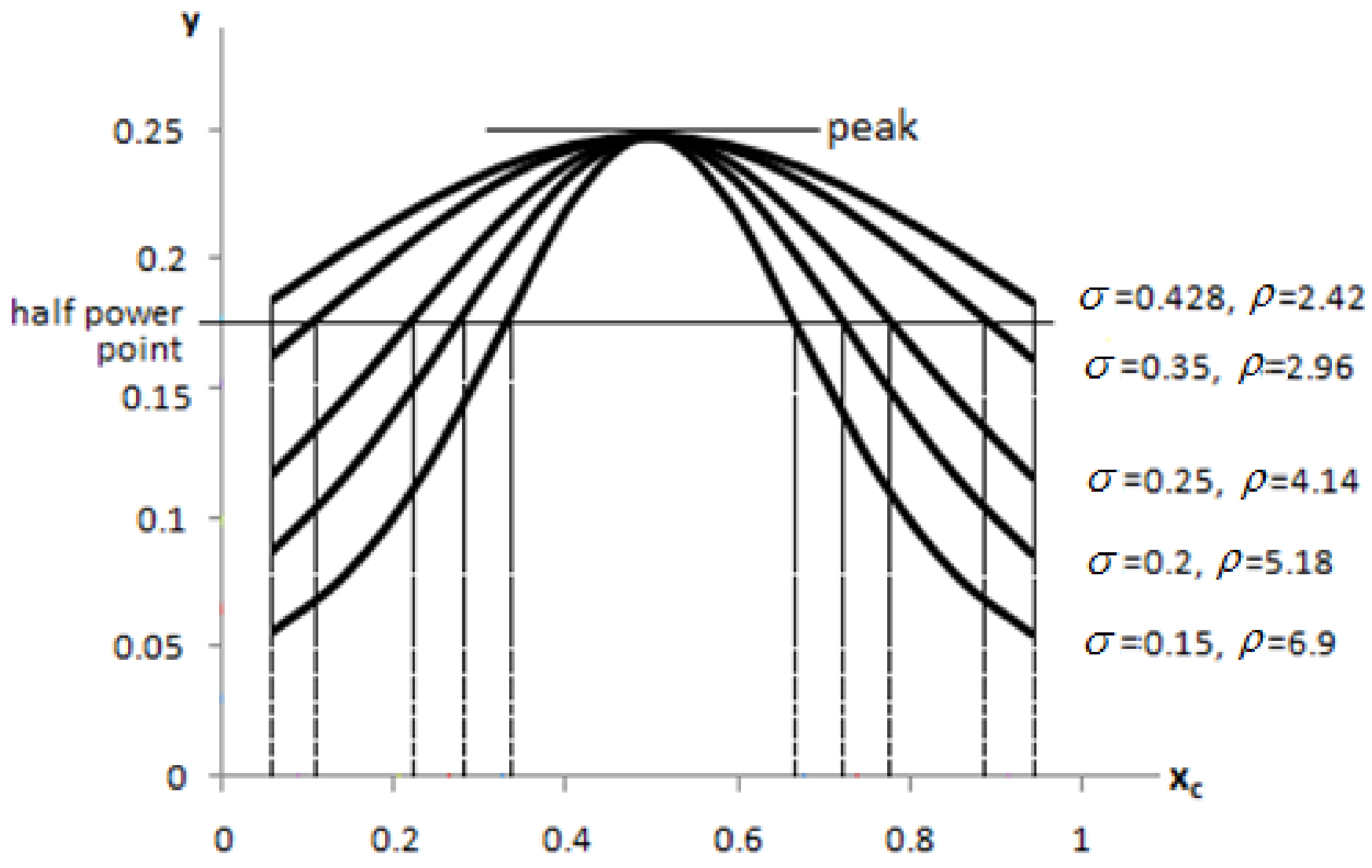

**Figure 4.6 Tuning Curves of ARN nodes**

## 4.9 Image Recognition

Image recognition and classification has become a routine task using CNNs. However, this should not stop us from exploring alternatives. For example, capsule networks suggested by Hinton et al. [2017 Sabour] incorporate a spatial relation among the features recognized by CNNs. Sensitivity of CNN to quantization has been explored in [2016 Wu]. Computational complexity of CNNs is very large, requiring support of large servers with special purpose accelerators. High performance implementations suitable for mobile devices are also being explored [2019 Niu]. Interestingly, Intel's new neuromorphic chip *Loihi* with on-chip learning capabilities does not implement CNN but a spiking neural net [2018 Davies]. These are few reasons why alternates need to be explored. In this section, we will demonstrate the capability of ARN for image recognition using MNIST data set. The work can be very easily extended to other image sets without much difficulty.

Some of parameters that need to be fixed while building ARN include the choice of resonator, spatial spread of neural input, neuronal density, selection of threshold and resonance control parameter ρ (rho), perturbation methods, number of layers, etc. To keep the network simple, I have assumed only two ARN layers, one type of perturbation and a symmetric 4x4-segmented image input with pixel values normalized to 0:1 range as these are found to be adequate for the data set used to test the learning ability of the ARN structure.

## 4.10 The Dataset

One of the first test benches to test basic capabilities of any image classifier is the Modified National Institute of Standards and Technology (MNIST) database of handwritten English numerals from 0 to 9, established by LeCun et al. [1998 LeCun]. It has 70,000 samples in total consisting of 60,000 training and 10,000 test samples. The database can also be downloaded from alternate sources like Google TensorFlow and Keras API web sites. The pixel intensities vary between 0 to 255, with 0 representing black and 255 as white. The size of each image in MNIST data set has a cleaned 28x28 (=784) pixels image and a label corresponding to the digit which image represents. The original data set is in Unicode, which may be converted to a suitable format.

In the following section, a simple 2-layer ARN has been demonstrated to be able to efficiently classify MNIST data set with a small number of training samples. For a quick proof of concept implementation, a set of 50 randomly chosen samples of each digit, making a total of 500 samples, were used for training. The performance of ARN was evaluated using 150 test samples, representing about 30% of training set size. An accuracy of about 94% was observed. Details of this implementation are presented below. Larger datasets were used after the algorithm was validated.

## 4.11 The Architecture of two-layer ARN

The algorithm and the architecture of two-layer ARN used for image recognition are shown in Figure 4.7 and Figure 4.8 respectively. Essentially there are two layers of ARN marked as L1 and L2. L1 receives parts of image as input and converts them to a feature index. These recognized indices are temporarily stored in a spatially ordered list. This list is applied to L2, which will recognize the digit. Nodes in L2 are labelled with the class or type of the image. For MNIST database, the labels are the values of digits, viz., zero to nine.

### 4.11.1 Tiling

Tiling the image allows the input image to be analyzed in parts. It is not necessary that the tiling be contiguous but is customary. The order in which the tiles are presented to L1 is not critical but certain spatial information is contained in that order. If the order of tiles is changed, the spatial relation between the features is altered and hence the output from the classifier may differ. A simple image tiling as shown in Figure 4.9 has been used. Tiles are numbered 1 to 16 from top left to bottom right. Presenting them to L2 in reverse order rotates the image by 180 degrees. Similarly, spatial reordered lists can be interpreted as mirror, shift, rotate etc. These operators are not significant for digit recognition but may improve recognition in several other use cases. Effectively, reordering of the feature list can serve as an internal synthesizer. Layer L2 can be trained with these altered lists to improve recognition accuracy.

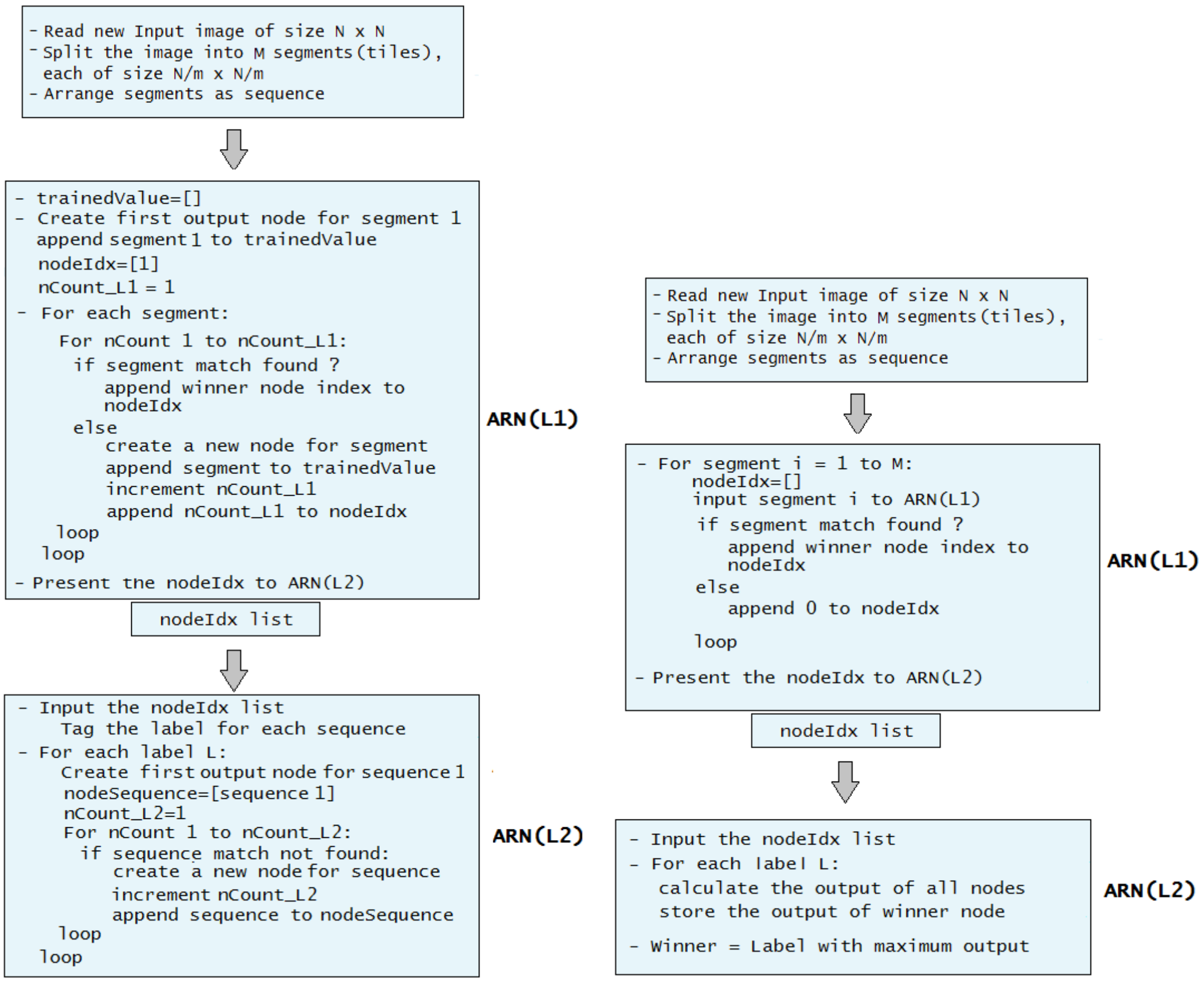

(a) (b)

**Figure 4.7 Algorithm for two-layer ARN for Image Recognition (a) training (b) testing**

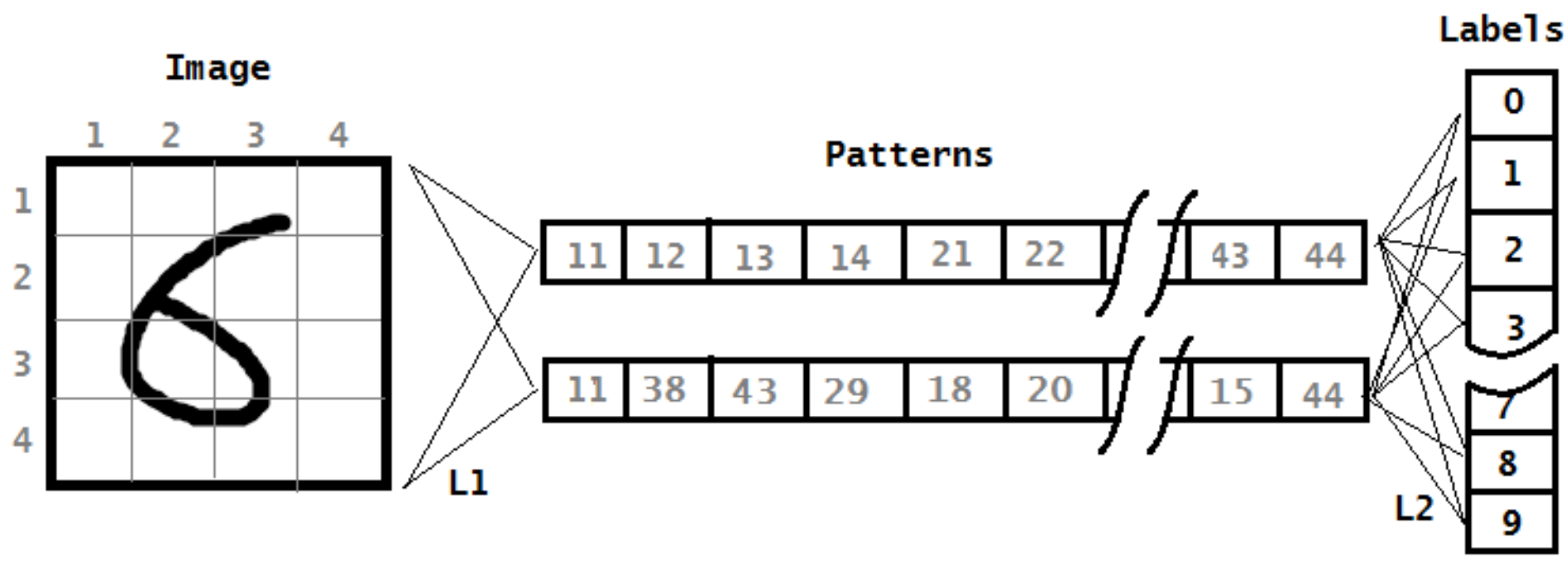

**Figure 4.8 An overview of Two-layer ARN for Image Recognition**

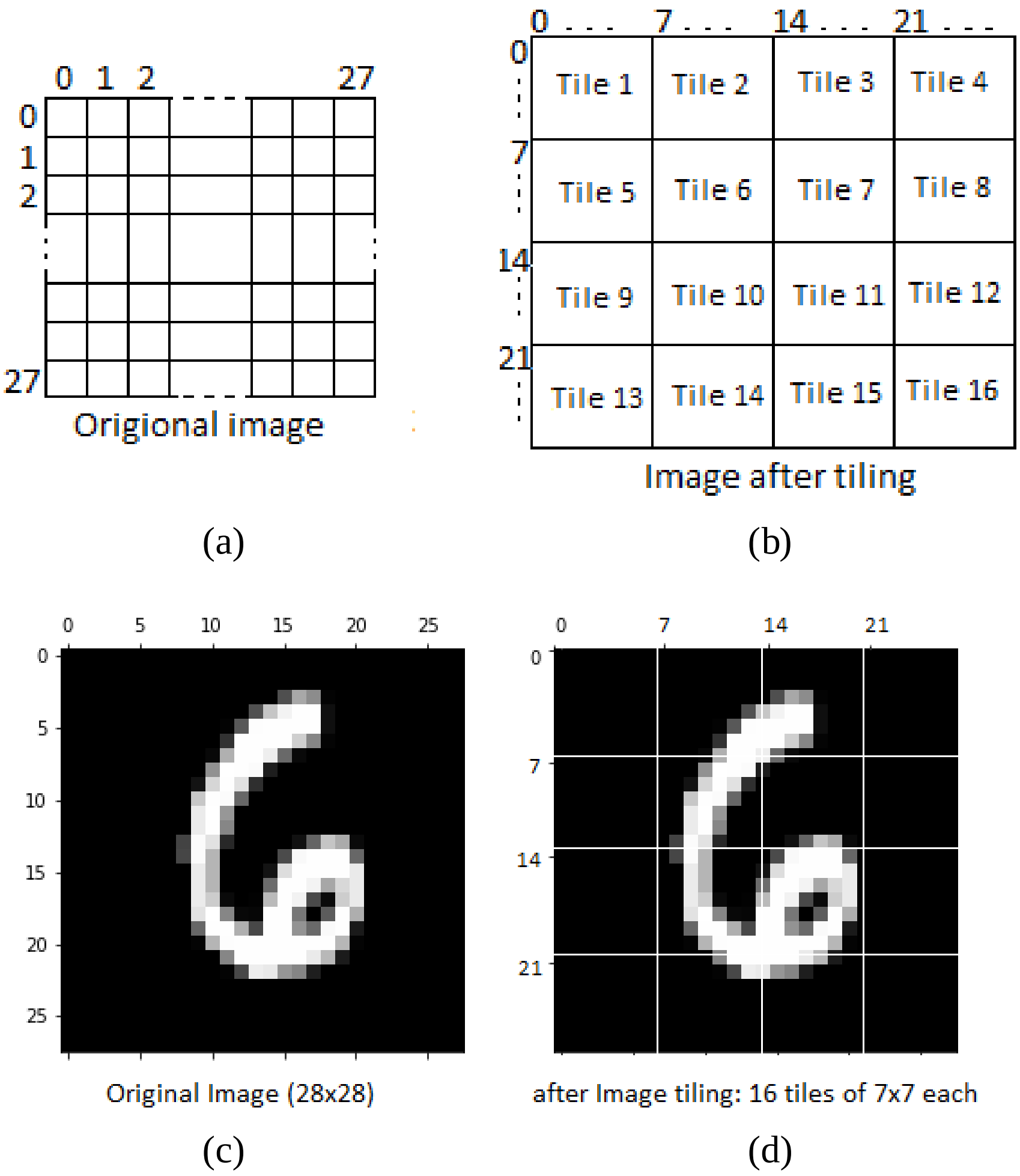

**Figure 4.9. Image tiling for feature extraction**
**(a) Original (b) Tiled Image (c) Sample and (d) Tiled Sample**

### 4.11.2 Training

The network is trained for different training sample size viz., 50, 100, 200, 300, 500. As the number of training samples increases, the accuracy of recognition also increases. Tunability of ARN nodes makes it possible to achieve accuracy up to 94% with very few training samples as 50x10. Both $\rho$ and T have effect on learning. For the given data set, $\rho = 2.42$ and $T = 0.9$ gave better overall results. These observations are discussed later in section 4.13.

There are three possible cases of recognition viz., Correct recognition, Wrong recognition and Multiple recognitions. When two or more number of digits looks similar, there is a possibility of multiple recognitions. One or more nodes will have the same output which is maximum and above threshold. Some of the examples of digits with similar visibility are shown in Figure 4.10

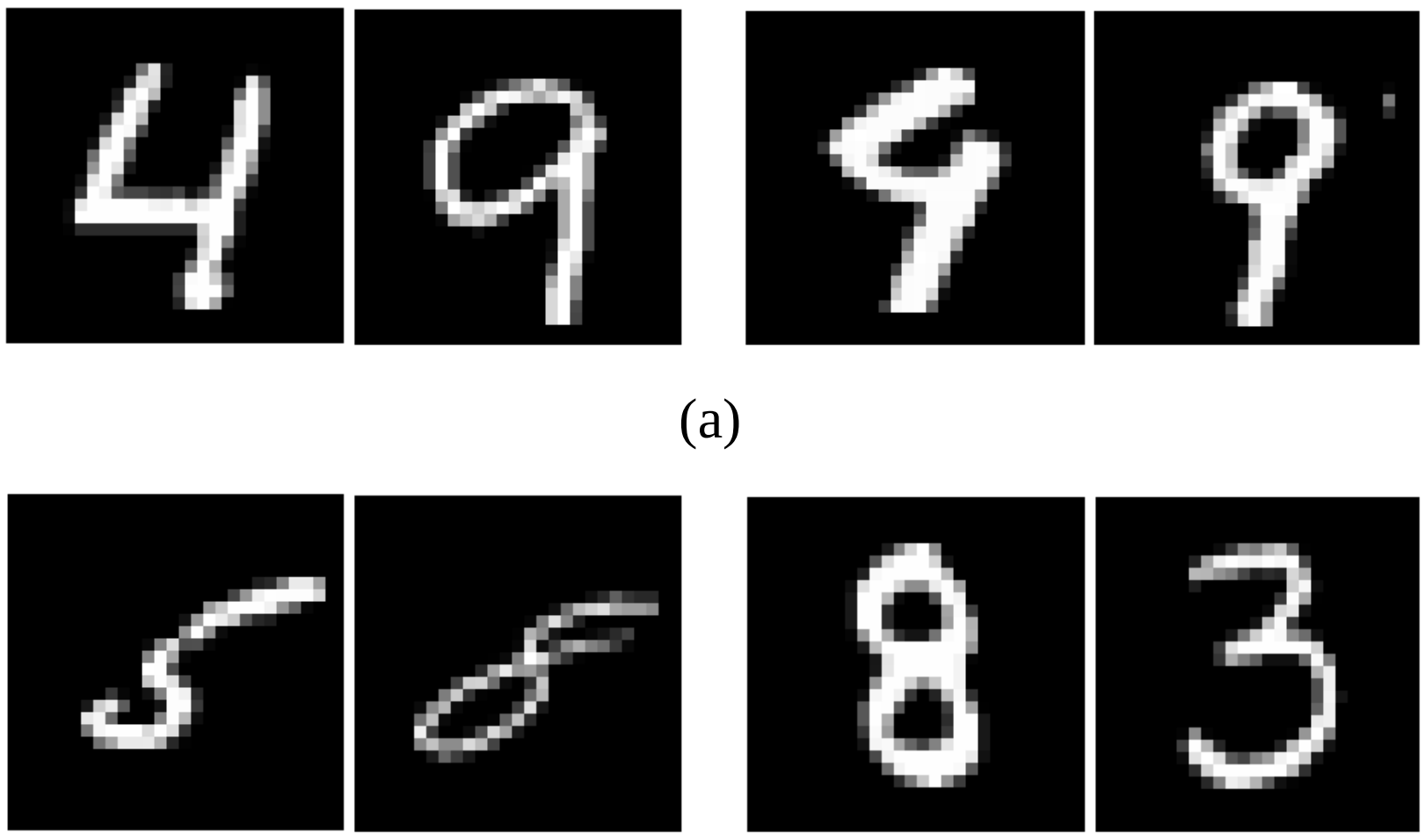

(a)

(b) (c)

**Figure 4.10 Visibly similar image samples from MNIST database**
**a) Digits 4 and 9, b) Digits 5 and 8 c) Digits 8 and 3**

### 4.11.3 Coverage of nodes

Coverage of ARN node provides an approximation of the features detected at L1. Therefore, the number of nodes is far less than the maximum possible number. For example, if a network is trained with 500 images and each image is broken into 16 tiles, there would be 8000 tiles. The number of nodes created depends on the coverage and the sequence of images input to the layer. For example, as ρ increases, or as the threshold (T) increases, nodes becomes more selective and hence the number of nodes increases. The number of nodes created in L1 is summarized in Table 4.2.

**Table 4.2 Number of nodes in 1st layer with respect to ρ, T and training sample size**

| Training sample size | T=0.9 | | | ρ=2.42 | | |
|---|---|---|---|---|---|---|
| | ρ=2.42 | ρ=2.7 | ρ=3 | T=0.85 | T=0.9 | T=0.95 |
| 50x10 | 318 | 446 | 578 | 93 | 318 | 1071 |
| 100x10 | 448 | 650 | 881 | 124 | 448 | 1777 |

### 4.11.4 The second layer

Second layer behaves similar to the first layer except that the nodes are now labelled as belonging to a specific class. Nodes in first layer do not have any labels but only a sequential index. As the image segments are scanned, one of the nodes in first layer resonates whose index is recorded. Set of 16 such indices per image are stored in a pattern array. Part of the Figure 4.8 marked as Patterns indicates reshaped output of layer 1, which is also the input to layer 2. Each node in layer 1 is able to identify several variations of its resonating pattern and hence represents not one but a set of input image segments, close to each other. Further similarities in segments need not be identified but some function that identifies indices with similar features may also be used. One such function named imagediff ( ) was implemented but its use was not found necessary.

The input to second layer consists of a pattern described above. Output labels assigned to nodes in the present case of MNIST images, indicates the number identified by the node, viz., 0 to 9. Several nodes will have the same label because a digit may have many different representations. Much like the first layer, the node with highest output is considered as winner in second layer, with its label indicating the recognized digit.

Specific rearrangements of patterns can indicate spatial operations on input image. For example, rotation of the image by 90 degrees to the right can be done at input of layer 2 by following operations: (i) reshaping the pattern vector as 4x4 matrix (ii) pivot on principal diagonal (iii) mirror on right vertical and (iv) reshape to vector. If such operations are applied on all test images, rotated images can also be identified. Alternately, a new node may be added at Layer 2 to identify rotated image. Small changes to orientation are tolerated at layer 1 itself and hence no new node may get added at layer 2.

### 4.11.5 Effect of Perturbation

Perturbation refers to introduction of small changes to the system for example, rotation by a small angle, translation, skew, etc. In a network with large number of nodes, appending new nodes that are related but different from existing nodes is equivalent to a learning step. Perturbed nodes may require validation during later stages of learning but it provides an opportunity to respond well to an unknown input. Recognition accuracy of ARN can be improved by adding perturbed nodes to the network, albeit at the cost of increased computation. Perturbation can occur in input, output or network parameters. Perturbation may be random but limited to small variations such that the estimated response is not wrong but within a limited range of acceptability. In some cases, it may even be possible to use locally correct analytical equations to synthesize perturbed nodes. Adding a perturbation layer to ARN, will also reduce the number of images required to train the network. Some of the samples after rotating the image by small angles are shown in Figure 4.11.

A neural network may be considered robust if it is able to classify the data even in presence of noise. Addition of noise to input images during training will improve the robustness to ARN systems. It is also necessary that the system should recognize slight changes to orientation and scaling of input. Certain degree of robustness of ARN is inherent in its resonance characteristics. It can be further improved by spawning new nodes from existing nodes and perturbing them. Often, this is done when a new node is added. It may also be done when the network is dormant, i.e., in the background when no input is expected. This is similar to dream sequence where actual inputs are altered in sequence, orientation, shape, size etc. Perturbation of input is meaningful in MNIST implementation but output perturbation is not meaningful as the output is discrete and labelled.

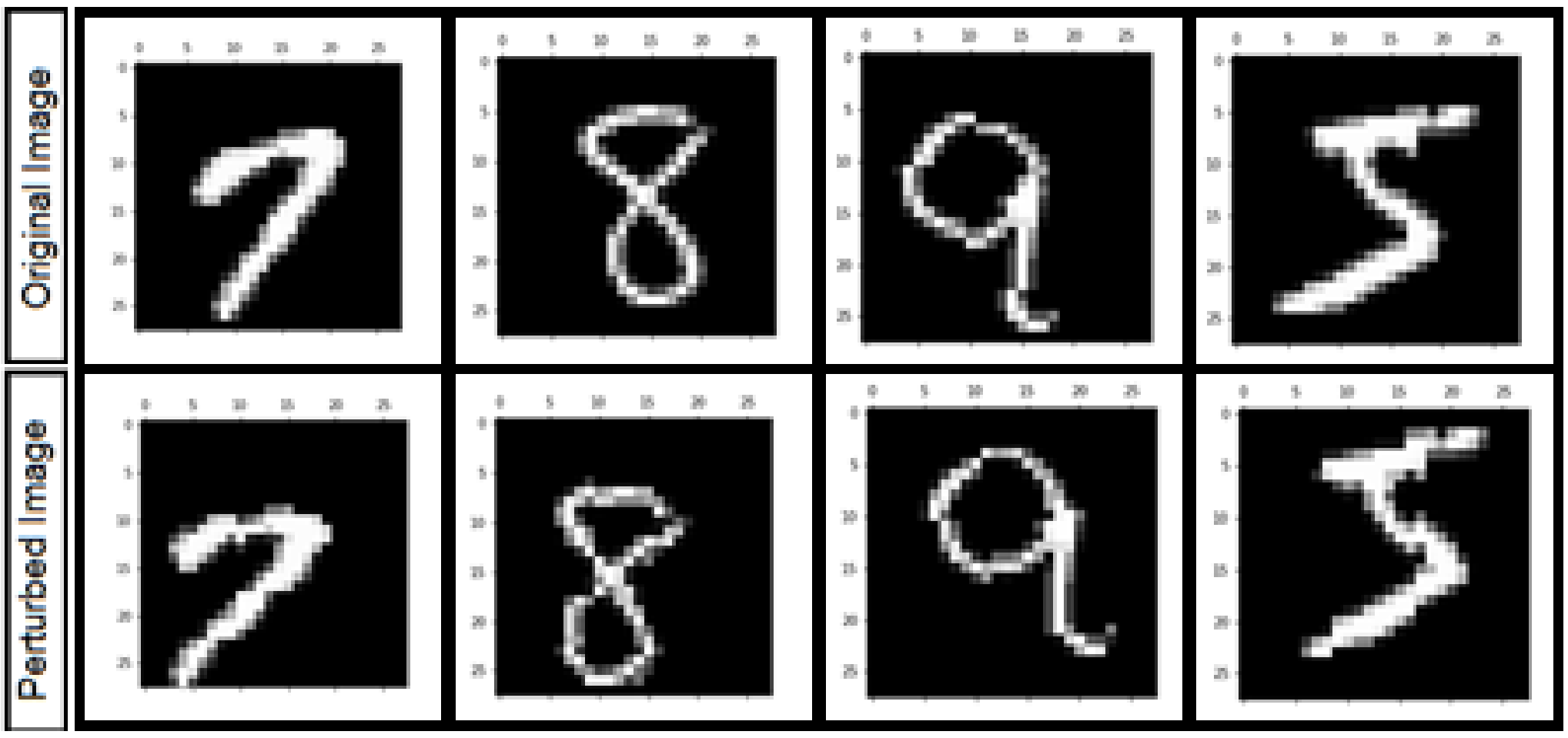

**Figure 4.11 Perturbed image samples**

However, in case of robotic path planning the output is a continuous function and perturbation of output of a mutated node performs useful interpolation or extrapolation of results [2017 Aparanji]. However, some shape functions need to be used to ensure smooth transition through mutated nodes. They will undergo retuning as and when they resonate. For discrete output values, output perturbation does not hold good. The neural networks are supposed to be noise tolerant; therefore if the input is perturbed by a small factor, some improvement in the performance is expected. However, if the input is perturbed by a large factor then it may mislead the network to completely misclassify the data.

### 4.11.6 Effect of Input Masking

On the first thought, it appears that the number of resonators, which constitute the bulk of computation, can reduce if some kind of input threshold is used to mark inputs that are significant and not using any resonators for inputs below a threshold. However, our experiments gave contrary results. Recognition accuracy reduced on thresholding at the input. Possible reason for this is low input is also as important as a high input. Both contribute equally to the recognition. The value of the input when the peak occurs is not important but the occurrence of the peak is important. Resonator peak can occur at any value of input. Therefore, thresholding reduces the information content from the input and hence it takes longer for ARN to learn. This results in reduced accuracy.

To further illustrate the effect, consider two numbers 3 and 8 (see Figure 4.10(c)). What distinguishes the two images is the absence of high valued pixels on the left side of 3. Alternately, one can say that presence of low value pixels on the left side is as critical to recognition of digit 3 as high valued pixels at similar location to recognition of digit 8. Absence of the resonators set to low value input will make it harder for ARN to distinguish between 3 and 8. Therefore, it is important to distinguish between absence of data and presence of low input data.

### 4.11.7 Effect of Tuning on Learning

Learning in ARN varies depending on the value of ρ, T and training sample size. As it is mentioned earlier, increase in value of ρ, reduces the coverage and vice-versa. However, for $\rho < 2$ the resonance curve shown in Figure 4.6 is almost flat. It means that there would be very few nodes created and coverage of one node may overlap with other, leading to multiple recognitions. This scenario reflects early stages of learning when the image is identified but not properly classified. As more learning occurs, resonance gets sharper and more images will get identified. For $\rho > 3$, the resonance curve gets very narrow. Lower coverage means more failures during test.

Therefore, the network should start with a reasonable value of ρ such that enough nodes are created to improve classification. The network should retune the nodes over time rather than starting with nodes having very flat or very narrow coverage.

Threshold also has a similar effect on learning. When the threshold is low, the coverage is large and hence most of the images are recognized but because of overlap of coverage, they will not be correctly identified. Increasing the threshold will reduce the coverage and increase the number of nodes in the ARN layers. Half power point is a reasonable threshold to start with. For improved performance, the threshold has to be increased slowly. Very high threshold reduces the coverage and thereby increasing the number of nodes. Threshold of around 0.9 (when peak is normalized to 1) yields good results.

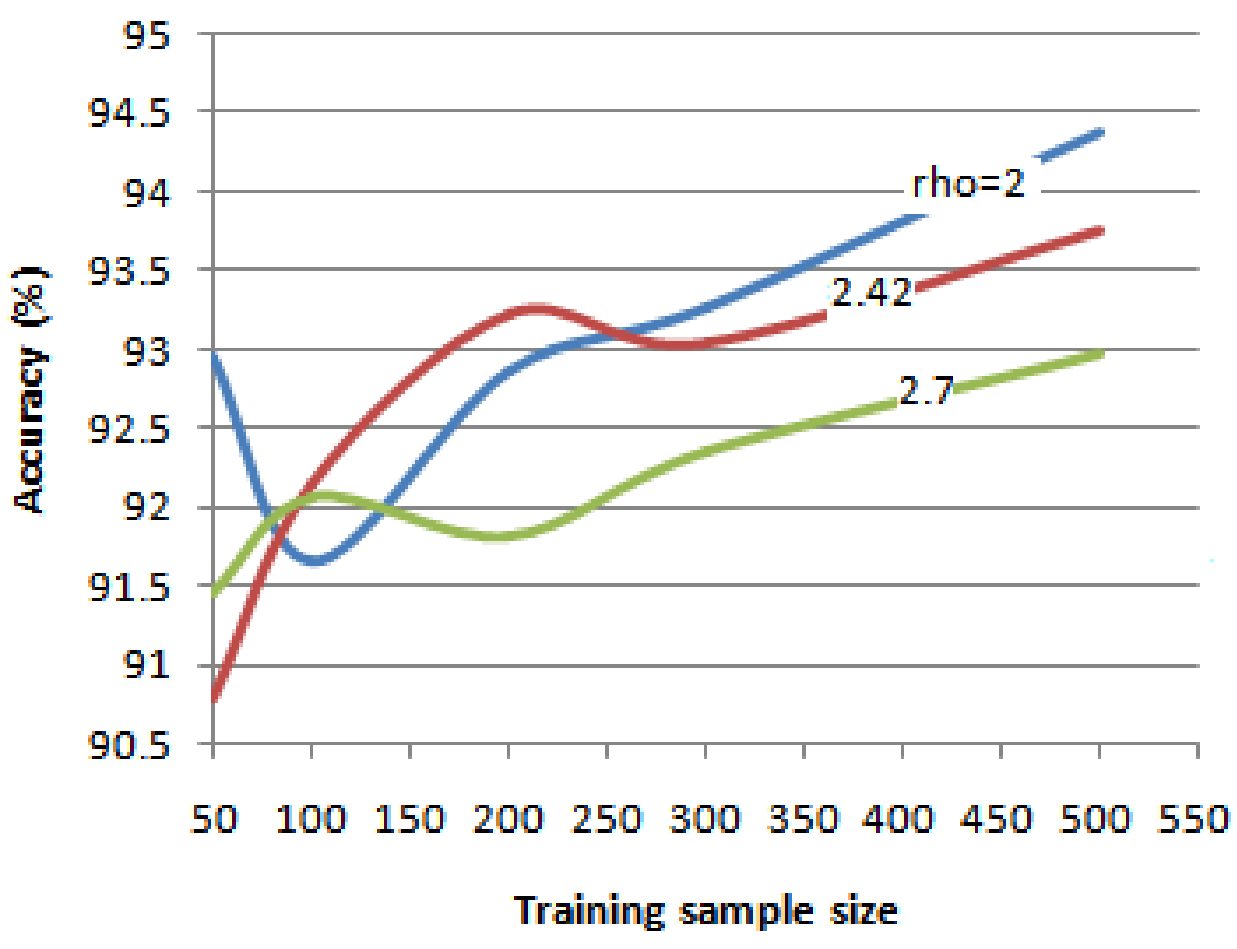

(a)

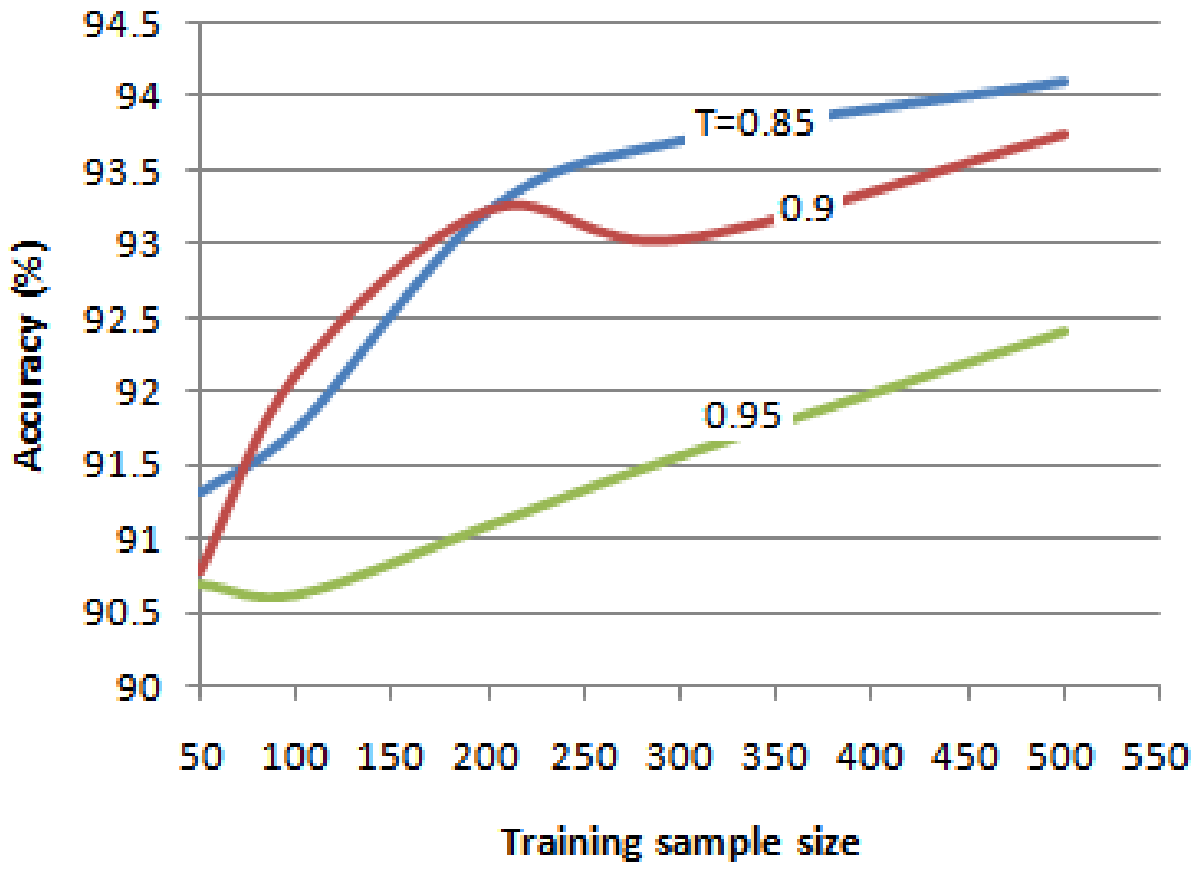

(b)

**Figure 4.12 Learning curves of Image Recognition using ARN**
**(a) Effect of Resonance Control Parameter on accuracy**
**(b) Effect of Threshold on accuracy**

As indicated in earlier sections, choice of resonance control parameter is data dependent and therefore needs experimentation with any given dataset. However, the results presented here give reasonable values for good learning (see Figure 4.12). Learning within a node happens by improving the coverage. Both dilation and contraction of coverage have their use. Dilation of coverage generalizes recognition while contraction creates a more specialized recognition.

### 4.11.8 Ambiguity Resolution

As discussed in previous sections, there are three possible cases of recognition viz., Correct recognition, Wrong recognition and Multiple recognitions. The case of multiple recognitions needs some attention. At the output of every layer, only one winner is expected. However, there will be cases when two or more nodes produce same output especially when low precision numbers are used. There are two possible approaches to resolve such ambiguities.

If the incoming input shifts the statistical mean towards the new input, resonance of such node may be shifted towards new input either by supervisory action of trainer or by reinforcement. This will reduce the distance between the peak and input, resulting in higher output from the node. Alternately, if the standard deviation of input is increasing, one of the nodes can relax the resonance control parameter to produce a slightly higher output. Relaxing the resonance will increase the coverage, producing a higher output.

Resolving the ambiguity in recognition is best implemented as a wrapper around every layer. The wrapper provides local feedback and ensures that only one winner is presented at all times. Functionality of such wrapper is best implemented as a small function that suggests local corrective action. Function of such wrappers may be compared to ganglion on nerve fibres that assist local reflex actions. In the current implementation, it is completely supervised.

## 4.12 Explainable Paths in ARN

In a multi layer ARN, each layer represents a level of abstraction. Each layer is independent and implements a partial recognition function. There is only one winner at every layer for any input. Output from a layer is accumulated over several iterations, reshaped and applied as input to next layer. Therefore, a relation between the sequentially firing neurons is input to the nodes in higher layer. This relation can be causal, temporal or spatial or a combination, depending on the function of the layer. The interface between two layers essentially records delays in the output from a lower layer and presents it as input to next layer(s). As the input to higher layers is a firing sequence, learning in ARN is similar to Hebbian learning that occurs in biological systems.

Further, output of a multi layer ARN can be traced across the layers up to the primary input layer, through a specific set of nodes and a temporal sequence. A node in every layer identifies a specific part of the whole recognition. An interface detects a relation. Therefore, every recognition is caused by a specific path from primary input up to the output layer, with as many segments as the number of layers in ARN. This segmented path is unique to every recognition. Therefore, input to the output can be easily interpreted in terms of partial recognitions at every layer. The path used to generate a specific output explains the reason for the output.

## 4.13 Results and Discussions

A network may be considered as robust if following conditions are met:

(a) Output is range bound and does not drift, oscillate, overshoot or vanish and

(b) Network recognizes the input in the presence of noise.

ARN is inherently robust because (i) output of a node or layer is always range bound and (ii) resonance characteristics gives noise tolerance. It is possible to add any number of feed forward layers without affecting the network stability. Further, the path taken for any specific recognition is always uniquely identifiable. This leads to a better understanding of the network and helps in performance improvement.

First of all, it is possible to track what the ARN actually did with the images. This traceability of image classification can be crucial in improving the network performance. For example, the images not likely to be used can be trimmed from the ARN resulting in a much smaller network, improving the overall performance of the network.

Three possible cases of recognition are shown in Table 4.3. The first row shows the images with correct recognition. Though the matched images are fairly different from the input, ARN matches the image correctly. On the other hand, the images in second row are different digits but the large part of the images is very similar. Therefore, there is a wrong recognition. We rush to state that the percentage of wrong recognition is less than 7% even for a small training set of 50x10 images. The last case of multiple recognitions is also similar. The images match in large part which caused two or more images to be recognized as potential winners.

Notice that wrong recognitions are due to strong similarities with the recognized digit. Often, (3, 5), (3, 8), (4, 9) etc. ( 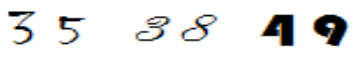 ) are wrongly recognized because of the way they are written (apriori probability). Both second and third cases can be resolved using larger training set and tuning the corresponding nodes to match better. It is also possible to add one or more layers of ARN to resolve such ambiguities. Recognition accuracy improves with increase in size of training set.

The confusion matrix for training sample size =200x10, $\rho$=2.42 and T=0.9 is given in Table 4.4 as an illustration. Notice that digit 1 has highest true positive and 3 has lowest true positive, indicating that a higher layer is required to resolve issues with recognition of digit 3.

**Table 4.3 Results for $\rho$=2.42, T=0.9, training sample size=200x10 & test sample size=60x10**

| Recognition type | Test image | Matched image from training set |
|---|---|---|
| Correct recognition | 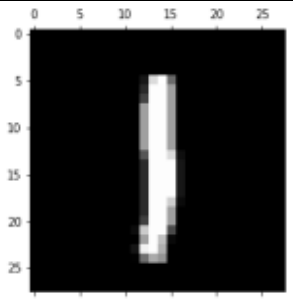 | 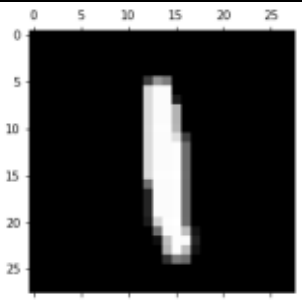 |
| | 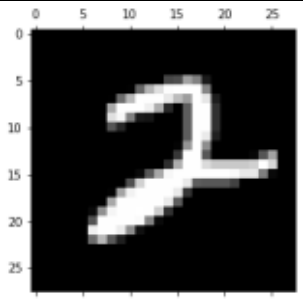 | 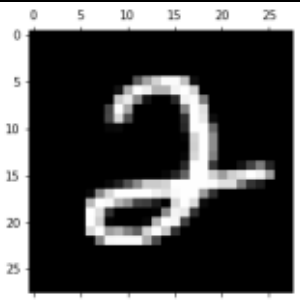 |
| | 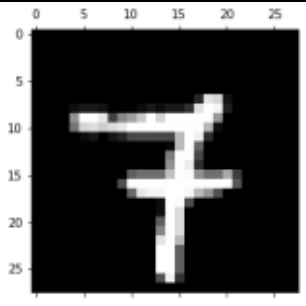 | 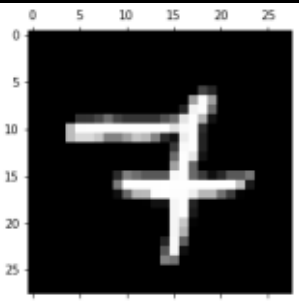 |
| | 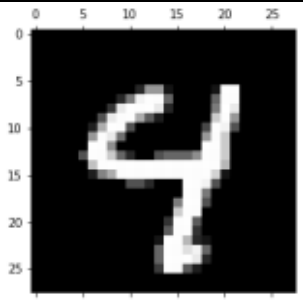 | 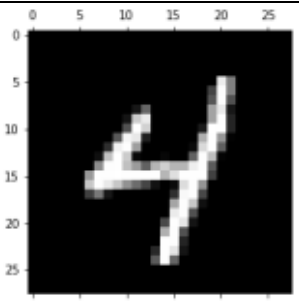 |
| Wrong recognition | 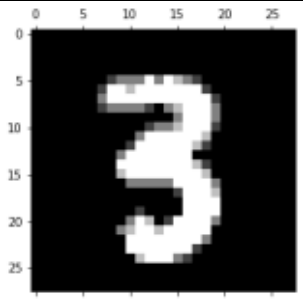 | 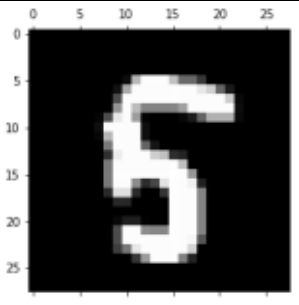 |
| | 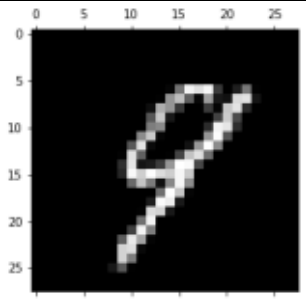 | 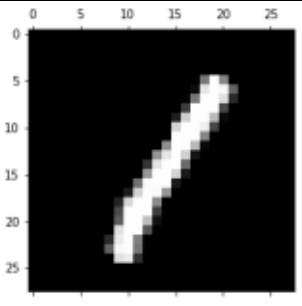 |
| | 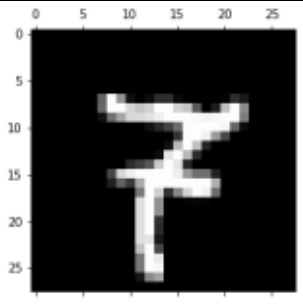 | 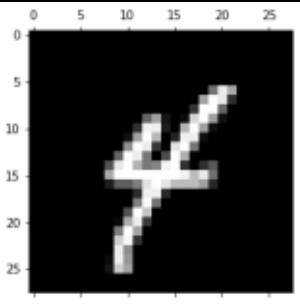 |
| Multiple recognitions | 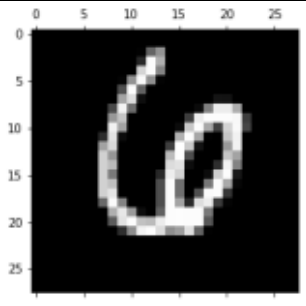 | 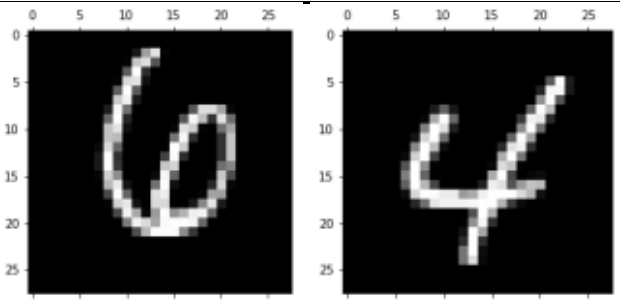 |
| | 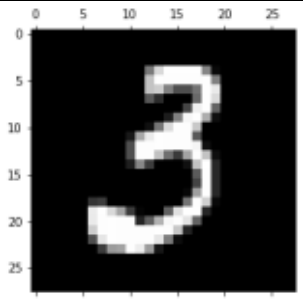 | 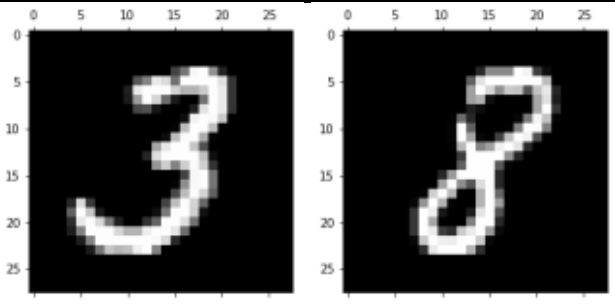 |
| | 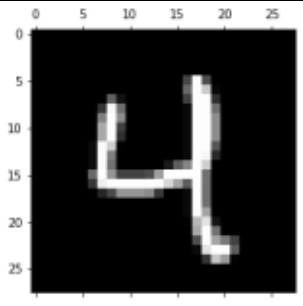 | 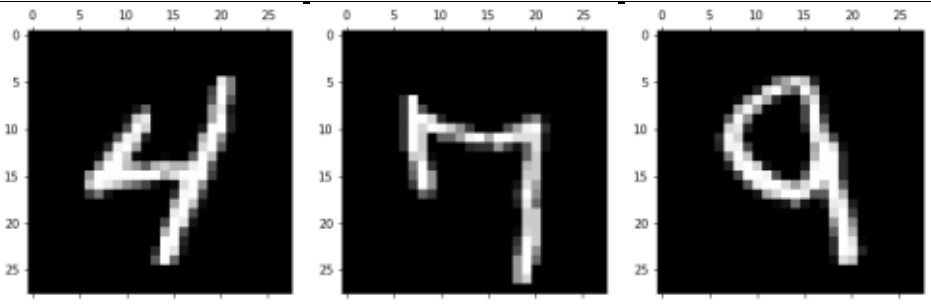 |

**Table 4.4 Confusion Matrix for $\rho$=2.42, T=0.9, training sample size=200x10 & test sample size=60x10**

| | 0 | 1 | 2 | 3 | 4 | 5 | 6 | 7 | 8 | 9 | Total |
|---|---|---|---|---|---|---|---|---|---|---|---|
| **0** | **41.75** | 1.00 | 2.50 | 2.75 | 0.00 | 3.75 | 2.00 | 0.00 | 6.25 | 0.00 | 60 |
| **1** | 0.00 | **59.00** | 0.00 | 0.00 | 0.00 | 0.00 | 0.00 | 1.00 | 0.00 | 0.00 | 60 |
| **2** | 1.31 | 4.14 | **35.97** | 8.03 | 1.61 | 2.11 | 1.11 | 2.61 | 1.98 | 1.11 | 60 |
| **3** | 0.00 | 6.17 | 3.33 | **25.83** | 0.333 | 13.33 | 1.17 | 5.00 | 4.50 | 0.33 | 60 |
| **4** | 2.00 | 5.92 | 1.42 | 0.00 | **37.58** | 0.17 | 0.16 | 2.17 | 0.48 | 10.16 | 60 |
| **5** | 3.25 | 1.70 | 0.70 | 3.25 | 1.67 | **37.95** | 3.50 | 0.20 | 4.78 | 3.00 | 60 |
| **6** | 3.19 | 1.36 | 2.20 | 0.00 | 4.69 | 2.36 | **44.30** | 0.11 | 1.61 | 0.11 | 60 |
| **7** | 0.00 | 4.03 | 0.00 | 1.33 | 3.20 | 0.00 | 0.00 | **40.03** | 0.70 | 10.70 | 60 |
| **8** | 5.37 | 2.50 | 5.20 | 2.70 | 1.58 | 4.37 | 1.20 | 1.58 | **32.41** | 3.08 | 60 |
| **9** | 0.25 | 1.00 | 0.00 | 1.25 | 8.67 | 0.25 | 0.00 | 2.83 | 2.58 | **43.16** | 60 |

## Summary

ARN looks like a promising solution to some problems in modern Artificial Intelligence. It has been used successfully in path planning which is one of the NP-hard problems to solve analytically. This work shows its ability to perform visual comparison, which is also hard to solve analytically. It is possible to implement temporal relations into ARN as described for pathnet, a version of ARN for motion control [2018 Aparanji]. Such networks can be useful for time series prediction and language processing. Parallel implementations of ARN with partitioned input space are possible especially in image recognition and classification. The problem of how features are shared among multiple partitions needs to be explored.

During adaptation of ARN for image recognition, some generalizations were made. The effect of tuning parameters such as resonance control parameter ($\rho$) and threshold (T) were studied. A method to estimate ρ was presented. The relation between ρ and standard deviation was derived. Some of the initial part of this work was presented in an International conference in May 2019 [2019 P7] and the details of theoretical derivations with generalized equations have been submitted to a SCOPUS indexed journal [2019 P3]. Currently review is in progress.

# Chapter 5

# Resonator for Auto Resonance Network

This research work started with plan to implement a basic compute core for hardware acceleration of neural processing, with a focus on ARN. Some of the open-source designs like MIPS [1981 Hennessy] were used to gain the initial direction for design of a processor (See https://www.mipsopen.com for code). Most of the open source soft core processors implement RISC architecture. MIPS softcore also offer several advanced features like SIMD extensions. Core designs for neural computation are very recent development and not available in open source.

Implementing the basic functional unit of ARN, i.e., the ARN neuron was the foucs of this work, similar to DaDianNao's focus on NFU (see section 2.5.8). As the resonator forms the core of ARN node (see section 4.3), initial focus was to implement the resonator. Attempts were made to optimize area and speed. All modules designed in this work have a well defined FSM and ports so that they are reliable and reusable.

One of the essential modules in realizing a resonator was a multiplier. Given the large number of multiplications required for any ANN, use of serial and parallel multipliers was compared. Interestingly, the serial multiplier has some distinct performance and area advantages in massively parallel environment, as presented in section 5.1.

It was clear from the experiments on ARN for image recognition that use of low precision can be acceptable, and infact advantageous for overall performance of ARN. Therefore, a 16-bit fixed point number format was used for all computations within the accelerator. Details of this format are presented in section 5.2. Details of the serial multiplier using this number format are presented in section 5.3.

Building a resonator for ARN has several options. Resonator requires a monotonic transform to map the input range to a constrained range. Among several options, the sigmoid, described by eqn. 4.7, has been used, as it is controllable and has uniform coverage characteristics over the entire input range of real numbers ($\mathbb{R}$). Overall design of the resonator is based on interpolation of a family of resonance curves (see Figure 4.1d) using input value and resonance control parameter as interpolation variables. Details of approximation using Piece-wise Linear (PWL) approximation and Second

Order Interpolation (SOI) methods are discussed in sections 5.4.1 & 5.4.2 respectively. Hardware implementation of an approximation unit based on these interpolation methods is given in section 5.5. Implementation of the resonator is presented in section 5.6.

## 5.1 Serial vs. Parallel Computation in Massively Parallel Environment

Almost every arithmetic computation can be implemented in parallel as well as serial mode. The number of serial units that can be realized in a given area is higher than the number of parallel units. However, parallel units compute faster. Assume that the parallel unit computes a result in 1 clock but the serial unit needs 10 clocks. If the area taken by parallel units is 15 times that of serial units, it should be possible to implement 15 serial units in the same area as one parallel unit (assuming no overheads). Given 10 clocks, the serial units will perform 15 computation while the single parallel unit can complete only 10 computations. Therefore, the throughput of a set of serial implementations can exceed the throughput of parallel implementation. This argument assumes that there are enough computations to keep the units busy. Such scenario exists in a massively parallel environment like that of a DL system. An observation is made here on the tradeoff between implementing serial and parallel units.

***Lemma**: In a massively parallel environment, the throughput of a set of serial execution units can exceed the throughput of parallel units occupying the same area as the set of serial units if the ratio of areas is greater than the ratio of execution times.*

**Proof**: In a massively parallel environment, number of pending operations will be in excess of available resources. Let $A_s$ and $A_p$ represent the areas of serial and parallel units. The parallel units take a larger area and hence the ratio $\frac{A_p}{A_s} = R_A > 1$. Similarly, let $T_s$ and $T_p$ represent the number of clocks required to execute serial and parallel units. As parallel units execute faster, the ratio of time taken to execute one instruction is $\frac{T_s}{T_p} = R_T > 1$. Considering area of one parallel unit, the number of serial units that can fit into the same area is $R_A$. For a given amount of time T, parallel unit can execute T/$T_p$ instructions. Similarly the set of serial units can execute $R_A(\frac{T}{T_s})$ instructions. Now, if the throughput from the set of serial units is larger than that of a parallel unit, we can write

$R_A(\frac{T}{T_s}) > (\frac{T}{T_p})$, or $R_A > (\frac{T_s}{T_p})$ , or

$$R_A > R_T.$$ **QED**

The interconnection and other overheads are ignored in the lemma as they can be absorbed into the area parameters. Some of the large modules like multiplier do satisfy this condition and hence it is better to implement more number of serial units than parallel units. Many open source soft core processors provide options to implement serial or parallel multipliers, which will find this lemma useful. For example, HF-RISC, a soft core RISC processor based on MIPS I architecture provides options for parallel, serial and software multipliers to suit the area/cost demands [2016 Johann]. Comparative performance of serial and parallel implementations of a hypothetical module are presented in Figure 5.1. The speed ratio $R_T$ is assumed to be 17:1. The figure shows the number of operations performed (represented as throughput) for two area ratios, viz., $R_A$=12 and $R_A$=20. It can be easily seen that when the area ratio exceeds speed ratio, serial units perform better.

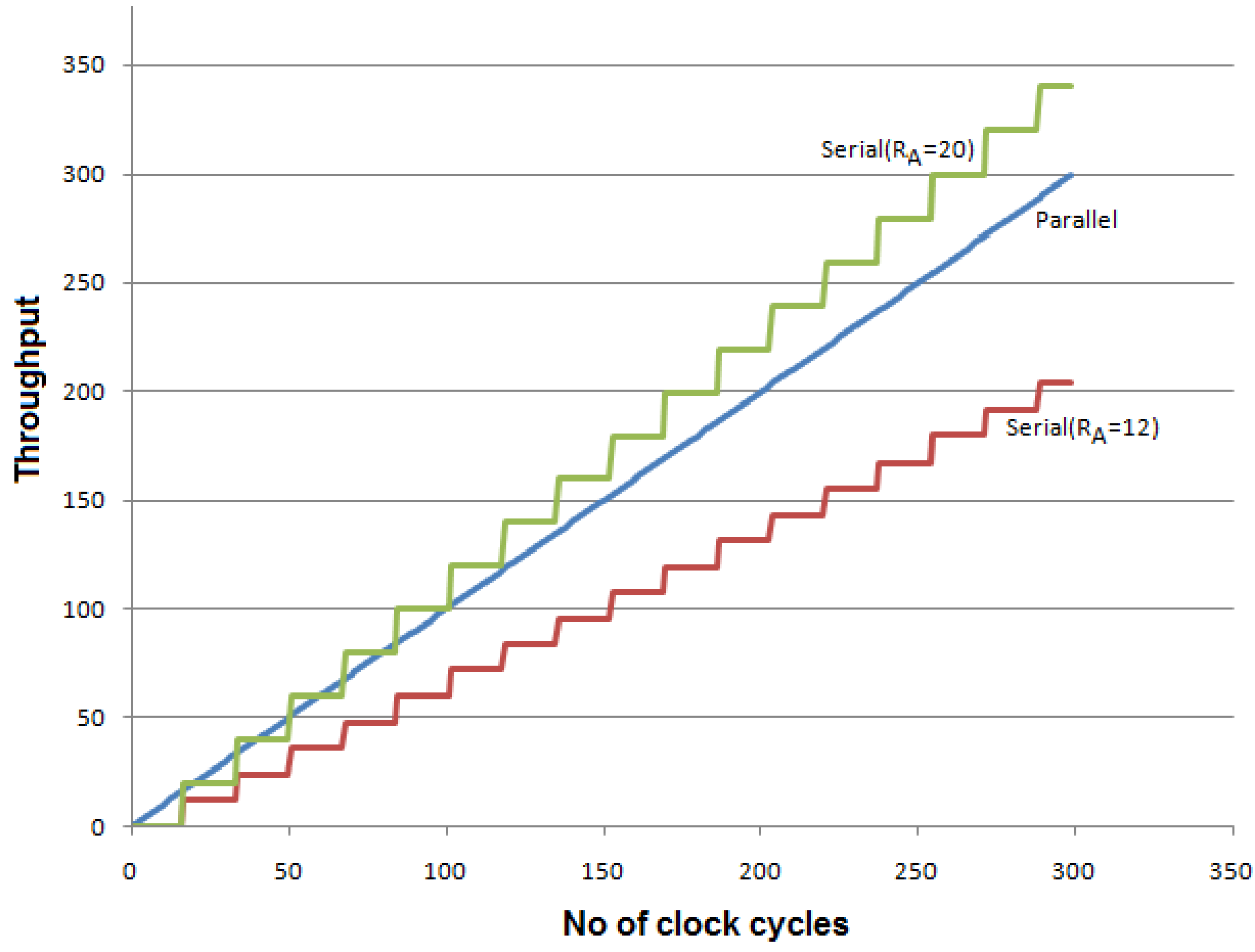

**Figure 5.1 Performance of Serial and Parallel modules in massively parallel environment with Serial to Parallel area ratio of 1:12 and 1:20. Speed ratio is 17:1.**

We implemented serial multipliers for the resonators because the estimated serial to parallel area ratio was around 1:32, while the speed ratio was 16:1. Details of the implementation are discussed in the following sections.

It is to be noted that the overhead of distributing the data to the modules also increases fast as the number of modules increases. Therefore, actual evaluation of serial versus parallel should include computation and peripheral overhead.

## 5.2 Number Representation

Speed and area performance of a module depends on the number representation and the algorithm used. Lowering the bit count will considerably reduce the silicon area but results in loss of accuracy. As the ANNs are noise tolerant, computations can be performed using low precision. A 16-bit number format was sufficient to achieve satisfactory numerical accuracy. Following points influenced the selection of 16-bit format.

(1) In many situations, it is sufficient to represent real world values with two to three fractional decimals of accuracy, or can be transformed to such precision.

(2) The range of numbers used in ARN is very small. Sigmoid saturates outside range of -5:+5. Output from the sigmoid is in the range of (0:1). ARN output is also scaled to be in the range of 0:1. Sum of resonators at the output of ARN node can be limited to small range by rearranging the order of execution.

(3) Resonance of ARN nodes ensures that the node responds correctly to values which are within the coverage of the node. So, an approximate value of input values is sufficient for ARN neurons. In other words, input values close to the locus of resonance will also produce output above threshold because of coverage. Therefore, loss in precision is acceptable.

(4) Area required for a 16-bit implementation of the resonator is considerably smaller than the area required for a 32-bit implementation.

Reducing the area per unit is important as DNNs are massively parallel networks requiring large number of neurons. Every neuron requires several multiplication operations and hence reduction in size of the number has a cascading effect on total area. Using a two or three decimal accuracy is often sufficient. The proposed 16-bit number format is shown in Figure 5.2. One of the design criteria was to allow accuracy of at least three decimals. Using 14 fractional bits would be more correct but that would not leave much space for integer part of the number. Twelve bits gives a reasonable accuracy to represent numbers that are accurate to three decimal places, as shown in Figure 5.3. It also helps reduce the effect of fourth decimal on the third decimal during computations. Please see Figure 5.3 for a bit count and decimal value of the bit for quick appreciation of accuracy.

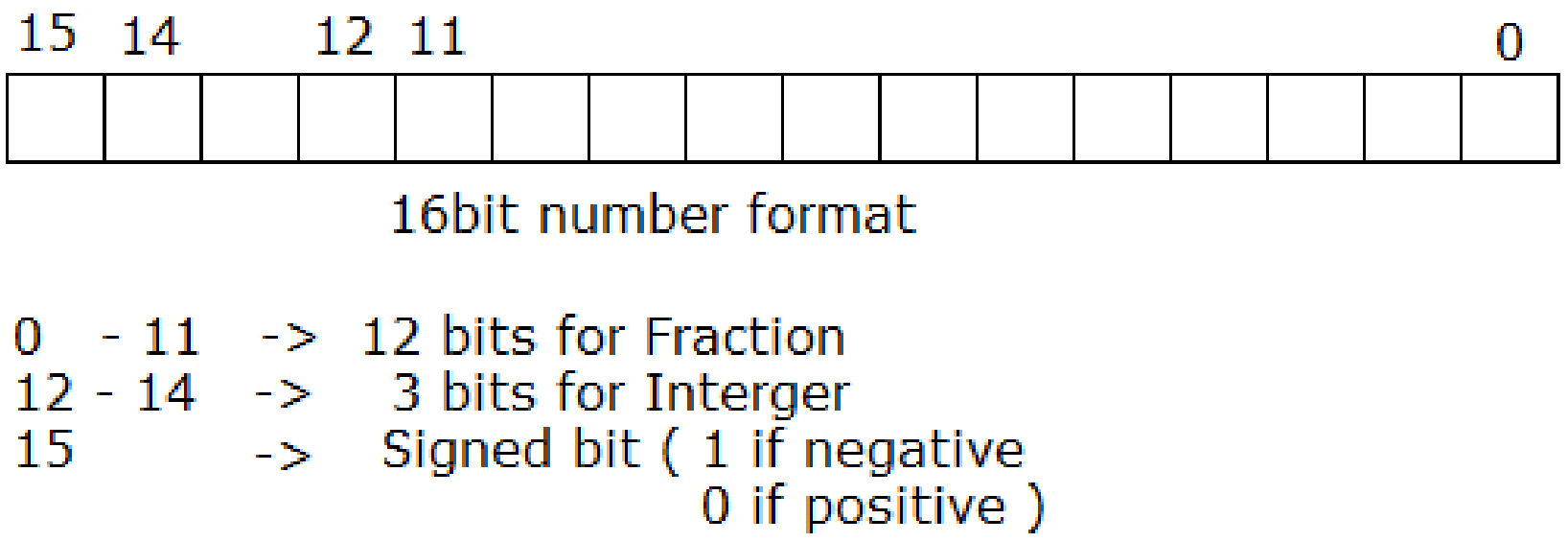

**Figure 5.2 16-bit Number format**

The number format uses twelve bits on the LSB side to represent the fractional part of the number. Next three (four) bits towards MSB are used for signed (unsigned) integer part. MSB is used to indicate the sign bit for signed integers. The decimal point is assumed to be fixed between 11$^{th}$ and 12$^{th}$ bits, as shown in Figure 5.2. The range of this number format is -7.99 to +7.99 for signed and 0 to 15.99 for unsigned. As the computations in ARN can be done using unsigned numbers, that format is frequently used. This range is adequate for ARN computations. As an example, 0.012 is represented as 0 000 0000 0011 $0001_b$. Some examples of numerical operations (unsigned) using 16-bit number format are shown in Table 5.1. It may be observed that limiting the number of bits to 12 has truncation effects but limited to third fractional place in decimal notation. Rounding is not implicit in any of the implemented operations.

**Table 5.1 Examples using Number format**

| Operation | Operand 1 | | Operand 2 | | Resultant | |
|---|---|---|---|---|---|---|
| | Actual | Number format | Actual | Number format | Actual | Number format |
| + | 3.25 | 0011 0100 0000 0000 | 1.76 | 0001 1100 0010 1000 | 5.0100 | 0101 0000 0010 1000 = 5.0097 |
| - | 1.09 | 0001 0001 0111 0000 | 0.28 | 0000 0100 0111 1010 | 0.8100 | 0000 1100 1111 0110 = 0.8100 |
| × | 2.63 | 0010 1010 0001 0100 | 1.59 | 0001 1001 0111 0000 | 4.1817 | 0100 0010 1110 0101 = 4.1809 |

All the basic components of a resonator viz., adder, multiplier, scaled shifted sigmoid activation etc. are implemented using this number format. These modules are discussed in the following sections.

| bit position | decimal fraction |
|---|---|
| 11 | 0.5 |
| 10 | 0.25 |
| 9 | 0.125 |
| 8 | 0.0625 |
| 7 | 0.03125 |
| 6 | 0.015625 |
| 5 | 0.0078125 |
| 4 | 0.00390625 |
| 3 | 0.001953125 |
| 2 | 0.0009765625 |
| 1 | 0.00048828125 |
| 0 | 0.000244140625 |

**Figure 5.3 Selection of 12 fractional bits to achieve 3-fractional digit accuracy**

## 5.3 Multipliers

An overview of low precision serial multiplier implemented using the proposed number format, for use in ARN, is presented here. For comparison, a Wallace tree multiplier using the same number format as described in section 5.2 was implemented. Details of this parallel multiplier and other basic building blocks are given in Appendix – 2. Some of these are listed here.

a) Priority Encoder (PE): The logic ensures that only the highest priority signal is allowed.

b) Bit Selector Logic (BSL): This is essentially a demultiplexer but designed specifically for gating inputs based on the signals from PE.

c) Command Control & Logic Unit (CCLU): It may be seen as a customized ALU when implementing higher functions for accelerator. CCLU has an accumulator (register) whose contents are controlled by the CCLU operations like load, clear, shift right, shift left, add and subtract etc.

### 5.3.1 Serial Multiplier

Serial multiplier is implemented using the classical method of shift and add. Overall structure of the multiplier is given in Figure 5.4. The unit is designed to work on both positive and negative edges of the clock.

The multiplier has two input registers corresponding to two operands. Both are independently controlled. *Load* operation will transfer the data on the bus to *sReg*. *Mul* operation will copy the bus to *slReg* and multiply it with contents of *sReg*. Both registers remain unaltered after the multiplication operation. Therefore, it may be used for single or streamed multiplication, if the multiplier is constant, e.g. scaling a vector or matrix by a constant. End of multiplication is indicated by the *endFlag* signal.

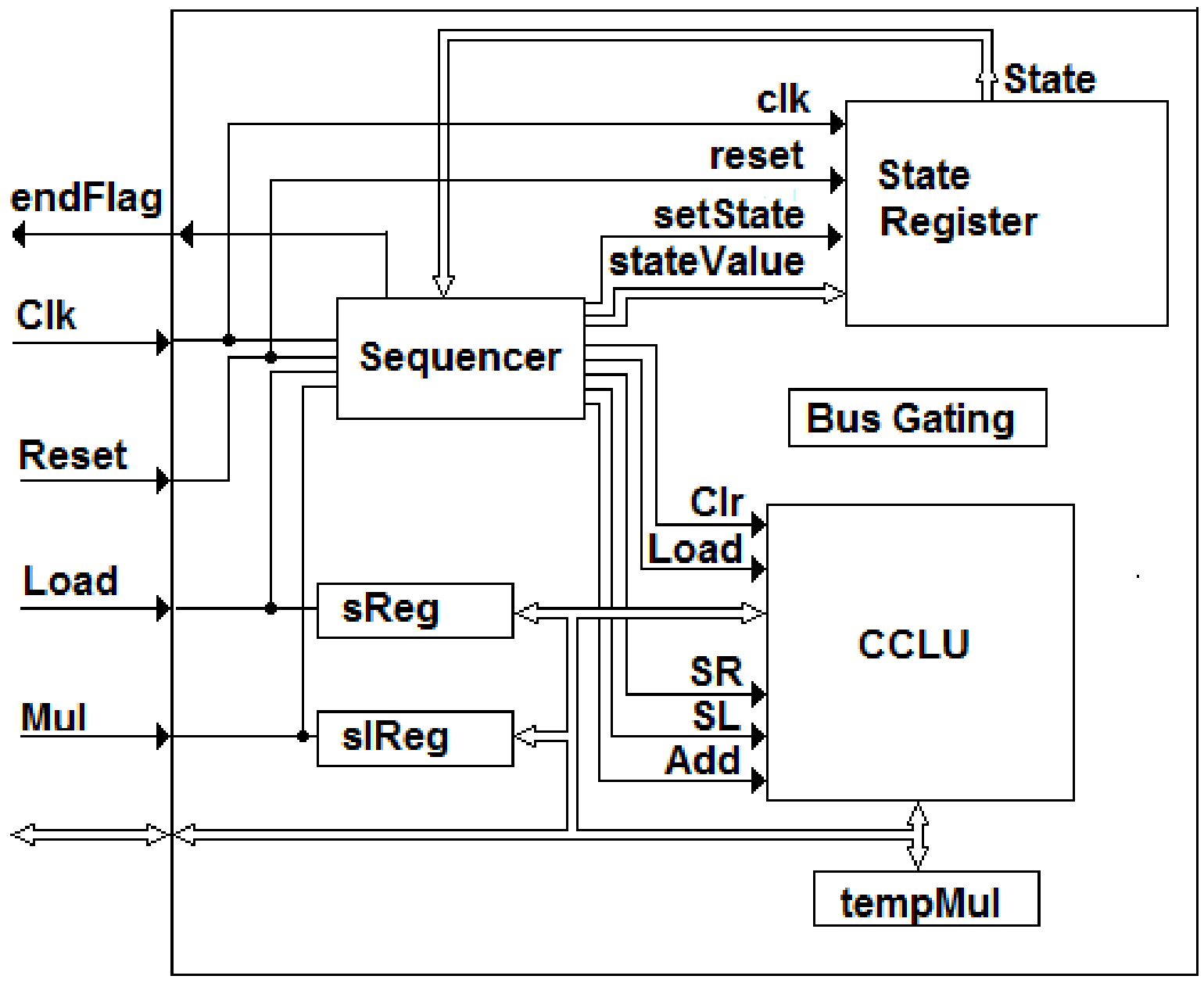

(a)

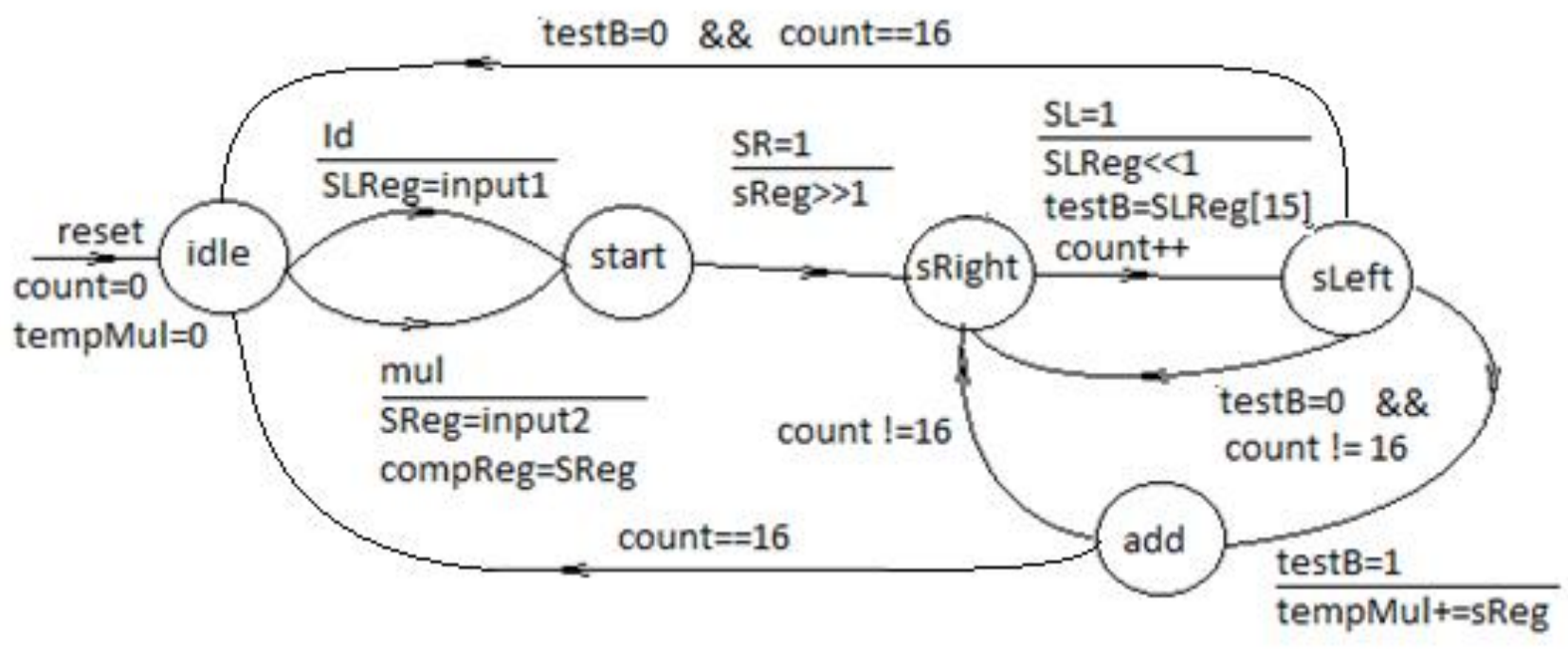

(b)

**Figure 5.4 Serial Multiplier (a) Block Diagram**
**(b) Finite State Machine (edge triggered)**

On arrival of *Mul* signal, a bit counter is initialized to 0 and the product register *tempMul* is cleared. Multiplication can be performed starting from LSB or MSB. Shifting of the multiplicand to left or right will depend on the preferred method. Let us assume LSB first as it is more commonly used method. On positive edge of the clock, the bit counter is incremented, *sReg* is shifted right and *slReg* is shifted left into *testBit* register. If *testBit* is 1 then *sReg* is copied to *tempMul* register. On arrive of next positive or negative edge *sReg* is added to *tempMul* if *testBit* is 1. Otherwise, *sReg* is not added and bit counter is incremented. The process is repeated till bit counter is 16. Once the bit counter reaches 16, the result is transfered to bus.

Full product will be 32-bit wide and hence truncated to 16-bits. As the number format assumes 12 bits for the fractional part, only 12 higher bits of the 24 bit LSB side are to be used. Of the remaining bits, next 6 (8) bits will correspond to integer part, of which, only lower 3 (4) bits will be significant for signed (unsigned) numbers. If any of the higher bits [29:27] (or [31:28] for unsigned) have 1, there is an overflow. MSB of two inputs is xor'ed to get the sign bit of the resultant for signed numbers. The method of truncation is shown in Figure 5.5.

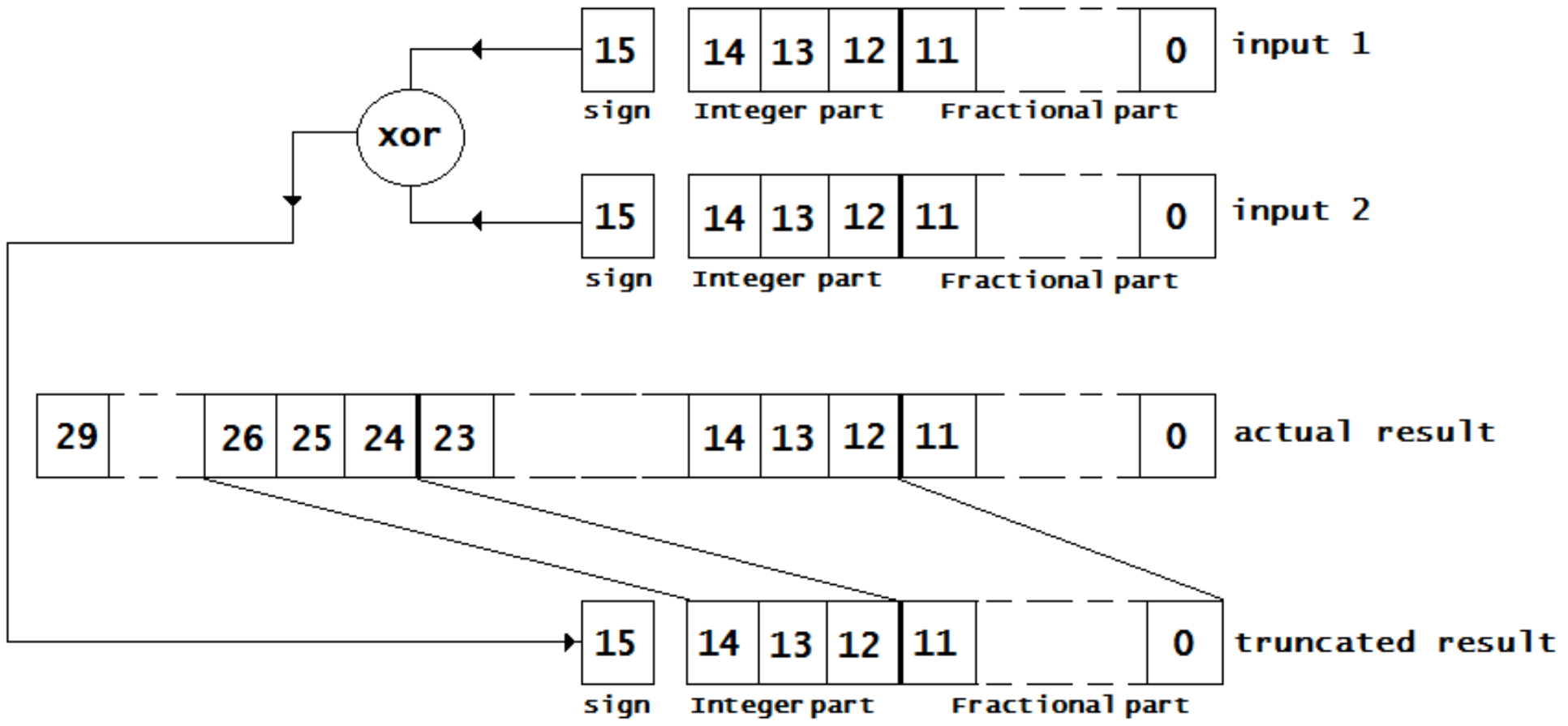

**Figure 5.5 Truncating the 32-bit result into 16-bit**

For signed numbers, the multiplication is performed only for 15-bits and hence the maximum number of clock cycles required for 16-bit multiplication would be 16 clocks (17 clocks for unsigned numbers), considering one extra clock for loading the inputs. The simulation result of serial multiplication is shown in Figure 5.6.

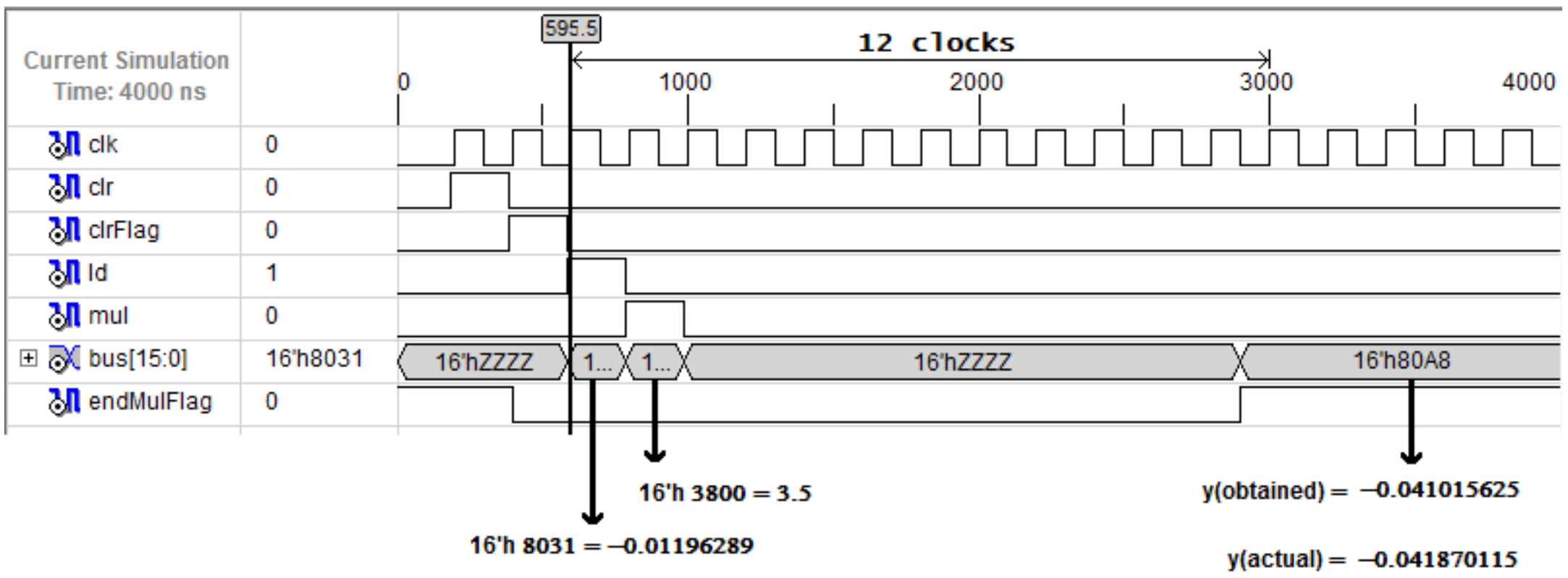

**Figure 5.6 Simulation result of 16-bit Serial Multiplication**

Presence of zeros in the multiplicand saves an edge of the clock. In Figure 5.6, multiplication is completed within 12 clock cycles (including load time) for the 15-bit multipication and the accuracy is up to 3 fractional digits (i.e., 0.001), which is often sufficient for ANN implementations. The accuracy can be further improved by using different techniques for truncating/roundng the result.

### 5.3.2 Performance Comparison of Serial and Parallel Multipliers

Performance comparison between serial and parallel multiplier is given in Table 5.2. The computational speed is in terms of clock cycles and the area is in terms of gate counts. It may be noticed that the serial to parallel speed ratio is 17:1 (16:1 for signed), while the area ratio is 1:32. Therefore, the serial multiplier performs better, both in terms of area and speed as described in lemma (section 5.1): This is true in case of massively parallel environment like neural computations. From the results presented, it may be observed that serial organization of processor including data paths may be preferred over parallel implementations.

**Table 5.2 Area and Performance comparison of Serial and Parallel Multiplier for 16-bit multiplication**

| For 16-bit multiplication | Serial Multiplier | Parallel Multiplier |
|---|---|---|
| No. of Clock cycles | 17 | 1 |
| Gate count (CLBs, IOs) | 66 | 2144 |

## 5.4 Approximation of non-linear Activation functions

In ANN, a non-linear activation function computes the action potential of a neuron. Several types of activation functions are used in ANNs, viz., sigmoid, tanh, ReLU etc. each with different capabilities. Graphs of these non-linear activation functions are shown in Figure 5.7. Exponentiation is a frequently used function among activation functions.. An exponential function can be expanded using Taylor series as given in eqn. (5.1).

$$e^{-x} = 1 - x + \frac{x^2}{2!} - \frac{x^3}{3!} + \frac{x^4}{4!} - \ \dots \qquad (5.1)$$

Direct implementation of exponential function would require a large number of computations and need very large silicon area. Software based implementation would be slow. Therefore, efficient hardware implementation of activation function is critical to real time performance of a NNP.

Activation functions are sensitive to inputs in a limited range. For example, the gradient of sigmoid is significant for inputs in the range $\{x \in \mathbb{R} | -5 \le x \le 5\}$. For $x > 5$, the output is near to 1 and for $x < -5$, the output is near to 0 (see Figure 5.7). Therefore, it is sufficient to implement a hardware module that implements the active region of the function efficiently. As mentioned in the earlier discussions, neural networks are noise tolerant and hence the computations need not be highly accurate. Therefore, use of approximation methods for neural computations is preferred.

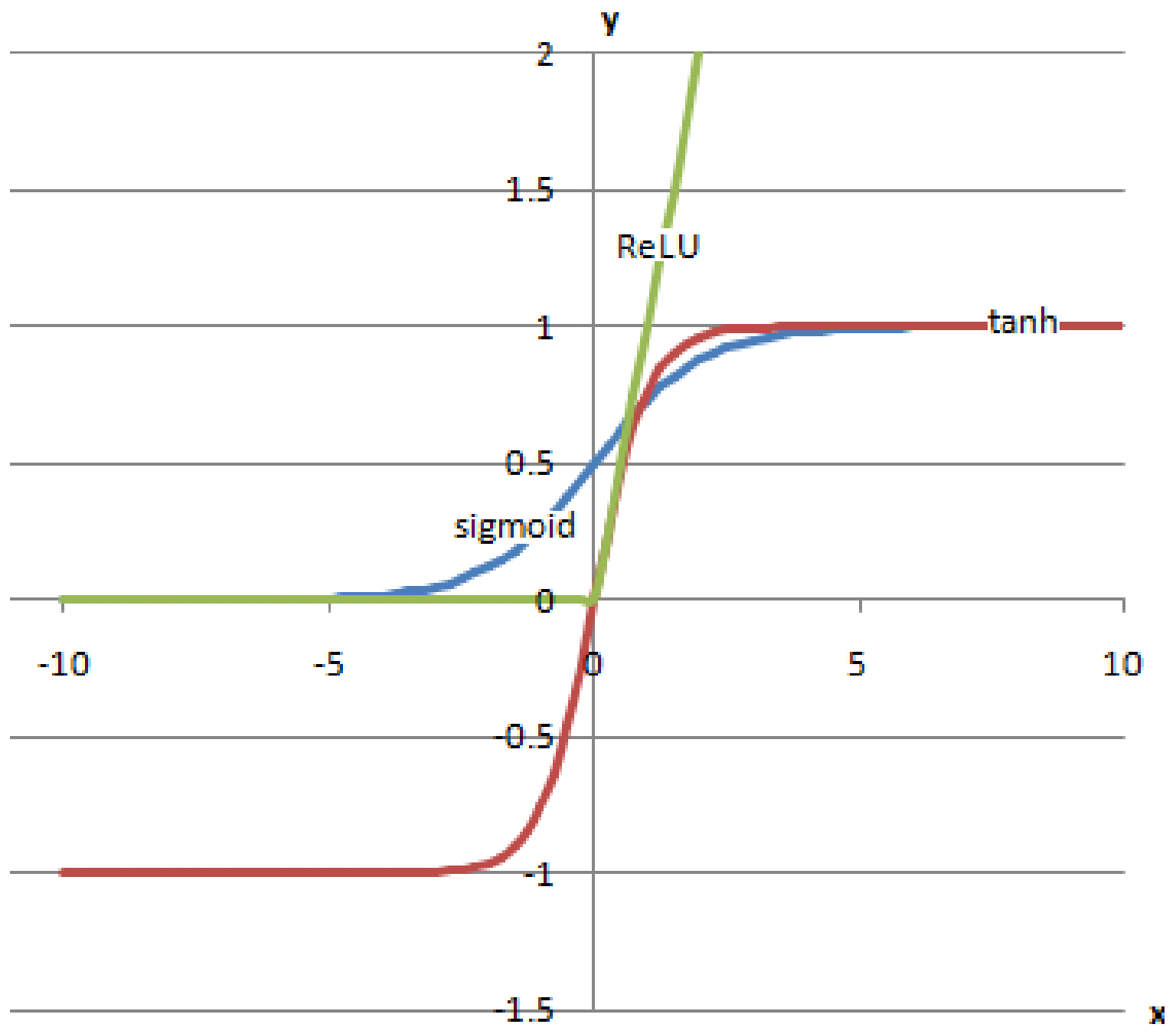

**Figure 5.7 Activation Functions in ANN**

Use of approximation methods like PWL and SOI will reduce the computational overhead, especially in multi-layer networks. These methods have been used to implement sigmoid and ARN resonator functions.

The approximation methods use known points on the curve to calculate the output of a given input at a time. The known points on the curve are stored in look up table (LUT). Computational accuracy and efficiency of using uniform and non-uniform spacing between known points have been compared. When the spacing is uniform and small, there are more number of known points stored in the LUT and hence increased LUT size and computational accuracy. It is also noticed that, with non uniform spacing it is possible to achieve better accuracy as compared with uniform spacing.

### 5.4.1 Piece-wise Linear (PWL)

Piece-wise linear is a well known method, in which a non-linear curve is approximated by a set of linear segments. Deviation of the linear segments from actual curve may be controlled by adjusting the spacing between end points of the segments.

For a quick look at PWL, consider the line in x-y plane shown in the Figure 5.8. Two points ($x_1$, $y_1$) and ($x_2$, $y_2$) are end points of the linear segment closely approximating a small section of a non-linear curve. The value of the curve at any point bounded by the end points of the linear segment can be calculated as given in eqn. (5.2)

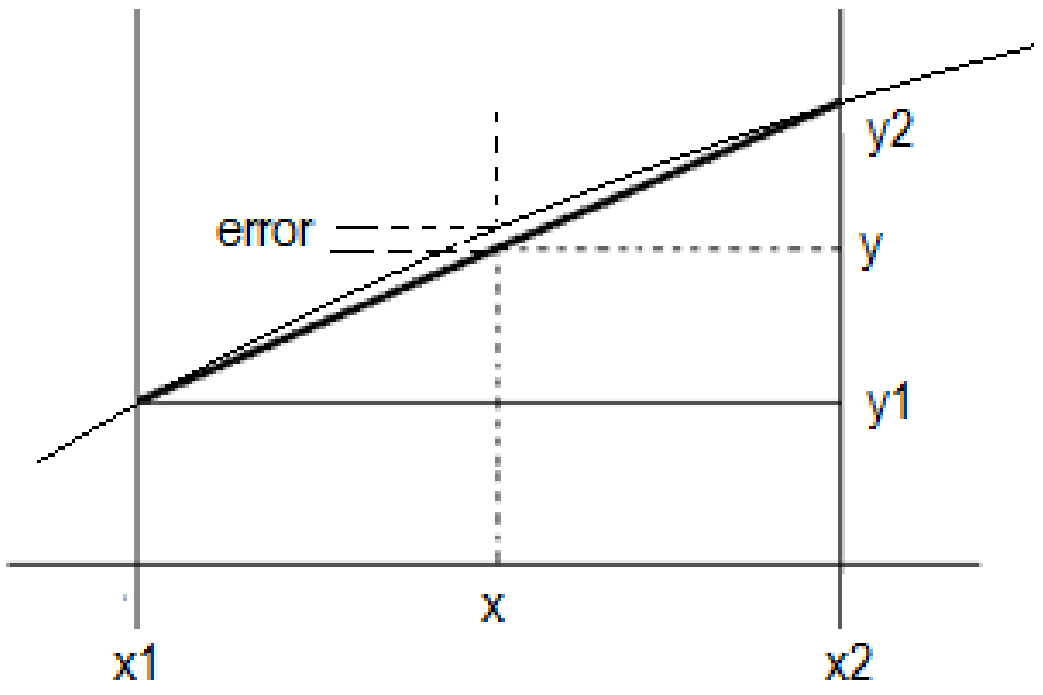

**Figure 5.8 PWL approximation of a non-linear curve**

$$y = \left(\frac{y_2 - y_1}{x_2 - x_1}\right)(x - x_1) + y_1$$
$$= m\,(x - x_1) + y_1 \qquad (5.2)$$

where, m in eqn. (5.2) indicates the local slope of the curve. The values of ($x_1$, $y_1$) and ($x_2$, $y_2$) are known and therefore, m can be pre-computed. Error is the difference between the expected value and the approximated value given by eqn. (A2.1) (see Appendix-2). In a hardware implementation, a set of known values (m, $x_1$, $y_1$) will be stored in the LUT. Note that error at these points is zero. To compute the output of any given input on the curve, two points $x_1$ and $x_2$ need to be identified such that $x_1 < x < x_2$. The corresponding values of m, $x_1$ and $y_1$ are fetched from LUT to perform the computation as given in eqn. (5.2). It can be easily seen from this equation that error is zero at end points of the line segment. Implementation of this equation requires three fetches (m, $x_1$, $y_1$), one multiplication, one subtraction and one addition (subtraction can be implemented as an addition operation with 2's complement). Therefore, number of computations is considerably reduced as compared to that of direct computation using Taylor expansion.

### 5.4.2 Second Order Interpolation (SOI)

Another popular approximation method is Second Order Interpolation which uses three points on the curve as shown in Figure 5.9. These three known points viz., ($x_1$, $y_1$), ($x_2$, $y_2$), ($x_3$, $y_3$) are stored in LUT and can be used to compute the output of given function.

The output of a non-linear curve shown in Figure 5.9 may be expressed as given in eqn. (5.3)

$$\begin{aligned} y &= ax^2 + bx + c \\ &= (ax + b)\,x + c \end{aligned} \qquad (5.3)$$

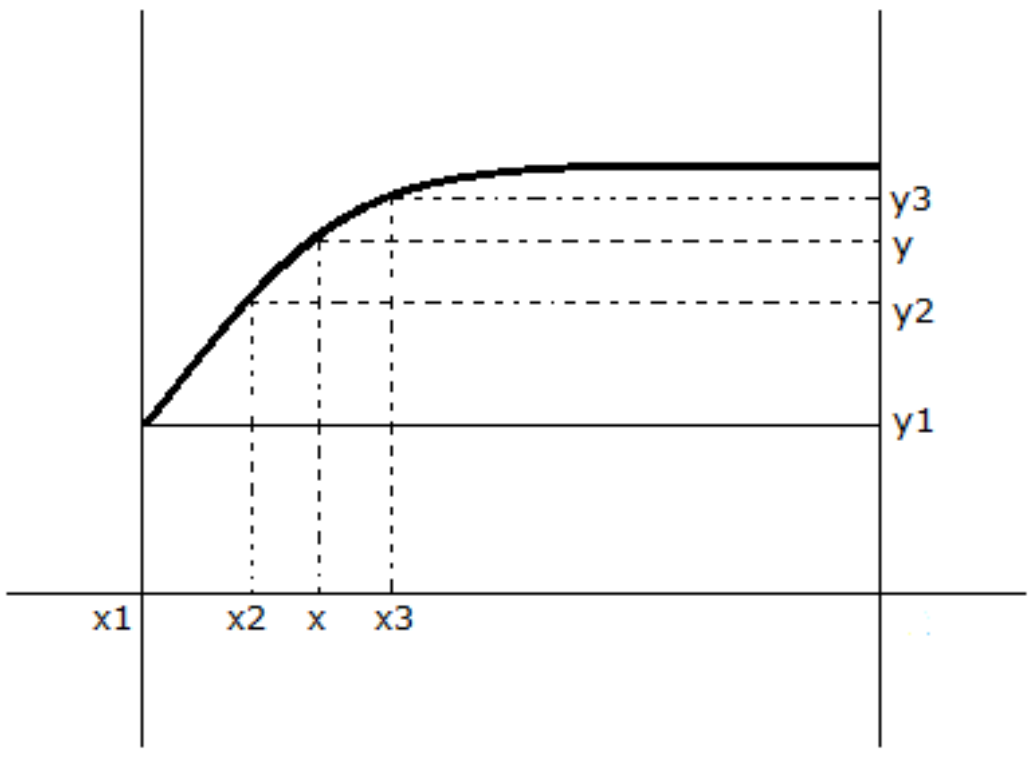

**Figure 5.9 SOI approximation of a non-linear curve**

The coefficients a, b and c are calculated by solving the equation (5.3) for the points $(x_1, y_1)$, $(x_2, y_2)$ and $(x_3, y_3)$ simultaneously as

$$a = \frac{\left((x_2-x_1)(y_3-y_1)\right)-\left((x_3-x_1)(y_2-y_1)\right)}{(x_2-x_1)(x_3-x_1)(x_3-x_2)} \qquad (5.4)$$

$$b = \frac{(y_2-y_1)-a\left(x_2^2-x_1^2\right)}{(x_2-x_1)} \qquad (5.5)$$

$$c = y_1 - ax_1^2 - bx_1 \qquad (5.6)$$

As the values of a, b and c can be pre-computed using the known points, they can also be stored in LUT along with x and y values. The values of a, b and c may be fetched from the look up table and the output of the function is computed using eqn. (5.3). Therefore, three fetch, two addition and two multiplications are required to perform this operation.

**Table 5.3 Maximum errors in computing non-linear functions using approximation methods**

| | | **Sigmoid** | | **Tanh** | | **ARN Resonator** | |
|---|---|---|---|---|---|---|---|
| | | PWL | SOI | PWL | SOI | PWL | SOI |
| **Uniform spacing** | 0.5 | 0.3838% | 0.1478% | 7.4995% | 2.4880% | 6.5575% | 0.8209% |
| | 0.25 | 0.2983% | 0.0226% | 1.9509% | 0.4750% | 2.1042% | 0.1165% |
| **Non-uniform spacing** | 0.125, 0.03125 | 0.0238% | 0.0006% | 0.4347% | 0.1814% | 0.3478% | 0.0314% |

The spacing between the known points has an impact on the computational accuracy. When the spacing is reduced, there would be more points to be stored in a memory and hence the accuracy is better. However, it increases the size of LUT. Maximum errors in computing some of the non-linear functions using approximation methods are given in Table 5.3. The result of approximation of a sigmoid function and the effect of spacing between known points are shown in Figure 5.10 to Figure 5.12.

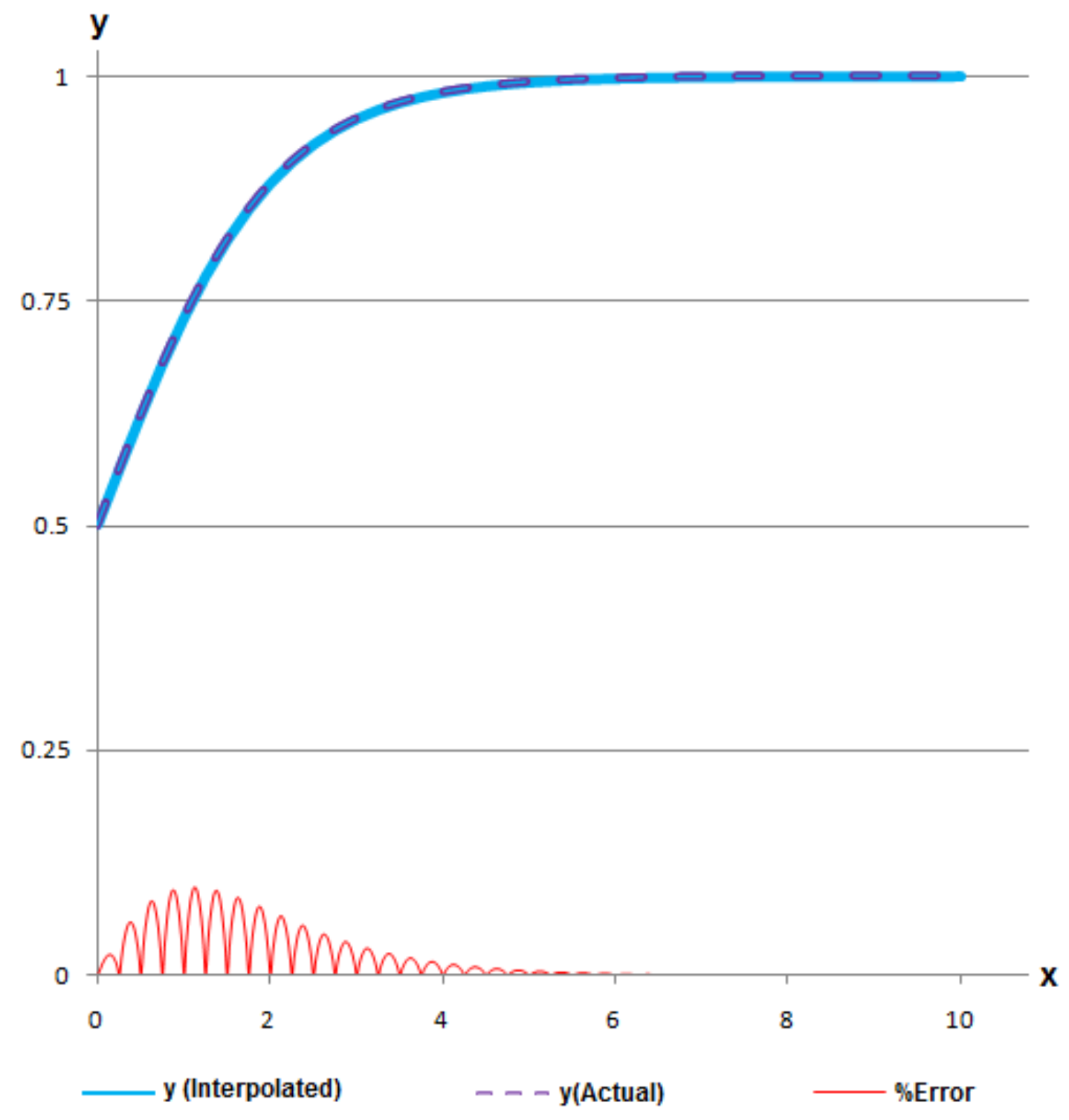

**Figure 5.10 PWL approximation of sigmoid at uniform spacing of 0.25**

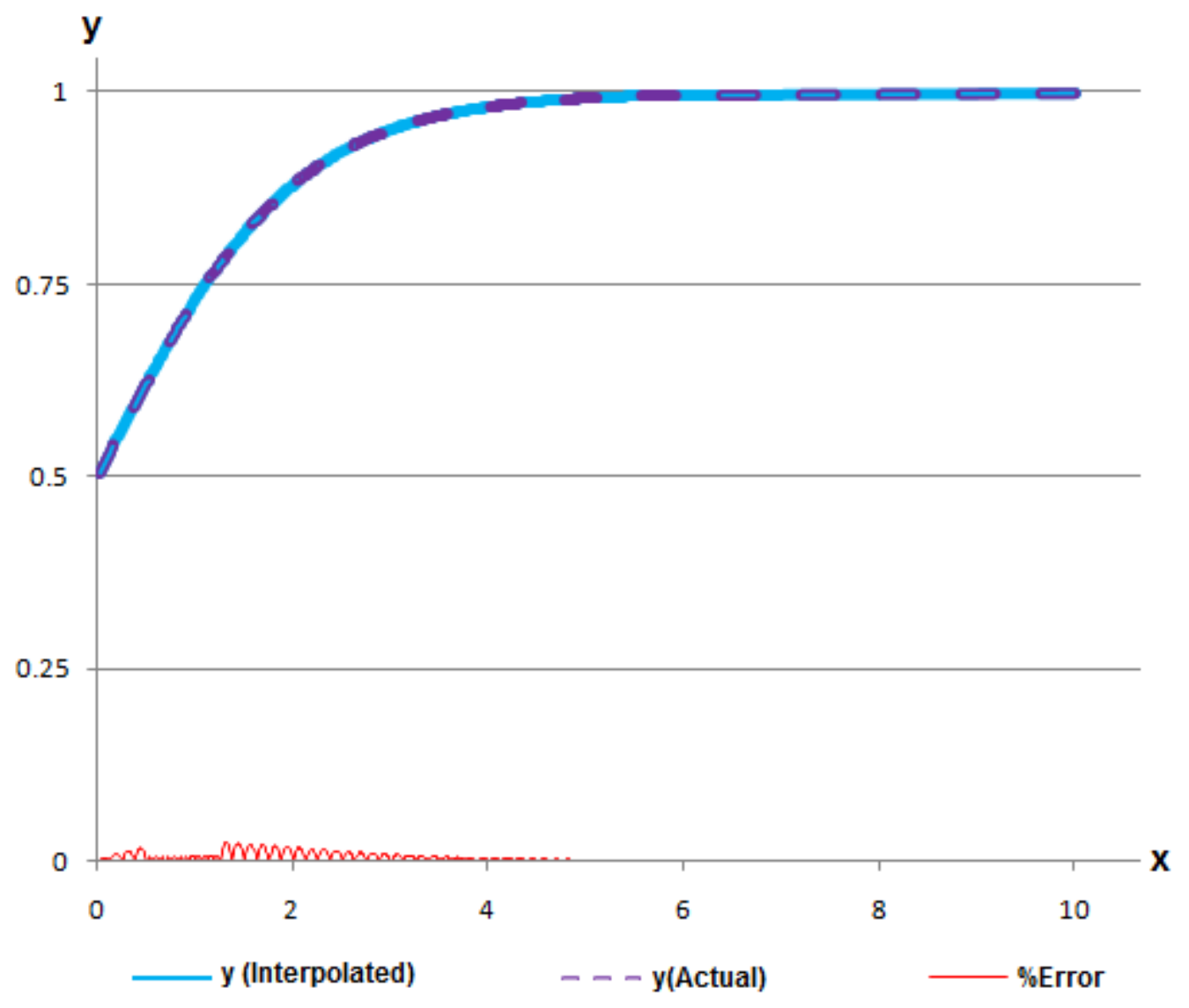

**Figure 5.11 PWL approximation of sigmoid at non-uniform spacing of 0.125, 0.03125**

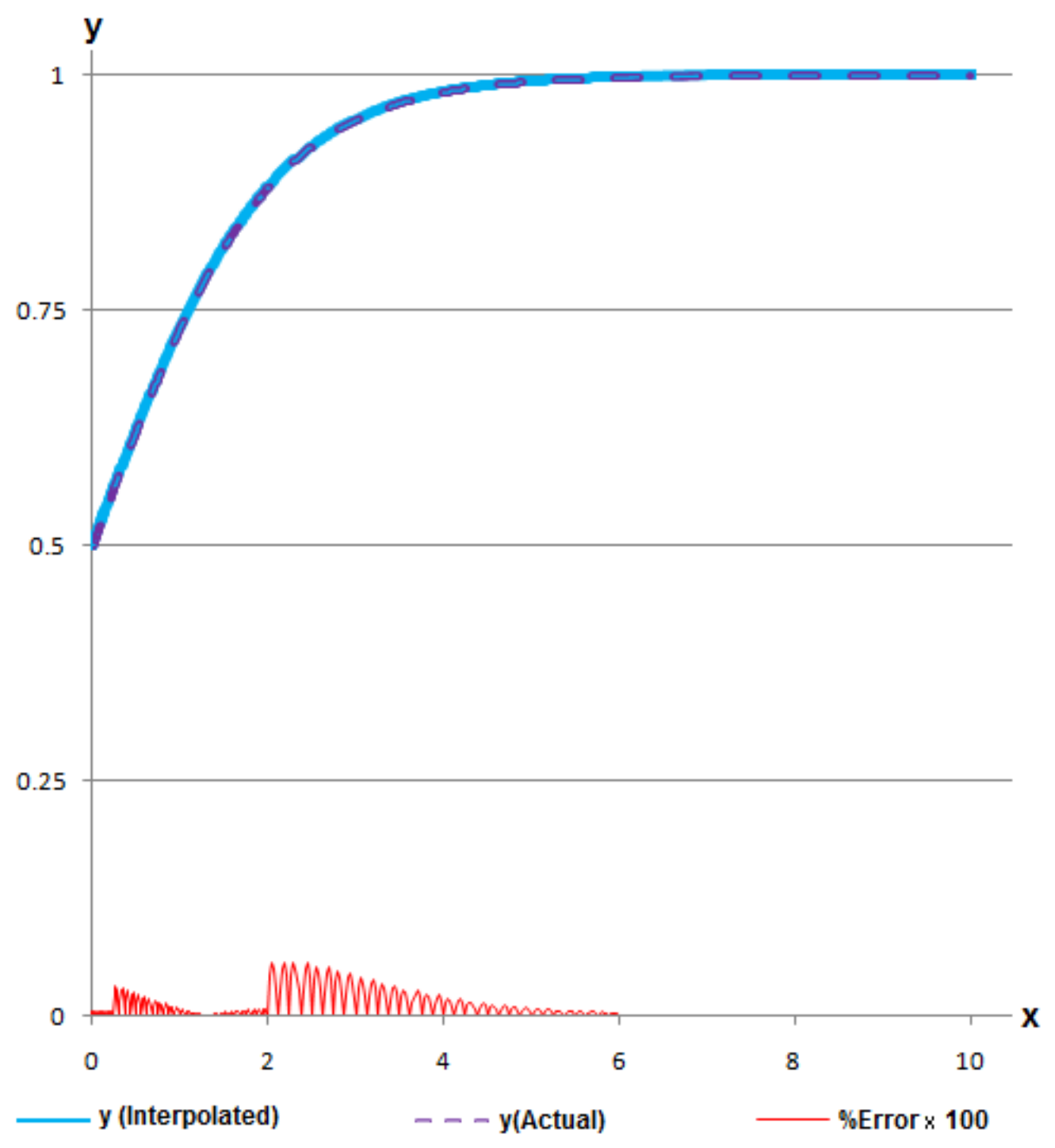

**Figure 5.12 SOI approximation of sigmoid at non-uniform spacing of 0.125, 0.03125**

## 5.5 Implementation of Activation functions (Sigmoid)

Implementing a sigmoid or any other similar non-linear function using approximation methods involves sampling the nonlinear curve at specific inputs and store the values in a $(x, y)$ table where $x$ is the input variable and $y$ is the computed value on the curve to be approximated. In case of PWL method values of x, y, m will be stored in LUT, while for SOI method a, b, c values will be stored. Therefore three registers per row are sufficient. The input (x) is compared with the stored table and the corresponding values from LUT are fetched. These values are applied to interpolator to compute the output of a function. Note that spacing of sample points need not be uniform, as long as the underlying curve is differentiable.

The block diagram of the proposed approximation module is shown in Figure 5.13 (a). The module uses an interpolator with a capability to perform PWL and SOI approximations. The interpolator block is shown in Figure 5.13 (b) and the FSM of implementation (edge triggered) is shown in Figure 5.13 (c).

On the *load* command, input value is loaded from the bus and stored in an internal register *xReg*. On the start command, content of this register is compared with the $x$ value of LUT. The values of a LUT are fetched and stored in registers $R_1$, $R_2$, $R_3$.

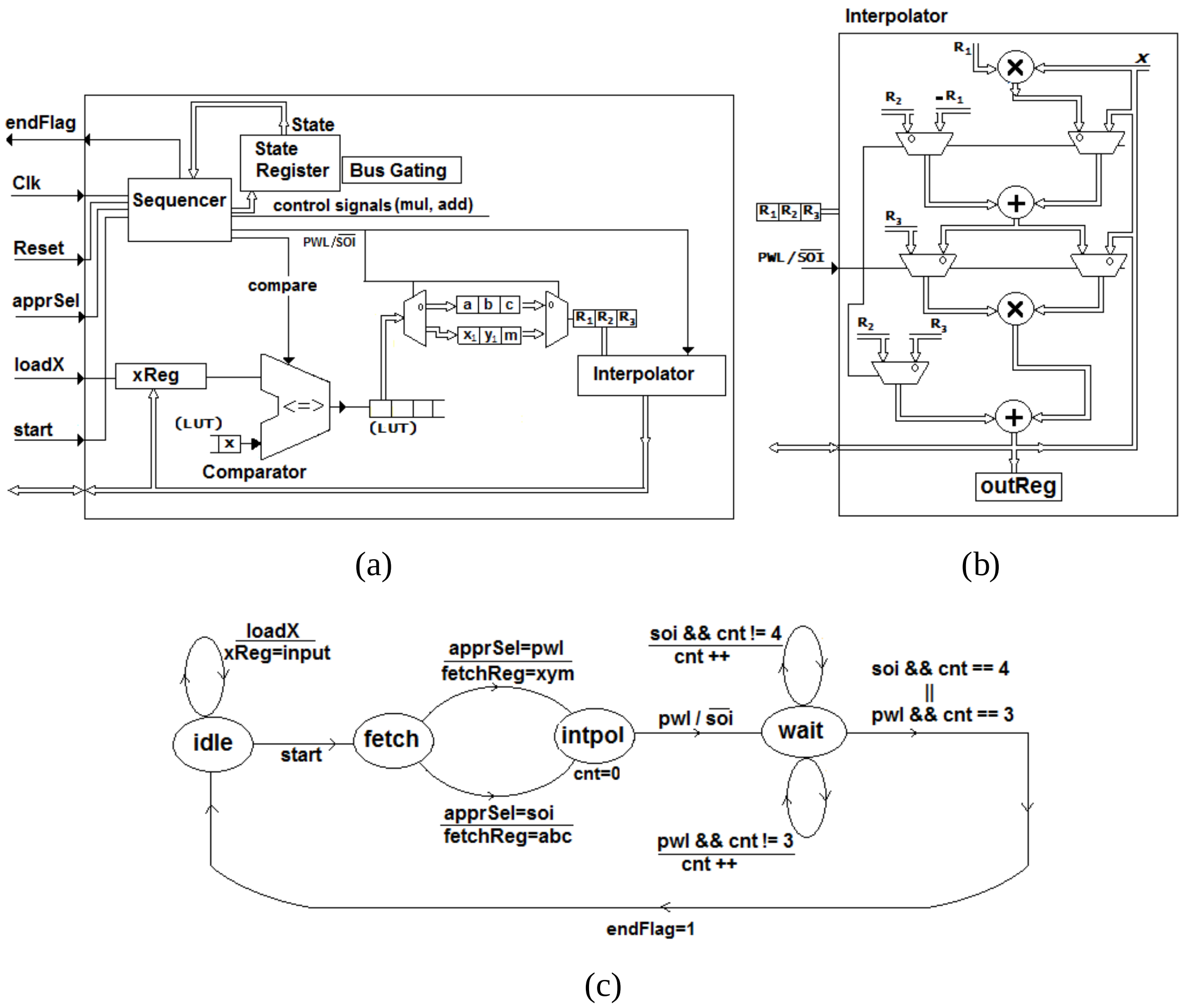

**Figure 5.13 a) Block diagram of the approximator b) Interpolator block for the approximator c) Finite State Machine of approximator**

Interpolator performs following actions depending on the method of approximation chosen by the approximator (which in turn has to get it from resonator or other higher module):

a) If the selected method is PWL, then the output register of a LUT will contain values of m, $x_1$, $y_1$. These register values are used by interpolator block to perform two additions and one multiplication as required by eqn. (5.2).

b) If the selected method is SOI, then the output register of a LUT will contain the values a, b and c. These are further applied to interpolator to perform two multiplications and two additions as required by equation 5.3. The block diagram of interpolator is shown in Figure 5.9 (b). This interpolator is designed to calculate the output of a function using either of eqn. (5.2) and eqn. (5.3)

c) After completion, the result is written to the bus and the *endFlag* signal is set to 1.

The approximator is responsible for pushing correct LUT entries to interpolator and storing the result till calling module clears it. Similar method can be used for other functions such as *tanh(x)* and resonator, except that LUT values need to be changed in correspondence with the function being implemented. Therefore, the concepts can be easily extended to other activation functions (like *tanh(x)*) that are functions of a single variable. The simulation results of sigmoid implementation using PWL and SOI method are shown in Figure 5.14 and Figure 5.15 respectively.

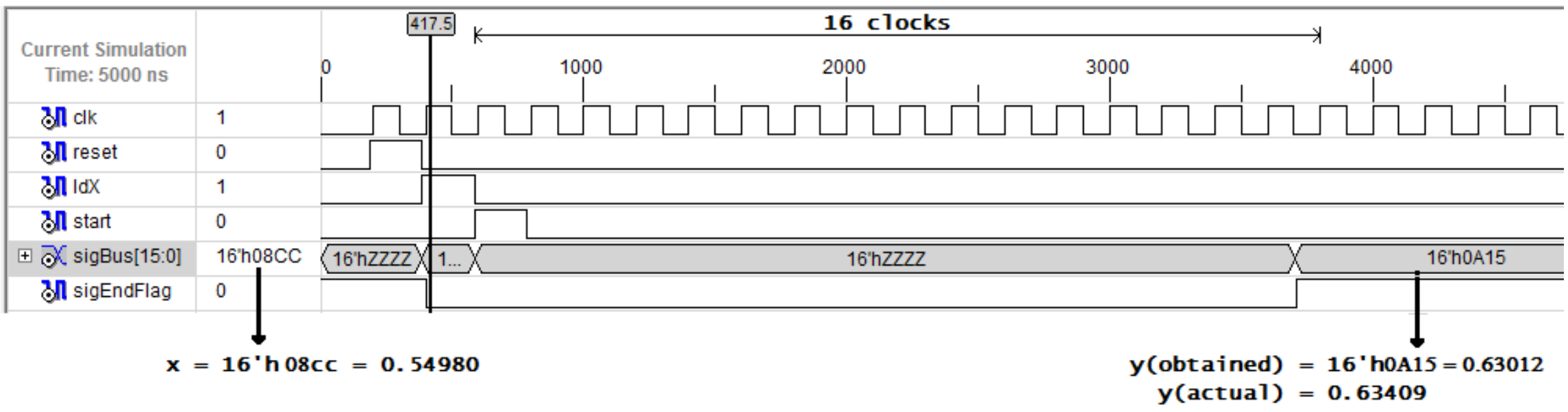

**Figure 5.14 Simulation result of sigmoid using PWL: non-uniform spacing**

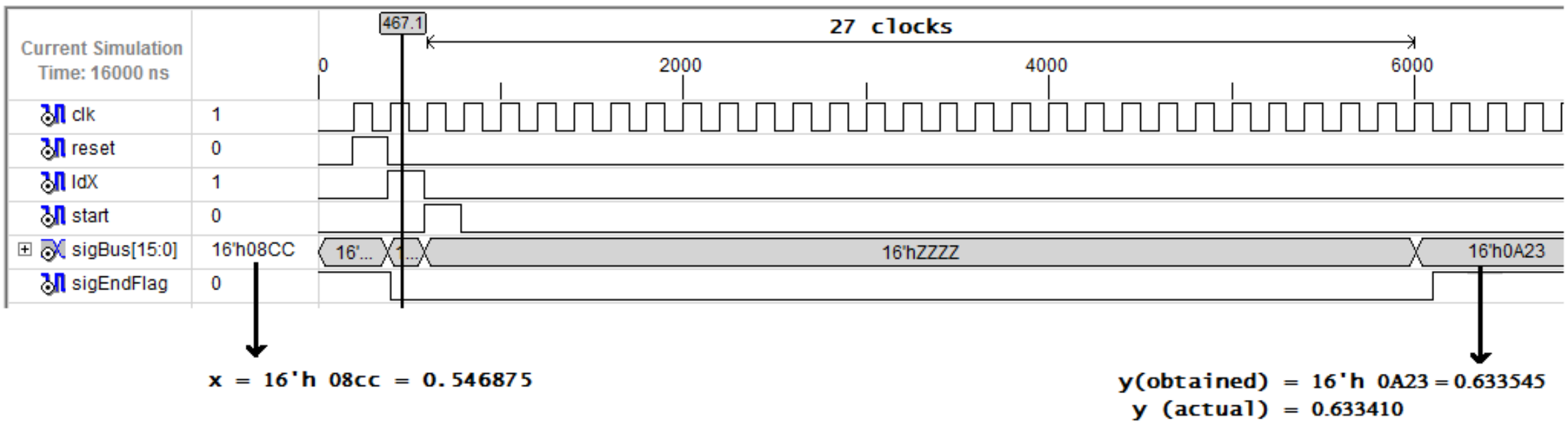

**Figure 5.15 Simulation result of sigmoid using SOI: non-uniform spacing**

### 5.5.1 Errors due to 16-bit number format

The number format shown in Figure 5.2 has 12-bits to represent the fraction part of fixed point number. Due to this restriction there will be small amount of numerical errors in the calculated result. The following tables (Table 5.4 and Table 5.5) summarize the errors due to number format for both approximation methods.

It is important to note that the computational accuracy can be controlled by selecting the interval. Decreasing the spacing improves accuracy but also increases the size of LUT, providing a trade-off between the two. However, using non-uniform spacing of points, this problem can be overcome to a certain extent.

**Table 5.4 Analysis of error due to number format for PWL method**

| x | $y_a$ (theoretical) | $y_i$ (interpolated) | $y_o$ (number format) | $Error_i$% | $Error_o$% |
|---|---|---|---|---|---|
| 0.0125 | 0.503124959 | 0.503131838 | 0.503119714 | 0.001367 | 0.001443 |
| 0.025 | 0.506249674 | 0.506261254 | 0.506233052 | 0.002287 | 0.003284 |
| 0.0375 | 0.509373902 | 0.509388248 | 0.509346058 | 0.002816 | 0.005466 |
| 0.05 | 0.512497396 | 0.512512819 | 0.512456695 | 0.003009 | 0.007942 |
| 0.0625 | 0.515619916 | 0.515634967 | 0.515625954 | 0.002919 | 0.001171 |

**Table 5.5. Analysis of error due to number format for SOI method**

| x | $y_a$ (theoretical) | $y_i$ (interpolated) | $y_o$ (number format) | $Error_i$% | $Error_o$% |
|---|---|---|---|---|---|
| 0.0125 | 0.503124959 | 0.503120937 | 0.503088474 | 0.000799 | 0.007252 |
| 0.025 | 0.506249674 | 0.506241875 | 0.506176948 | 0.001541 | 0.014366 |
| 0.0375 | 0.509373902 | 0.509362812 | 0.509265423 | 0.002177 | 0.021297 |
| 0.05 | 0.512497396 | 0.512483749 | 0.512353875 | 0.002663 | 0.028004 |
| 0.0625 | 0.515619916 | 0.515604687 | 0.515502930 | 0.002954 | 0.022688 |

The other aspect of calculation of the activation function is the speed with which computations can be carried out. Table 5.6 shows that the number of mathematical operations reduces considerably for approximation methods over direct computation using Taylor expansion. Therefore, the time of computation and area required are considerably reduced.

**Table 5.6 Computational complexity of direct implementation over approximation methods**

| Method | Operations | | | | | Max. No. of clock cycles |
|---|---|---|---|---|---|---|
| | Fetch | Add | Sub | Mul | Div | |
| Taylor Series Expansion (for accuracy up to 3 fractional digits) | 0 | 6 | 6 | 65 | 10 | 1212 (apprx.) |
| PWL | 3 | 1 | 1 | 1 | 0 | 17 |
| SOI | 3 | 2 | 0 | 2 | 0 | 35 |

## 5.6 ARN Resonator

Resonators in an Auto Resonance Networks (ARN) are described in chapter 4. In this section, we will demonstrate the fast implementation of resonators using the approximation methods.

While the sigmoid is a single curve, tunable nodes in ARN require a family of curves to be implemented (see Figure 4.6). However, tuning of ARN nodes depends on the input, threshold and a resonance control parameter ($\rho$), requiring a family of curves to be approximated. The equation of a resonator is given in eqn. (4.7) and the resonance curves of ARN nodes are shown in Figure 4.1(d). Among several types of resonators that can be used in ARN, the scaled shifted sigmoid envelop function shown in Figure 4.1(d) was used for implementation as it is found to be best in terms of controllability.

Each resonating curve shown in Figure 4.6 will have its own LUT. For example each row in the LUT of SOI method will have the values of input (x) and the corresponding values of coefficients (a, b and c in case of SOI and m, $x_1$, $y_1$ in case of PWL). Output of a resonator is calculated using eqn. (5.2) or eqn. (5.3) depending on the approximation method selected. The block diagram and FSM for implementing resonator is shown in Figure 5.16. The method of implementation is similar to that of approximator, except that the LUT must be selected for the given value of rho. These LUT values and the input x are applied to approximator to perform the computation.

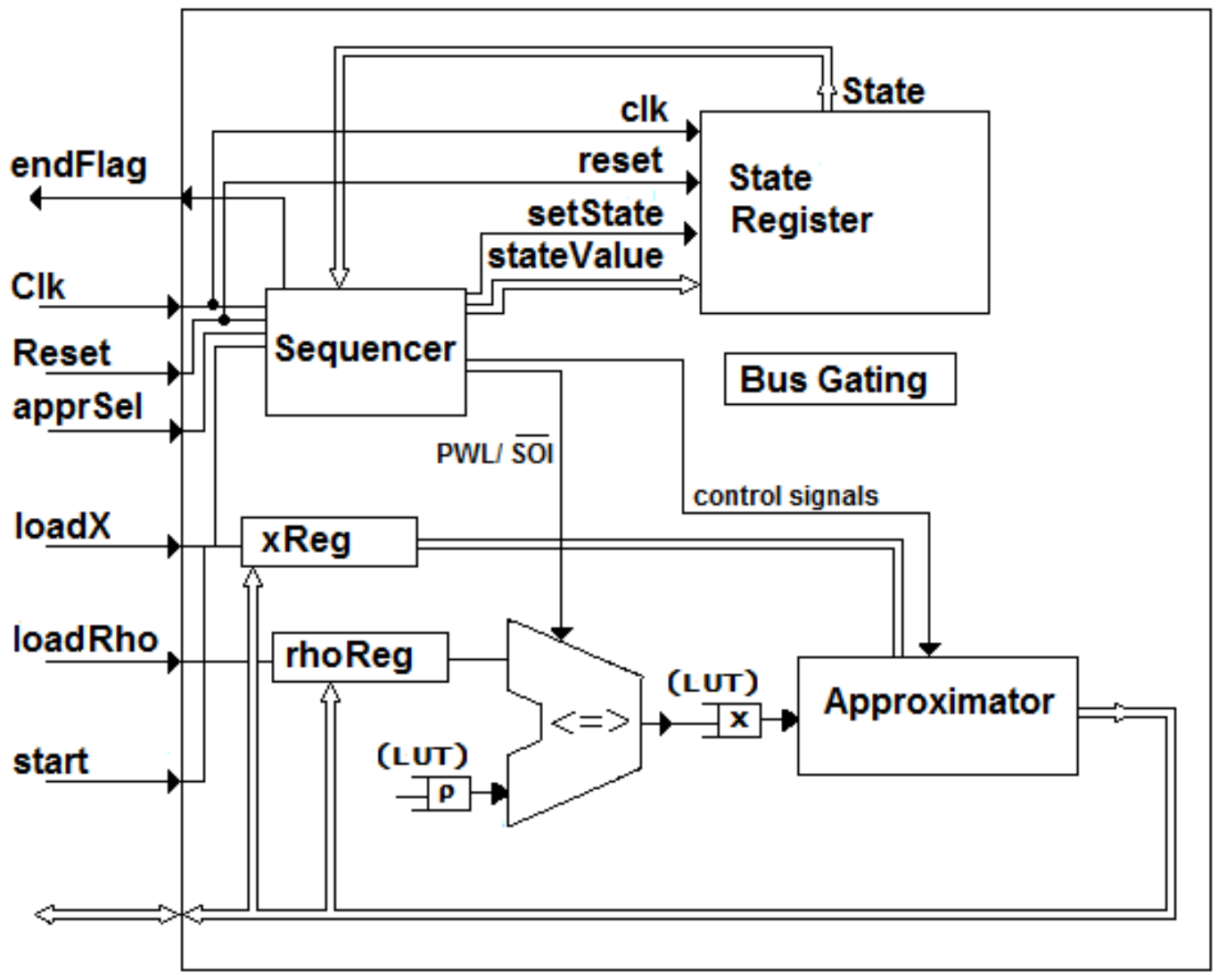

(a)

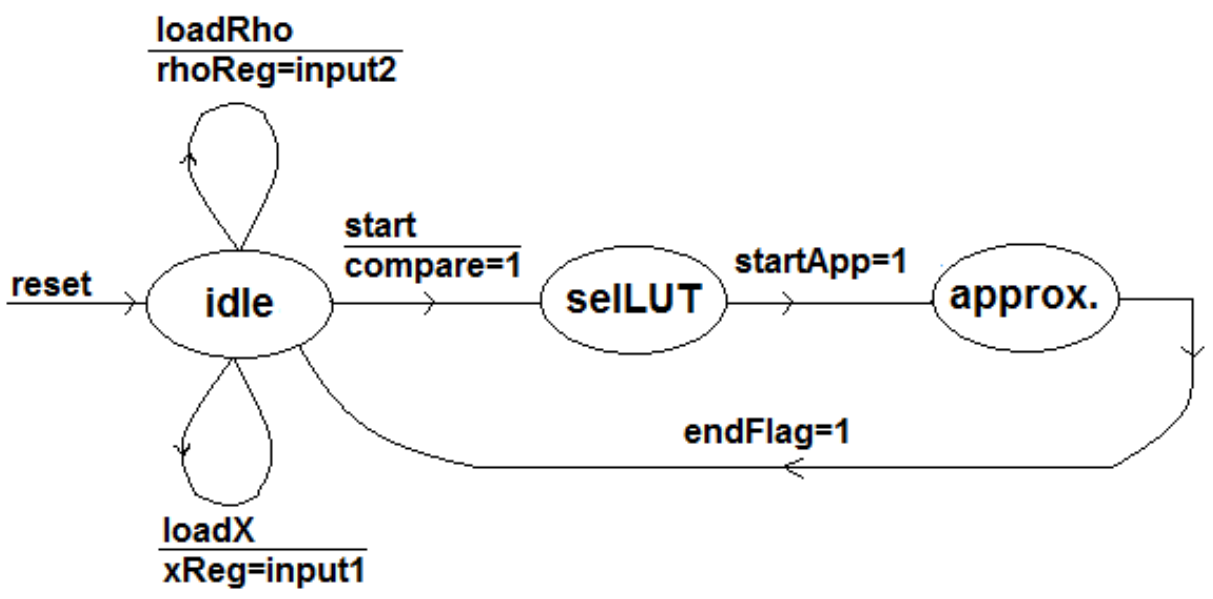

(b)

**Figure 5.16 Resonator (a) Block Diagram and (b) Finite State Machine**

The area and performance comparison of resonator implementation using PWL and SOI methods considering non-uniform spacing are summarized in Table 5.7. Simulation results of resonator for ρ = 2.42 using PWL and SOI are shown in Figure 5.17 and Figure 5.18 respectively.

**Table 5.7 Area and Performance of resonator implementation using approximation methods**

| Method (non-uniform) | Storage register (2 bytes each) | Number of LUT values | Bytes | Total bytes | Clock cycles (max.) | Max Error % |
|---|---|---|---|---|---|---|
| **SOI** | Input x | 11 | 2 | 22 | 35 | 0.3478 |
| | a (For 4 sets of $\rho$) | 10*4 | 2 | 80 | | |
| | b (For 4 sets of $\rho$) | 10*4 | 2 | 80 | | |
| | c (For 4 sets of $\rho$) | 10*4 | 2 | 80 | | |
| | **Total** | **262 bytes of storage** | | | | |
| **PWL** | Input x | 11 | 2 | 22 | 18 | 0.0314 |
| | m (for 4 sets of $\rho$) | 10*4 | 2 | 80 | | |
| | y (for 4 sets of $\rho$) | 10*4 | 2 | 80 | | |
| | **Total** | **182 bytes of storage** | | | | |

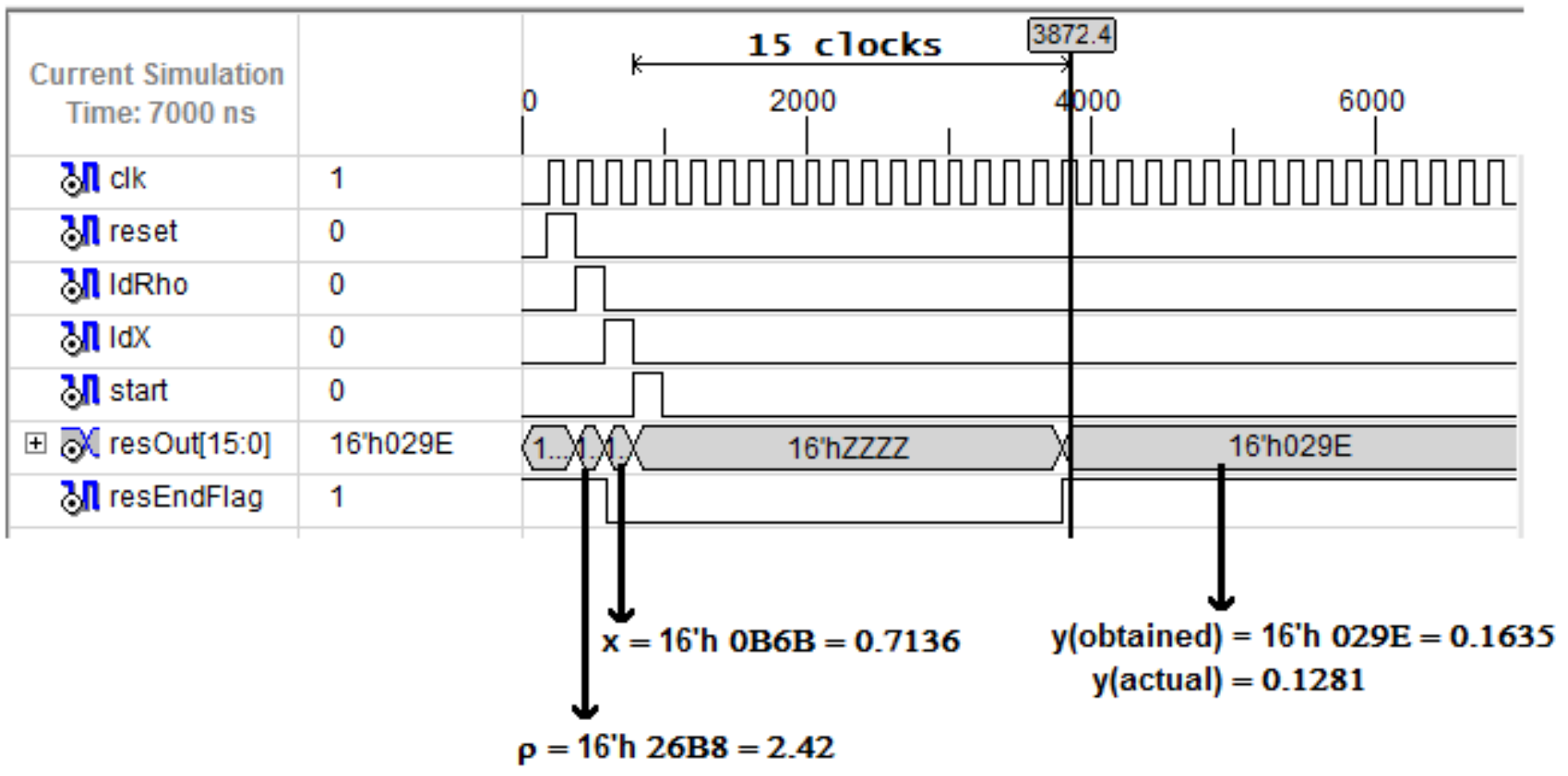

**Figure 5.17 Simulation result of Resonator using PWL method for rho=2.42**

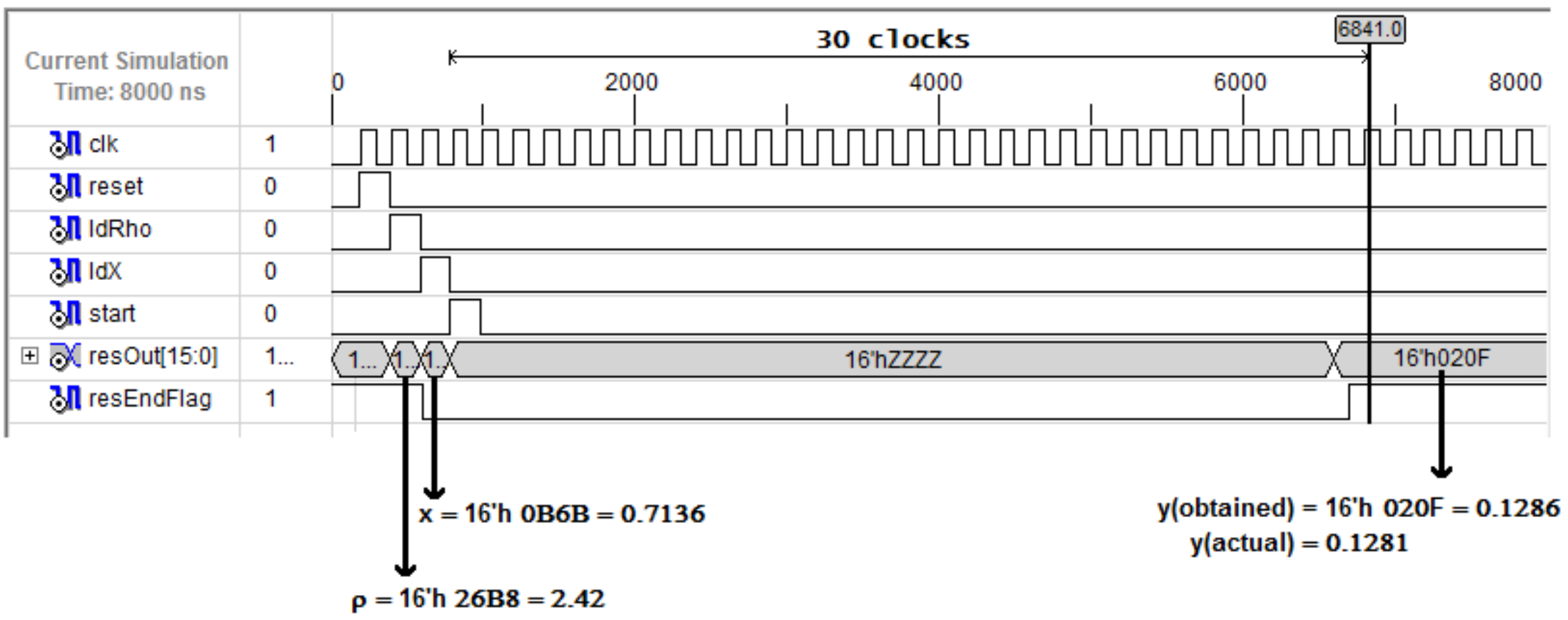

**Figure 5.18 Simulation result of Resonator using SOI method for rho=2.42**

## Summary

Analysis of serial vs. parallel computations in massively parallel environment was conducted. Justification for use of serial multipliers instead of parallel multipliers has been presented as a lemma. To test the lemma, both serial and parallel multipliers were implemented but serial multipliers showed better performance as expected and hence selected for implementation. Software implementation of ARN presented in chapter 4 had shown that low precision can be beneficial for ARN implementations. Therefore, a new low precision 16-bit fixed point number format was defined and used in all hardware implementations. Computing activation functions requires calculation of exponents and hence their implementations can be slow. Hence approximation methods with low error rate were implemented. Approximation methods to implement non-linear activation functions have been discussed in this chapter with sigmoid as an illustration. These methods are further used to implement *tanh* and ARN resonator. The details of several other units used to implement the resonator and the parallel multiplier used to compare the performance are given in Appendix-2.

# Chapter 6

# Multi-Operand Adder for Neural Processing

In the previous chapter, implementation of resonator was discussed. Output from several resonators needs to be added to implement complete ARN nodes. This chapter describes a new approach to build multi-operand adders for these structures.

Human neurons are very densely connected, with as many as a hundred thousand synapses on a single neuron. There may be around 45 thousand synaptic connections per neuron in mice, another prolific mammal species (see Figure 6.1). Present day DNNs do not come any close to these numbers but are limited only by the hardware capabilities of present day accelerators. There is a need to review arithmetic computations to be implemented. There is also a need to evolve inter-neuronal connection models.

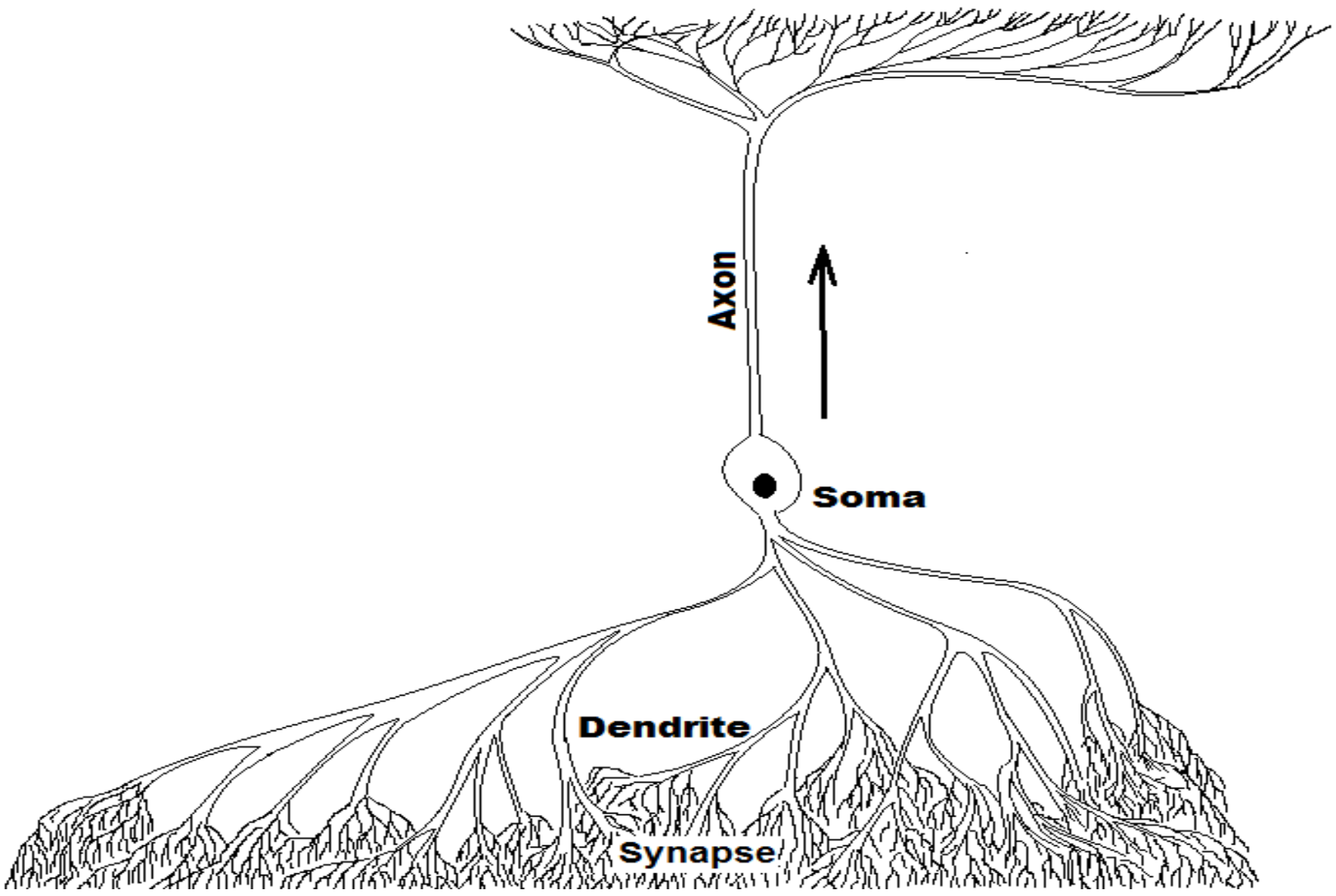

**Figure 6.1 Densely connected neuron with very large number of synapse**

Many modern DNNs use large number of inputs. For example, MNIST image recognition using ARN as described in chapter 4 has 49 inputs in the first layer and 16 inputs in the second layer. Alexnet has 363 inputs in the first layer [2017 Sabour]. Therefore, the number of operands in a network will vary depending on the requirements of the end application. In several models implementing DNNs inputs are scaled by a weight factor and then summed to compute the output, e.g. Multi-Layer Perceptron (MLP). Such operations can be combined in to a *Streamed Multiply and Accumulate* (SMA) instruction as shown in Figure 6.2(a) or implemented in hardware as a systolic array using cells for each operation, gradually combining the results. An arrangement similar to the SMA for ARN using a single resonator repeatedly over all the inputs is also possible, as shown in Figure 6.2(b). However, transfer of resonator parameters will be an overhead as resonator uses several node specific parameters.

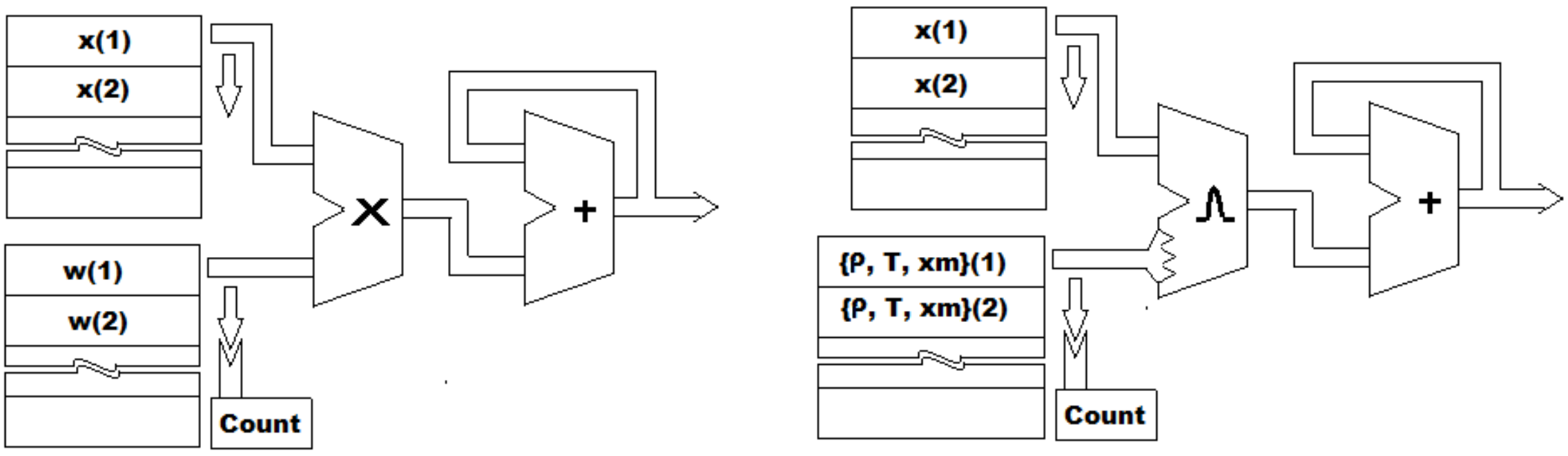

(a) (b)

**Figure 6.2 Implementing a Streamed Instruction**
**(a) MLP neuron (b) ARN neuron**

A hardware intensive design with a set of resonators, each with its own small internal memory enough to store several node configurations at a time could be faster. These units can communicate over a network of internal serial buses with a message passing architecture, as shown in Figure 6.3. Output of several resonators may be added all at a time using a multi-operand adder.

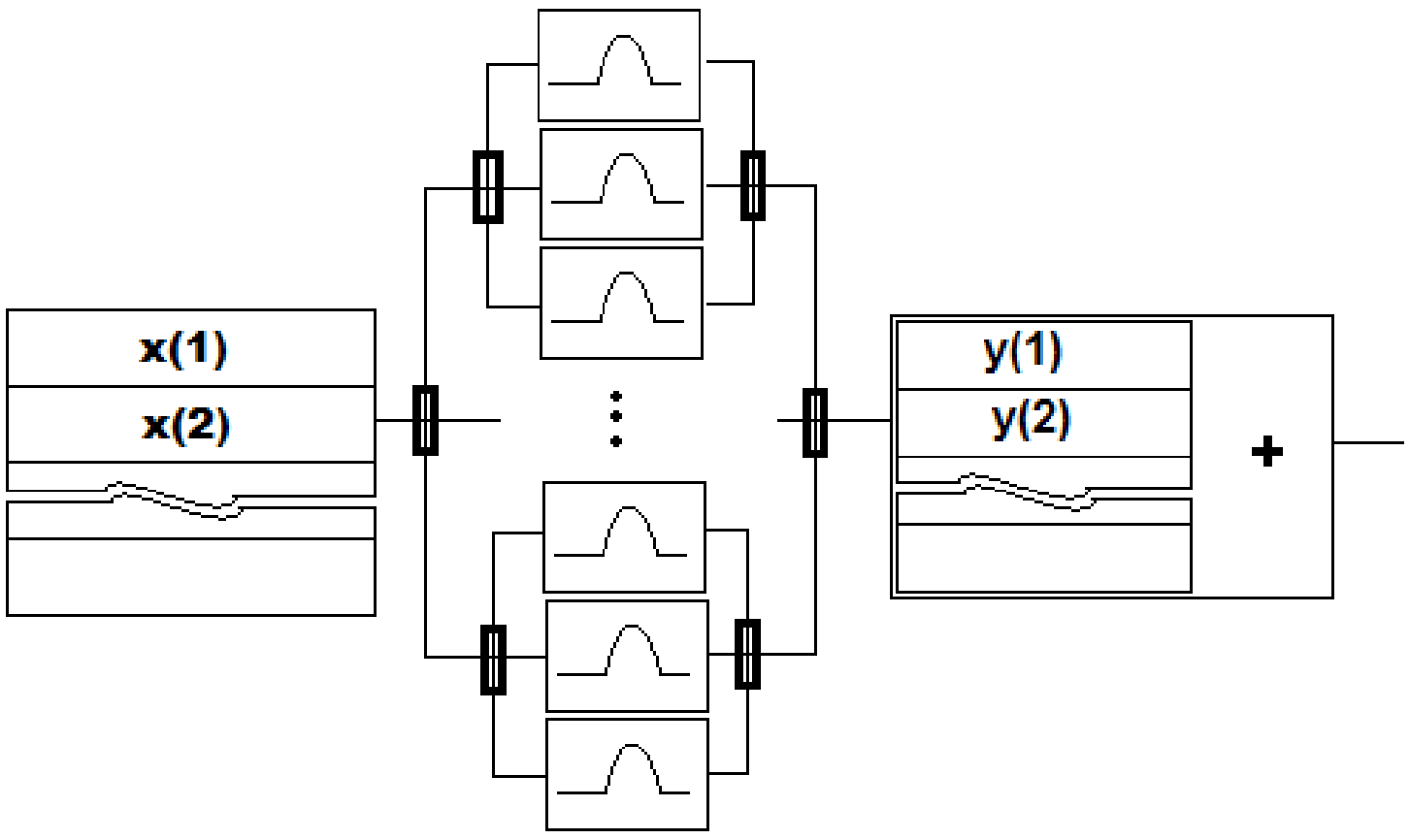

**Figure 6.3 Implementing structured ARN nodes with multi-operand adder**

In this chapter, we present an implementation of a reconfigurable multi-operand adder, necessary to meet the application specific needs. It is possible to pipeline modules to achieve a high throughput adder. A method to estimate the size of carry in such adder is described. Interestingly, an upper bound on the size of carry depends only on the number of operands and is independent of the size (bit-width) of the number and the number base used to compute the sum. The theory has been extended to obtain the condition for change in the number of columns required to represent the carry and when an overflow or carry column transition occurs. These results have been used to design a modular 4-operand adder. A fast implementation of 4-operand parallel adder is presented. Procedure to reconfigure set of 4-operand adders to implement larger adders is described. As an example, a 16-operand adder is built using 4-operand adders.

## 6.1 Theoretical construction and derivations

Some of the issues related to the design of multi-operand arithmetic units are selection of number of operands, modularity in design, computation of carry, area optimization, etc. Systolic arrays for performing multiple operations over several steps have been described in literature [1981 Kung]. However, not enough attention has been given to multi-operand arithmetic in modern processor designs. One of the possible problems is lack of theoretical work in digital implementation of multi-operand arithmetic. This section will present some theoretical work done during this research.

In a multi-operand addition, carry will be more than a single bit. Using a large number of carry bits makes it harder to know if the computation produces any overflow. Each node in a DNN may have varying number of inputs and therefore the number of operands is not always fixed. If computations are done with a fixed number of operands, some of these operands will not contribute to the result but drain the power and can introduce delays. Therefore a capability to dynamically set the number of operands will be required. As the number of operands increases, computation of carry bits becomes harder. To alleviate these problems, an algorithm to compute multi-operand addition is required. A prerequisite for such an algorithm would be an estimate of the upper bound on the carry for arbitrary number of operands. For number system in base $b$, size of carry C increases as the number of operands N increases but does not increase at the integer boundary of $\log_b \mathrm{N}$, $\mathrm{N} > 1$ as expected but only for somewhat higher number of operands. Therefore, it is also necessary to estimate the number of bits required to represent the carry of the final sum, which represents a much tighter bound. It is interesting to know when the size of carry increases by a column.

We describe some theoretical background required for implementing a multi-operand addition. The results are used to develop a new type of adder for multi-operand addition. The adder has been simulated to verify the theory. These adders have also been used in implementation of ARN node with multiple resonators.

Following discussions can be applied to any number base (binary, decimal, octal, etc.), for any number of columns (bits, digits, octets etc.) and for any number of operands.

### 6.1.1 Basic Notations

Let N be the number of operands, M be the number of columns (bits, digits, etc.), $b$ be the number base, Z be the total sum, C be the value of a carry and S be the column sum.

The total sum Z, in relation to $b$, C and S can be expressed as, $\mathrm{Z} = b\mathrm{C} + \mathrm{S}$, where $0 \leq \mathrm{S} < b$. S may also be expressed as

$$\mathrm{S} = \mathrm{Z} \bmod b \qquad (6.1)$$

or

$$b\mathrm{C} = (\mathrm{Z} - \mathrm{S}) \qquad (6.2)$$

Therefore, if $(\mathrm{Z} - \mathrm{S}) > 0$, it is a multiple of $b$. If $\mathrm{C} > (b - 1)$, C will extend to more than one column. Therefore, as a general form, C may be expressed as

$$\mathrm{C} = \sum_{\mathrm{i}=0}^{\mathrm{p}} \mathrm{c_i}\, b^{i}, \text{ where } 0 \leq \mathrm{c_i} < b \qquad (6.3)$$

In eqn. (6.1), when Z is a multiple of $b$, we get, $\mathrm{S} = 0$ and therefore, Z can be expressed completely in terms of C. Number of rows, N, is independent of $b$ but without loss of any generality, it may also be expressed as a number in base $b$, similar to C, as

$$\mathrm{N} = \sum_{\mathrm{i}=0}^{\mathrm{p}} \mathrm{n_i}\, b^{i}, \text{ where } 0 \leq \mathrm{n_i} < b \qquad (6.4)$$

First we will study a single column addition and then extend the results to multiple columns. For any base $b$, single column addition, the maximum value of an operand is $(b - 1)$.

Therefore, for N-operand addition, maximum value of total sum is $\mathrm{Z} = \mathrm{N}(b - 1)$. This is also the condition for maximum carry. In the following discussion, base of a number is assumed to be $b$ or written as $(b)$ where necessary.

***Lemma 1:*** *For two-operand 1-column addition,* $N = 2$, $M = 1$ *and for a base of* $b$, *maximum carry is* $C_{(b)} = 1$ *and column sum is* $S_{(b)} = b - 2$.

***Proof:*** Total sum when all the operands are $(b - 1)$ is

$$\mathrm{Z} = \mathrm{N}(b-1) = 2(b-1) = 2b - 2 = b + (b - 2)$$

These terms are separated such that multiples of $b$ and remainder expressed as addition. We can therefore write $\mathrm{Z} = 1b^1 + (b-2)b^0$. $\therefore \mathrm{C} = 1$ and $\mathrm{S} = (b-2)$.

Alternately, S and C may be verified using eqn. (6.1) and eqn. (6.2) as follows:

$$\mathrm{S} = \mathrm{Z} \bmod b = (-2) \bmod b = b - 2, \ \forall b \geq 2$$

$$\mathrm{bC} = \{2b - 2 - b + 2\} = b, \quad \therefore \mathrm{C} = 1 \qquad \textbf{QED}$$

This result holds for all $b$. This can be verified easily in trivial cases, e.g.,

for $b = 2$, $\mathrm{Z} = 1_{(2)} + 1_2 = 10_{(2)}$ $(\mathrm{C} = 1, \mathrm{S} = 2 - 2 = 0)$

for $b = 8$, $\mathrm{Z} = 7_{(8)} + 7_{(8)} = 16_{(8)}$ $(\mathrm{C} = 1, \mathrm{S} = 8 - 2 = 6)$

for $b = 10$, $\mathrm{Z} = 9_{(10)} + 9_{(10)} = 18_{(10)}$ $(\mathrm{C} = 1, \mathrm{S} = 10 - 2 = 8)$

for $b = 16$, $\mathrm{Z} = \mathrm{F}_{(16)} + \mathrm{F}_{(16)} = 1\mathrm{E}_{(16)}$ $(\mathrm{C} = 1, \mathrm{S} = 16 - 2 = 14 = \mathrm{E})$

As we add more rows to this basic operation, $(b - 1)$ terms are added to Z, S will be decreased by 1 and C will be increased by 1, for every additional row. Interestingly, when $N \bmod b = 0$, $\mathrm{S} = 0$, adding next row adds another $b - 1$ term, which is absorbed completely by S and therefore C remains unchanged. Therefore, we can state this as follows:

***Lemma 2:*** *For maximum carry condition, as* $N$ *increases by* $1$, $S$ *decreases by* $1$ *and* $C$ *increases by* $1$, *except when* $N = nb + 1$.

***Proof:*** For a given N rows each containing a number $(b - 1)$, let Z be the sum of N rows, which has numeric value

$$\mathrm{Z_N} = b\mathrm{C_N} + \mathrm{S_N} \qquad (6.5)$$

where the suffix $(\cdot)_\mathrm{N}$ represents the values when N rows are considered. Increasing the number of rows from N to N + 1, we get

$$\mathrm{Z}_{\mathrm{N}+1} = \mathrm{Z}_\mathrm{N} + (b-1)$$

$$= b\mathrm{C}_\mathrm{N} + \mathrm{S}_\mathrm{N} + b - 1$$

Or

$$\mathrm{Z}_{\mathrm{N}+1} = b\,(\mathrm{C}_\mathrm{N} + 1) + (\mathrm{S}_\mathrm{N} - 1) \tag{6.6}$$

$$= b\mathrm{C}_{\mathrm{N}+1} + \mathrm{S}_{\mathrm{N}+1}$$

By induction, the lemma holds for any N. When $\mathrm{Z}_\mathrm{N} = \mathrm{n}b$, we get $\mathrm{S}_\mathrm{N} = 0$. Adding one more row will not result in addition of carry C because

$$\mathrm{S}_{\mathrm{N}+1} = (\mathrm{S}_\mathrm{N} - 1) \bmod b = -1 \bmod b$$

$$= (b-1)$$

$\mathrm{S}_{\mathrm{N}+1} < b$ and therefore, additional row value will be absorbed into S and hence there is no increment to C. **QED**

Consider the following numeric examples of 1-column addition:

i) $b = 2$: $\mathrm{N} = 3$: $\mathrm{Z} = 3 = 01\,1_{(2)}$, $\mathrm{C} = 01_{(2)} = 1, \mathrm{S} = 1_{(2)}$

$\mathrm{N} = 4$: $\mathrm{Z} = 4 = 10\,0_{(2)}$, $\mathrm{C} = 10_{(2)} = 2, \mathrm{S} = 0_{(2)}$

$\mathrm{N} = 5$: $\mathrm{Z} = 5 = 10\,1_{(2)}$, $\mathrm{C} = 10_{(2)} = 2, \mathrm{S} = 1_{(2)}$

$\mathrm{N} = 6$: $\mathrm{Z} = 6 = 11\,0_{(2)}$, $\mathrm{C} = 11_{(2)} = 3, \mathrm{S} = 0_{(2)}$

ii) $b = 8$: $\mathrm{N} = 15$: $\mathrm{Z} = 105 = 15\,1_{(8)}$, $\mathrm{C} = 15_{(8)} = 13$, $\mathrm{S} = 1_{(8)}$

$\mathrm{N} = 16$: $\mathrm{Z} = 112 = 16\,0_{(8)}$, $\mathrm{C} = 16_{(8)} = 14$, $\mathrm{S} = 0_{(8)}$

$\mathrm{N} = 17$: $\mathrm{Z} = 119 = 16\,7_{(8)}$, $\mathrm{C} = 16_{(8)} = 14$, $\mathrm{S} = 7_{(8)}$

$\mathrm{N} = 18$: $\mathrm{Z} = 126 = 17\,6_{(8)}$, $\mathrm{C} = 17_{(8)} = 15$, $\mathrm{S} = 6_{(8)}$

iii) $b = 10$: $\mathrm{N} = 19$: $\mathrm{Z} = 17\,1_{(10)}$, $\mathrm{C} = 17_{(10)}$, $\mathrm{S} = 1_{(10)}$

$\mathrm{N} = 20$: $\mathrm{Z} = 18\,0_{(10)}$, $\mathrm{C} = 18_{(10)}$, $\mathrm{S} = 0_{(10)}$

$\mathrm{N} = 21$: $\mathrm{Z} = 18\,9_{(10)}$, $\mathrm{C} = 18_{(10)}$, $\mathrm{S} = 9_{(10)}$

$\mathrm{N} = 22$: $\mathrm{Z} = 19\,8_{(10)}$, $\mathrm{C} = 19_{(10)}$, $\mathrm{S} = 8_{(10)}$

iv) $b = 16$: $\mathrm{N} = 15$: $\mathrm{Z} = 225 = \mathrm{E}\,1_{(16)}$, $\mathrm{C} = \mathrm{E}_{(16)} = 14$, $\mathrm{S} = 1_{(16)}$

$\mathrm{N} = 16$: $\mathrm{Z} = 240 = \mathrm{F}\,0_{(16)}$, $\mathrm{C} = \mathrm{F}_{(16)} = 15$, $\mathrm{S} = 0_{(16)}$

$\mathrm{N} = 17$: $\mathrm{Z} = 255 = \mathrm{F}\,\mathrm{F}_{(16)}$, $\mathrm{C} = \mathrm{F}_{(16)} = 15$, $\mathrm{S} = \mathrm{F}_{(16)}$

$\mathrm{N} = 18$: $\mathrm{Z} = 270 = 10\,\mathrm{E}_{(16)}$, $\mathrm{C} = 10_{(16)} = 16$, $\mathrm{S} = \mathrm{E}_{(16)}$

As we add rows, $\mathrm{N} \bmod b$ iterates through $0$ to $b-1$ and C misses one count per iteration. As N increases, difference between C and N increases, making C smaller by 1 for every $b$ rows.

In all the cases, the carry C will not exceed $\mathrm{N}-1$. This indeed is an upper bound on the carry, and can be stated as a theorem as described below:

***Theorem***: *An upper bound on value of the carry is numerically equal to the number of operands minus one, irrespective of the number of columns or the number system used i.e., if there are $N$ operands, the upper bound on the value of carry is $N-1$.*

**Proof**: For $\mathrm{M}=1$,

$$\mathrm{S} = \mathrm{N}(b-1) \bmod b$$

$$b\mathrm{C} = \mathrm{N}(b-1) - (\mathrm{N}(b-1) \bmod b)$$

or $$b\mathrm{C} = \mathrm{N}(b-1) - (-\mathrm{N} \bmod b) \qquad (6.7)$$

Let us consider specific cases of eqn. (6.7) for N in relation to $b$:

For $\mathrm{N} < b$: $(-\mathrm{N} \bmod b) = (b-\mathrm{N})$

$$\therefore b\mathrm{C} = \mathrm{N}(b-1) - (b-\mathrm{N})$$

or $$\mathrm{C} = \mathrm{N}-1 \qquad (6.8)$$

For $\mathrm{N} = \mathrm{n}b$: $(-\mathrm{N} \bmod b) = 0$

$$\therefore b\mathrm{C} = \mathrm{n}b(b-1) - 0$$

or $$\mathrm{C} = \mathrm{N}-\mathrm{n} \qquad (6.9)$$

For $\mathrm{N} > b$: N can be expressed as

$$\mathrm{N} = \mathrm{n}b + \mathrm{r}, \quad \text{where } 0 \le \mathrm{r} < b \qquad (6.10)$$

$$(-\mathrm{N} \bmod b) = (-\mathrm{r} \bmod b) = b-\mathrm{r}$$

$$\therefore b\mathrm{C} = (\mathrm{n}b+\mathrm{r})(b-1) - (b-\mathrm{r})$$

$$b\mathrm{C} = \mathrm{n}b^2 - (\mathrm{n}-\mathrm{r}+1)b$$

or $$\mathrm{C} = \mathrm{N} - 1 - \mathrm{n} \qquad (6.11)$$

Therefore, from eqn. (6.8), eqn. (6.9) and eqn. (6.11), upper bound on C for all values of N is $N - 1$.

Once again, we will use induction to prove that the above theorem holds for any column. Let $Z_m$ be the column sum including carry. It can be expressed as

$$Z_m = N\,(b-1) + C_{m-1} \tag{6.12}$$

where $C_{m-1}$ represents the carry from previous column. Let us assume that the carry from previous column has an upper bound of $N - 1$. Substituting the value of carry from previous column, we can write

$$\begin{aligned} Z_m &= N(b-1) + N - 1 \\ &= Nb - 1 \end{aligned} \tag{6.13}$$

The column sum $S_m$, i.e., $Z_m \bmod b$ can be calculated as

$$\begin{aligned} S_m &= (Nb-1) \bmod b \\ &= -1 \bmod b \\ &= b - 1 \end{aligned} \tag{6.14}$$

The carry can now be written as

$$bC_m = Nb - 1 - (b-1)$$

$$\therefore\ C_m = N - 1 \tag{6.15}$$

Therefore, by induction, upper bound on the value of carry for any multi-operand addition is $N - 1$. **QED**

It may be noted that $N - 1$ is a generalized upper bound. It is possible to derive a tighter bound in specific cases like for $N > b,\ N = nb$, etc. As N increases, the upper bound reduces to compensate for the S term becoming zero for every $b$-th addition.

Before proceeding with these special cases, let us see some examples of multi-operand addition for different values of N and $b$ as shown in Table 6.1 and Table 6.2. Note that, $N < b$ is valid only for $b > 2$, as $N < 2$ is not an addition at all (See Table 6.1).

**Table. 6.1 An upper bound on the value of carry for single-column addition**

| Single-column addition | | | | | | Actual value of C | | Upper bound on C |
|---|---|---|---|---|---|---|---|---|
| *b* | N | M | Z | | n | $C_b$ | C | |
| | | | Carry | Sum | | | | |
| **For N < *b*** | | | | | | | | **N-1** |
| 10 | 2 | 1 | 1 | 8 | 0 | 1 | 1 | 1 |
| | 4 | 1 | 3 | 6 | 0 | 3 | 3 | 3 |
| 16 | 10 | 1 | 9 | 6 | 0 | 9 | 9 | 9 |
| | 15 | 1 | E | 1 | 0 | E | 14 | 14 |
| **For N > *b*** | | | | | | | | **N-1-n** |
| 2 | 5 | 1 | 10 | 1 | 2 | 10 | 2 | 2 |
| | 7 | 1 | 11 | 1 | 3 | 11 | 3 | 3 |
| 10 | 11 | 1 | 9 | 9 | 1 | 9 | 9 | 9 |
| | 18 | 1 | 16 | 2 | 1 | 16 | 16 | 16 |
| 16 | 20 | 1 | 12 | C | 1 | 12 | 18 | 18 |
| | 33 | 1 | 1E | F | 2 | 1E | 30 | 30 |
| **For N = n*b*** | | | | | | | | **N-n** |
| 2 | 4 | 1 | 10 | 0 | 2 | 10 | 2 | 2 |
| | 12 | 1 | 110 | 0 | 6 | 110 | 6 | 6 |
| 10 | 20 | 1 | 18 | 0 | 2 | 18 | 18 | 18 |
| | 50 | 1 | 45 | 0 | 5 | 45 | 45 | 45 |
| 16 | 16 | 1 | F | 0 | 1 | F | 15 | 15 |
| | 48 | 1 | 2D | 0 | 3 | 2D | 45 | 45 |

**Table. 6.2 An upper bound on the value of carry for multi-column addition**

| Multi-column addition | | | | | Actual value of C | | Upper bound on C |
|---|---|---|---|---|---|---|---|
| *b* | N | M | Z | | $C_b$ | C | N-1 |
| | | | Carry | Sum | | | |
| 2 | 2 | 3 | 1 | 110 | 1 | 1 | 1 |
| | 4 | 3 | 11 | 100 | 11 | 3 | 3 |
| | 7 | 3 | 110 | 001 | 110 | 6 | 6 |
| | 7 | 5 | 110 | 11001 | 110 | 6 | 6 |
| | 10 | 3 | 1000 | 110 | 1000 | 8 | 9 |
| | 64 | 3 | 111000 | 000 | 111000 | 56 | 63 |
| 10 | 2 | 3 | 1 | 998 | 1 | 1 | 1 |
| | 4 | 3 | 3 | 996 | 3 | 3 | 3 |
| | 10 | 3 | 9 | 990 | 9 | 9 | 9 |
| | 15 | 4 | 14 | 9985 | 14 | 14 | 14 |
| | 1112 | 3 | 1110 | 888 | 1110 | 1110 | 1111 |
| 16 | 2 | 3 | 1 | FFE | 1 | 1 | 1 |
| | 4 | 3 | 3 | FFC | 3 | 3 | 3 |
| | 18 | 3 | 11 | FEE | 11 | 17 | 17 |
| | 65520 | 2 | FEF0 | 10 | FEF0 | 65264 | 65519 |

### 6.1.2 Number of Carry Columns

Upper bound on C is an indication of number of columns required to perform multi-column addition. Often, the number of columns is more useful than the upper bound on carry. From eqn. (6.3) it is easy to see that the number of columns increases by one for every increase in p. The actual shift occurs when N is somewhat higher than $b^p$.

***Corollary:*** *The number of columns required to represent the carry is* $\log_b(\mathrm{N}-1)$.

This is an obvious result because the maximum value of carry is $\mathrm{N}-1$ by the above theorem.

This corollary can also be derived from an alternate observation. Result of an N-operand M-column addition can be expressed as

$$\mathrm{Z} = \mathrm{N}\,\{\,(b-1)\,b^{\mathrm{M}-1} + (b-1)b^{\mathrm{M}-2} + \cdots + (b-1)\} \qquad (6.16)$$

Or

$$\mathrm{Z} = \mathrm{N}\,(b^{\mathrm{M}}-1) \qquad (6.17)$$

The term $(b^{\mathrm{M}}-1)$ represents the largest number that can be represented using M columns. Therefore, for all $\mathrm{p} \geq 1$ and N in the range $b^{\mathrm{p}} > N \geq b^{(\mathrm{p}-1)}$, p columns are sufficient to represent the carry. This is a much higher bound than $\log_b(\mathrm{N}-1)$, but both will require the same number of columns. E.g., for $b = 10$, $\mathrm{p} = 2$, $100 > \mathrm{N} \geq 10$, $(\mathrm{N} = 10, \mathrm{Z} = 90;\ \mathrm{N} = 99, \mathrm{Z} = 891)$ two columns are sufficient to represent a carry, but the value of carry will have an upper bound of $\mathrm{N}-1$ as observed earlier.

It may be noted that the difference between the actual maximum carry and the above defined upper bound increases as N increases because of the effect discussed earlier. In the next section, a much better estimate of the bound is derived.

### 6.1.3 Column Transition

Let us assume $\mathrm{N}-1$ is the upper bound on size of carry, when N is the number of rows. Increasing the number of rows from $\mathrm{N} = b^{\mathrm{p}}$ to $\mathrm{N} = b^{\mathrm{p}} + 1$, will increase the size of carry from $(b^{\mathrm{p}}-1)$ to $b^{\mathrm{p}}$. If we note that $(b^{\mathrm{p}}-1)$ is the maximum number that can be represented by p columns, it is easy to see that $b^{\mathrm{p}}$ requires $\mathrm{p}+1$ columns. In other words, the number of columns required for carry will increase from p to $\mathrm{p}+1$ bits, i.e., by one column. For example, for $\mathrm{N} = 2^4 = 16$, upper bound on carry is $(\mathrm{N}-1) = 15$ or $1111_2$, which requires 4 bits of carry. For $\mathrm{N} = 17$, this upper bound is 16 ($10000_2$) suggesting 5 bits of carry. For any N in the range $2 \leq \mathrm{N} < b$, Z will require just one column for carry. For multi-column numbers, when $\mathrm{N} = b$, eqn. (6.17) becomes

$$\mathrm{Z} = b(b^{\mathrm{M}}-1) = (b^{\mathrm{M}+1} - b)$$

which is much less than $(b^{M+1} - 1)$ and therefore $M + 1$ columns are sufficient to represent Z (e.g., $b = 10, N = 10, M = 2, Z = 990$ , three columns are required to represent Z).

When $N = b + 1$,

$$Z = (b + 1)(b^{M} - 1)$$

$$= (b^{M+1} - 1) + (b^{M} - b)$$

The term$(b^{M} - b) > 1$ for all $M > 1$, or $Z > b^{M+1}$ and also $Z < b^{M+2}$. Therefore we will require $M + 2$ columns to represent Z, i.e., the number of columns increase by one (e.g., $b = 10$, $N = 11$, $M = 2$, $Z = 1089$, four columns are required to represent Z). Similar increment is expected at every increase in order of N, i.e., $N = b^{2}$ $b^{3}$ ... $b^{p}$ etc.

It is clear from discussions in previous section that, this increase in number of carry bits does not occur in most case immediately at $b^{p} + 1$ rows but few rows later. This is because multiplication by $b^{p}$ shifts the $(b^{M} - 1)$ by p columns, leaving room for few more rows of $(b^{M} - 1)$ terms. This gap becomes substantially wide when $p > M$.

For example, for $N = 16, M = 4, Z = 16x15 = 11110000_{2}$ allowing one more row before we need one more column, i.e., $N = 17 = 2^{4} + 1, Z = 11111111_{2}$ which is still $p + M = 8$ bit wide, which requires only 4 bits of carry. Clearly, for $N = 18$, 5 bits are needed to handle carry.

However, from eqn. (6.11), we observe that a tighter upper bound $(N - 1 - n)$ can be used, in which case, only 4 bits are sufficient. In fact, this number of carry bits will not change for next $n + 1$ rows. This difference of $(n + 1)$ grows as the number of rows crosses p. Therefore, it is possible to derive when the number of columns actually increases after $b^{p}$ boundary. Combining eqn. (6.17) and eqn. (6.4) we can write

$$Z = \left(\sum_{i=0}^{p} n_i\, b^{i}\right)(b^{M} - 1)$$

$$= n_p b^{M+p} - n_p b^{p} + \left(\sum_{i=0}^{p-1} n_i\, b^{i}\right)(b^{M} - 1) \qquad (6.18)$$

Changing N from $(b^{M+p} - 1)$ to $b^{M+p}$ increases the number of columns by 1. In eqn. (6.18) this can occur if $n_p = 1$ in $n_p b^{M+p}$ and all other terms are zero. However, existence of $-n_p b^{p}$ term in the eqn. (6.18) delays this slightly beyond $b^{p}$. For a quick check, let us assign $n_p = 1$ and rewrite eqn. (6.18) as

$$Z = b^{M+p} - b^{p} + \left(\sum_{i=0}^{p-1} n_i\, b^{i}\right)(b^{M} - 1) \qquad (6.19)$$

The transition occurs when

$$-b^{p} + \left(\sum_{i=0}^{p-1} n_i\, b^{i}\right)(b^{M} - 1) \geq 0$$

Or when

$$\left(\sum_{i=0}^{p-1} n_i\, b^{i}\right)(b^{M} - 1) \geq b^{p} \qquad (6.20)$$

This is the condition for N to be on the threshold of column transition: MSB of the carry will shift to next column to the left. A numerical example for column transition is shown in Table 6.3. For $b = 2$, $N = 15$, $p = 3$ and number of carry bits is 4, when N increases by 1, i.e., $N = 16$, we have $p = 4$ and number of carry bits is expected to be 5. However, the transition is delayed till N increases such that eqn. (6.20) is satisfied. Minimum coefficients that satisfy the equation

$(n_3 b^3 + n_2 b^2 + n_1 b^1 + n_0 b^0)(b^3 - 1) \geq b^4$ are:

$n_3 = 0, n_2 = 0, n_1 = 1, n_0 = 1$ or $0011_2 = 3_{10}$

Therefore transition occurs when N shifts by 3, i.e., $N = 16 + 3 = 19$. This can be seen easily in Table 6.3. Also notice from Table 6.3 that when $N = 16$, S=$000_{(2)}$, $(b^M - 1)$ shifts by 4 columns to the left, with S taking three zeros and C taking the remaining one zero.

**Table. 6.3 Example for Column Transition**

| b | M | $b^M$-1 | N | $N_{(b)}$ | p | $b^p$ | $Z_{(b)}$ | | $Z_{(10)}$ |
|---|---|---|---|---|---|---|---|---|---|
| | | | | | | | C | S | |
| 2 | 3 | $111_{(2)}$ | 15 | 1111 | 3 | $1000_{(2)}$ | 1101 | 001 | 105 |
| | 3 | $111_{(2)}$ | 16 | 10000 | 4 | $10000_{(2)}$ | 1110 | 000 | 112 |
| | 3 | $111_{(2)}$ | 19 | 10011 | 4 | $10000_{(2)}$ | 10000 | 101 | 133 |

## 6.2 Multi-Operand Binary Addition

### 6.2.1 Design of Look-Up Table

We will use binary numbers ($b = 2$) in the following sections. For a four operand, 1-column binary addition, the upper bound on C is 3, therefore, C has only four possible values $(00{:}11)$ and the total sum, Z has only 5 possible values $(000{:}100)$. Therefore, 2-bits are sufficient to represent the carry of 4-operand addition and 3-bits would be sufficient to represent column sum (including carry). Therefore, column sum for input data bits can be implemented as a look-up table of $N \times (1 + p)$ bits or as a hardwired 1's count logic. As N grows, size of the look-up table increases exponentially and therefore, the lookup table must be limited to a small N.

| Input | Output |
|---|---|
| 0 0 0 0 | 0 0 0 |

| Input | Output |
|---|---|
| 0 0 0 1 | 0 0 1 |
| 0 0 1 0 | |
| 0 1 0 0 | |
| 1 0 0 0 | |

| Input | Output |
|---|---|
| 0 0 1 1 | 0 1 0 |
| 0 1 1 0 | |
| 1 1 0 0 | |
| 1 0 0 1 | |
| 0 1 0 1 | |
| 1 0 1 0 | |

| Input | Output |
|---|---|
| 0 1 1 1 | 0 1 1 |
| 1 1 1 0 | |
| 1 1 0 1 | |
| 1 0 1 1 | |

| Input | Output |
|---|---|
| 1 1 1 1 | 1 0 0 |

**Figure 6.4 Structure of 4×3 LUT**

The I/O map of 4 × 3 LUT (4-bit input, 3-bit output) is shown in Figure 6.4. Output of LUT has only five of the possible eight outputs as the maximum output value from LUT is $100_{(2)}$. It is possible to build a RAM based LUT using the table given in Figure 6.4 or build an optimized combinatorial circuit (1's count logic) as shown in Figure 6.5.

The advantage of using this circuit instead of memory based LUT is that the adder can be implemented without any need for clocking. This implementation reduces overall time for computation and would consume a minimum silicon area as compared to a memory-like implementation. Longest path in Figure 6.5 has four gates. So the speed performance of the LUT is good.

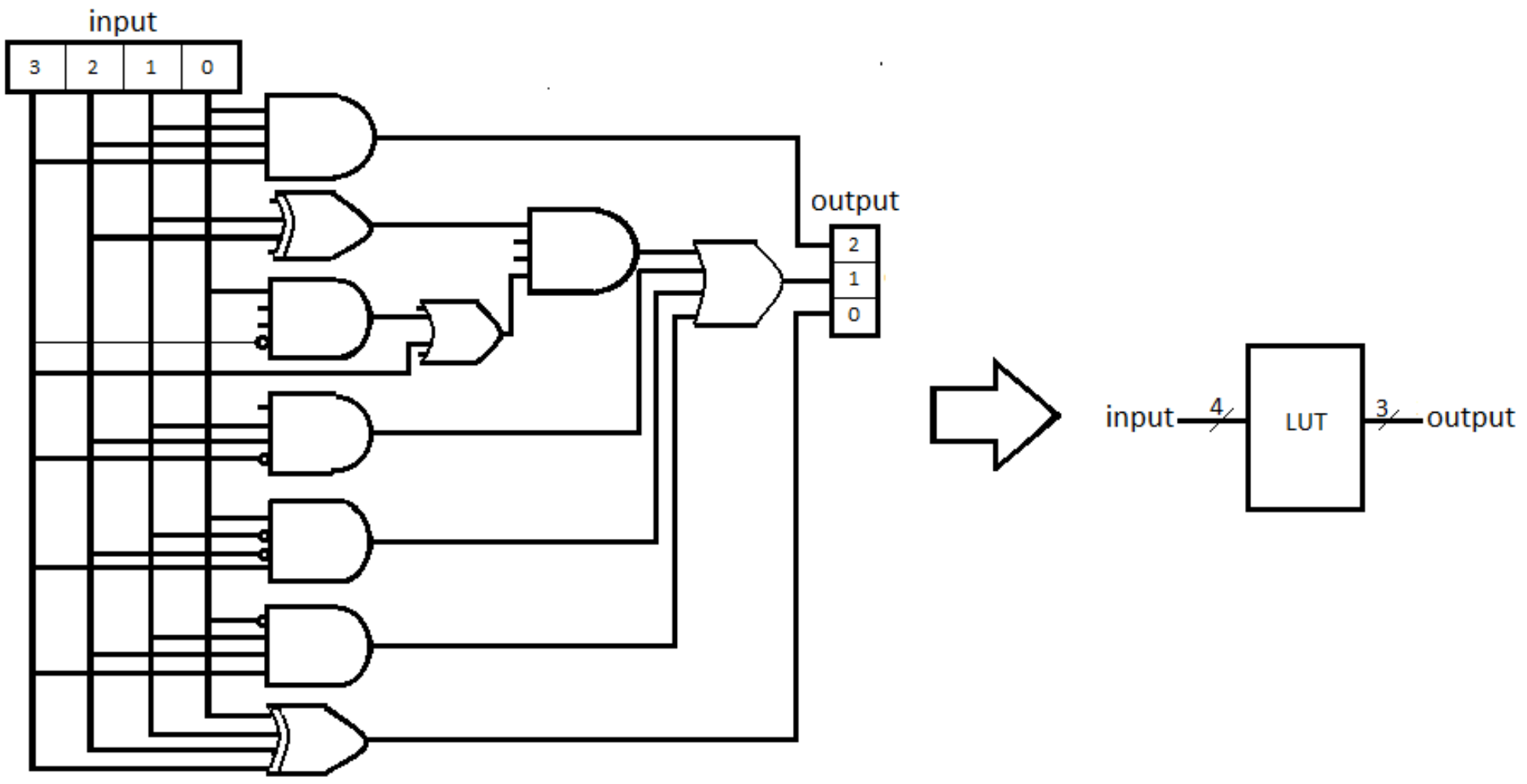

**Figure 6.5 One's count logic equivalent to 4x3 LUT**

### 6.2.2 Proposed Serial Algorithms

Following algorithms implement column wise addition propagating the carry towards left side (higher) columns as shown in Figure 6.6. Two algorithms discussed below differ in the way they handle the carry. Both are serial implementations which require multiple clocks. However, as the gate count is small, there may be a performance advantage in case of massively parallel operations. A Fast adder is discussed later in this section.

From corollary, $M + p$ bits are sufficient to compute N-operand addition for M-bit data, as in Figure 6.6(a).

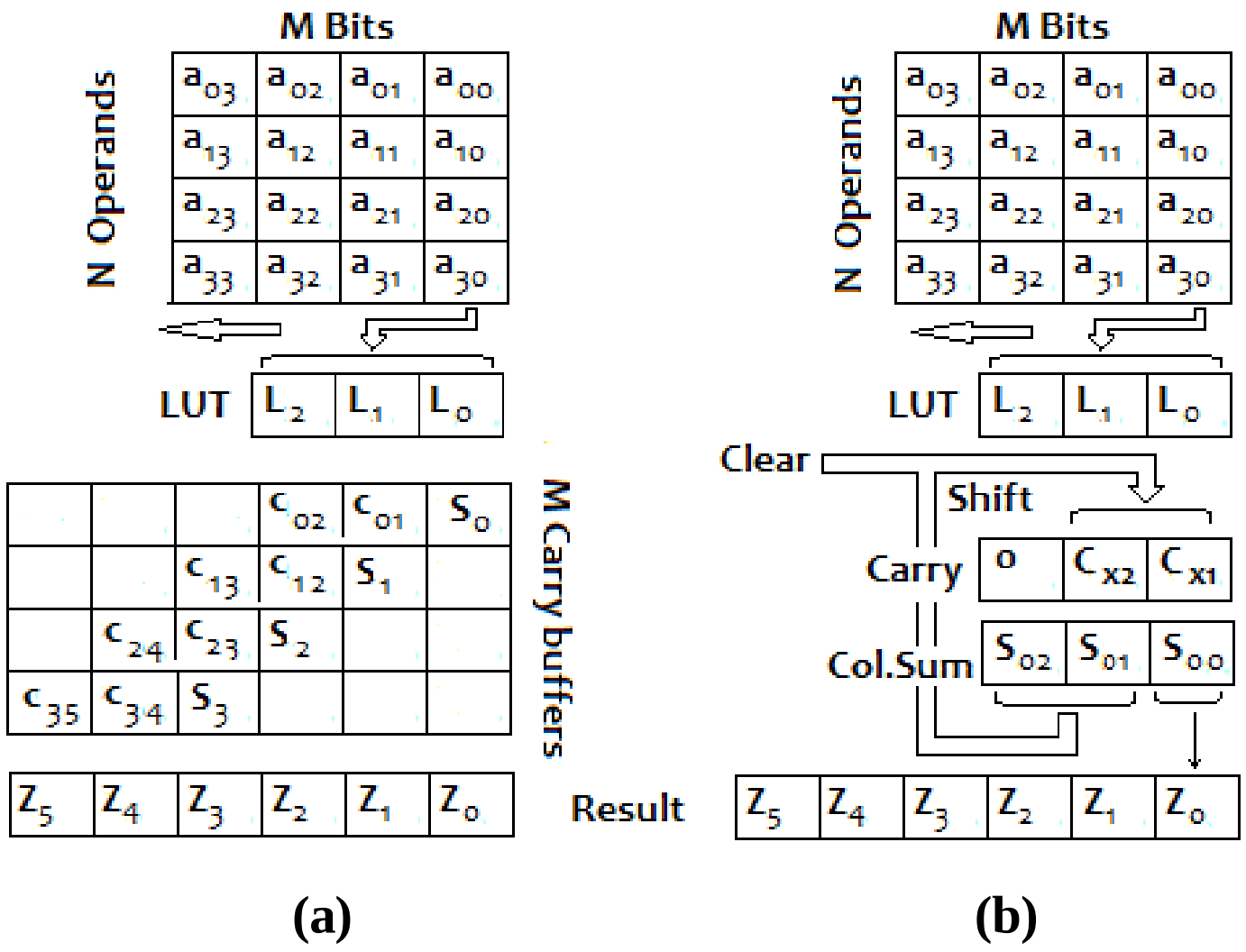

**Figure 6.6 4x4 Serial Addition using (a) algorithm-1 (b) algorithm-2**

***Algorithm-1***: This algorithm is similar to hand calculation. The algorithm is as follows:

a) Starting from LSB, each i-th column will have N data rows and $i - 1$ carry rows to be added.

b) The partial sum is copied to carry buffer, starting at column i and extending to the left.

c) Move to next column to the left and repeat till all $M + p$ columns are computed.

d) At the end of the iteration, result is available in output buffer.

It is important to note that the value of carry in every column is in the range $0 \leq c_i < b$. Figure 6.6(a) shows data flow for this algorithm. When input data is loaded, first column can be computed, generating the part of the data required for p-columns on left side. Second column can be computed now, and so on till all $M + p$ columns are computed. Column sums of input data can be computed in a single clock. However, total sum Z has to wait till the data from previous p column is computed. The algorithm can compute a column addition in one clock cycle. Hence for M-bit addition, we will require at least $M + 1$ clocks.

***Algorithm-2:*** This algorithm is similar to the first but carry generated in a column is fully added to the column on the left. The advantage here is that there is no need to store the columns of the partial sum in p carry buffers. The algorithm is as follows:

a) For every i-th column from LSB to MSB, apply N-row data in the column to LUT.
b) Output of the LUT is added with carry buffer. LSB of this addition is copied to i-th bit of Z buffer.
c) Remaining bits are shifted 1 column to the right and copied to carry buffer.
d) Increment i. Repeat the partial sum computation for all columns.

After all the columns are added, remaining bits in the carry buffer are copied to Z register. Figure 6.6 (b) shows data flow of this algorithm. This algorithm also can compute M-bit wide N operand addition in $M + 1$ clocks, but it has lower memory requirements.

### 6.2.3 Optimizing the Algorithms

Both algorithms can compute in $M + 1$clocks. Implementing large LUTs may require large area. Hierarchical implementations with several levels of LUTs also may offer some solution. However, as the number of rows increases, the path delay in calculating column sum may approach or exceed one clock interval, effectively forcing a reduction in clock frequency.

Some simple optimization steps can be incorporated into the above algorithm. For example, it is possible to keep the size of carry buffer limited to p bits instead of using $M + \log_2(N - 1)$ bits. It is also possible to merge S and Z buffers. Partial column sums can be grouped together and may be computed in parallel.

## 6.3 Design of Multi-Operand Adder

Keeping in mind the need to reconfigure the adder hardware, a modular $4 \times M$ (4 row, M columns) adder was built first. This module has been used to build a $16 \times 16$ adder. A general method to build modules with more rows or columns is also described in this section. The implementation shows a sequential adder performing column by column addition, serially. This will have a small area. It is possible to convert it to a combinatorial design where the operation may take very small number of clocks (gate delays) but at the expense of larger area used by more instances of the LUT, discussed in section 6.2.1. As indicated in chapter 5, serial adders may have an advantage in a massively parallel environment.

### 6.3.1 4-operand, M-bit (4×M) Serial Adder

Figure 6.7 shows block diagram of a $4 \times M$ serial adder, implemented using Algorithm-2 described above.

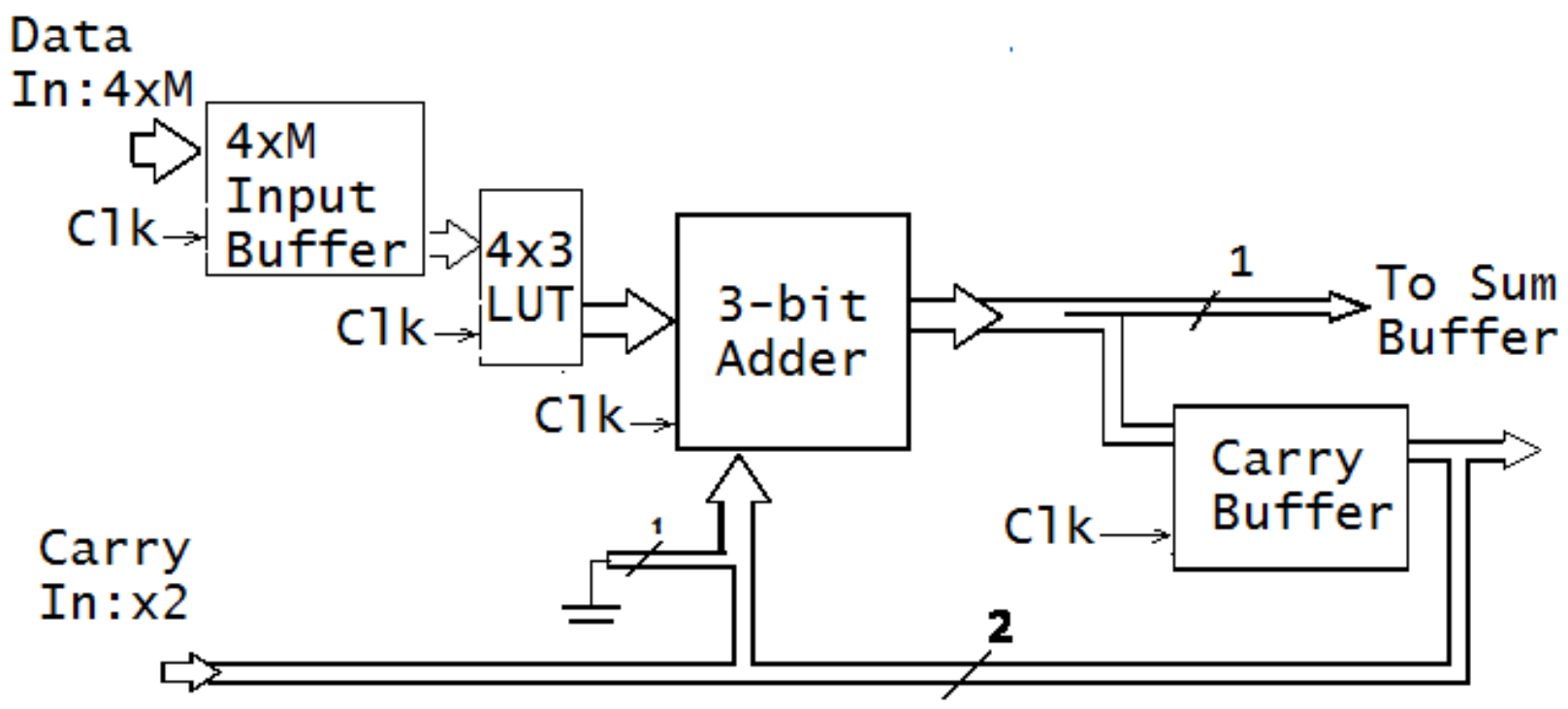

**Figure 6.7 Implementation of 4×4 serial adder using algorithm-2**

From the above theorem, maximum carry for 4 operand addition is 3, i.e., $11_{(2)}$. We prefix it with a zero to make it 3 bit wide or $011_{(2)}$. Maximum output from the 3-bit adder is therefore $100_{(2)} + 011_{(2)} = 111_{(2)}$. Therefore, there is no overflow beyond two bits. Working of the adder shown in Figure 6.7 is given below:

i) Load input buffer with data to be added and clear carry buffer.

ii) For column $i$ from $0$ to $M-1$, apply column as input to LUT and get the 3 bit output. Add it with contents of carry buffer. Shift LSB to $i$-th bit of output buffer. Shift higher 2 bits to right and copy to carry buffer.

iii) increment $i$ and repeat ii).

Step iii) can be integrated into step ii) to reduce a clock, by transferring all three bits of 3-bit adder.

Implementation of 3-bit adder in Figure 6.7 requires some attention. As it is a 2-operand addition, a 3-bit full adder may be used. However, knowing that C has only four possible values $(00{:}\,11)$ and L (output of a LUT) has only 5 possible values $(000{:}\,100)$, only 20 out of 64 possible values will be applied as input to 3-bit adder. Therefore, full 3-bit addition will not be required any time.

### 6.3.2 4-operand, 4-bit (4×4) Parallel Adder

A fast parallel implementation of 4x4 adder is given in Figure 6.8. It uses multiple copies of LUT given in Figure 6.5. Longest path has 4 LUTs and hence a delay of 16 gates. However, modularity of the design supports pipelining of the adder, allowing much faster performance. The simulation result of this fast adder is given in section 6.5.

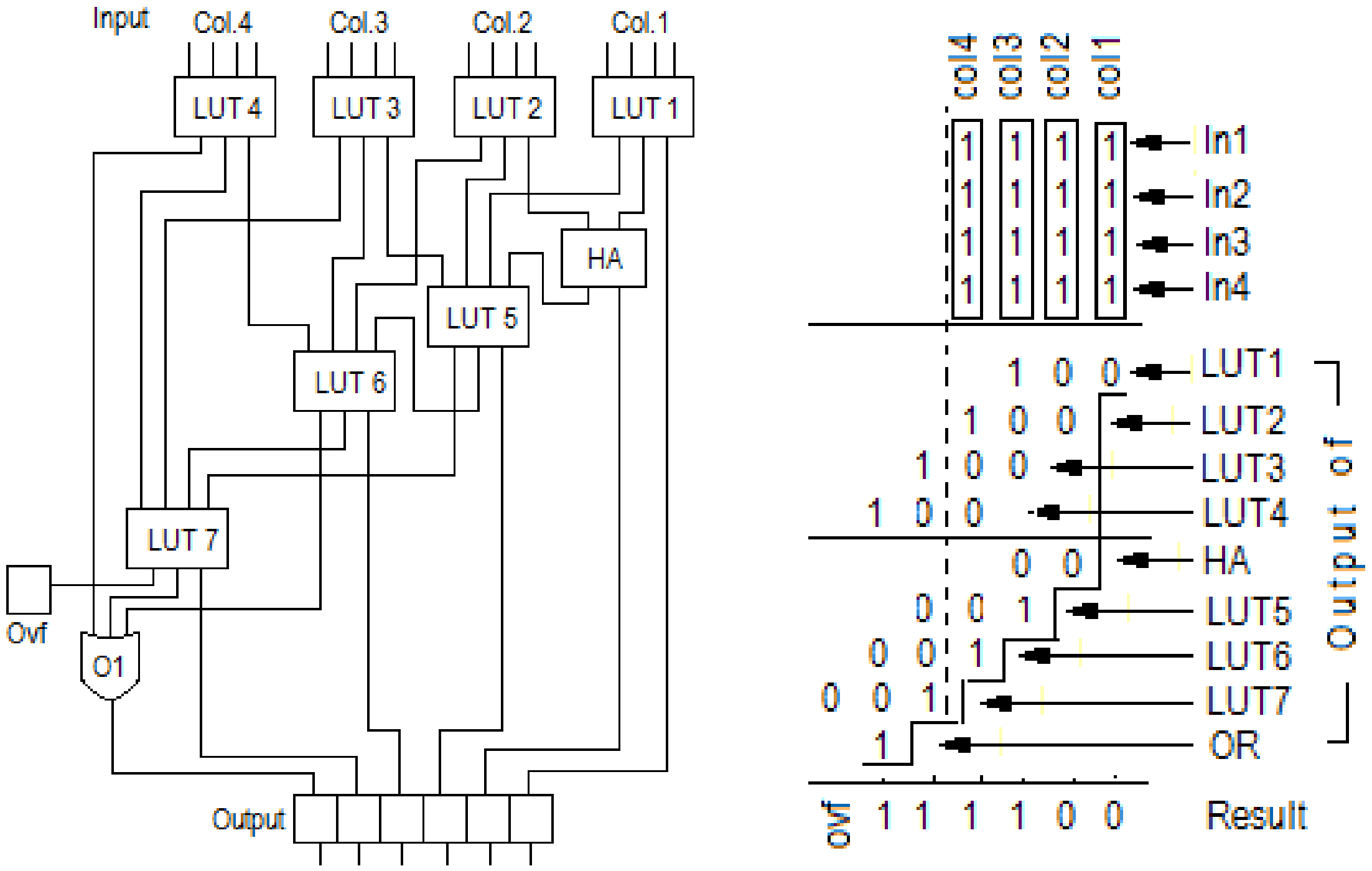

(a) (b)

**Figure 6.8 Fast Implementation of 4x4 adder (a) Schematic and (b) Example**

Construction of a $16 \times 16$ adder using $4 \times 16$ adder module is discussed in the next section. While implementing arrays of multi-operand adders, e.g. for neural networks, each $4 \times M$ serial adder may have one LUT. As only one column is added at a time, only one LUT is sufficient for the 4-operand adder. Size of LUT is fixed by choice of N (= 4) and is independent of width of input i.e., M bits. As N increases, size of LUT decoder grows geometrically. Therefore while constructing adders with more than 4 operands, it is better to use 4-operand adder as a basic module. As the implementation does not require large area, use of one LUT per $4 \times M$ adder is a good practice.

## 6.4 Generalized Algorithm for Reconfiguration

The number of operands in a neural network is dynamic and it varies depending on several parameters like the learning algorithm, type of the input, end application and etc. Therefore, reconfiguring the existing hardware becomes necessary. In the following section, an algorithm to implement larger modules using 4-operand adder as a basic unit is explained (Table 6.4). The algorithm is as follows:

i) $N_1 = N$

ii) Calculate the number of levels required: $L = \text{ceil}(\log N_1 / \log 4)$

iii) For all the levels from 1 to L:

  a. Calculate the number of rows (number of 4-operand modules): $R = N_1/4$

  b. For all the rows from 1 to R:

    1) Generate $S$ & $C$ for each adders

    2) Sum $S$ from each adder have to be added together till a single $S$ term is obtained. Each such addition will generate additional $C$ terms

  c. $N_1 = R$, repeat iii

iv) Add all $C$ terms and concatenate with the $S$ term to obtain the final sum

As illustration, implementation of 16×16 adder using this algorithm is presented in the following section.

### 6.4.1 Implementing a 16×16 adder

A set of $4 \times 16$ and $4 \times 4$ adders can be used to construct a $16 \times 16$ adder. Overall procedure of implementing $16 \times 16$ adder considering the algorithm described in Table 6.4 will be as follows (See Figure 6.9):

i) Divide the operands in to 4 groups of 4 operands each and input each group to a $4 \times 16$ adder. Separate the sum and carry parts to obtain four sum values S3:S0 and carry values C3:C0.

ii) Add four sum values S3 to S0 to generate the final sum S and carry C4.

iii) Add all carry values C4 to C0, to generate final carry C.

iv) Concatenate {C, S} to get the final output.

**Table 6.4 Algorithm for reconfiguring adder modules**

```
Assume 4x16 and 4x4 adder modules are available for reconfiguration.

Objective
   To configure a 16 bit adder with N operands

Structure
   Module Add4x16(in[4][16], s, c)
   Module Add4x4 (in[4][4], s, c)

   Configuration {
      Add4x16 A[1...L][1...R]
      Add4x4  C[1...L][1...R]
      Add4x4  B[1][1]
   }

Algorithm
   N1 = N
   L = ceil(log(N1)/log(4))

   For i = 1 to L
      R = N1/4
      For j = 1 to R
         For k = 1 to 4
            p = (j-1)*4 +k
            //Place sum Adders
            If (i == 1)
               //Adders
               A[i][j][k] = in[p]
            Else if (i == 2)
               A[i][j][k] = A[i-1][p].S
            //Place carry adders
            If (i >= 2)
               C[i][j][k]= A[i-1][p].S
            If (i >= 3)
               C[i][j+R][k] = C[i-1][p].S
      N1=R

   B[1][1] = A[L][1].C
   B[1][2] = C[L][1].S
   Result = {B[1].S, A[L][1].S}
```

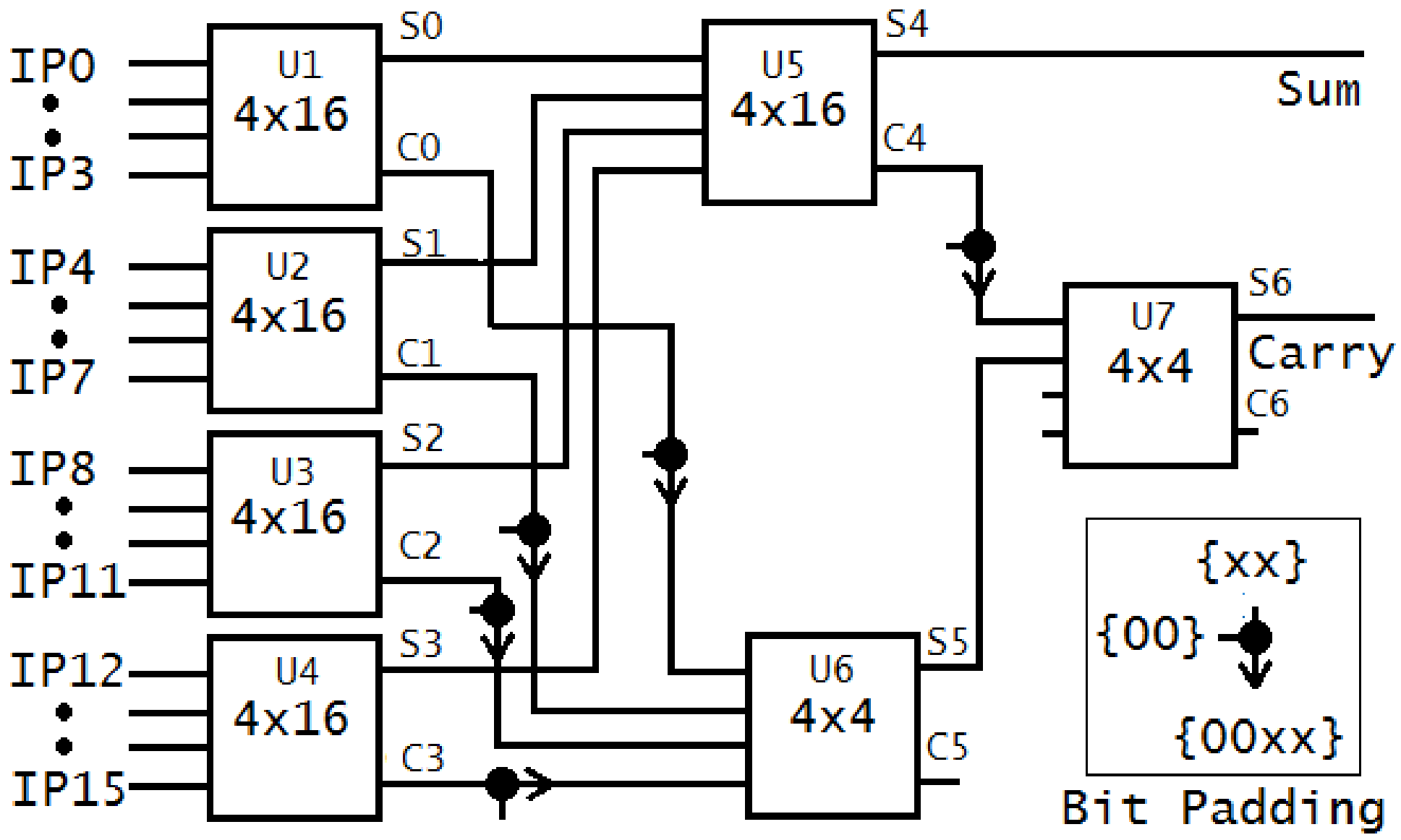

**Figure 6.9 Modular implementation of 16×16 adder**

Few details need attention. Carry output from U1 to U5 are zero padded to match the $4 \times 4$ adders U6 and U7. S5 from U6 has an upper bound of 12, i.e., $1100_{(2)}$. So there is no carry from U6 i.e., C5 = 0, always. On the other hand, carry C4 has an upper bound of 3, i.e., $0011_{(2)}$ as U5 has only 4 inputs. Therefore, Sum S6 from U7, will have a maximum value of 15, i.e., $1111_{(2)}$. Therefore, there is no carry from U7, i.e., C6 = 0. This is as expected from Theorem – maximum carry for 16 operand addition is 15. It is possible to replace U6 by a $4 \times 2$ adder. In that case, S5 and C5 have to be concatenated as {C5, S5} and apply it as input to U7. Further, as U7 has only two inputs, a $2 \times 4$ adder is sufficient. Carry C6 will be only one bit but may be ignored as it will always be zero. We have used 4-operand adder only for modularity of implementation.

## 6.5 Simulation Results

The simulation result of $4 \times 4$ serial addition using single LUT is shown in Figure 6.10, which shows the addition of four numbers of base 16 i.e., A+F+1+2=$1C_{(h)}$. Output of LUT for four columns is {2, 3, 1, 2}. Four columns will require four clock cycles. Stable data is available at the fifth clock cycle. The simulation result of fast implementation of parallel 4x4 adder described in section 6.3.2 is shown in Figure 6.11.

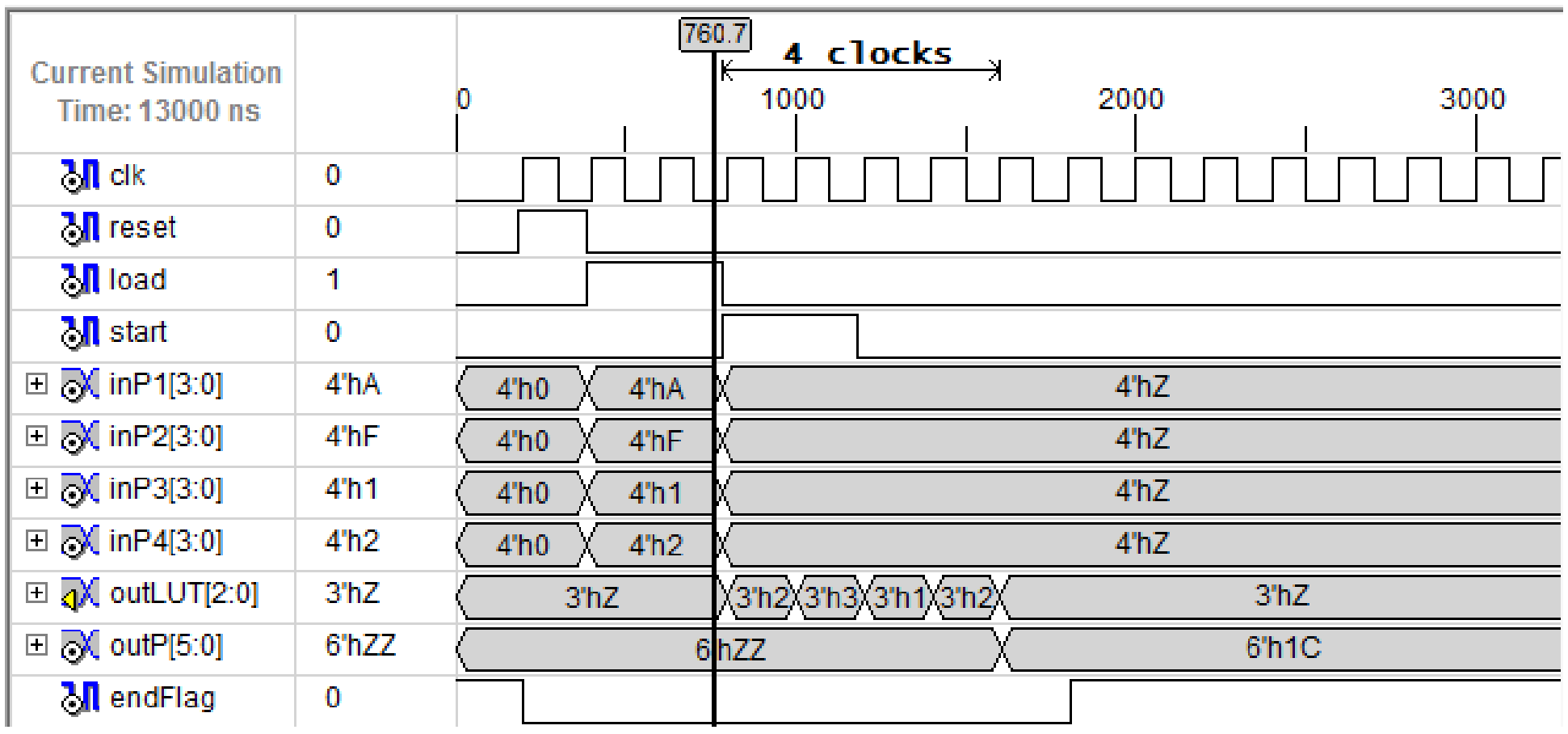

**Figure 6.10 Simulation result of 4×4 serial addition**

The $4 \times 4$ serial adder module can be easily extended to $4 \times 16$ adder simply by increasing the number of iterations to 16. Correspondingly we will perform 16 column additions. The simulation result of $4 \times 16$ serial addition is shown in Figure 6.12, it shows the addition of four 16-bit numbers represented using 16-bit number format i.e., 0A1A+0FFF+0A2D+01CD=2613. The operation requires a maximum of 16 clock cycles. Stable data is available at the $17^{th}$ clock cycle. The simulation result of $16 \times 16$ adder is shown in Figure 6.13.

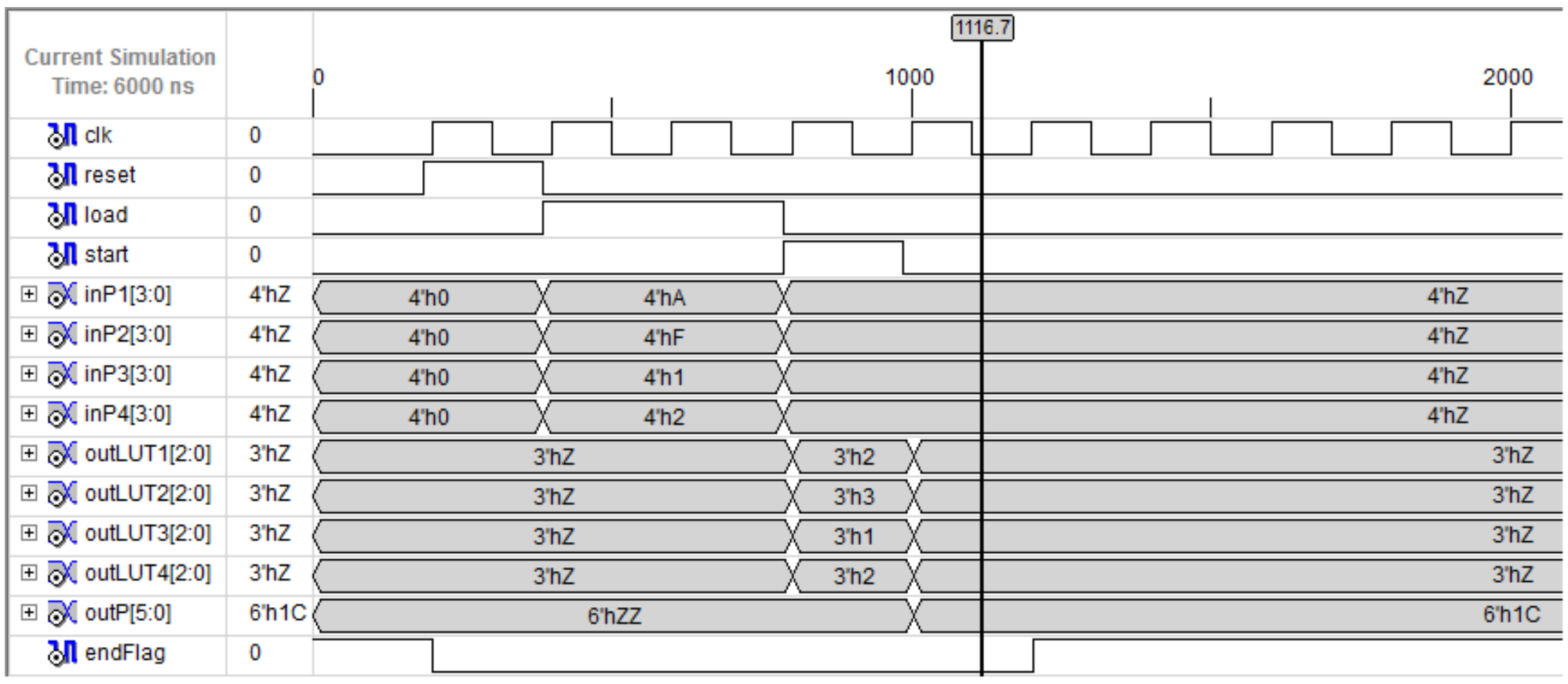

**Figure 6.11 Simulation result of Fast 4×4 parallel adder**

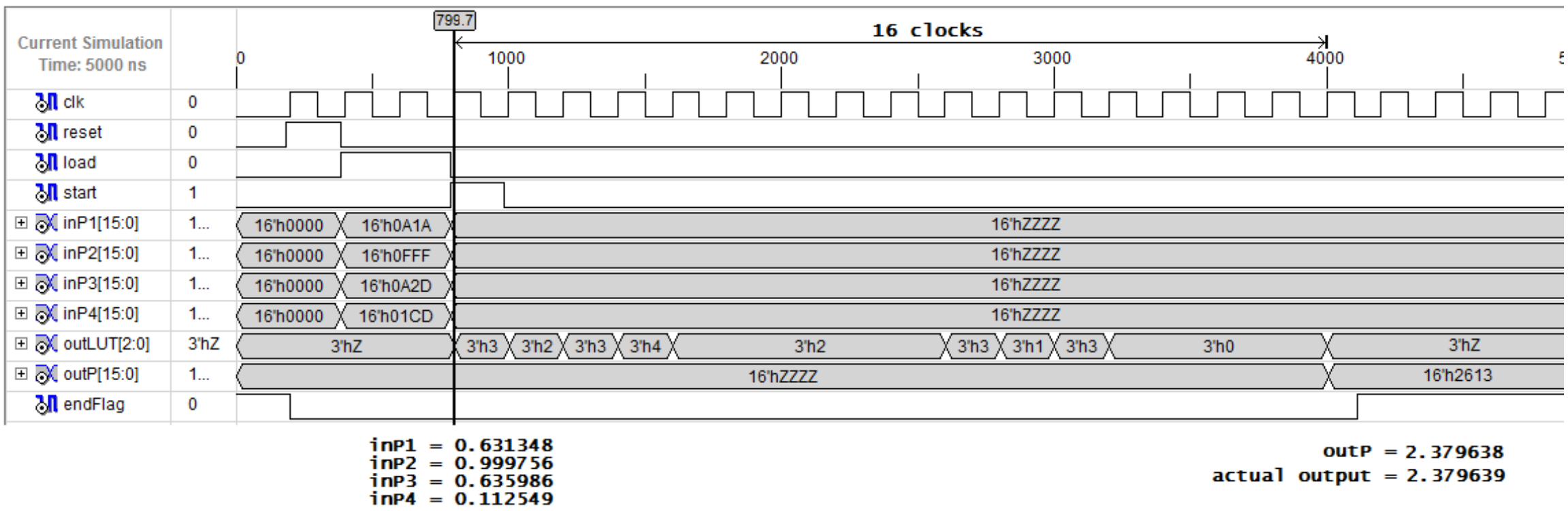

**Figure 6.12 Simulation result of 4×16 addition using 16-bit number format**

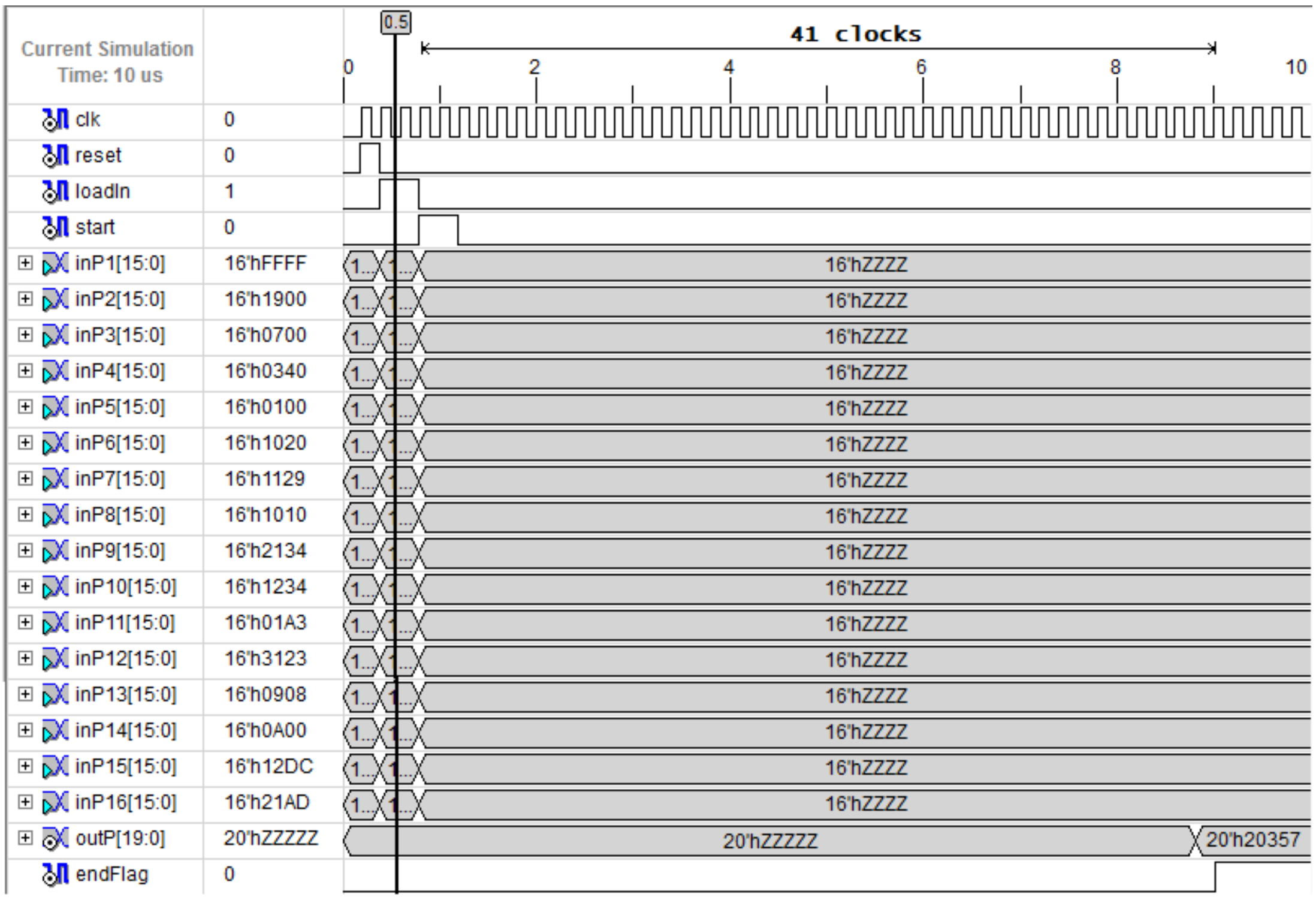

**Figure 6.13 Simulation result of 16×16 addition**

## Summary

Multi-operand addition is a frequent requirement in ANN hardware. However, there was a need to compute the number of carry bits required to handle multi-operand addition. The necessary theoretical studies have been carried out and the results thereof have been presented as lemmas and a theorem. A modular multi-operand adder has been built. An algorithm to implement larger adders using a modular approach has been presented. All these modules have been integrated to build a ARN neuron and a Perceptron, as discussed in chapter 7.

# Chapter 7

## Implementation of ARN nodes

The modules discussed in chapter 5 and chapter 6 are integrated to realize a 16-input neuron for ARN. 16 input Perceptron also has been realized.

In an ARN, as discussed in chapter 4, every ARN node has as many resonators as the number of inputs. The output of every resonator is added and scaled to compute the output of ARN node. So, the number of resonators available to the node is the number of inputs supported by the ARN node. However, if more inputs are required, the partial outputs from adder need to be temporarily stored. These can be added later after all inputs are processed. In the current implementation, addition is performed using a multi-operand adder. This adder can be configured to match the actual number of inputs to save computational time. A regularization unit at the output of ARN nodes may be used to normalize the outputs to be in the range as required. As an illustration, an image recognition system using a two-layer ARN is considered.

The node with 16 inputs followed by resonators, multi-operand adder is shown in Figure 7.1. The inputs to the resonator are x, $x_m$ and control parameters ($\rho$ and T). It is possible to use same values of $\rho$ and T for the all resonators, or they can also use different values. The output of a node with N inputs is calculated using the formula given in equation (4.8).

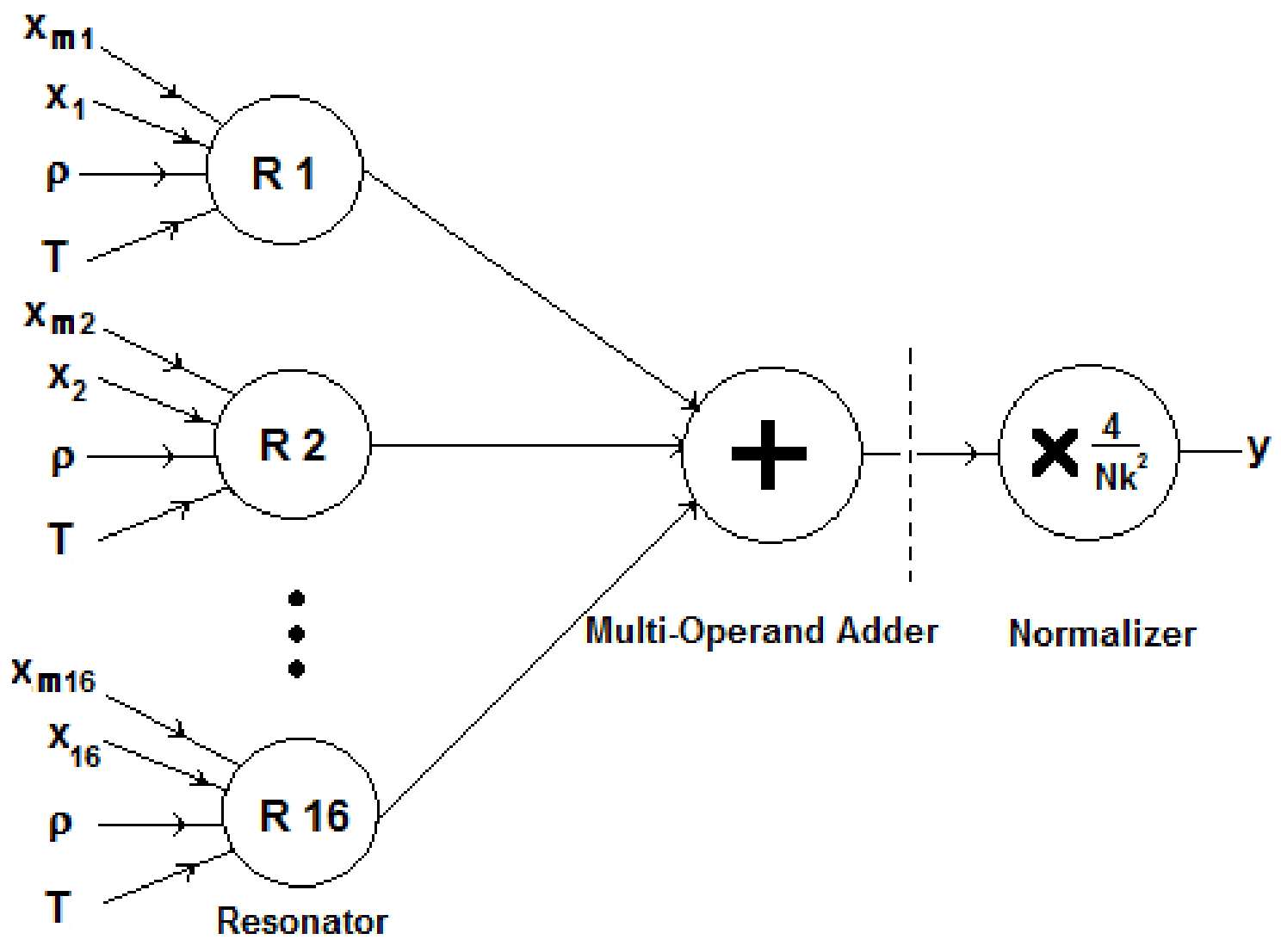

**Figure 7.1 Structure of image recognition using ARN with 16-inputs**

The normalizer (regularization unit) used at the end of an adder block may be implemented as a serial multiplier with the value of one operand fixed to $4/Nk^2$. However, as the number of inputs to a network may vary, this value should be fixed only at the end of arrival of inputs.

The block diagram of a 16-input ARN node is shown in Figure 7.2(a). FSM of implementation is shown in Figure 7.2(b). Current implementation assumes k=1 and identical threshold and control parameter values for all the resonators ($\rho$=2.42, T=0.9).

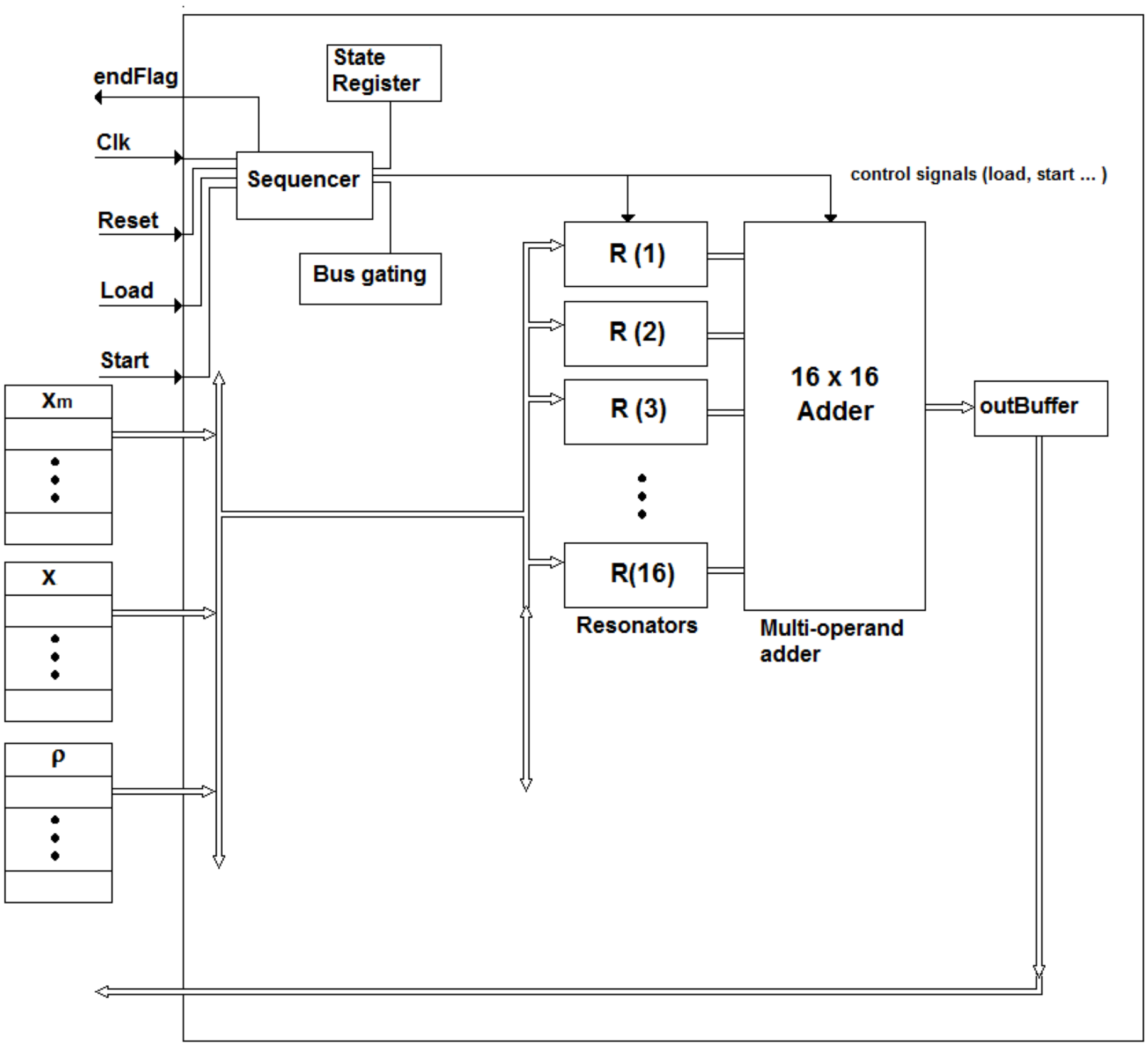

(a)

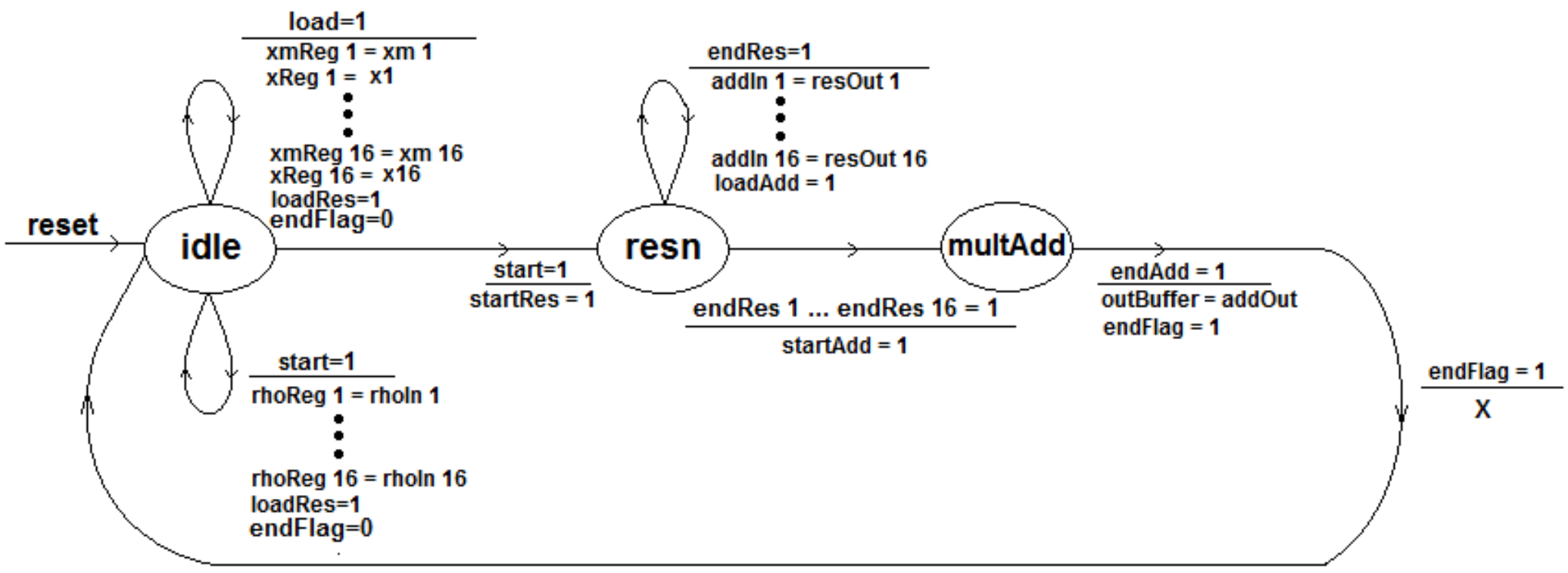

(b)

**Figure 7.2 16-input ARN node a) Block diagram b) FSM**

On the positive or negative edge of the clock, when *load* command is active, input values will be loaded into the internal registers. On *start* command, the contents of internal registers will be copied to the registers of individual resonators. When all the necessary values are copied, the sequencer sends a *start* signal to the resonators and waits for the completion. When the sequencer receives the *endRes* signal from all the resonators, it will copy the results of each resonators into the input registers of multi-operand adder. Upon loading all the input values, sequencer will send the *start* signal to the multi-operand adder and waits for the *endAdd* signal. Once the *endAdd* signal is received, sequencer will transfer the result of adder to output buffer and sets *endFlag* =1, to indicate the end of operation. The simulation result of 16-input ARN node is shown in Figure 7.3 (output normalizer is not used). The input and output values of simulation result are summarized in Table 7.1.

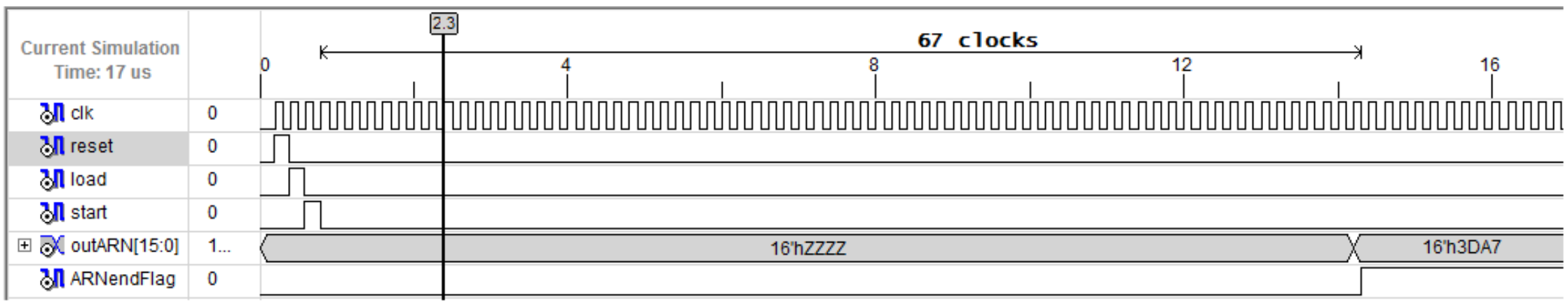

**Figure 7.3 Simulation result of 16-input ARN node**

**Table 7.1. Input and output values of 16-input ARN node shown in Figure 7.3**

| x | | $x_m$ | | Resonator output | | | Node output (y) | | |
|---|---|---|---|---|---|---|---|---|---|
| Actual value | 16-bit number format | Actual value | 16-bit number format | Actual value | Obtained value | 16-bit number format | Actual value | Obtained value | 16-bit number format |
| 0.5 | 0800 | 0.25 | 0400 | 0.2284 | 0.2297 | 03AD | 3.8047 | 3.8549 | 3DA7 |
| 0.2 | 0333 | 0.15 | 0266 | 0.2490 | 0.25 | 0400 | | | |
| 0.33 | 0547 | 0.13 | 0214 | 0.2359 | 0.2399 | 03D7 | | | |
| 0.78 | 0C7A | 0.5 | 0800 | 0.2233 | 0.2297 | 03AD | | | |
| 0.14 | 027D | 0.05 | 00CC | 0.2470 | 0.2480 | 03F8 | | | |
| 0.26 | 0428 | 0.21 | 035C | 0.2490 | 0.25 | 0400 | | | |
| 0.2 | 0333 | 0.1 | 0199 | 0.2463 | 0.2480 | 03F8 | | | |
| 0.56 | 08F5 | 0.43 | 06E1 | 0.2439 | 0.2446 | 03EA | | | |
| 0.34 | 0570 | 0.2 | 0333 | 0.2429 | 0.2446 | 03EA | | | |
| 0.9 | 0E66 | 0.7 | 0B33 | 0.2359 | 0.2399 | 03D7 | | | |
| 0.59 | 0970 | 0.4 | 0666 | 0.2372 | 0.2399 | 03D7 | | | |
| 0.7 | 0B33 | 0.35 | 0599 | 0.2100 | 0.2141 | 036D | | | |
| 0.5 | 0800 | 0.23 | 03AE | 0.2251 | 0.2297 | 03AD | | | |
| 0.66 | 0A8F | 0.61 | 09C2 | 0.2490 | 0.25 | 0400 | | | |
| 0.94 | 0FDA | 0.82 | 0D1E | 0.2448 | 0.2446 | 03EA | | | |
| 0.86 | 0DC2 | 0.77 | 0C51 | 0.2470 | 0.25 | 0400 | | | |

Similar concept of implementing ARN node may be extended to other neural architectures like MLP. A typical neuron in MLP will have set of inputs corresponding to the strength of synaptic connections. Each of these inputs are multiplied with their synaptic weights and added together followed by a non-linear activation function. Therefore, a structure of an 16-input perceptron will contain set of multipliers, adders and an activation function. In addition to these basic modules it can include several other modules depending on its end application, input type, learning algorithm and many other parameters.

The block diagram of 16-input perceptron is shown in Figure 7.4. There are sixteen serial multipliers, which will multiply the inputs and their synaptic weights. The results of the serial multipliers are added using $16 \times 16$ adder followed by a non-linear activation function. The type of activation function used in any DLNNs depends on the end application and the learning algorithm. In this implementation sigmoid activation function is considered. The details of implementation of sigmoid using approximation methods were described in chapter 5. LUT for Hyperbolic tangent (*tanh*) has been implemented and may be used in realizing other types of DNNs. The approximation of *tanh* function is described in Appenidx-2.

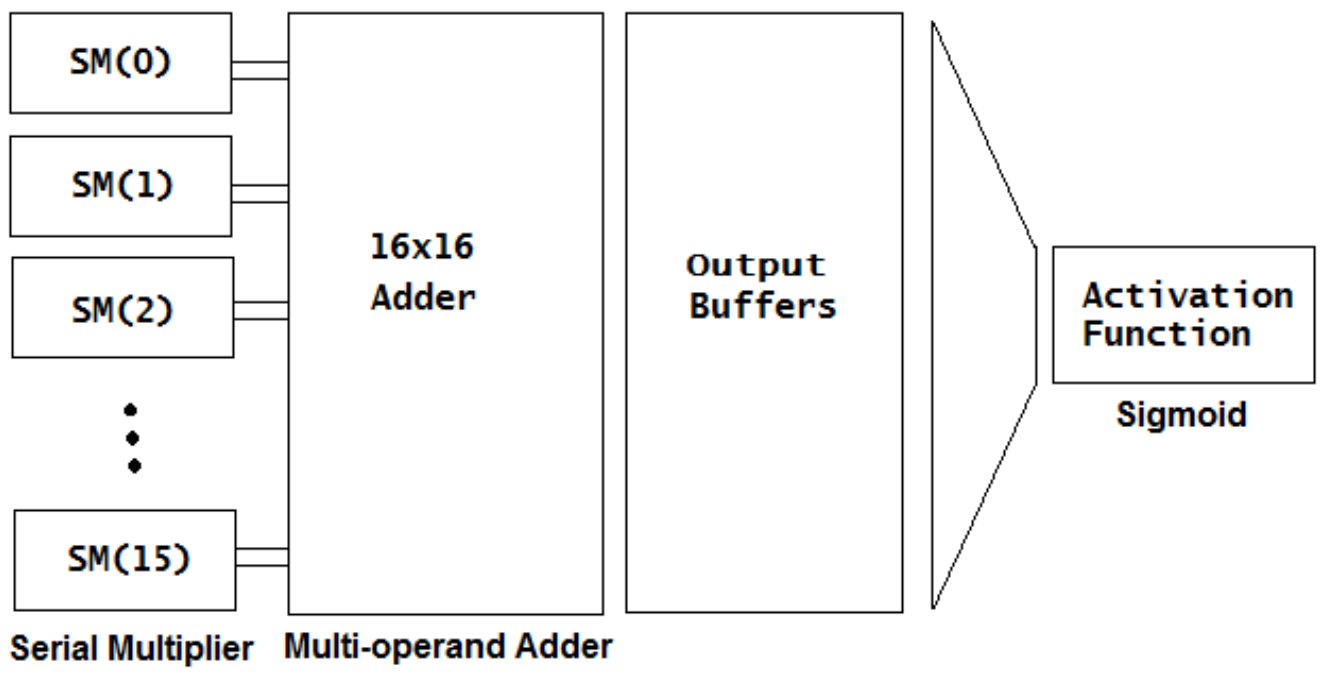

**Figure 7.4 Structure of 16-input perceptron**

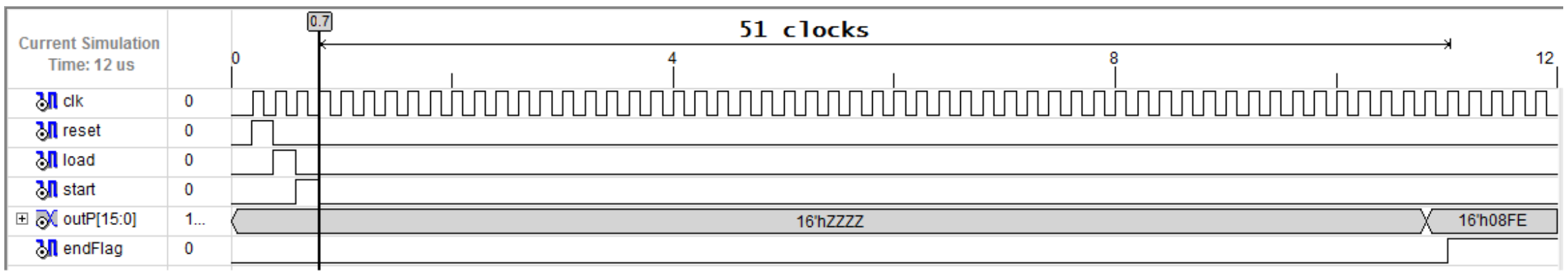

**Figure 7.5. Simulation result of 16-input perceptron**

The simulation result of 16-input perceptron is shown in Figure 7.5, which shows 16-inputs and 16-wieghts and the output as $08FE_{(h)}$. Inputs and outputs are represented using a 16-bit number format as summarized in Table 7.2. The expected output is 0.5942 and the obtained output is 0.5620 giving an absolute error of 0.0322.

**Table 7.2 Input and output values of 16-input perceptron shown in Figure 7.5**

| x | | w | | $\Sigma y_n$ | | | Node output (y) | | |
|---|---|---|---|---|---|---|---|---|---|
| Actual value | 16-bit number format | Actual value | 16-bit number format | Actual value | Obtained value | 16-bit number format | Actual value | Obtained value | 16-bit number format |
| 0.09 | 0170 | 0.1 | 0199 | 0.38151 | 0.3281 | 0540 | 0.5942 | 0.5620 | 08FE |
| 0.15 | 0266 | 0.01 | 0028 | | | | | | |
| 0.12 | 01EB | 0.2 | 0333 | | | | | | |
| 0.05 | 00CC | 0.5 | 0800 | | | | | | |
| 0.009 | 0024 | 0.16 | 028F | | | | | | |
| 0.123 | 01F7 | 0.21 | 035C | | | | | | |
| 0.087 | 0164 | 0.19 | 030A | | | | | | |
| 0.201 | 0337 | 0.09 | 0170 | | | | | | |
| 0.05 | 00CC | 0.25 | 0400 | | | | | | |
| 0.15 | 0266 | 0.02 | 0051 | | | | | | |
| 0.27 | 0451 | 0.12 | 01EB | | | | | | |
| 0.02 | 0051 | 0.26 | 0228 | | | | | | |
| 0.1 | 0199 | 0.36 | 05C2 | | | | | | |
| 0.07 | 011E | 0.6 | 0999 | | | | | | |
| 0.054 | 00DD | 0.29 | 02A3 | | | | | | |
| 0.18 | 02E1 | 0.63 | 0A14 | | | | | | |

## Summary

The hardware modules discussed in chapter 5 and 6 viz., resonators, approximation functions, serial multiplier and multi-operand adders were integrated to design a 16-input ARN neuron and MLP neuron (perceptron). All the implementations were carried out using 16-bit number format. Several other modules required to implement a complete 16-input neuron are described in Appendix-2.

# Chapter 8

# Conclusion

A brief summary of the work reported in the previous chapters is presented here. A small introduction to how and why I selected the topic will help me to align the readers with my thought processes then.

The original goal was to start working on processor design and focus along the industry challenges. In 2016, support for IoT at chip level was missing and many companies were working on integrating communication modules into the processor chip. However, in about a year, new AI applications were being introduced, especially in image recognition using CNN. In a short while, language translation tools using LSTM and similar ANN structures started appearing. Then industry shifted attention to AI as these techniques could address real life problems hitherto considered as not feasible. In the mean while, most of the major chip makers had integrated essential modules for web based IoT. GSM and Bluetooth modems also provided the required functionality. Web services then could take the IoT data to servers efficiently and then processing could be done on the servers. So, a level of saturation in IoT was attained. Most of the industry therefore put IoT to the back burner. Interestingly, next level of IoT is waiting for developments in AI.

It was still early period of my research and I was advised to shift focus to AI instead of IoT. Our research group had just implemented Auto Resonance Networks, a new ANN structure, for robotic path planning, which could be extended to several other areas like image processing and time series prediction. The challenges posed by AI were far more complex and therefore, it was wise to focus attention to AI. The work now was focussed on implementing ARN for image recognition and implementing hardware modules necessary to implement ARN at processor level.

Section 8.1 presents a review of the complete work done during the period of this research. Results of the work have been presented in individual chapters but a combined discussion is presented in section 8.2. Interpretation and application of these results is also presented in the same section, as appropriate. Last section of the thesis, Section 8.3 presents scope for extending the work reported in this thesis.

## 8.1 Review of the Work Done

Auto Resonance Network is an approximating hierarchical feed forward network. ARN is also an explainable network. Nodes learn from input samples (supervised learning) and by perturbation of existing nodes (reinforcement learning). All learning is based on local information. However, by arranging ARN in a hierarchy, it is possible to reach higher levels of data abstraction. Every neuron approximates each of its inputs and adapts its resonance parameters in response to changes in data.

Some of my contributions to ARN are presented in chapter 4. Firstly, I have presented a generalization of the earlier model available in literature. This allows better control over the range of turning parameters. Sections 4.3 to 4.5 describe the generalized model and its control graphs. Section 4.8 describes the relation between the resonance parameters and variance of the input data. This result is useful in dynamically adjusting the properties of ARN neurons automatically, without any supervision. Supervision is only required in labelling the recognized pattern and not to improve the recognition accuracy. I also standardized use of sigmoid function for resonance as it allowed better mathematical modelling. Sections 4.9 to 4.11 describe use of ARN for image recognition. A two layer ARN and a small sample set of 500 images was enough to reach recognition accuracy of above 90%. Section 4.11 also discusses more general issues like data masking, node tuning and resonance characterization. It is to be noted that the resonance in ARN is different than that of an electrical LC circuit. The peak of ARN resonance is bound to a predefined value, much like the activation level of a biological neuron. Quality of resonance is largely controlled by the coverage, which is similar to bandwidth of a signal filter. Like a biological neuron, ARN neurons respond to a set of input values. One way of addressing such situations is to add one more layer of neurons to the network. Section 4.11 also discusses how such ambiguities may be resolved within the layer using a local supervisory wrapper function. Such wrapper action may be happening in ganglia of biological neural networks.

Hardware implementation of ARN was primary goal of this research. Some of the key issues encountered during conceptual design stages have been presented in chapter 3. Implementation of various hardware modules has been discussed in chapter 5 to chapter 7. Chapter 5 discusses three preliminary issues in implementation of ANNs. Firstly, the advantage of using serial computations over parallel computations when the ratio of areas is larger than the ratio of compute times is presented in section 5.1. Serial computation can also have communication advantages. Most of the computations in ANN can benefit from use of lower precision. Numbers used in ARN span a limited range of values. Therefore, an approximate representation, limited to about 3 fractional decimal places, is sufficient for good ARN performance. Hence a new fixed point number format was defined to take advantage of this fact. Also, using fixed point numbers instead of real values improves computational speed. All numeric modules use this number format. First of the numeric modules to be implemented was a serial multiplier. The ratio of gates used for serial and parallel multipliers was 66:2144 (~1:32). Speed advantage of parallel multiplier over serial multiplier was around 1:17. Clearly, according to lemma presented in section 5.1, using serial multiplier is better for improved performance.

Computation of non-linear activation functions (except ReLU) using Taylor series expansion is computationally expensive. Therefore approximations accurate to less than 0.1% (0.001) using piece-wise linear and second order interpolation have been implemented. Error performance of the approximation is discussed in section 5.4 and 5.5. Design and implementation details of ARN resonator are given in section 5.6.

Most of the neuronal processing involves a multi-operand adder at the output. The number of inputs to a neuron can be very large. Most of the adders can handle only two numbers at a time and hence the multi-operand additions can take considerable time. One of the problems in implementing multi-operand addition operation is that of carry. For example, addition of 16 numbers requires 4 bit carry while using 17 numbers requires 5 bits. Therefore, we looked into the theoretical basis for size of carry. Section 6.1 elaborates this exploration to an analytical solution. This has been used to create modular multi-operand adders, which can be reconfigured to handle arbitrary number of operands (limited only by the number of modules available on chip). Details of these modules are discussed in section 6.2.

Chapter 7 contains implementation details of a 16-input ARN neuron and MLP neuron. These units can be combined to form larger neurons.

## 8.2 Quick Overview of Results

At the time when this research work was taken up, ARN had been used only in control systems. It was clear that ARN could be used to address several AI tasks but no experiments had been conducted. Implementing an image recognition algorithm using

ARN for MNIST data set is only a proof of such concept. The implementation partitions the input in to small segments, similar to CNN but the similarity ends there. However there are significant differences from CNN. Firstly, ARN does not perform any gradient search but only a local optimization based on statistical characteristics of input data. These experiments have confirmed the possibility of using small ARN structures in edge processors, for local AI support.

Hardware implementation presented some interesting results. In case of massively parallel operations typically seen in ANN, the number of multipliers is very large. Hence hardware accelerators have to implement a large number of multipliers. Multipliers are known to consume a large silicon area. Hence, we tried to use serial multipliers. On simulation, I observed that group of serial multipliers gives a better throughput than parallel multipliers, occupying a similar area. A lemma that justifies this observation was an important result.

Use of low precision arithmetic for ANN was only to gain speed advantage. This work was possibly one of the first to notice that the performance actually improves with low precision arithmetic. This was more so in ARN as low precision actually helped implement resonance in an easier way. Therefore defining and using a low precision arithmetic was an important observation. Many modern AI processors implement low precision arithmetic, but at the time of publication of our work, it was not very well reported.

Multi-operand addition is generally handled as an iterative procedure or data flow structure, considering two operands at a time. A better option was to perform multi-operand addition as a single integrated step. However, we could not find any published literature that addressed the issue. The challenge of implementing multi-operand adder was overcome by developing necessary theoretical framework. This allowed development of modular, reconfigurable multi-operand adder. One of the difficulties was to know the exact number of carry bits. Using more number of bits slows down the computation while using smaller number of bits can produce wrong results. Procedure to compute the number of bits was a significant result of this research. As the number of carry bits can be estimated from the theorem, overflow will not occur at any point of time.

Several other modules were required to implement a complete 16-input ARN node. These modules have been presented as an appendix. The primary objective of the research was successfully achieved by implementing this complete unit.

## 8.3 Scope for Future Work

As any other research, the final goal post keeps shifting as we reach our initial targets. ARN is a relatively new concept and there is a need to extensively study and prove its adaptability in various application environments. Several neuronal features need to be studied further. For example, I believe that real neural networks work sequentially but asynchronously. Current ARN models hold the neuronal activation for a certain time depending on the strength of excitation. This assumes that there exists some form of local synchronization among neurons in a neighbourhood within and across layers. Further work is necessary to verify if clock jitter along neuronal paths has much effect on the accuracy of recognition. Second issue in neuronal modelling is use of spiking models within ARN. Instead of holding the activation for a length of time, spikes of constant value (corresponding to neuronal excitation level) but of varying frequency may be generated. These spikes can further trigger changes in network output over a period of time, corresponding to a single excitation input. Experiments need to be conducted to see if merging ARN with spiking improves recognition accuracy.

Tuning of ARN needs further improvements in implementation. Higher layers of ARN receive data which is not ordered in any specific way. Finding hash functions that retain similarity of inputs at ARN neuronal output could represent significant research effort. Implementation specific solutions can be immediately presented but generalization will require further study.

Ambiguity resolution requires some algorithmic approach. Currently, use of wrappers around sets of neurons and around a layer of neurons is suggested. However, there could be alternate possibilities to resolve ambiguities. For example, early work on ARN for robotic motion planning, a concept of *folds* was introduced to explore alternate paths to joint motion. Such folds occur naturally in robotic systems. It may offer some new directions to ambiguity resolution. Supervised learning in ARN is facilitated by use of class labels. However, in case of ambiguities, labels do not provide any hints to any possible solution. Diffusion functions have been described in ARN literature but need further exploration.

On the hardware side, this research can be extended till a commercial IP is implemented. The modules developed here are only preliminary designs. Significant contributions have been made in the area of multi-operand adder. The accelerator can be used as it is but further improvements are required in terms of design of instructions to control the accelerator, error handling, etc. Physical layout has not been considered because of resource limitations at the research facility. Current design uses shared memory between the CPU and accelerator. This needs further work on data exchange mechanisms and protocols like resource locking to avoid loss of data. Note that multi-operand multiplications are not very common and hence conventional designs can be used.

Deep learning will continue to influence processor design in years to come. Learning models will evolve through better understanding of central and peripheral nervous systems. As days pass, artificial general intelligence gets closer to reality.

# Publications

**International Journals (Published)**

[2019 P1] **Shilpa Mayannavar and Uday Wali**. "*Hardware Accelerators for Neural Processing",* Int. J. of Computer Information Systems and Industrial Management Applications (IJCISIM), ISSN 2150-7988, Volume 11, pp. 046-054, (May 2019)

[2016 P2] **Shilpa Mayannavar and Uday Wali.** "*Design of Modular Processor Framework*", Int. J. of Technology and Science (IJTS) Volume 9 No. 1, pp.36-39, (May 2016)

**International Journals (Communicated)**

[2019 P3] **Shilpa Mayannavar, Uday Wali and V M Aparanji**. "*A Novel ANN Structure for Image Recognition*", submitted to IETE Journal of Research (Nov. 2019)

[2019 P4] **Shilpa Mayannavar and Uday Wali**. "*Design of Reconfigurable Multi-operand Adder for Neural Processing*", submitted to IETE Journal of Research (Nov. 2019)

**Book Chapter (Published)**

[2019 P5] **Shilpa Mayannavar and Uday Wali.** "*Fast Implementation of Tunable ARN Nodes*", In: Abraham A., Cherukuri A., Melin P., Gandhi N. (eds.) Advances in Intelligent Systems and Computing (AISC), vol. 941. Springer, Cham, (April 2019). Presented in $18^{th}$ Int. Conf. on Intelligent System Design and Applications, VIT, Vellore, (Dec. 2018)

[2019 P6] **Shilpa Mayannavar and Uday Wali**. "*Design of Multi-Operand Adder for Neural Processing*", In Saroj Kaushik, Daya Gupta, Latika Kharb, Deepak Chahal (eds.) Information, Communication and Computing Technology (CCIS), Springer. Communications in Computers and Information Science, vol 1025, pp. 167-177, Nov. 2019, Springer, Singapore

[2019 P7] **Shilpa Mayannavar and Uday Wali**. "*A Noise Tolerant Auto Resonance Network for Image Recognition*", In Saroj Kaushik, Daya Gupta, Latika Kharb, Deepak Chahal (eds.) Information, Communication and Computing Technology (CCIS), Springer. Communications in Computers and Information Science, vol 1025, pp. 156-166, Nov. 2019, Springer, Singapore

**International Conferences (Published)**

[2018 P8] **Shilpa Mayannavar, Bahubali Shiragpur and Uday Wali**. "*Saturating Exponential Law for Reduction of Power and PAPR in OFDM Networks*", In 2018 Fourth International Conference on Computing, Communication Control and Automation (ICCUBEA), pp. 1-6, IEEE Xplore Digital Library, (Aug. 2018)

**International Conferences (In press : Accepted for Publication)**

[2018 P9] **Shilpa Mayannavar and Uday Wali**. "*Performance Comparison of Serial and Parallel Multipliers in Massively Parallel Environment*", In 3rd International Conference on Electronics, Communication, Computer Technologies and Optimization Techniques, Mysore, IEEE Xplore Digital Library (In press), (Dec. 2018)

[2018 P10] **Shilpa Mayannavar and Uday Wali**. "*Hardware Implementation of an Activation Function for Neural Network Processor*", In International Conference on Electrical, Electronics, Computers, Communication, Mechanical and Computing, Vaniyambadi, Vellore. IEEE Xplore Digital Library (In press), (Jan. 2018)

# References

## (Print and On-line)

1. [1960 Widrow] **B Widrow,** *An Adpative Adaline Neuron using chemical memristors*, Stanford Electronics Laboratory Technical Reports, No. 1553-2, Oct. 1960
2. [1962 Rosenblatt] **F Rosenblatt,** *Principles of neurodynamics – perceptrons and the theory of brain mechanisms*, Cornell University, Report No. 1196-G-8, March 1962
3. [1964 Wallace] **C S Wallace**, *A Suggestion for a fast Multiplier,* IEEE Transactions on Electronic Computers, vol.13 issue1, pp.14-17, IEEE, Feb. 1964
4. [1969 Minsky] **M Minsky, S Papert**, *Perceptron: an introduction to computational geometry* The MIT Press, Cambridge, expanded edition, 1969
5. [1973 Grossberg] **S Grossberg,** *Contour enhancement, short term memory and constancies in reverberating neural networks,* Studies in applied mathematics, vol. 52, pp.217-257, 1973
6. [1973 Singh] **Shanker Singh and Ronald Waxman**, *Multiple operand addition and multiplication*, IEEE Transactions on computers, vol.c-22, no.2, pp. 113-120. Feb. 1973
7. [1978 Atkins] **D E Atkins and S C Ong**, *A comparison of two approaches to multi-operand binary addition*, In IEEE 4th Symposium on Computer Arithmetic (ARITH) pp. 125-139, 1978
8. [1980 Fukushima] **K Fukushima**, *Neocognitron, A Self-organizing Neural Network Model for a Mechanism of Pattern Recognition Unaffected by Shift in Position*, Biol. Cybernetics, vol. 36, pp. 193-202, Springer-Verlag, 1980
9. [1981 Hennessy] **J Hennnessy, N. Jouppi, F. Baskett, J. Gill**, *MIPS: a VLSI processor architecture,* in VLSI Systems and Computaion, pp. 337-346, 1981
10. [1981 Kung] **H. T. Kung and S. W. Song**, *A systolic 2-D convolution chip,* Carnegie Mellon University, Document A104872, March 1981
11. [1985 Ackley] **David H Ackley, Geoffrey E Hinton, Terrence J Sejnowski, Terrence J**, "*A learning algorithm for Boltzmann machines*", Cognitive Science, Vol. 9 No. 1, pp. 147–169, 1985 doi:10.1207 /s15516709cog0901_7
12. [1985 Rumelhart] **D Rumelhart, G Hinton, R Williams,** *Learning internal representations by error propagation* Technical report No. ICS-8506, California University, San Diego and LA Jolla Inst. for Cognitive Science, 1985
13. [1987 Grossberg] **Stephen Grossberg**, *Competitive Learning: From Interactive Activation to Adaptive Resonance*, Cognitive Science, vol. 11, no. 1, pp.23-63, 1987
14. [1987 Malsburg] **C.von der Malsburg**, *Synaptic Plasticity as Basis of Brain Organization* in The Molecular Basis of Learning, John Wiley & Sons Ltd, pp 1-24, 1987

15. [1989 Waibel] **Alexander Waibel, Toshiyuki Hanazawa, Geoffrey Hinton, Kiyohiro Shikane, Kevin J Lang**, *Phoneme Recognition using Time-Delay Neural Networks* **IEEE** Trans. On Acoustics, Speech and Signal Processing. 37(3), 1989

16. [1990 Kohonen] **T Kohonen,** *The Self-organizing Map,* Invited paper, Proceedings of the IEEE, vol. 78, no. 9, pp.1464-80, 1990

17. [1990 Thorpe] **S J Thorpe,** *Spike Arrival Times: A highly efficient coding scheme for neural networks*, In R Eckmiller, G Hartman & G Hauske (Eds), Parallel processing in neural systems and computers, pp. 91-94, North Holland Elsevier, 1990

18. [1990 Wang] **D Wang and M Arbib,** *Complex temporal sequence learning based on short term memory,* Proceedings of IEEE, vol 78, no. 9, pp.1536-42, 1990

19. [1991 Siegelmann] **Hava T Siegelmann, Eduardo D Sontag**, *Turing computability with Neural nets",* Applied Mathematics Letters, Vol.4, no. 6, pp. 77-80, 1991

20. [1993 Wawrzynek] **John Wawrzynek, Krste Asanovic**, *The design of a Neuro-Microprocessor*", IEEE transactions on neural networks, vol.4, no.3, May 1993

21. [1994 Glover] **Michael A Glover, W Thomas Miller**, *A massively parallel SIMD processor for neural network and machine vision applications,* Advances in Neural Information Processing Systems 6, NIPS, 1994

22. [1994 Hazewinkel] **Hazewinkel, Michiel, ed. (2001) [1994],** *Turing machine,* Encyclopedia of Mathematics, Springer Science+Business Media B.V. / Kluwer Academic Publishers, *ISBN 978-1-55608-010-4,* 1994

23. [1994 Verleysen] **Michel Verleysen, Philippe Thissen, Jean-Luc Voz**, *An analog processor architecture for a Neural Network Classifier*", IEEE Micro, Vo. 14, Issue 3, pp. 16-28, June 1994

24. [1997 Hochreiter] **Sepp Hochreiter, Jurgen Schmidhuber**, *Long Short-Term Memory,* Neural Computation 9(8): 1735-1789, 1997

25. [1998 LeCun] **Yann LeCun, Corinna Cortes, Christopher J C Burges**, *The MNIST Database of hand written digits,* http://yann.lecun.com/exdb/mnist

26. [1998 Nilsson] **Nils J Nilsson**, *Artificial Intelligence, A new Synthesis*, Morgan Kaufmann Publishers, Elsevier, San Francisco, ISBN: 81-8147-190-3, 1998

27. [1999 Doetsch] **Fiona Doetsch, Arturo Alvarez-Buylla,** *"Sub Ventricular Zone Astrocytes are Neural Stem Cells in Adult Mammalian Brain,"* Cell, vol. 97, Iss. 6, pp. 703-716, June 1999,

28. [1999 Malsburg] **Von der Malsburg**, The *What and Why of Binding: The Modeler's Perspective,* Neuron, 24, 95–104, Copyright ©1999 by Cell Press, Sep. 1999

29. [1999 Yegna] **B Yegnanarayana,** *Artificial Neural Networks,* ISBN 81-203-1253-8, Prentice Hall of India, New Delhi, 1999

30. [2000 Omondi] **Amos R Omondi**, *Arithmetic-unit and processor design for neural networks,* Neural networks for signal processing. Proceedings of the

2000 IEEE Signal processing society workshop (Cat. no. 00TH8501), IEEE, 2000

31. [2006 DeHon] **Andre DeHon, Yury Markivsky, Eylon Caspi, Michael Chu, Randy Huang, Stylianos Perissakis, Laura Pozzi, Joseph Yeh, John Wawrzynek**, *Stream computations organized for reconfigurable execution*, Microprocessors and microsystems 30 pp. 334-354, 2006

32. [2006 Goldberg] **David E Goldberg**, *Genetic Algorithms in search, optimization and machine learning*, Pearson Education Inc. Noida, India, ISBN. 978-81-775-8829-3, 2006

33. [2007 Fujii] **Robert Fujii and Taiki Ichishita,** *Effect of synaptic weight assignment on spiking neural response*, proc. 6th Int. Conf. on Machine Learning and Applications, pp. 217-222, IEEE Computer Society 2007

34. [2008 Che] **Shuai Che, Jiayuan Meng, Jeremy W Sheaffer, Kevin Skadron**, *A performance study of general purpose applications on graphics processors with CUDA*", Journal of Parallel and Distributed Computing, 68(10), 1370-1380, Oct. 2008

35. [2008 Lindholm] **Erik Lindholm, John Nickolls, Staurt Oberman and John Montrym**, *NVidia Telsa: A unified graphics and computing architecture*", IEEE Micro Vol. 28, Issue 2, pp. 39-55, DOI 10.1109/MM.2008.31 Mar-Apr, 2008

36. [2009 Carpenter] **Gail Carpenter,** *Adaptive Resonance Theory,* Open Boston University, Cognitive and Neural Systems Technical Report, 2009

37. [2009 Farabet] **Clement Farabet, Cyril Poulet, Jefferson Y Han, Yann LeCun,** *CNP: An FPGA based processor for convolutional networks*, International Conference on Field Programmable Logic and Applications, 31st Aug. – 2nd Sep. 2009, IEEE Xplore, 2009

38. [2010 Bergstra] **James Bergstra, Olivier Breuleux,Frederic Bastien, Pascal Lamblin, Razvan Pascanu, Guillaume Deshardians, Joseph Turian, David Warde-Farley, Yousha Bengio**, *Theano: a CPU and GPU math compiler in Python*", proceedings of the 9th Python in science conference, SCIPY, 2010

39. [2010 Nickolls] **John Nickolls, William J Dally**, *The GPU computing era*, IEEE Micro, 30(2), 56-69, DOI 10.1109/MM.2010.41 March-April 2010

40. [2010 Rinkus] **Gerard J. Rinkus**, *A Cortical sparse distributed coding model linking mini- and macrocolumn-scale functionality*, frontiers in Neuroanatomy, Hypothesis and Theory Article, Vol.4, No.17, pp.1-13, June 2010

41. [2011 Hawkins] **J. Hawkins and D. George**, *Hierarchical temporal memory (HTM)* whitepaper, 2011
[Online: last accessed 26-07-2019]
https://numenta.org/resources/HTM_CorticalLearningAlgorithms.pdf

42. [2011 Herrero] **J R Herrero**, *Special issue: GPU computing*", Concurrency and Computation: Practice and Experience, 23(7), 667-668, 2011

43. [2011 Keckler] **Stephen W Keckler, William J Dally, Brucek Khailany, Michael Garland, David Glasco**, *GPUs and the future of parallel computing*", IEEE Micro, 31(5), 7-17, Oct. 2011

44. [2012 Krizhevsky] **Alex Krizhevsky, Ilya Sutskever, Geoffrey E Hinton,** *ImageNet Classification with Deep Convolutional Neural networks*", Advances in Neural Information Processing Systems 25, pp. 1097-1105, 2012

45. [2012 Lotric] **Uros Lotric, Patricio Bulic**, *Applicability of approximate multipliers in hardware neural networks*, Neurocomputing Journal Elsevier, 96(1), 57-65, Nov. 2012

46. [2013 Saha] **Anindya Saha, Manish Kumar, Hemant Mallapur, Santhosh Billava**, **Viji Rajangam**, *Mechanism for efficient implementation of software pipelined loops in VLIW processors*", United States Patent, US 8447961B2. May 21 2013

47. [2014 Chen] **Yunji Chen, Tao Luo, Shaoli Liu, Shijin Zhang, Liqiang He, Jia Wang, Ling Li, Tianshi Chen, Zhiwei Xu, Ninghui Sun, Olivier Temam**, *DaDianNao: A Machine-learning supercomputer*", 47th annual IEEE/ACM international symposium on microarchitecture, pp. 609-622, 2014

**48.** [2014 Goodfellow] **Ian J. Goodfellow, Jean Pouget-Abadie, Hendi Mirza, Bing Xu, et al.,** *Generative Adversarial Nets*, arXiv: 1406.2661vl [stat.ML], 2014

49. [2014 Graves] **Alex Graves, Greg Wayne, Ivo Danihelka**, *Neural turing machines*, Google DeepMind, London, UK 2014, arXiv:1410.5401v2 [cs.NE], 10 Dec. 2014

50. [2014 Intel AVX] **Intel**® *Advanced Vector Extensions 2015/2016*; GNU Tools Cauldron 2014; Presented by Kirill Yukhin of Intel, July 2014

51. [2014 Jia] **Yangqing Jia, Evan Shelhammer, Jeff Donahue, Sergey Karayev, Jonathan Long, Ross Girshick, Sergio Guadarrama, Trevor Darrell**, *Caffe, convolutional architecture for fast feature embedding*," Cornell University archives, arXiv: 1408.5093v1 [cs.CV], 20 June 2014

52. [2014 Lin] **Min Lin, Qiang Chen, Shuicheng Yan**, *Network in Network*, arXiv:1312.4400v3 [cs.NE], Cornell university, Mar. 2014

53. [2014 Sak] **Hasim Sak, Andrew Senior, Francoise Beaufays**, *Long Short-Term Memory based recurrent neural network architectures for large vocabulary speech recognitio*n, arXiv:1402.1128v1 [cs.NE], Cornell University Library, Feb. 2014

54. [2014 Schmidhuber] **Jurgen Schmidhuber**, *Deep learning in Neural Networks: An overview*, Neural Networks, Vol 61, pp 85-117, Jan 2015. Also available from arXiv, Cornell Univ., Oct. 2014

55. [2014 Szegedy] **Christian Szegedy, Wei Liu, Yangqing Jia, Pierre Sermanet, Scott Reed, Dragomir Anguelov, Dumitru Erhan, Vincent Vanhoucke, Andrew Rabinovich**, *Going deeper with convolutions*, arXiv:1409.4842v1 [cs.CV], Cornell University, Sep. 2014

56. [2015 Abadi] **Martin Abadi, Ashish Agardwal, Paul Barham, Eugene Brevdo, Zhifeng Chen, Craig Citro, Greg S Corrado, Andy Davis,**

**Jeffrey Dean, Matthieu Devin et al.**, *TensorFlow: Large-Scale Machine Learning on Heterogeneous Distributed Systems”*, Preliminary White Paper, Nov. 2015

57. [2015 Akopyan] **Michael V DeBole, Brian Taba, Arnon Amir, Filipp Akopyan, Alexander Andreopoulos**, *TrueNorth: Design and tool flow of a 65mW 1 million neuron programmable neurosynaptic chip***”,** IEEE Transactions on Computer Aided Design of Integrated Circuits and Systems, 34(10), 1537-1557, DOI: 10.1109/ TCAD.2015.2474396, Oct. 2015

58. [2015 Courbariaux] **Matthieu Courbariaux and Jean-Pierre David,** “*Training Deep Neural Networks with Low Precision Multiplications*”, Workshop contribution at ICLR 2015, arXiv:1412.7024v5 [cs.LG], 23 Sep. 2015

59. [2015 He] **Kaiming He, Xiangyu Zhang, Shaoqing Ren, Jian Sun**, *Deep Residual Learning for Image Recognition*, arXiv:1512.03385v1 [cs.CV], Cornell University, Dec. 2015

60. [2015 LeCun] **Yann LeCun, Yann LeChun, Yoshua Bengio and Geoffrey Hinton**, *Deep Learning*, Review paper, Nature, Vol. 521, May 2015

61. [2015 Simonyan] **Karen Simonyan, Andrew Zisserman**, *Very deep convolutional networks for large-scale image recognition*, conference paper at ICLR (2015), arXiv:1409.1556v6 [cs.CV], 10 Apr. 2015

62. [2016 Aparanji] **V M Aparanji, Uday Wali and R Aparna**, *A Novel Neural Network Structure for Motion Control in Joints*, In 1st International Conference on Electronics, Communication, Computer Technologies and Optimization Techniques, Mysore, pp. 227-232, IEEE Xplore Digital Library, 2016

63. [2016 Chi] **Ping Chi, Shuangchen Li, Cong Xu, Tao Zhang, Jishen Zhao, Yongpan Liu, Yu Wang, Yuan Xie**, *PRIME: A Novel processing-in-memory architecture for neural network computation in ReRAM based main memory*, ACM/IEEE 43rd international symposium on computer architecture, 2016

64. [2016 Johann] **Sergio F Johann, Matheus T Moreira, Ney L V Calazans, Fabiano P Hessel**, *The HF-RISC processor: Performance assessment*, In 2016 IEEE 7th Latin American Symposium on Circuits & Systems (LASCAS), Feb.-March 2016

65. [2016 Liu] **Shaoli Liu**, **Zidong Du, Jinhua Tao, Dong Han, Tao Luo, Yuan Xie, Yunji Chen, Tianshi Chen,** *Cambricon: An Instruction Set Architecture for Neural Networks*, ACM/IEEE 43rd Annual International Symposium on Computer Architecture, pp. 393-405, 2016

66. [2016 Mrazek] **Vojtech Mrazek, Syed Shakib Sarwar, Lukas Sekanina, Zdenek Vasicek, Kaushik Roy**, *Design of Power-Efficient Approximate Multipliers for Approximate Artificial Neural Networks*, In International Conference on Computer-Aided Design, Austin, TX, USA, November 2016

67. [2016 NVIDIA] **White paper, NVIDIA** GeForce 1080, NVIDIA https://international.download.nvidia.com/geforce-

com/international/pdfs/GeForce_GTX_1080_Whitepaper_FINAL.pdf, 2016 (Last accessed 17-06-2019)

68. [2016 Sawada] **Jun Sawada, Filipp Akopyan, Andrew S Cassidy, Brain Taba, Michael V Debole et al.**, *TrueNorth Ecosystem for Brain-Inspired Computing: Scalable Systems, Software, and Applications*, In SC'16: Proceedings of the International Conference for High Performance Computing, Networking, Storage and Analysis, 13-18, Nov. 2016

69. [2016 Schabel] **Joshua Schabel, Lee Baker, Sumon Dey, Weifu LI, Paul D Franzon**, *Processor-in-memory support for artificial neural networks*, IEEE International Conference on Rebooting Computing (ICRC), 17-19 Oct. 2016

70. [2016 Silver] **David Silver, Aha Huang, Chris J Maddison, Arthur Guez, Laurent Sifre, George van den Driessche, Julian Schrittwieser, Ioannis Antonoglou, Veda Panneershelvam, Marc Lanctot, Sander Dieleman, Dominik Grewe, John Nham, Nal Kalchbrenner, Ilya Sutskever, Timothy Lilicrap, Madeleine Leach, Koray Kavukcuoglu, Thore Graepel, Demis Hassabis**, *Mastering the game of Go with deep neural networks and tree search*, Article Nature, 529, pp 484-504, Jan. 2016

71. [2016 Tang] **Jiexiong Tang, Chenwei Deng, Guang-Bin Huang**, *Extreme Learning Machine for Multilayer Perceptron*", IEEE transactions on Neural Networks and learning systems, Vol. 27, Issue 4, pp 809-821, 2016

72. [2016 Tsai] **Wei-Yu Tsai, Davis R Barch, Andrew S Cassidy, Michael V DeBole et al.**, *Always-on Speech Recognition using TrueNorth, a Reconfigurable*, Neurosynaptic Processor, IEEE Transactions on Computers, Vol. 66, No. 6, pp. 996-1007, 2016. DOI: 10.1109/tc.2016.2630683

73. [2016 Wu] **Jiaxiang Wu, Cong Leng, Yuhang Wang, Qinghao Hu, Jian Cheng**, *Quantized Convolutional Neural Networks for Mobile Devices*, arXiv:1512.06473v3 [cs.CV], 16 May 2016

74. [2017 Ajayi] **Tutu Ajayi, Khalid A1-Hawaj, Aporva Amarnath, Steve Dai, Scott Davidson, Paul Gao, Gai Liu, Anuj Rao, Austin Rovinski, Ningxiao Sun, Christopher Torng, Luis Vega, Bandhav Veluri, Shaolin Xie, Chun Zhao, Ritchier Zhao, Christopher Batten, Ronald G Dreslinski, Rajesh K Gupta, Michael B Taylor, Zhiru Zhang**, *Experiences using the RISC-V Ecosystem to Design an Accelerator-Centric SoC in TSMC 16nm*, In 1st workshop on Computer Architecture Research with RISC-V (CARRV 2017), Aug. 2017

75. [2017 Aparanji] **V M Aparanji, U V Wali and R Aparna,** *Robotic motion control using machine learning techniques*, Int Conf Communications and Signal Processing, Apr 2017, pp 1241-45 Melmarvathur, India,DOI: 978-1-5090-3800-8/17, IEEE (2017)

76. [2017 Chen] **Yu-Hsin Chen, Tushar Krishna**, *Eyeriss: An energy-efficient reconfigurable accelerator for deep convolutional neural networks,*

IEEE Journal of Solid-State Circuits, Vol. 52, Issue 1, pp. 127-138, Jan. 2017

77. [2017 Cadence] **Cadence Blog**, Vision *C5 DSP for Standalone Neural Network Processing* https://community.cadence.com/cadence_blogs_8/b/breakfast-bytes/posts/vision-c5, May 2017

78. [2017 Gunning] **David Gunning,** *XAI- Explainable Artificial Intellignece,* DARPA/I20, 2017

79. [2017 Intel] **Intel,** *Choosing the Right In-Memory computing solution,* white paper, Intel, Sep. 2017

80. [2017 Jouppi] **Norman P. Jouppi, Cliff Young, Nishant Patil, David Patterson, Gaurav Agarwal, Raminder Bajwa et.al**, *In-Datacenter performance analysis of a Tensor Processing Unit*, In 44th International Symposium on Computer Architecture (ISCA), pp. 1-12, ACM, Toronto, ON, Canada, June 2017

81. [2017 Judd] **Patrick Judd, Jorge Albericio and Andreas Moshovos**, *Stripes: bit-serial deep neural network computing*, IEEE computer architecture letters, vol.16, no.1, Jan.-June 2017

82. [2017 Koster] **Urs Koster, Tristan Webb, Xin Wang, Marcel Nassar, Arjun Bansal, William Constable, Oguz H Elibol, Scott Gray, Stewart Hall, Luke Hornof, Amir Khosrowshahi, Carey Kloss, Ruby J Pai, Naveen Rao**, *Flexpoint: An adaptive numerical format for efficient training of Deep Neural Networks*, arXiv:1711.02213v2 [cs.LG], Cornell University, Dec. 2017

83. [2017 Li] **Guanpeng Li, Siva Kumar Sastry Hari, Michael Sullivan, Timothy Tsai, Karrthik Pattabiraman, Joel Emer, Stephen W Keckler**, *Understanding Error Propagation in Deep Learning Neural Network (DNN) Accelerators and Applications*, In ACM/IEEE Supercomputing Conference SC"17, 2017, Denver, Co, USA, Nov. 2017

84. [2017 NVIDIA] **NVIDIA**, *NVIDIA Tesla V100 GPU Architecture*, white paper, Aug. 2017

85. [2017 Rao] **Naveen Rao,** *Intel® Nervana™ Neural Network Processors (NNP) Redefine AI Silicon*, October 2017 https://www.intel.ai/intel-nervana-neural-network-processors-nnp-redefine-ai-silicon/?_ga=2.202615705.900532956.1508268435- [Last accessed 15-06-2019]

86. [2017 Sabour] **Sara Sabour, Nicholas Frosst and Geoffrey E Hinton**, *Dynamic Routing Between Capsules*, In: 31st conference on Neural Information Processing Systems, Long Beach, CA, USA (2017)

87. [2017 Sze] **Vivienne Sze, Yu-Hsin Chen, Joel Emer, Amr Suleiman, Zhengdong Zhang**, *Hardware for machine learning: challenges and opportunities*, Invited paper, 2017 IEEE Custom integrated circuits conference (CICC), 30th April – 3rd May 2017

88. [2017 Tao] **Tao Luo, Sholi Liu, Ling Li, et al.,** *DaDianNao: A Neural Network Super Computer*, IEEE Transactions on Computers, vol. 66, iss. 1, pp. 73-88, Jan 2017

89. [2017 Wathan] **Govind Wathan,** *ARM dynamIQ – Technology for the next era of compute, 21 March 2017*, Processors blog, ARM Community.

https://community.arm.com/developer/ip-products/processors/b/processors-ip-blog/posts/arm-dynamiq-technology-for-the-next-era-of-compute [Last accessed 15-06-2019]

90. [2017 Wang] **Chao Wang, Lei Gong, Qi Yu, Xi Li, Yuan Xie, Xuehai Zhou**, *DLAU: A scalable deep learning accelerator unit on FPGA*, IEEE transactions on computer-aided design of integrated circuits and systems, vol. 36, no.3, March 2017

91. [2018 Wave] *Wave Computing Extends AI Lead by Targeting Edge of Cloud Through Acquisition of MIPS*, June 15, 2018

92. [2018 Abdelouahab] **Kamel Abdelouahab, Maxime Pelcat, Francois Berry**, *The challenge of Multi-Operand Adders in CNNs on FPGAs; How Not to Solve It!*, SAMOS XVIII, Pythagorion, Samos Island, Greece Association for Computing Machinery, July 15–19, 2018

93. [2018 Aparanji] **V M Aparanji, Uday Wali and R Aparna**, *Pathnet: A neuronal model for Robotic Motion Planning*, Cognitive Computing and Information Processing, Springer Singapore, 2018

94. [2018 Davies] **Mike Davies, Narayan Srinivasa, Tsung-Han Lin, Gautham Chinya, Yonqiang Cao, Sri Harsha Choday, et al.**, *Loihi,A Neuromorphic ManyCore Processor with on-chip learning*, IEEE Micro, Vol. 38, No. 1, pp 82-99, 2018

95. [2018 Elliott] **Robert Elliott and Mark O‟Connor**, *Optimizing Machine Learning Workloads on Power-efficient Devices*, White paper, ARM, 2018

96. [2018 Hill] **Parker Hill, Babak Zamirai, Shhengshuo Lu, Yu-Wei Chao, Michael Laurenzano, Meharzad Samadi Marios Papaefthymiou, Scott Mahlke, Thomas Wenisch, Jia Deng, Lingjia Tang, Jason Mars**, *Rethinking Numerical Representations for Deep Neural Networks*, arXiv:1808.02513v1 [cs.LG], Cornell Digital Library. Available online at https://arxiv.org/abs/1808.02513, Aug. 2018

97. [2018 Johnson] **Jeff Johnson**, *Rethinking floating point for deep learning*, arXiv:1811.01721vl [cs.NA], Nov. 2018

98. [2018 Juefei-Xu] **Felix Juefei-Xu, Vishnu Naresh Boddeti, Marios Savvides** *Perturbative Neural Networks*, arXiv digital library, Cornell University, 2018

99. [2018 Ortiz] **Marc Ortiz, Adrian Cristal, Eduard Ayguade, Marc Casas,** *Low-Precision Floating-Point Schemes for Neural Network Training*, arXiv:1804.05267v1 [cs.LG], 14 April 2018

100. [2018 Rodriguez] **Andres Rodriguez, Eden S, Etay Meiri, Evarist Fomenko, Young Jin K, Haihao S, Barukh Z,** *Lower numerical precision Deep Learning inference and training***,** White paper, Intel AI Academy, Jan. 2018

101. [2018 Vaughan] **Joel Vaughan, Agus Sudjianto, Erind Brahimi, Jie Chen and Vijayan N Nair**, *Explainable Neural Networks based on Additive Index Models*, arXiv:1806.01933v1 [stat.ML] Cornell University Library, June 2018

102. [2018 Videantis] **Videantis**, MP6000UDX Processor for deep learning, Available at https:/www.videantis.com/products, Jan. 2018

*v-MP6000UDX – Deep learning and Vision Processor* [Last accessed 27-07-2019]
http://www.drembedded.com/product/ai_v-mp6000udx

103. [2019 Bouvier] **Maxence Bouvier, Alexandre Valentian, Thomas Mesquida, Francois Rummens, Marina Reyboz, Elisa Vianello, Edith Beigne**, *Spiking Neural Networks hardware implementations and challenges: a survey*, ACM journal on emerging technologies in computing systems, vol. 15 issue 2, April 2019

104. [2019 DeBole] **Michael V DeBole, Brian Taba, Arnon Amir, Filipp Akopyan, Alexander Andreopoulos**, *TrueNorth: Accelerating from zero to 64 million neurons in 10 years*, Computer, Vol. 52, No. 5, pp. 20-29, 2019

105. [2019 Farrukh] **Fasih Ud Din Farrukh, Tuo Xie, Chun Zhang, Zhihua Wang**, *A Solution to Optimize Multi-Operand Adders in CNN Architecture on FPGA*, 2019 IEEE International Symposium on Circuits and Systems (ISCAS), Sapporo, Japan, 26-29 May 2019

106. [2019 Imani] **Mohsen Imani, Mohammad Samragh, Yeseong Kim, Saransh Gupta, Farinaz Koushanfar, Tajana Rosing,** *RAPIDNN: In-Memory Deep Neiral Network Acceleration Framework,* arXiv:1806.05794v4 [cs.NE], 11 April 2019

107. [2019 Kirin] Available online at http://www.hisilicon.com/en/Media-Center/News/Key-Information-About-the-Huawei-Kirin970
[Last accessed 06-09-2019]

108. [2019 Nagendra] **Manoj Nagendra**, *Samsung On-Device AI technology for AI Deep Learning announced*, News, Samsung Technology, 2019.
Available online at:
http://www.fonearena.com/blog/286030/samsung-on-device-ai-technology-for-ai-deep-learning.html, [Last accessed 06-09-2019]

109. [2019 Niu] **Wei Niu, Xiaolong Ma, Yanzhi Wang, Bin Ren**, *26ms Inference Time for ResNet-50: Towards Real-Time Execution of all DNNs on Smartphone*, arXiv:1905.00571v1 [cs.LG], 2 May 2019

110. [2019 NXP] **NXP Semiconductor,** *i-MX-RT600:RT600: Arm Cortex-M33-basedFamily of crossover processors*, Available at:
https://www.nxp.com/products/processors-and-microcontrollers/
arm-based-processors-and-mcus/i.mx-applications-processors/i.mx-rt-series/arm-cortex-m33-based-family-of- crossover- processors: i.MX-RT600 [Last accessed 06-09-2019]

# Additional References

1. [1989 Blum] **Avrim Blum, Ronald L Rivest,** *Training a 3-node neural network is NP-Complete,* Proceedings of Neural Information Processing Systems NIPS-1989, Vancouver, Canada, pp. 494-501, 1989
2. [1995 LeCun] **Y LeCun, L Jackel, L Bottou, A Brunot, C Cortes, J Denker, H Drucker, I Guyon, U Muller, E Sackinger, P Simard, and V Vapnik**, *Comparison of Learning algorithm for handwritten digit recognition,* In Int. conference on artificial neural networks, pp. 53-60, 1995
3. [1995 Siegelmann] **Hava T Siegelmann, Eduardo D Sontag**, *On the computational power of neural nets,* Journal of computer and system sciences 50, 132-150, 1995
4. [2002 Defelipe] **Defelipe J, Alonso-Nanclares L, Arellano J**, *Microstructure of the neocortex: Comparative aspects*, Journal of neurocytology. 31, 299-316, 2002, DOI: 10.1023/A:1024130211265.
5. [2003 Siegelmann] **H T Siegelmann**, *Neural and super-Turing computing,* Minds and machines 13, 103-114, Kluwer academic publishers, Netherlands, 2003
6. [2006 Hinton] **Geoffrey Hinton**, To *recognize shapes, first learn to generate images*", 2006 Progress in brain research, 165:535–547, 2007
7. [2007 Hinton] **G. E. Hinton**, *Learning* multiple layers of representation, Trends in cognitive science, vol. 11 no. 10, pp:428-433, 2017
8. [2007 Vassiliadis] **N Vassiliadis, G Theodoridis, S Nikolaidis**, *Exploring opportunities to improve the performance of a reconfigurable instruction set processor*, International journal of electronics, Taylor & Francis, Vol. 94, issue 5, pp. 481-500, 2007
9. [2012 Esmaeil] **Hadi Esmaeilzadeh, Adrian Sampson, Luis Ceze, Doug Burger**, *Neural acceleration for general purpose approximate programs,* IEEE/ACM 45th international symposium on microarchitecture, 2012
10. [2014 Fischer] **Asja Fischer and Christian Igel.** *Training Restricted Boltzmann Machines: An Introduction.* Pattern Recognition Vol. 47, pp.25-39, 2014
11. [2015 Lin] **Tsung-Yi Lin, Michael Maire, Serge Belongie, Lubomir Bourdev, Ross Girshick, James Hays, Pietro Perona, Deva Ramanan, C Lawrence Zitnick, Piotr Dollar**, *Microsoft COCO: Common Objects in Context*", arXiv:1405.0312v3 [cs.CV], 21 Feb. 2015
12. [2015 Zhou] **Young Zhou, Jingfei Jiang**, *An FPGA-based Accelerator Implementation for Deep Convolutional Neural Networks*, In 2014 4th International Conference on Computer Science and Network Technology ICCSNT. 2015
13. [2016 Bo Wu] **Bo Wu, Steve Cox, Markus Willems**, *Software Development Kits (SDKs) for Proprietary Processors,* White paper, Synopsys, Jun. 2016

14. [2016 Cheung] **Kit Cheung, Simon R Schultz, Wayne Luk**, *NeuroFlow: A general purpose spiking neural network simulation platform using customizable processors*, Frontiers in Neuroscince, Vol. 9, Article 516, Jan.2016

15. [2016 Shafiee] **Ali Shafiee, Anirban Nag, Naveen Muralimanohar, Rajeev Balasubramonian, John Paul Strachan, Miao Hu, R Stanley Williams, Vivek Srikumar**, *ISAAC: A Convolutional Neural Network Accelerator with In-Situ Analog Arithmetic in Crossbars*, In ACM/IEEE 43rd Annual International Symposium on Computer Architecture, 2016

16. [2017 Lan] **Hui-Ying Lan, Lin-Yang Wu, Xiao Zhang, Jin-Hua Tao, Xun-Yu Chen, Bing-Rui Wang, Yu-Qing Wang,QiGuo, Yun-Ji Chen**, *DLPlib: a library for deep learning processor*, Journal of computer science and technology, March 2017

17. [2017 Chang] **Andre Xian Ming Chang, Aliasger Zaidy Purdue, Vinayak Gokhale Purdue, Eugenio Culurciello**, *Compiling deep learning models for custom hardware accelerators*, arXiv:1708.00117v2 [cs.DC], 10 Dec. 2017

18. [2017 Gokhale] **Vinayak Gokhale, Aliasger Zaidy, Andre Xian Ming Chang, Eugenio Culurciello**, *Snowflake: A model agnostic accelerator for deep convolutional neural networks*, arXiv:1708.02579vl [cs.AR], Aug. 2017

19. [2017 Venkata] **Swagath Venkataramani, Ashish Ranjan, Subarno Bamerjee, Dipankar Das, Sasikanth Avancha, Ashok Jagannathan, Ajaya Durg, Dheemanth Nagaraj, Bharat Kaul, Pradeep Dubey, Anand Raghunathan**, *SCALEDEEP: A scalable computer architecture for learning and evaluating deep networks*, In ACM/IEEE 44th annual international symposium on computer architecture, ISCA, 2017

20. [2017 Yasoubi] **Ali Yasoubi, Reza Hojabr**, *Power-efficient accelerator design for neural networks using computation reuse*, IEEE Computer architecture letters, vol.16, No.1, Jan.-June 2017

21. [2018 Bahou] **Andrawes AI Bahou, Geethan Karunaratne, Renzo Andri, Lukas Cavigelli, Luca Benini,** *XNORBIN: A 95 Tops/s/W Hardware Accelerator for Binary Convolutional Neural Network*, arXiv:1803.05849v1 [cs.CV], Mar. 2018

22. [2018 Cai] **Ruizhe Cai, Luhao Wang, Yanzhi Wang et al.,** *VIBNN: Hardware acceleration of Bayesian Neural Networks*, arXiv:1802.00822vl [cs.LG], 2 Feb. 2018

23. [2018 Carbon] **A Carbon, J M Philippe, O Bichler, R Schmit, B Tain, D Briand, N Ventroux, M Paindavoine et al.,** *PNeuro: a scalable energy-efficient programmable hardware accelerator for neural networks*, In 2018 Design, automation and test in Europe conference and exhibition (DATE), 19-23 March 2018

24. [2018 Elsayed] **Gamaleldin F Elsayed, Jascha Sohl-Dickstein and Ian Goodfellow**, "*Adversarial reprogramming of neural networks*", arXiv:1806.111145v2 [cs.LG], 29 Nov. 2018

25. [2018 Fischl] **Kate Fischl, Terrence C Stewart, Kaitlin Fair and Andreas G Andreou,** *Implementation of the Neural Engineering framework on the TrueNorth neurosynaptic system*, In 2018 IEEE Biomedical circuits and systems conference (BioCAS), 17-19 Oct. 2018

26. [2018 Guo] **Kaiyuan Guo, Shulin Zeng, Jincheng Yu, Yu Wang and Huazhong Yang,** *A Survey of FPGA Based neural network accelerator*, arXiv:1712.08934v2 [cs.AR], 15 May 2018

27. [2018 Kang] **Kang, Hyeong-Ju**, *Efficient fixed-point representation for ResNet-50 Convolutional Neural Network*, Journal of the Korea Institute of Infromation and Communication Engineering, Vol. 22, Issue 1, pp. 1-8, 2018

28. [2018 Li] **Zhizhong Li, Derek Hoiem**, *Learning without Forgetting*, IEEE Transactions on pattern analysis and machine intelligence, vol.40, no.12, Dec. 2018

29. [2018 Liu] **Liu, X., Kim, D. H., Wu, C., and Chen, D. (2018),** *Resource and data optimization for hardware implementation of deep neural networks targeting FPGA-based edge devices***,** Proceedings of the 20th system level interconnect prediction workshop on – SLIP'18, 2018

30. [2018 Park] **Eunhyeok Park, Dongyoung Kim, Sungjoo Yoo**, *Energy-efficient Neural Network Accelerator based on outlier-aware low-precision computation*, in ACM/IEEE 45$^{th}$ International Symposium on Computer Architecture, 2018

31. [2018 Saletore] **Vikram Saletore, Deepthi Karkada, Vamsi Sripathi, Kushal Datta, Ananth Sankaranarayanan**, *Boosting Deep Learning training and inference performance on Intel Xeon and Intel Xeon Phi processors*, Intel White paper, 2018

32. [2018 Schott] **Lukas Schott, Jonas Rauber, Matthias Bethge and Wieland Brendel**, *Towards the first adversarially robust neural network model on MNIST*, arxiv, Sep. 2018

33. [2018 Tajasob] **Sarvenaz Tajasob, Morteza Rezaalipour, Maoud Dehyadegari, Mahdi Nazm Bojnordi**, *Designing efficient imprecise adders using multi-bit approximation building blocks*, Presented at ISLPED'18, Seattle, WA, USA, July 23-25, 2018

34. [2018 Wijerante] **Sasindu Wijeratne, Sandaruwan Jayaweera, Mahesh Dananjaya and Ajith Pasqual,** *Reconfigurable Co-Processor architecture with limited numerical precision to accelerate deep convolutional neural networks*, In IEEE 29$^{th}$ International Conference on Application specific systems, architectures and processors (ASAP), 2018

35. [2018 Xilinx] **Xilinx**, *Accelerating DNNs with Xilinx Alveo Accelerator Cards*, white paper, October 2018

36. [2018 Zhu] **Yuhao Zhu, Matthew Mattina, Paul Whatmough,** *Mobile machine learning hardware at ARM: A systems-on-chip (SoC) perspective*, arXiv:1801.06274v2 [cs.LG], Feb. 2018

37. [2019 Henry] **G Gnenn Henry, Terry Parks**, *Processor with Architectural Neural Network Execution Unit*, US Patent, US 10,275,394 B2, Apr. 30, 2019

38. [2019 LeCun] **Yann LeCun**, *Deep Learning Hardware: Past, Present, and Future*, ISSCC 2019 IEEE International Solid-State Circuits Conf., pp 12-19, 2019

39. [2019 Shin] **Dongjoo Shin, Hoi-Jun Yoo**, *The heterogeneous deep neural network processor with a non-von Neumann architecture*, Proceedings of the IEEE, pp 1-16, Feb. 2019

40. [2019 Wu] **Ephrem Wu, Xiaoqian Zhang, David Berman, Inkeun Cho, John Thendean**, *Compute-Efficient Neural-Network Acceleration*, FPGA’19 proceedings of the 2019 ACM/SIGDA International Symposium on Field-Programmable Gate Arrays, pp 191-200, USA, Feb. 2019

41. [2019 Yang] **Zebin Yang, Aijun Zhang and Agus Sudjianto**, *Enhancing explainability of Neural Networks through architecture constraints*, arXiv:1901.03838v1 [stat.ML] Cornell University Library, Jan. 2019

42. [2019 Yu] **Ye Yu, Yingmin Li, Shuai Che, Niraj K. Jha, Weifeng Zhang**, *Software-Defined Design space exploration for an efficient AI accelerator architecture*, arXiv:1903.07676v1 [cs.DC], 18 Mar. 2019

# Appendix – 1

## Python Library for Image Recognition using ARN

As discussed in chapter 4, two-layer ARN is used for MNIST image recognition. All the simulations were conducted in a python language. As ARN is a new network model, no existing libraries were used. Libraries for ARN were built using python. *ARN.train()* module is used to train the ARN: the control parameters for this module are $\rho$ (rho), threshold (T) and training sample size. *ARN.test()* is used to test an image for recognition. *ARN.nodeTuning()* is used to select the resonance curves depending on the values of $\rho$, threshold and input value. It is worth noting that the output of a node is calculated using the approximation methods described in section 5.4.

In the first layer, input image of 28×28 pixels is divided into 16 tiles of 7×7 each. Therefore, the first layer has 49 inputs. Each tile is applied to first layer and a new node is created for every unique tile identified by ARN. Therefore, the output of first layer has 16 values, indicating the node index to which the tile is matched. These 16 values are used as input to second layer and again a new node is added for every unique combination of 16 values. As there are 10 digits (0 to 9), there would be 10 classes in second layer. The resonant values of each node in layer 1 and layer 2 are stored in a file as trained values. The same file is used during test.

During test, a test image is divided into 16 tiles of 7×7 each. Every tile is applied to existing nodes of ARN in first layer and the sequence of matched node indices is obtained. This sequence is applied to second layer and the output of all the existing nodes in each class is calculated. The class, to which this sequence is maximally matched, is considered as the winner class. Therefore, the test image will be recognized as the digit represented by the winner class.

I have also included pseudo-code for image recognition using ARN. The actual code is available on github. It is possible to change the values of training sample size, resonance control parameter and threshold before starting the training. The code is divided into smaller segments for easy understanding. I have also included the comments wherever necessary.

The pseudo codes are given in following tables.

| Sl. No. | Table label | Description |
|---|---|---|
| 1 | Table A1.1 | Pseudo code to train ARN for image recognition in MNIST database |
| 2 | Table A1.2 | Pseudo code for ARN library (train, test and node tuning) |
| 3 | Table A1.3 | Pseudo code to test ARN |

**Table A1.1 Pseudo code to train ARN for image recognition in MNIST database**

```
# Import the libries to perform basic I/O operations viz., reading a file,
displaying the images etc.
import matplotlib.pyplot as plt
import matplotlib.cm as cm
import numpy as np
import pandas as pd
from xlrd import open_workbook
import random
import pylab as pl
import ARN         #The code for this library is available in Table A1.2
import xlrd

#set the parameters for ARN
sampleSize= No. of training samples per class
threshold=  value of threshold
N_train=sampleSize*10                 #total number of training samples
rho= value of resonance control parameter
N_test= total number of test samples
imgIdx=random.randint(0,N_test-1)           #selecting an image for test

#file location for training samples
rows_train=pd.read_csv(("E:/sortedTrain50.csv"),header = None, nrows=N_train)


#Define a function to convert 28x28 image into 16 tiles of 7x7 each
def getTile(img, count,row,col,pixels):
    initialization
    #set the origin for each tile
    origin=28*row*count + 7*col
    repeat {
            r=0
            repeat {
                    #set the row origin for each tile
                    row_origin=origin+(r*28)
                    repeat {
                            append the pixel values
                           } until 7 columns are over
                    Increment r
                   }until r is less than 7
           }until origin is less than 784
    Return the tile

#present each image from the training samples and convert each image into 16
tiles using the getTile function
Repeat {
        For 4 rows
         For 4 columns
          getTile
```

```
        append the tile to array of size 49
        increment the imageRow
       }until all the images from training sample are over

#train ARN for these images
ARN.train()
#the code for this library is given in Table A1.2
```

**Table A1.2 Pseudo code for ARN library (train, test and node tuning)**

```
def train(sampleSize, samples_train,rho,N_train,threshold):
    #import the required libraries
    #define a function of creating a new node and checking the output of a node
    #the code for nodeTuning is given in Table A1.3

    def checkOutNode(xt,xm,rho):
       yPat=nodeTuning((xt-xm),rho)
       return yPat

_______#training in layer 1______________________
    #initialize the nodeIdx array of size 16 to store the node indices
    #create first output node for the first tile of an training image
    #present the next tiles to created nodes
    Repeat {
            For all the existing output nodes
              checkOutNode
            if there is no maximal and above threshold output
              store the input tile
              add a new node index to the nodeIdx array
              increment the node count
            else
              add the node index with maximum and above threshold output
           }until all the tiles are over
    #Store the nodeIdx and input tile to which a new node was created
    return

_______#training in layer 2______________________
    #divide the nodeIdx into 10 groups, each of size = sampleSize
    #create first output class of '0' for the first nodeIdx arrays
    #initialize the output array
    #Initialize the image count to 0

    Repeat {
             #Initialize the tile count to 0
             Repeat {
                      If nodeIdx match
                       append the node value to the output array
                      else
                       append '0' to the output array
                      increment the tile count
                    }until all 16 tiles are over
             Divide the sum of output array by 16
             If sum is < threshold
                Increase the node count
                Store the image as trained image
             Increment the image count
           }until all the images are over
```

```
    Repeat the above procedure for all 10 classes

    #at the end of training, excel file will be generated which has the pixel
values of each tile used for training in first layer and the nodeIdx for which
a new node was created for each class in second layer.  This can be used during
test.

#define function for test
define test():
           #import the required libraries
          #initialization
     nodeIdx_test=[]
     Present the test input

__________#first layer____________________
     nodeIdx_test=getTile()
__________#second layer___________________
     Apply nodeIdx_test to second layer
     Repeat{
            Check the output of all existing nodes
            Identify the node with maximum output value
           } for all 10 classes
     Identify one class, whose node has maximum output among others
     Declare the winner class as recognized digit

#define a function for node tuning
def nodeTuning(x, xm, rho):
           #Perform the following using PWL and SOI
              #Calculate the output of a node using y = X (1 - X), with X =
1/(1+exp(-rho*(x - xm)))
             #Store the output of a node for different values of rho

        choose the method of approximation
        compare the input x and rho with the LUT values
        if SOI:
           get the values of a, b and c
           calculate the value of output using y=(ax+b)x+c
        else:
           get the values of m, x1 and y1
           calculate the value of output using y=m(x-x1)+y1
        return y
```

**Table A1.3 Pseudo code to test ARN**

```
import the required libraries
set the values for the parameters rho and T (layer 1 and layer 2)
import the test file
select an image to test
get the trained values from the excel file generated after training
present the test image
    get 16 tiles of this image using getTile()
    apply these tiles to first layer
    get the node indices for test image
    apply the node index array to second layer
    compute the output of all the existing nodes for each class
    identify the node with maximum and above threshold output in each class
    identify the class with a node having maximum output
declare the winner class as the identified digit
```

# Appendix – 2

# Verilog library for Hardware modules

In this appendix, implementation details of the hardware modules as discussed in chapter 5 and chapter 6 including the pseudo code and the actual code wherever possible are given. The code was written in verilog Hardware Description Language. Xilinx ISE 9.2i webpack was used for code editing and simulation.

The codes/pseudo codes are given in following tables.

| Sl. No. | Table label | Description |
|---|---|---|
| 1 | Table A2.2 | Code for Priority Encoder (PE) |
| 2 | Table A2.3 | Code for Bit Select Logic (BSL) |
| 3 | Table A2.4 | Code for State transition |
| 4 | Table A2.5 | Code for busIO |
| 5 | Table A2.6 | Pseudo code for CCLU |
| 6 | Table A2.7 | Pseudo code for Serial Multiplier |
| 7 | Table A2.8 | Pseudo code for Sigmoid/tanh/ARN resonator |
| 8 | Table A2.9 | Code for count one's logic/ single column adder |
| 9 | Table A2.10 | Pseudo code for 4×M adder |
| 10 | Table A2.11 | Pseudo code for 16×16 adder using 4×16 adder |

## A2.1 Serial Multiplier

Serial multiplier is implemented using a classical method of shift and add. The basic operations required to implement serial multiplier are clear (CLR), load (LD), shift right (SR), shift left (SL), add (ADD) and subtract (SUB). The input CLR is used to clear the contents of multiplier, LD is used to load the inputs for multiplication. Once the two inputs are loaded, the multiplication will start. The algorithm of multiplication using shift and add is given in Table A2.1.

**Table A2.1 Algorithm for 16-bit Serial Multiplication**

```
Step1: Reset all the registers.  Initialize count=0

Step2: On load command, load the first input to SRReg (shift right register)

Step3: On mul command, load the second input to SLreg (shift left register)

Step4: Shift right the SReg and shift left the SLReg by one bit. Store the MSB
of SLReg in testB. Increment the count

Step5: if count=16, go to step 7, else If testB=1 go to step 6; else go to step
4 until count is equal to 16

Step6: Add SReg to tempMul, go to step 4 and continue until count is equal to
16

Step7: When count equals 16, transfer the contents of tempMul to the bus and
raise the mulEndFlag.
```

Serial multiplier has several sub modules viz., Priority Encoder (PE), Bit Selector Logic (BSL), state transition module, Command Control & Logic Unit (CCLU) and bus design.

a) The PE module was required to ensure that only input command is active at a time.
b) BSL module was required to select the input value corresponding to the active input command.
c) CCLU is used to perform the basic operations viz., CLR, LD, SR, SL, ADD, SUB.

These modules will be discussed in the following sections.

## A2.1.1 Priority Encoder

Priority Encoder (PE) is used to select the input command depending on the priority. When more than one input command is active at a time, the one with higher priority will be selected by the PE module. The block diagram of PE is as shown in Figure A2.1.

The design is such that the input commands are given a priority in an increased order: CLR being the first input, given a highest priority and the last input SUB is given lowest priority. When multiple inputs are active at a time, the first input will only be activated: subsequent gates are enabled only when the previous inputs are inactive. When none of the input commands are active, it is considered as NOP.

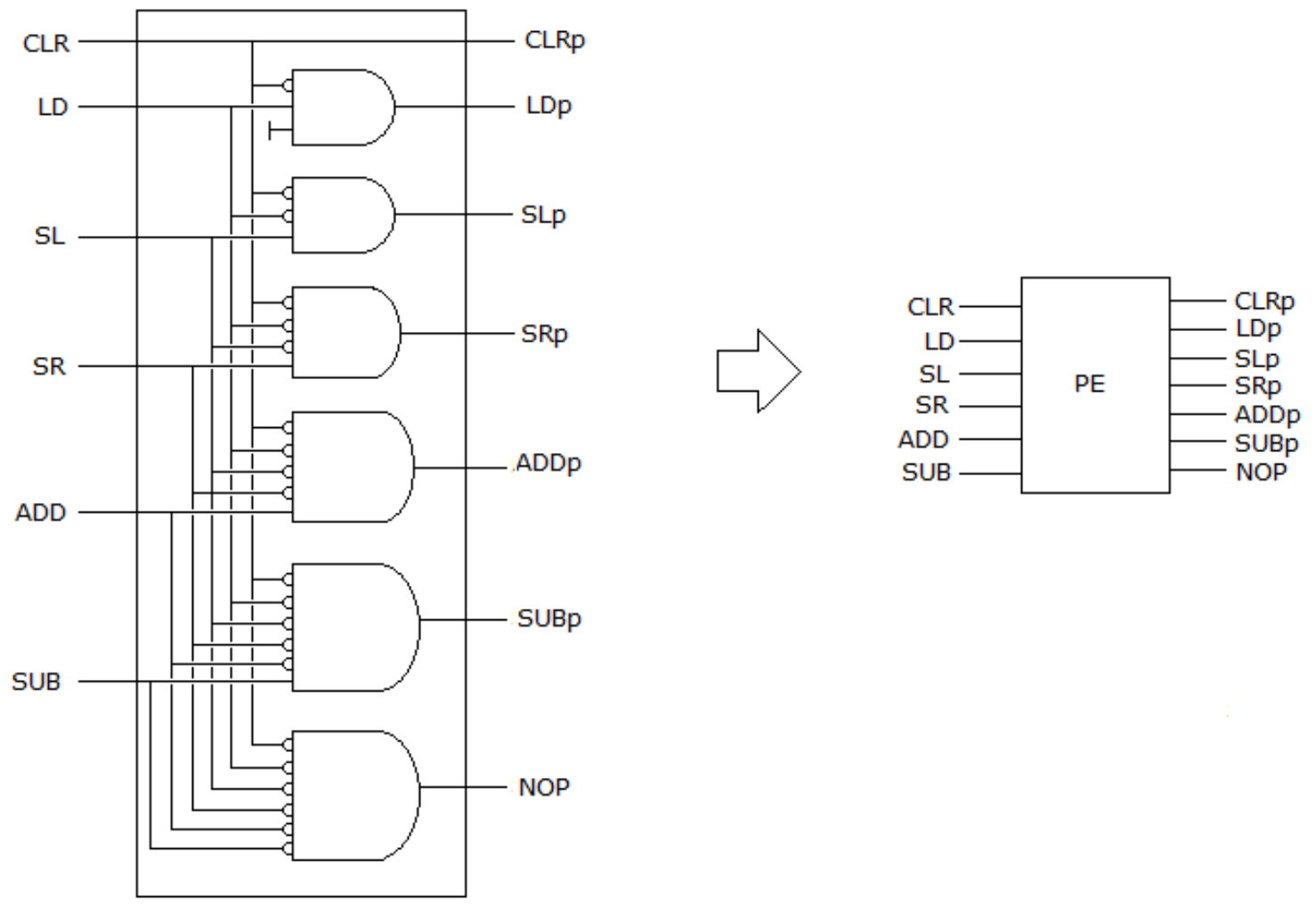

**Figure A2.1 Block Diagram of Priority Encoder (PE)**

Note that, the use of priority encoder is not required but is used only to resolve the conflict in case multiple lines are selected. Generally this should not occur but having a priority encoder will ensure that only one command is executed at a time.

**Table A2.2 verilog code for Priority Encoder (PE)**

```
/* Description: This module is a part of serial multiplier module. It is used
to select only one input when more than one inputs are active at a time. Input
is selected based on the priority. The priority is in the following order:
CLR, LD, SL, SR, ADD, SUB. When none of the inputs are active, it is
considered as No-operation (NOP).
The logic for this module is given in the Figure A2.1 */

//Instance of the module
PE(CLR,LD,SL,SR,ADD,SUB,CLR_p,LD_p,SL_p,SR_p,ADD_p,SUB_p,NOP_p);

//Implementation
PE(in1,in2,in3,in4,in5,in6,out1,out2,out3,out4,out5,out6,out7);

//IO declaration
//Constructing a logic for Figure A2.1
   buf b1(out1,in1);
   and a1(out2,in1,~in1,1'b1);
   and a2(out3,in3,~in2,~in1);
   and a3(out4,in4,~in3,~in2,~in1);
   and a4(out5,in5,~in4,~in3,~in2,~in1);
   and a5(out6,in6,~in5,~in4,~in3,~in2,~in1);
   and a6(out7,~in6,~in5,~in4,~in3,~in2,~in1);
endmodule
```

## A2.1.2 Bit Selector Logic

Bit Selector Logic (BSL) is used to select the input value corresponding to the active input command. The inputs to BSL are the prioritized outputs from PE. Input values $D_i$, $D_{i-1}$ and $D_{i+1}$ represent the present, previous and next bit of the N bit input respectively. Input command CLR will select the input '0' to clear the values, SR command will select the previous bit represented as $D_{i-1}$, SL command will select the next bit represented as $D_{i+1}$, ADD and SUB will select the sum and difference values: sum and difference are calculated using the full adder and full subtractors. NOP will select the present input i.e., $D_i$. Therefore, the output of BSL will have any one of the above inputs corresponding to the active input command. The block diagram of BSL is as shown in Figure A2.2.

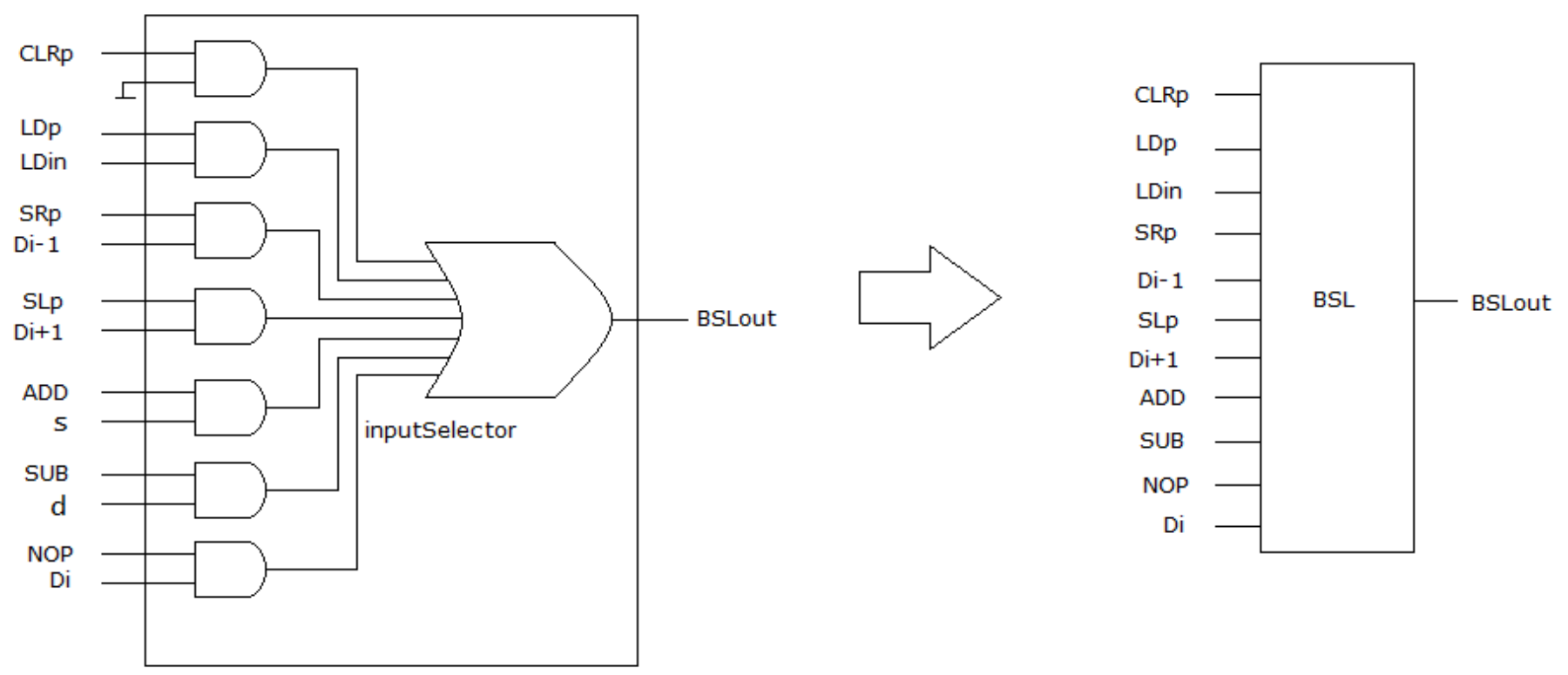

**Figure A2.2 Block Diagram of Bit Select Logic (BSL)**

**Table A2.3 Bit Select Logic (BSL)**

```
/* Description: This module is a part of serial multiplier module.  It is used
to select the input value corresponding to the active input command.  The logic
for this module is given in the Figure A2.2 */

//Module instance
module BSL(CLRp, LDp, LDi, SLp, D_next, SRp, D_pre, ADD, s, SUBp, d, NOP, Di,
BSLout);

//Implementation
module BSL(in1, in2, in3, in4, in5, in6, in7, in8, in9, in10, in11, in12, in13,
out1);

//IO declaration
//constructing logic
   wire c,l,s,r,a,d,n;
   and a7(c,in1,1'b0);
   and a8(l,in2,in3);
   and a9(s,in4,in5);
   and a10(r,in6,in7);
   and a11(a,in8,in9);
   and a12(d,in10,in11);
   and a13(n,in12,in13);
   or o1(out1,c,l,s,r,a,d,n);
endmodule
```

## A2.1.3 State transition module

State transition is one of the important modules in the hardware design. The block diagram and state transition diagram of state transition is shown in Figure A2.3.

It has four inputs viz., *clk*, *reset*, *setState* and *setStateValue*, with *reset* as the higher priority input. When a *setState* command is active, the value stored in *setStateValue* will be assigned to the state. The clock input is dual edge triggered. D flipflop is used to store the values of present state (*preState*) and next state (*nextState*).

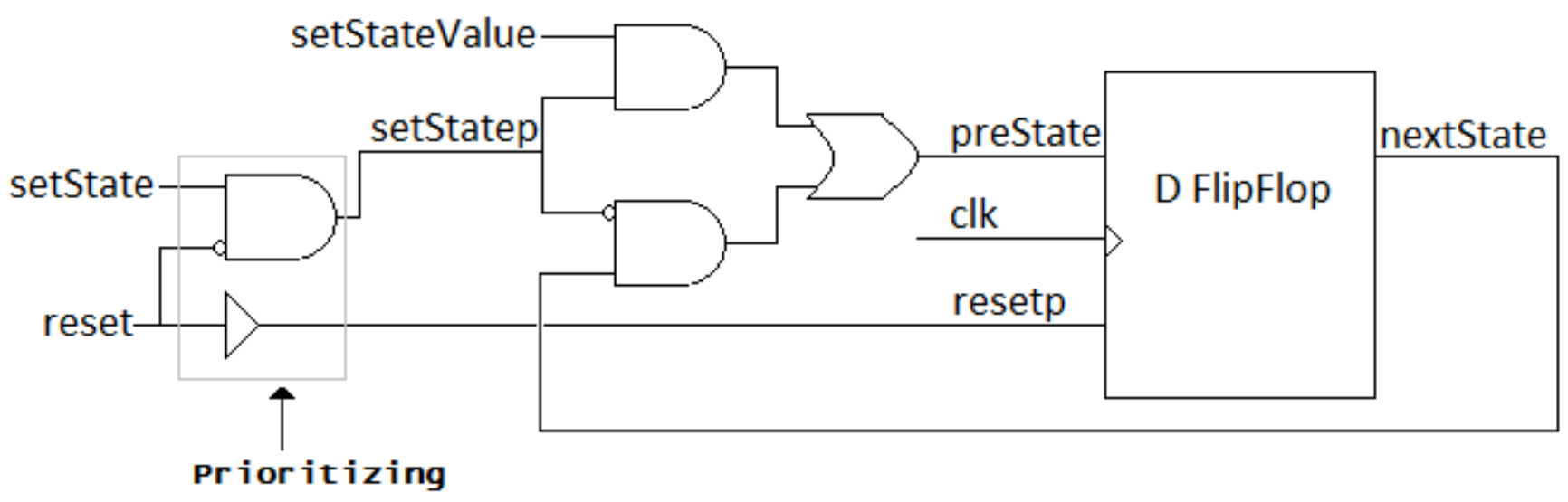

**(a)**

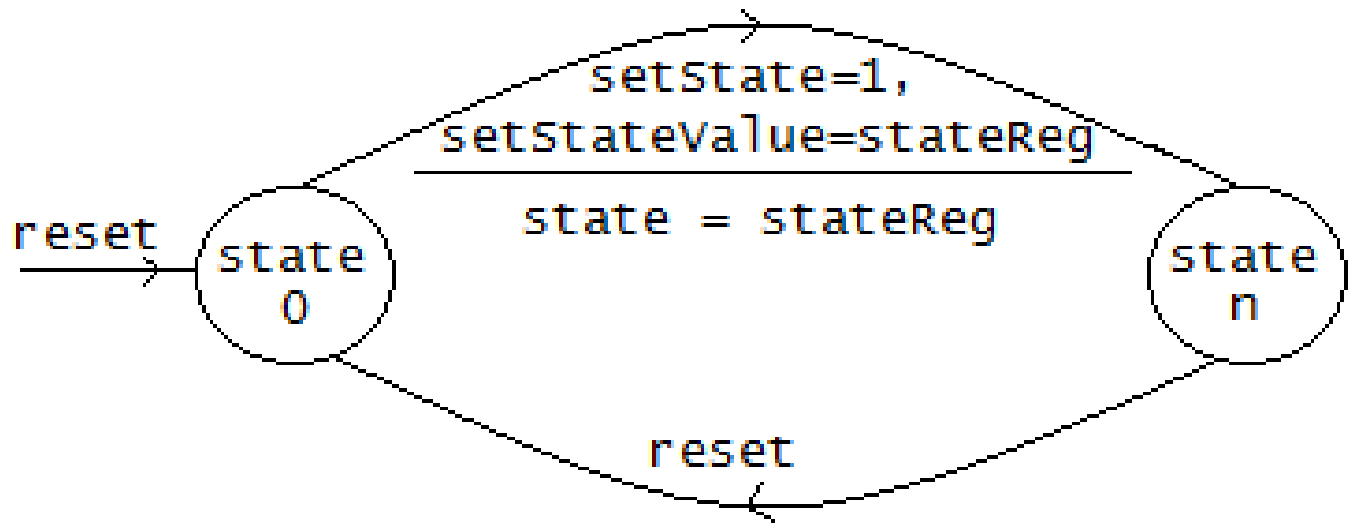

**(b)**

**Figure A2.3 State transition module a) block diagram b) FSM**

The design is such that, on *reset* command the state will be reset to 0 and on *setState* command, state will be the value of a *stateReg*. With this method it is possible to have any number of states.

**Table A2.4 State Transition module**

```
/* Description: This module is used for state transition.  It is used in most
of modules like serial multiplier, activation function etc.  The logic for this
module is given in the Figure A2.3(a) */
//Module instance
module stateT(clk, reset, setState, setStateValue, state);

//Implementation
module stateT(in1, in2, in3, in4, out1);
//IO declaration
//Proritization
   buf b1(in2_p, in2);
   and a2(in3_p, ~in2, in3);

   wire [2:0] PS;
   bufif1 b1[2:0](PS, in4, in3_p);
   bufif0 b2[2:0](PS, out1, in3_p);
   dffState d1[2:0](PS, in1, in2_p, out1);
endmodule

# D- Flipflop
//Module instance
module dffState (clk,reset,d, st);

//Implementation
module dffState (in1, in2,in3,out);
```

```
//IO declaration
   initial
   begin
      out = 0;
   end
   always@(posedge in1 or negedge in1 or posedge in2)
   begin
      if(in1)
         out = 0;
      else
      begin
         #10
         out = in3;
      end
   end
endmodule
```

### A2.1.4 Prioritized Bus Design

Bus is a common module used for data transfer. It will have some input ports and output ports. Input ports will write the data to the bus, while output ports will read the data from the bus. In order to eliminate the conflicts between multiple inputs ports, priority has been applied to the input ports such that first input port will have the highest priority, while the last input port will have least priority.

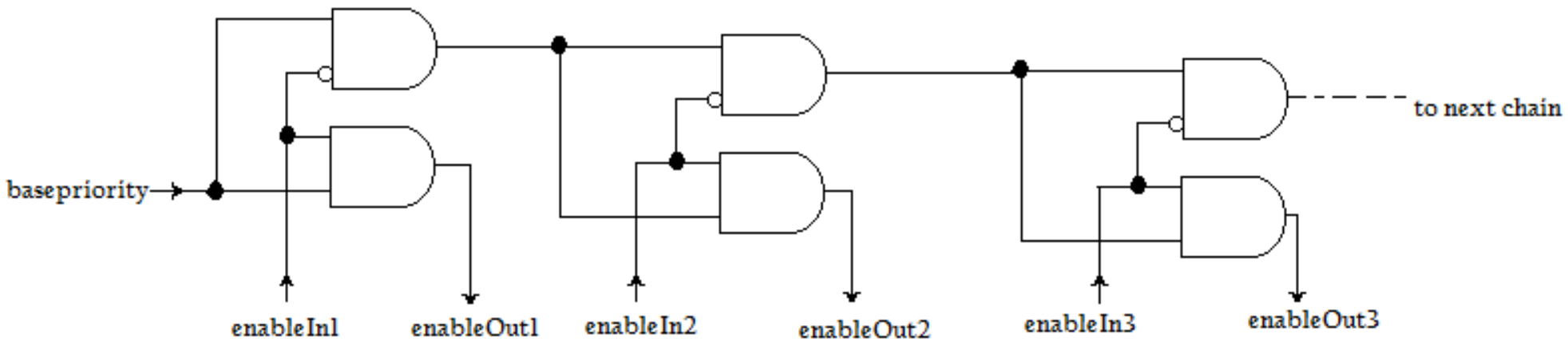

**Figure A2.4 Structure of a priority chain**

Consider the priority chain shown in Figure A2.4, there are three enable signals (*enableIn1* to *enableIn3*) corresponding to three input ports. Base priority is hardwired to logic 1. When the first input port is enabled i.e., *enableIn1*=1, *enableOut1* will be 1, which can then be used as enable signal for port 1 to transfer the data to bus. When *enableIn1* is disabled, second chain will be activated and second input port will be enabled to write the data when *enableIn2* is enabled. When more than one enable signals are active at a time, say for example, *enableIn1* and *enableIn2* are active, only first chain is activated, while the second chain is disabled (because the first input port has highest priority). Therefore, even though multiple enable signals are active a time, the port with highest priority will be allowed to transfer the data to bus.

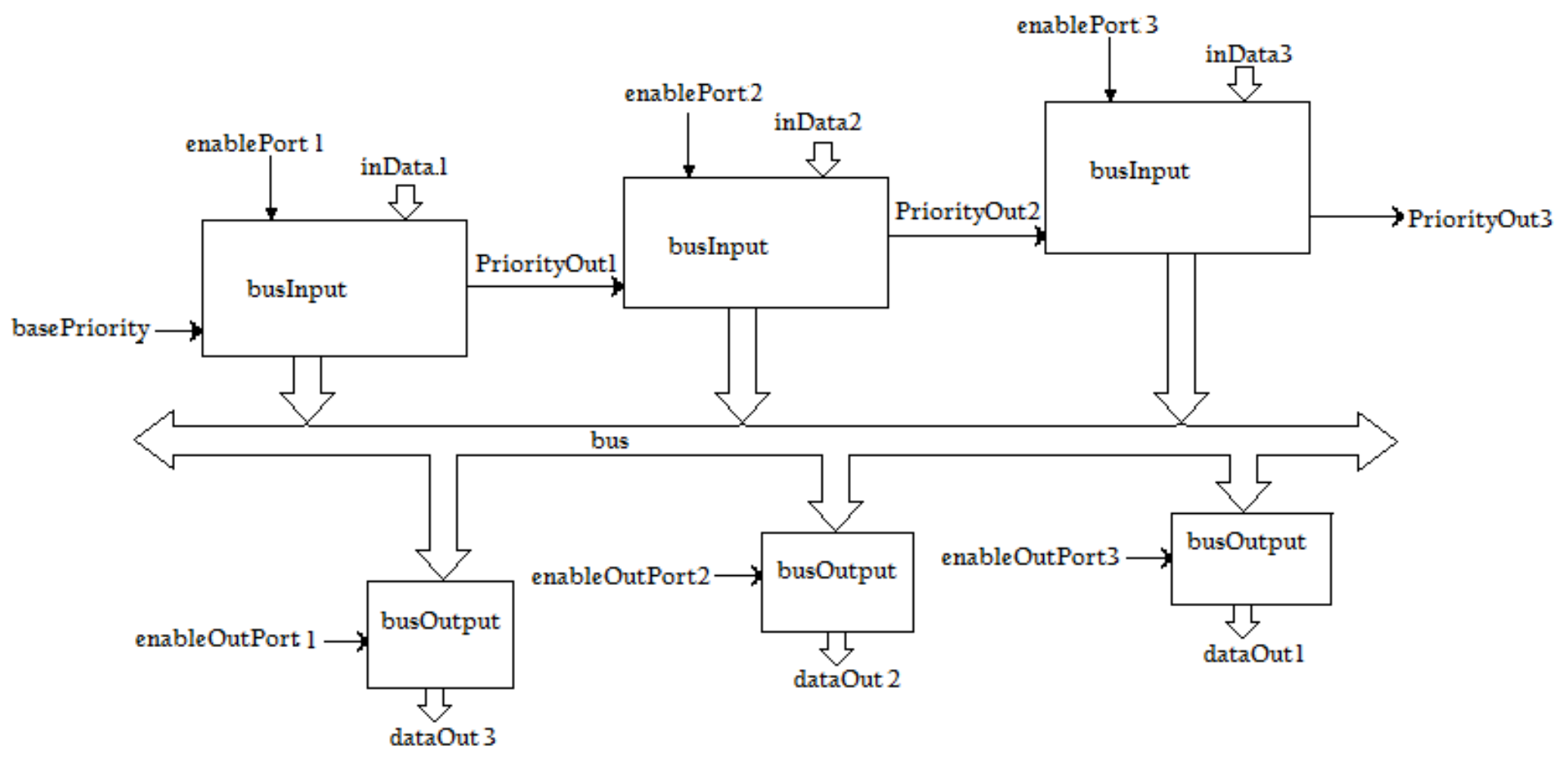

**Figure A2.5 Structure of a bus with input and output ports**

On the other hand, this priority design is not required for the output ports: reading of a data from a bus will not overwrite the existing data on the bus. The Figure A2.5 shows the bus with 3 input ports and 3 output ports. The number of ports can be changed during the design.

**Table A2.5 Bus IO**

```
/* Description: This module is used for bus IO.  The number of input and output
ports to bus can be changed as required.  The Figure A2.5 is used as the
reference for this code */

// module instance
module busIO(bus, inP1, inP2, writeEn1, writeEn2, outP1, outP2, readEn1,
readEn2);

//Implementation
module busIO(out, in1, in2, in3, in4, in5, in6, in7, in8);
//IO declaration

   busIn in1 (out, in1, in2, in3, in4);
   busOut out1 (out, in5, in6, in7, in8);
endmodule

// sub modules instance
busIn(bus, inData1, inData2, enablePort1, enablePort2);
busInput(bus, inData, enablePort, priorityIn, priorityOut);
PriorChain(priority_in, enable_in, priority_out, enable_out);
busOut(bus, dataOut1, dataOut2,enableOutPort1, enableOutPort2);
busOutput(dataOut, bus, enableOutPort);

// implementation
module busIn(out, in1, in2, in3, in4);
//IO declaration
   wire priorityOut1, priorityOut2;
   inout tri [16:0]out;
   reg basePriority;

   initial
```

```
   begin
      basePriority = 1'b1;
   end

   busInput inPort1 (out, in1, in3, basePriority, priorityOut1);
   busInput inPort2 (out, in2, in4, priorityOut1, priorityOut2);
endmodule

module busInput(out1, in1, in2, in3, out2);
//IO declaration
   wire pEnable;
   wire [1:0]fanout;
   PriorChain p(in3, in2, in4, pEnable);
   assign fanout = {pEnable, pEnable};
   bufif1 b0[15:0] (out1, in1, pEnable);
endmodule

module PriorChain(in1, in2, out1,out2);
//IO declaration
   and a1(out1, ~in2, in1);
   and a2(out2, in2, in1);
endmodule

module busOut(out, in1, in2, in3, in4);
//IO declaration

   busOutput bo1(in1, out, in3);
   busOutput bo2(in2, out,  in4);
endmodule

module busOutput(out, in1, in2);
//IO declaration

   bufif1 buf1[15:0](out, in1, in2);
endmodule
```

## A2.1.5 Command Control & Logic Unit

Command Control & Logic Unit is designed to perform the basic operations like load (LD), clear (CLR), shift right (SR), shift left (SL), addition (ADD) and subtract (SUB). It is also used to control individual operations of the serial multiplier module. The block diagram of CCLU is as shown in Figure A2.6.

**Table A2.6 Pseudo code for CCLU**

```
module declaration
    I/O declaration
    Initialization
    Instantiate the following modules
....Figure A2.6 can be used a reference for this code
    PE module to prioritize the input commands
    BSL module to select the input corresponding to the active input command
    D flip-flops to store the result of BSL
    cascade connection of above modules for n-bit CCLU
    wait for output to be available on output port and set endF=1
endmodule
```

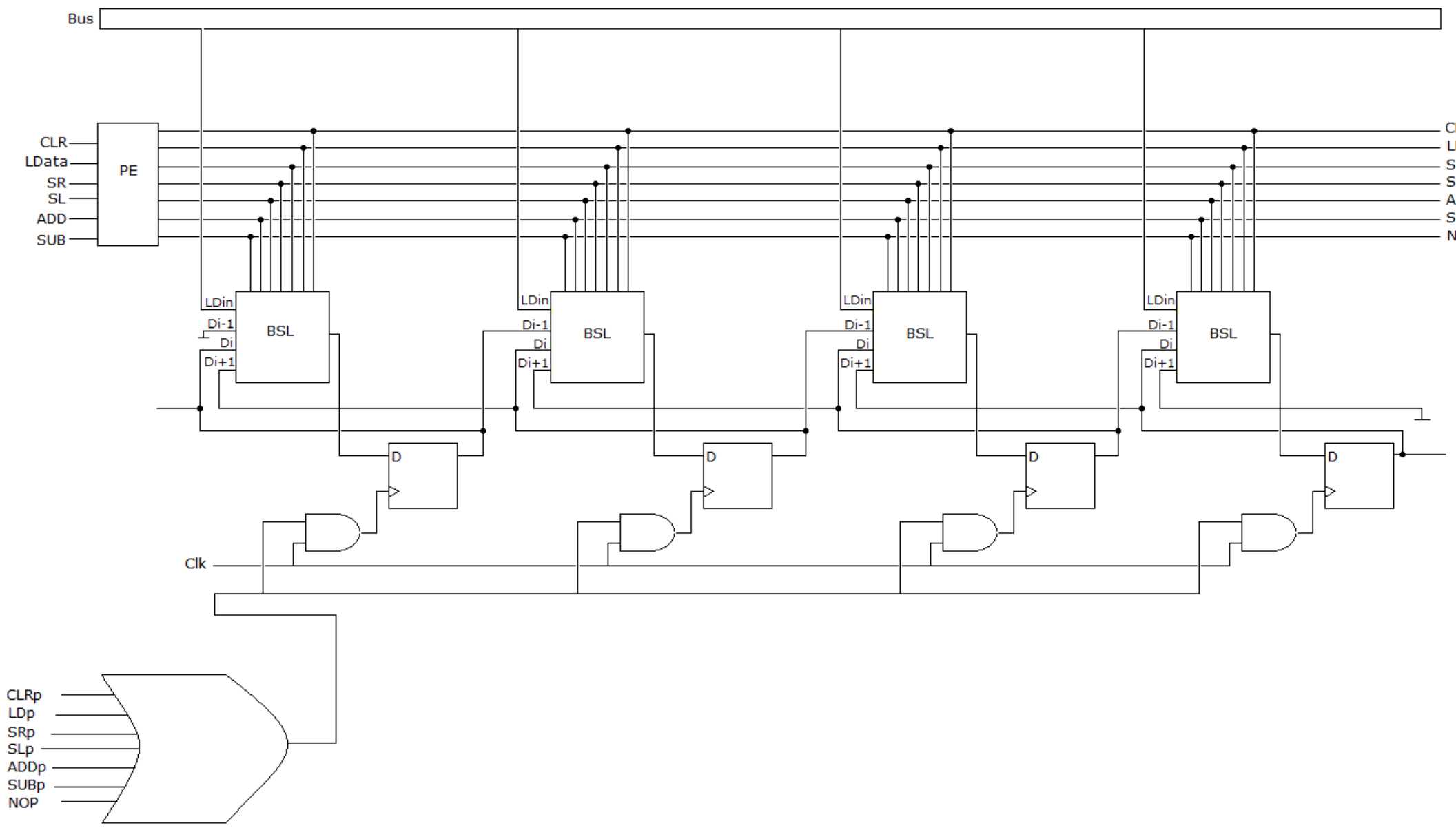

**Figure A2.6 Block Diagram of 4-bit Command Control Logic Unit (CCLU)**

All the above discussed modules are combined to make a serial multiplier. The block diagram and FSM of serial multiplier is shown in Figure A2.7. The input commands to the multiplier are *clk*, *reset*, *ld* and *mul*. When the reset command is active, all the registers of the multiplier module will be reset to '0', endFlag will be cleared to perform the next multiplication. When *ld* command is active, first input is loaded from the bus. When *mul* command is active, second input is loaded and the multiplication will start. On completion, multiplied result is stored in *tempMul* and the *endFlag* is set to indicate that the result is available on the bus. The psuedocode to implement serial multiplier is shown in Table A2.7.

**Table A2.7 Pseudo code for Serial Multiplier**

```
module declaration
    I/O declaration
    Initialization
    Instantiate the following modules
         BusIO
         State transition
         CCLU
...Figure A2.7 can be used as reference for this code
    Initialization, SReg, SLReg, tempMul, result=0
    Take two inputs
    Load the first input on load command, SReg=input1
    Load the second input on mul command and start the multiplication,
SLReg=input2
    Repeat{
           Shift right the first input
           Shift left the second input
```

```
            If the MSB of left input = 1:
                tempMul+=SReg
           } for N-bits
    Transfer the result of tempMul to bus
Endmodule
```

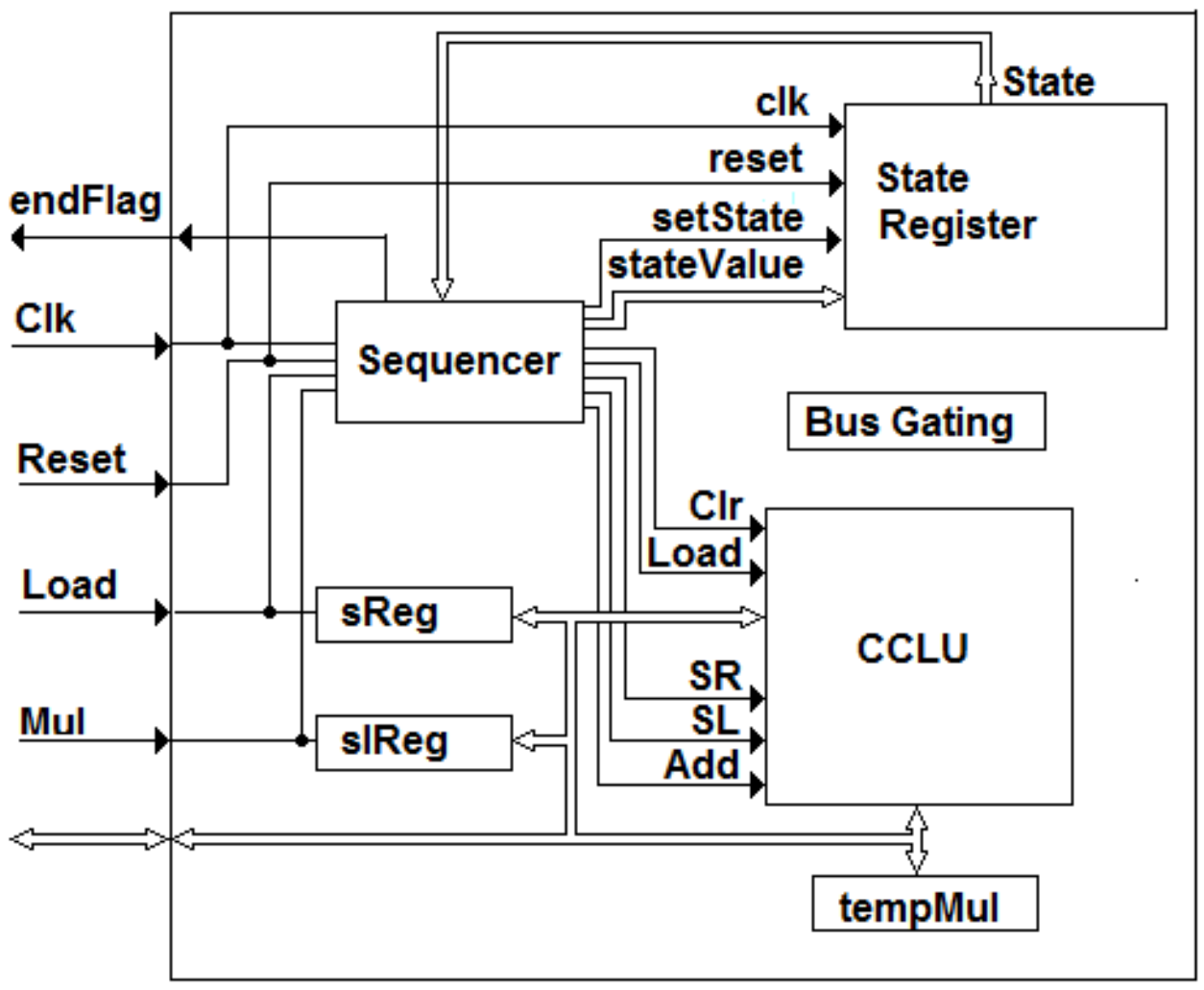

(a)

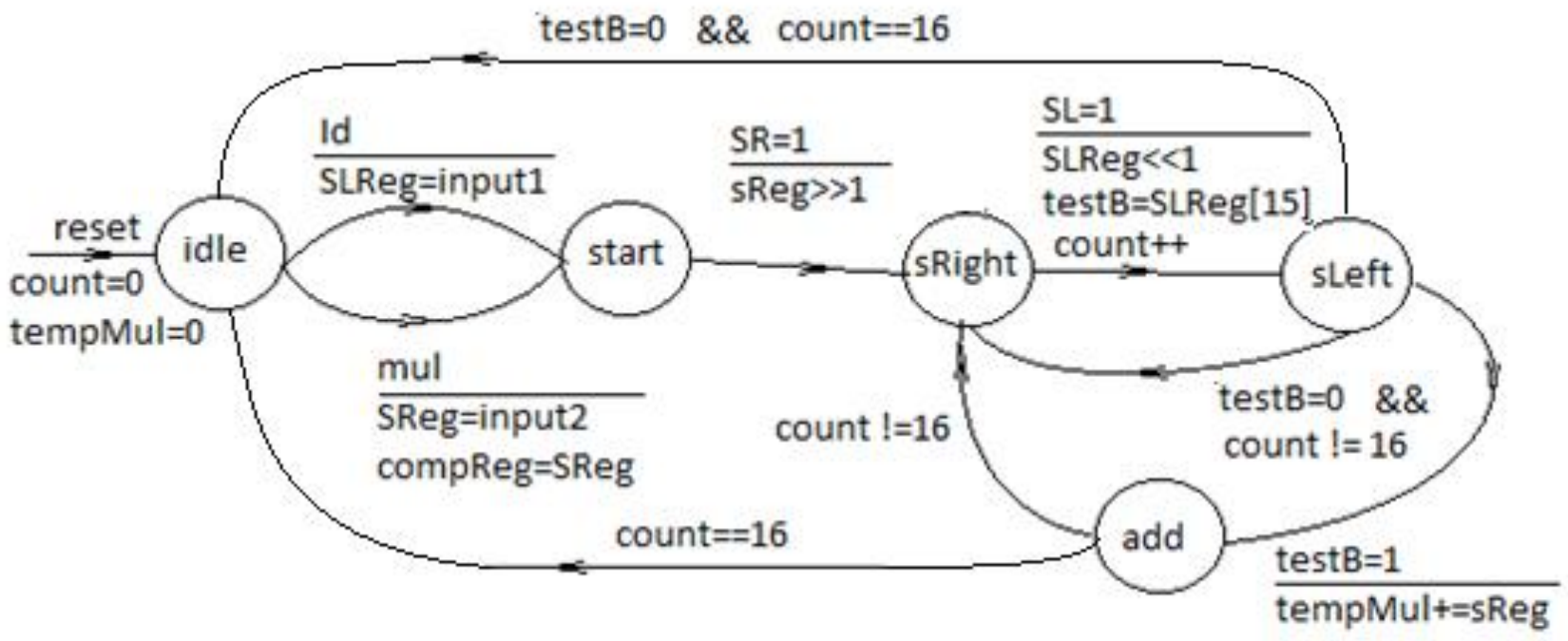

(b)

**Figure A2.7 Serial Multiplier a) Block diagram b) FSM**

## A2.2 Wallace Tree Multiplier

Parallel multipliers, also considered as fast multipliers, use combinational circuits like Full Adders and Half Adders for the addition of partial products (see Figure A2.8). The performance of parallel multipliers is determined by the number of partial products. The structue shown in Figure A2.8 is used to implement 16-bit wallace multiplier [1964 Wallace]. The implmentation uses full adders and half adders. The inputs and outputs of this implementation are represented using the 16-bit number format discussed in section 5.2. The simulation result of Wallace tree multiplier (16-bit) is shown in Figure A2.9.

The performance comparison of serial and parallel multipliers is discussed in section 5.3.2.

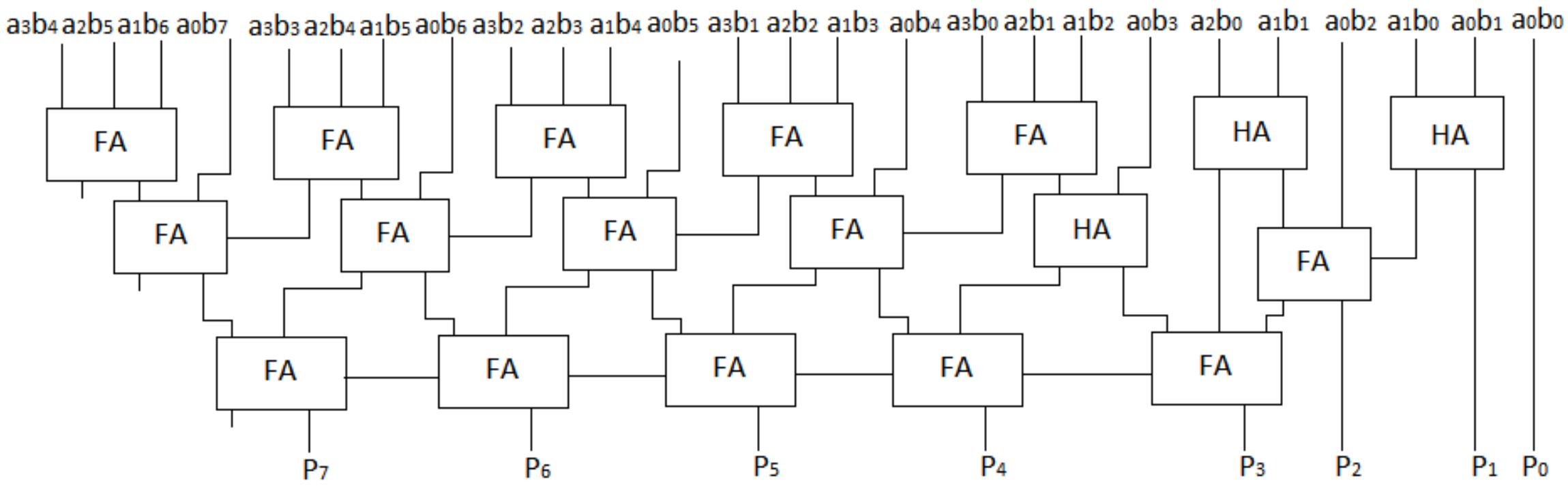

**Figure A2.8 Structure of 4-bit Wallace tree Multiplier**

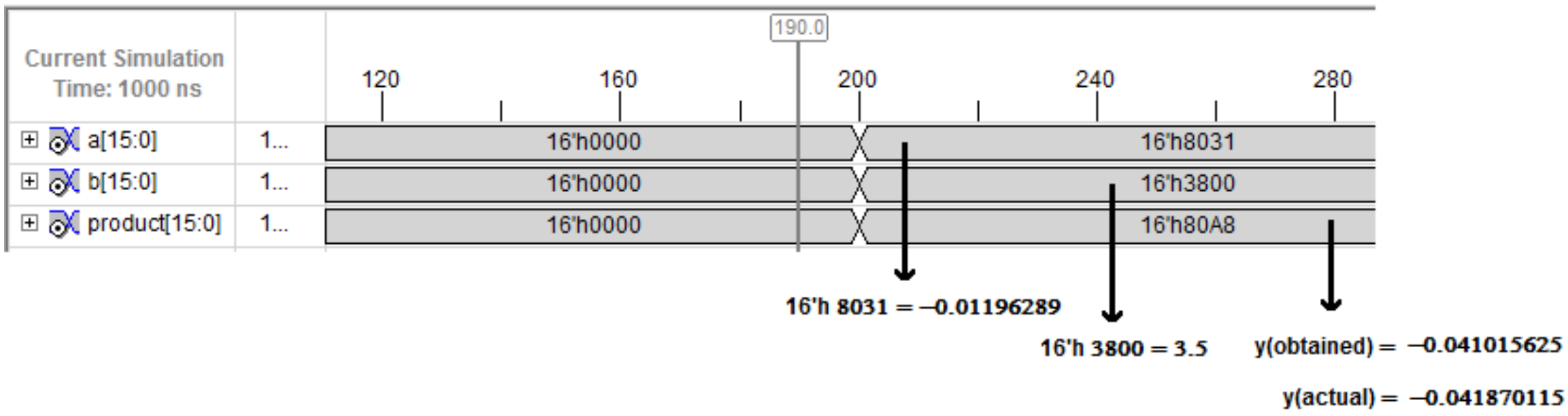

**Figure A2.9 Simulation result of 16-bit Parallel Multiplication**

## A2.3 Sigmoid Approximation using PWL

The approximation methods described in section 5.4 are used to calculate the output of a sigmoid function. In this section, PWL method is described in detail by considering different spacing between two known points. The graph of sigmoid shown in Figure 5.7 is symmetric about x axis and therefore, for simplicity, the curve in the first quadrant is considered for all the analysis and implementation.

### A2.3.1 Uniform Spacing of 0.5

Two known points on the curve are taken at uniform spacing of 0.5. For example, if the range of x is 0:1, then the known points are at x=0, 0.5 and 1. The output of these points are calculated using the eqn. (5.2). For example, to calculate the output when x=0.25, we need to use known points $x_1$=0, $y_1$=0.5 and $x_2$=0.5, $y_2$=0.6225, as x is in the range $x_1 < x < x_2$. The approximation curve of sigmoid at a uniform spacing of 0.5 is shown in Figure A2.10. This has maximum relative error of 0.3838%, calculated using the formula given in eqn. (A2.1).

$$\%Error = \frac{|Approximated\ \ value - Theoretical\ \ value\ |}{Theoretical\ \ value} * 100 \qquad (A2.1)$$

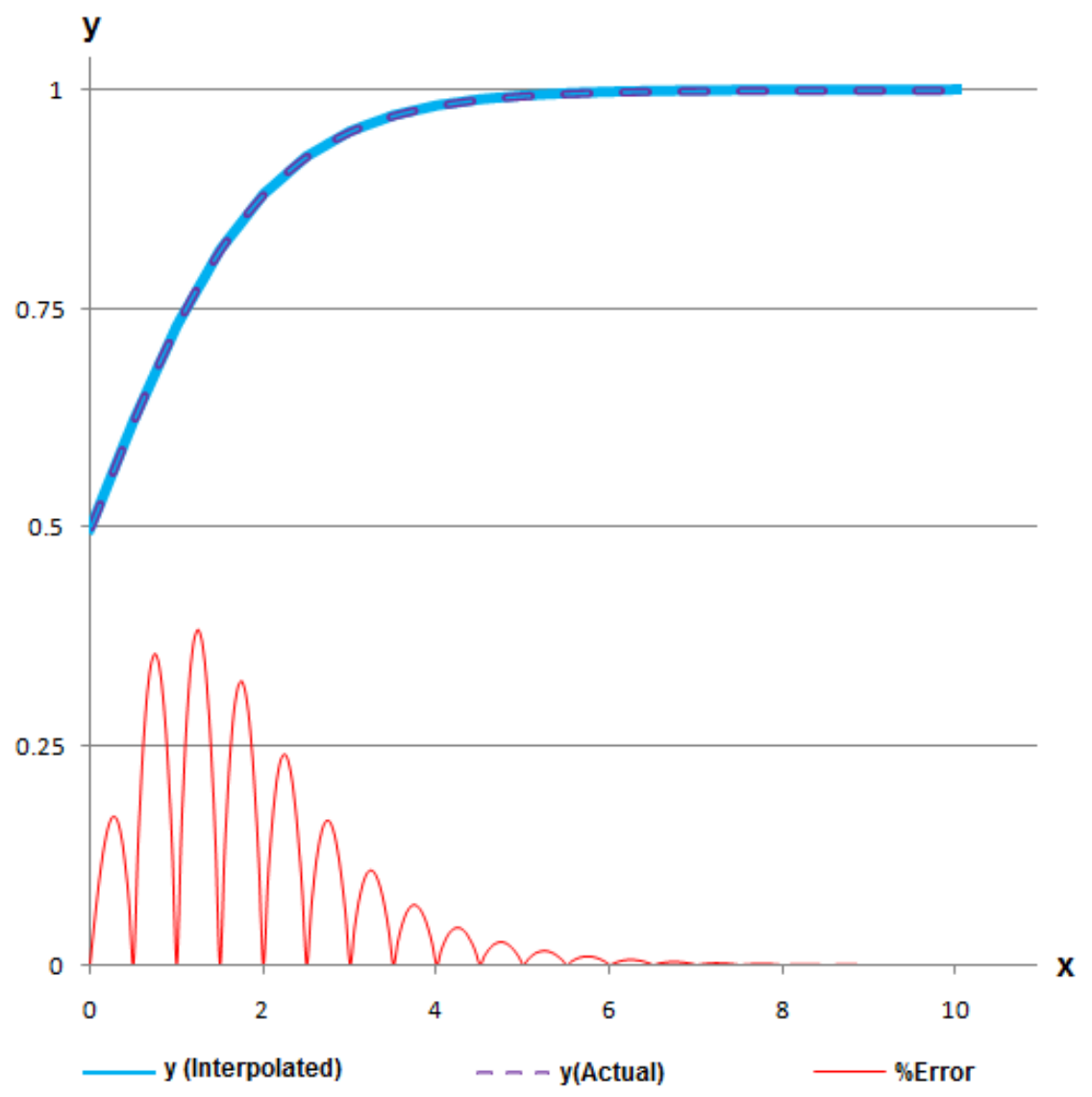

**Figure A2.10 PWL approximation of Sigmoid function at uniform spacing of 0.5**

### A2.8.2 Uniform Spacing of 0.25

The computational error can be further reduced by reducing the spacing between two known points. For example, the spacing between two known points is reduced from 0.5 to 0.25 i.e., the known points are at $x_1$=0, $x_2$=0.25, $x_3$=0.5, $x_4$=0.75 and so on. This has maximum error deviation of 0.0974%, an improvement of nearly 75% over the 0.5 spacing (see Figure A2.11).

### A2.8.3 Non-uniform Spacing

From the Figure A2.10 and A2.11, it is noticed that, with less spacing between the known points, it is possible to improve the accuracy. However, it also increases the memory size of LUT. Therefore, to balance this trade off, a non-uniform spacing can be used. It can be noticed from Figure A2.11 that, the error is maximum between x=0.75 and 1.25. This error can be further reduced by reducing the spacing only between these points and keeping the other points unaltered. For this reason, it is called as non-uniform spacing. The Figure A2.12 shows the approximation of a curve with non-uniform spacing and a maximum error of 0.0238% is observed.

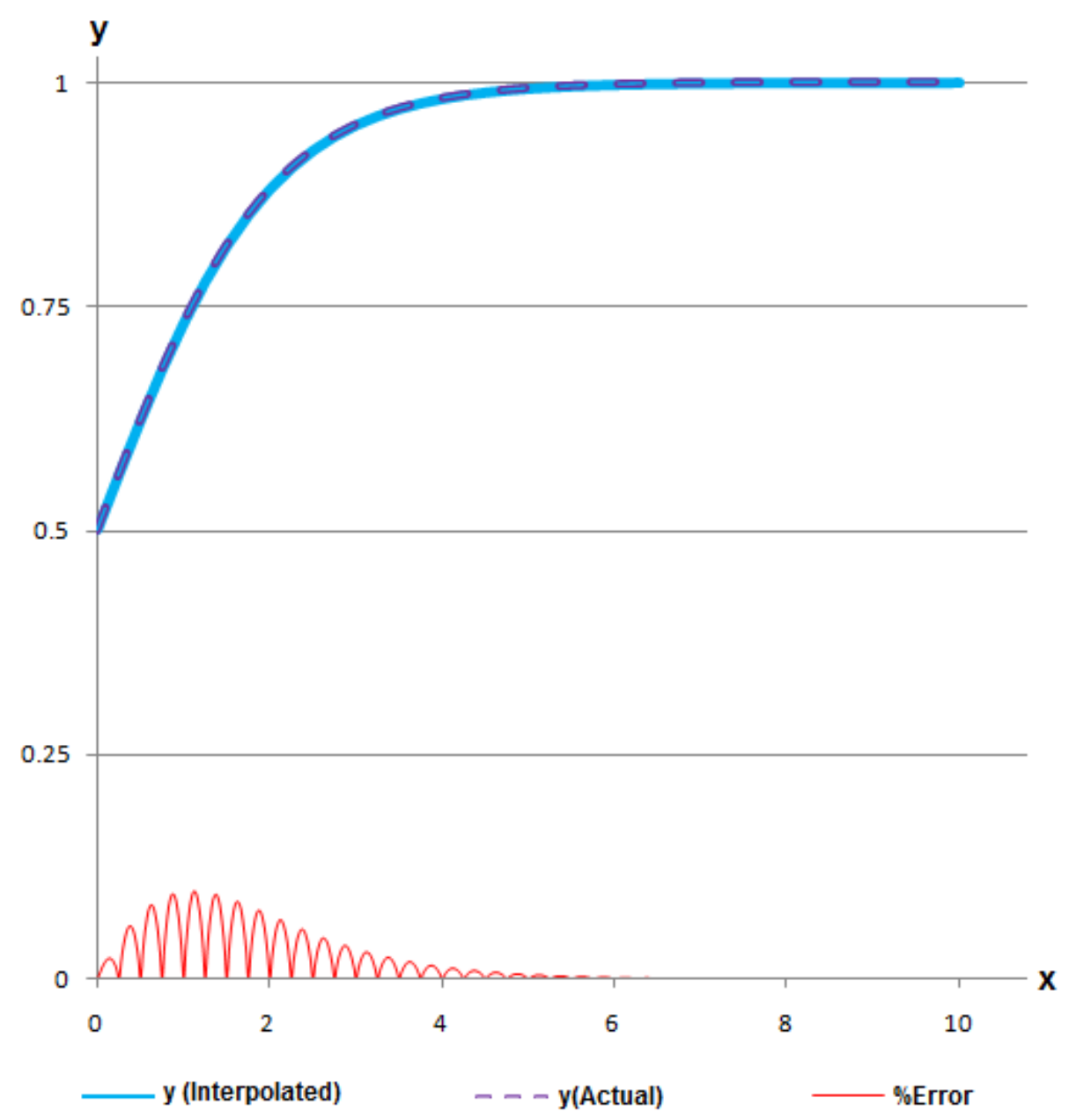

**Figure A2.11 PWL approximation of Sigmoid function at uniform spacing of 0.25**

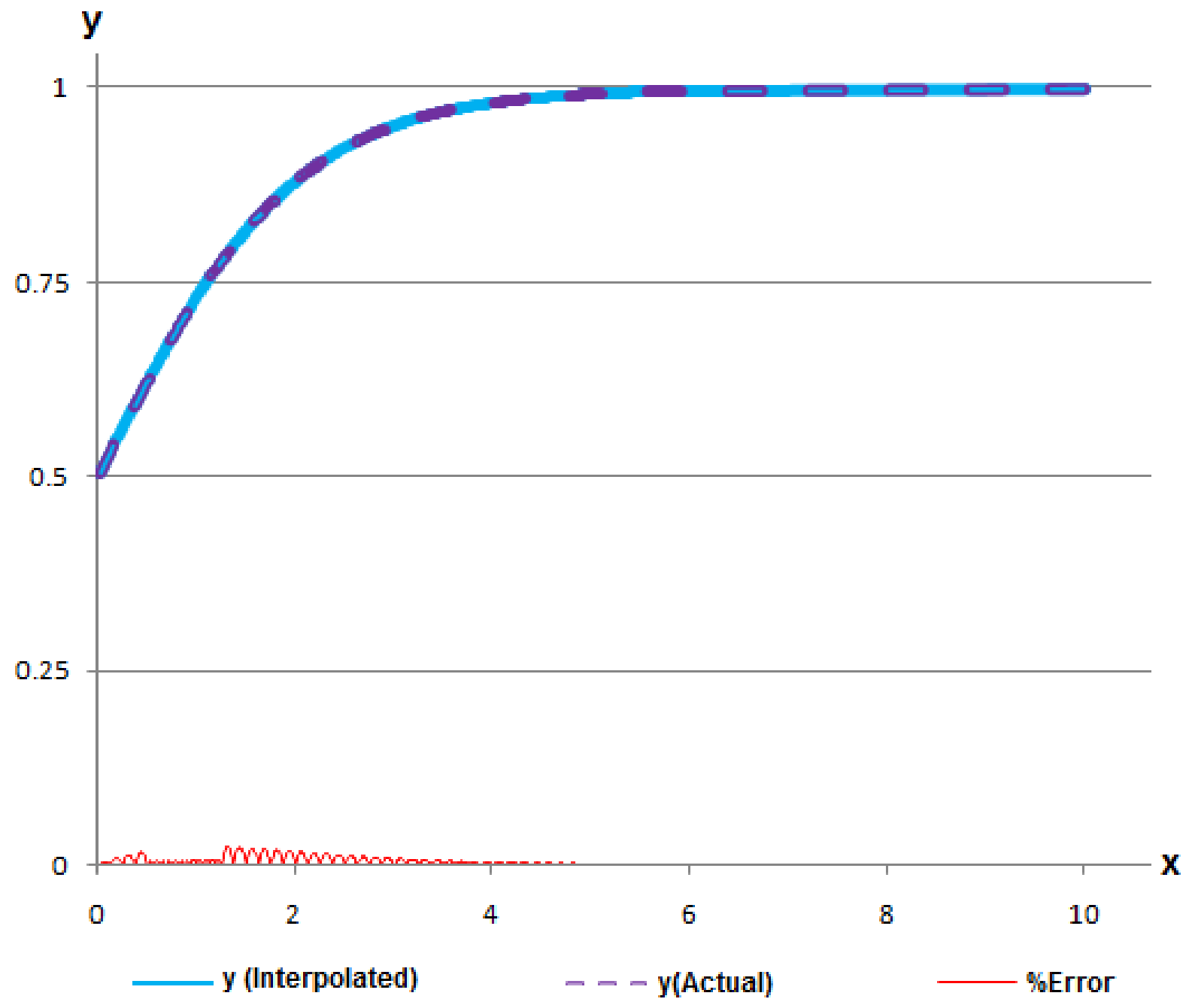

**Figure A2.12 PWL approximation of Sigmoid function at non-uniform spacing**

## A2.9 Sigmoid Approximation using SOI

In this section, another approximation method called Second Order Interpolation (SOI) is discussed. This method uses three points on the curve to approximate the output of a non-linear function. Similar to PWL method, both uniform and non-uniform spacing methods are applied to SOI method. The accuracy of computation is higher in this method but with an additional cost of increased LUT size.

### A2.9.1 Uniform Spacing of 0.125

The known points are stored at uniform spacing of 0.125, i.e., x=0, 0.125, 0.25, 0.375 and so on. The graph of SOI with uniform spacing of 0.125 is shown in Figure A2.13 and the maximum error of 0.0030% was noted. For visibility purpose the error is amplified by 100.

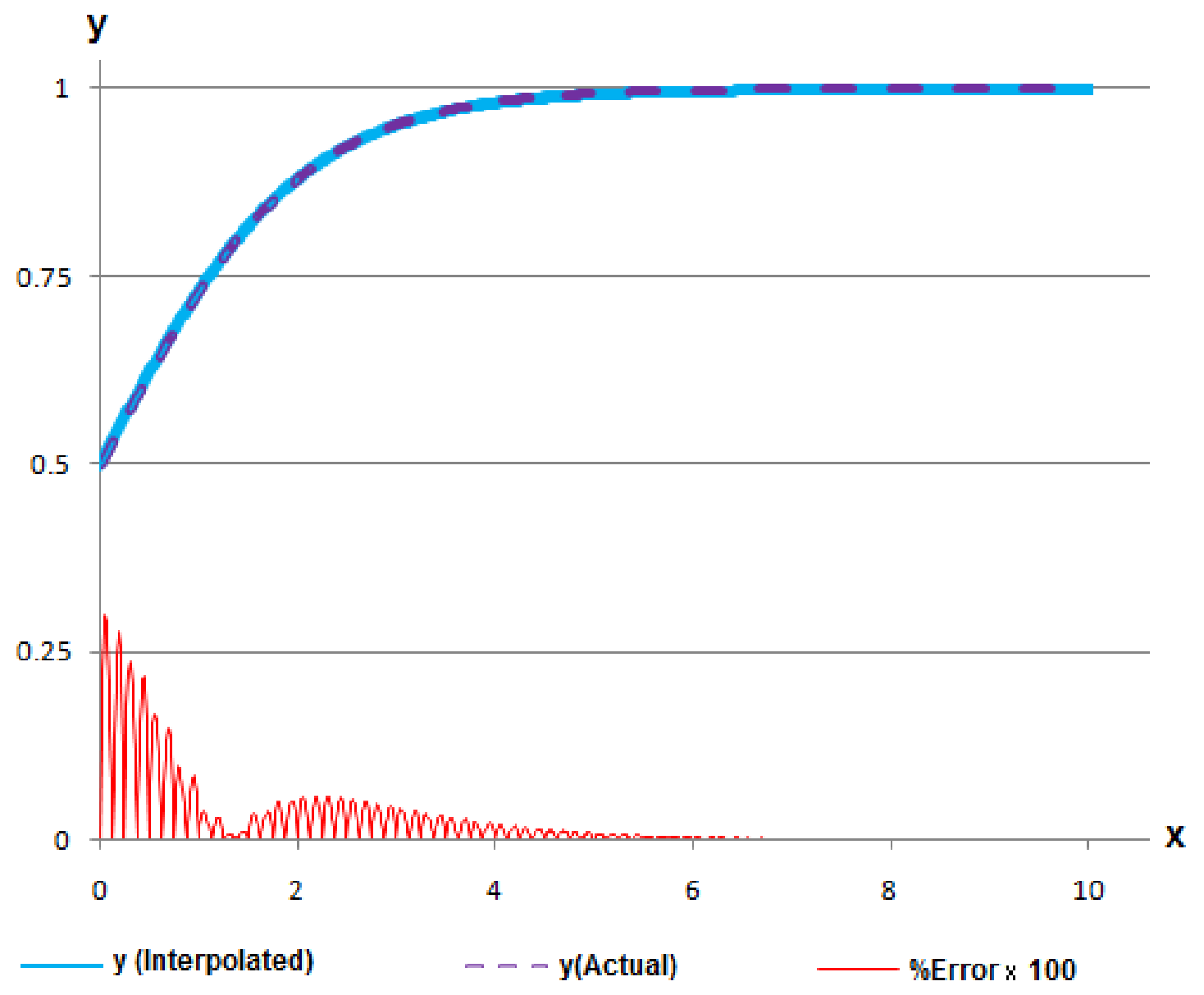

**Figure A2.13 SOI approximation of Sigmoid function at uniform spacing of 0.125**

### A2.9.2 Non-uniform Spacing

It may be noticed from Figure A2.13 that the error is maximum at x=0.05. This can be reduced by decreasing the spacing between 0 and 1 to 0.03125 and by keeping spacing of 0.125 outside this range. This non-uniform spacing reduces the error down to 0.00057% as shown in Figure A2.14. The error is amplified by 100.

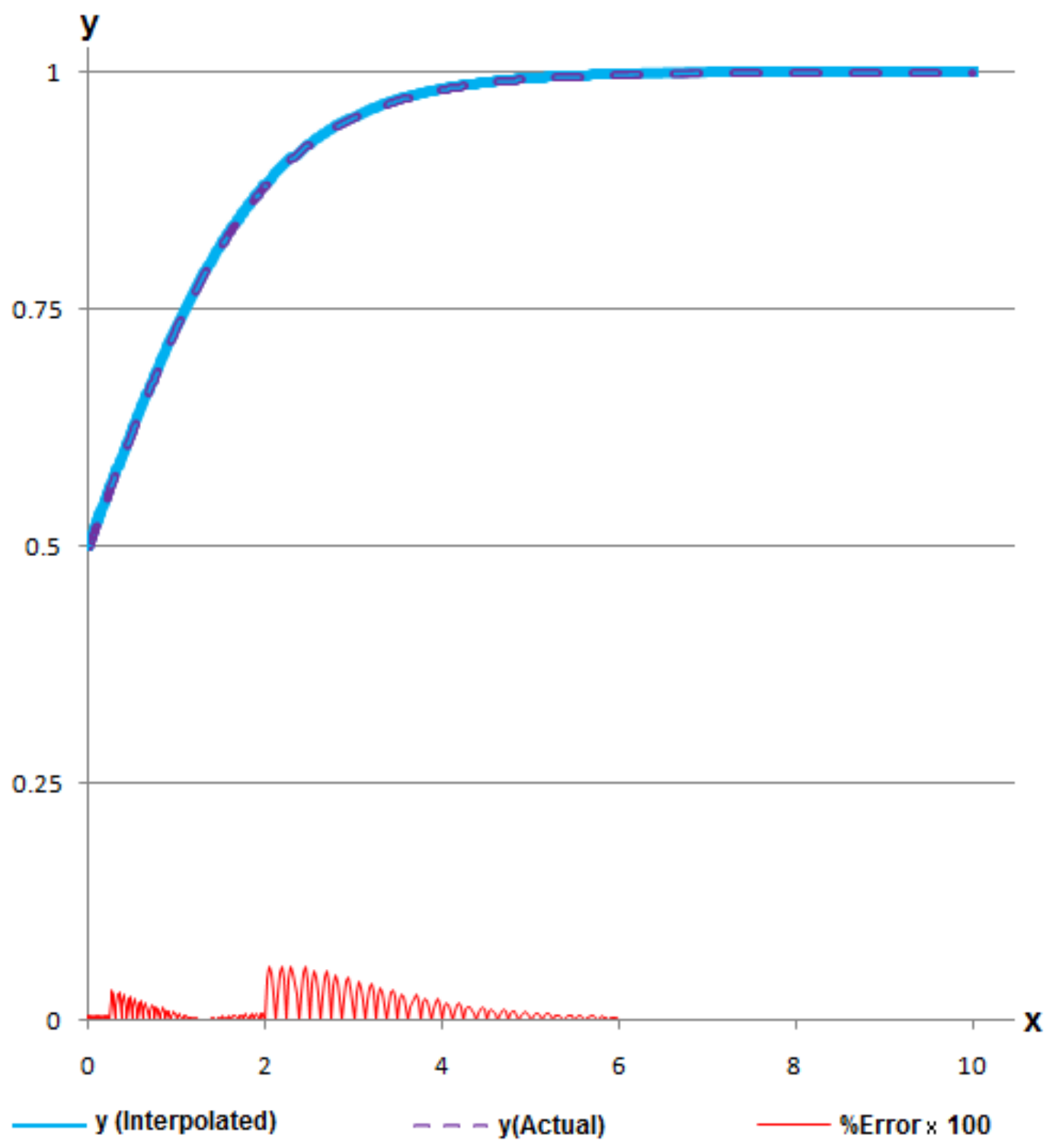

**Figure A2.14 SOI approximation of Sigmoid function at non-uniform spacing**

## A2.10 Hyperbolic tangent (tanh) Activation function

A hyperbolic tangent (*tanh*) is another activation function that is widely used in neural networks. It also has an S-shaped curve like a sigmoid function, but with the ℝ between -1:1 as shown in Figure A2.15.

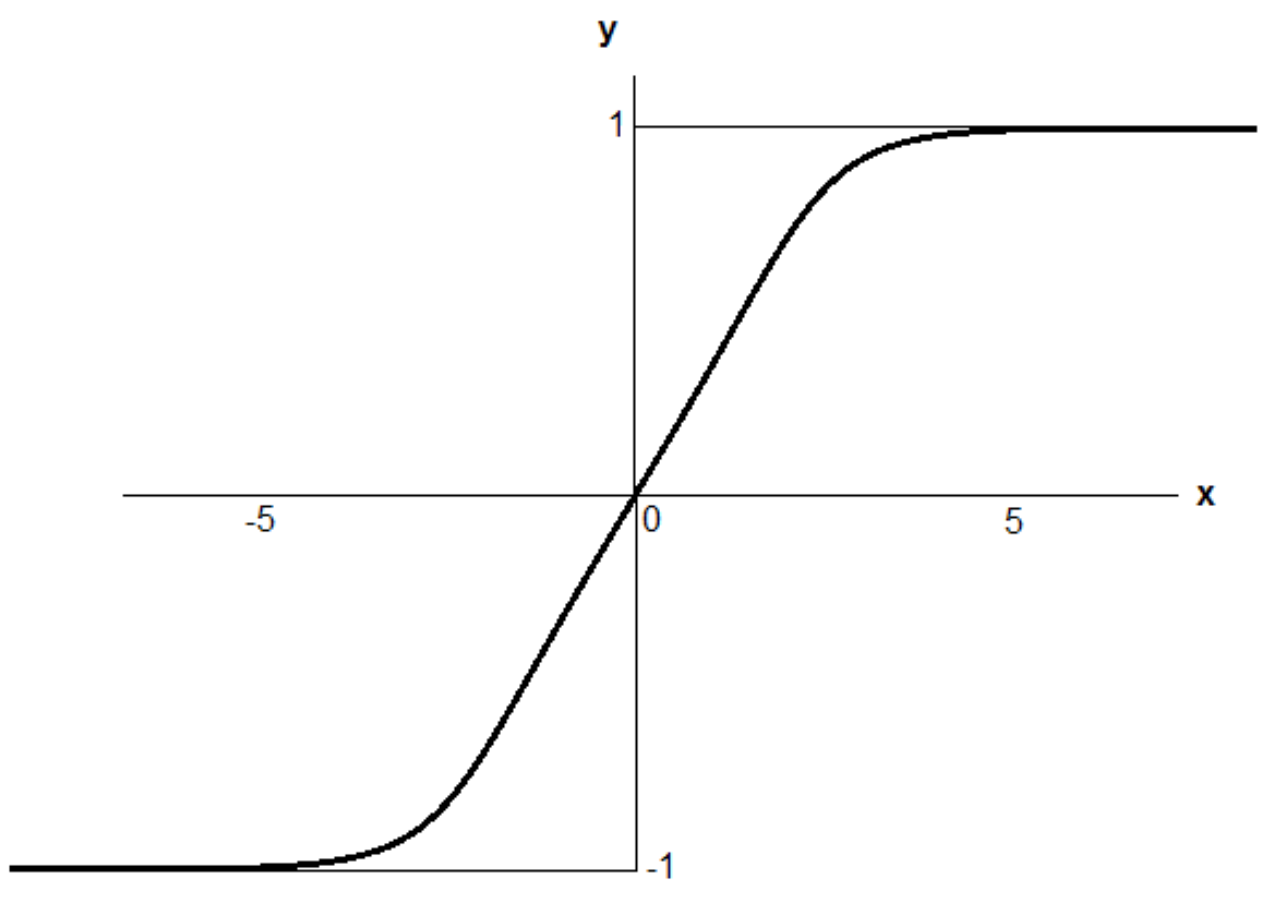

**Figure A2.15 Hyperbolic Tangent (*tanh(x)*) curve**

The equation for a *tanh* function is given by eqn. (A2.2).

$$y = \frac{e^x - e^{-x}}{e^x + e^{-x}} \tag{A2.2}$$

The approximation methods described in previous section are applied to tanh function as well. The graph for approximating the tanh curve for PWL with maximum error of 0.4347% and SOI with maximum error of 0.1814%, for no-unifrom spacing are shown in Figure A2.16 and Figure A2.17 respectively.

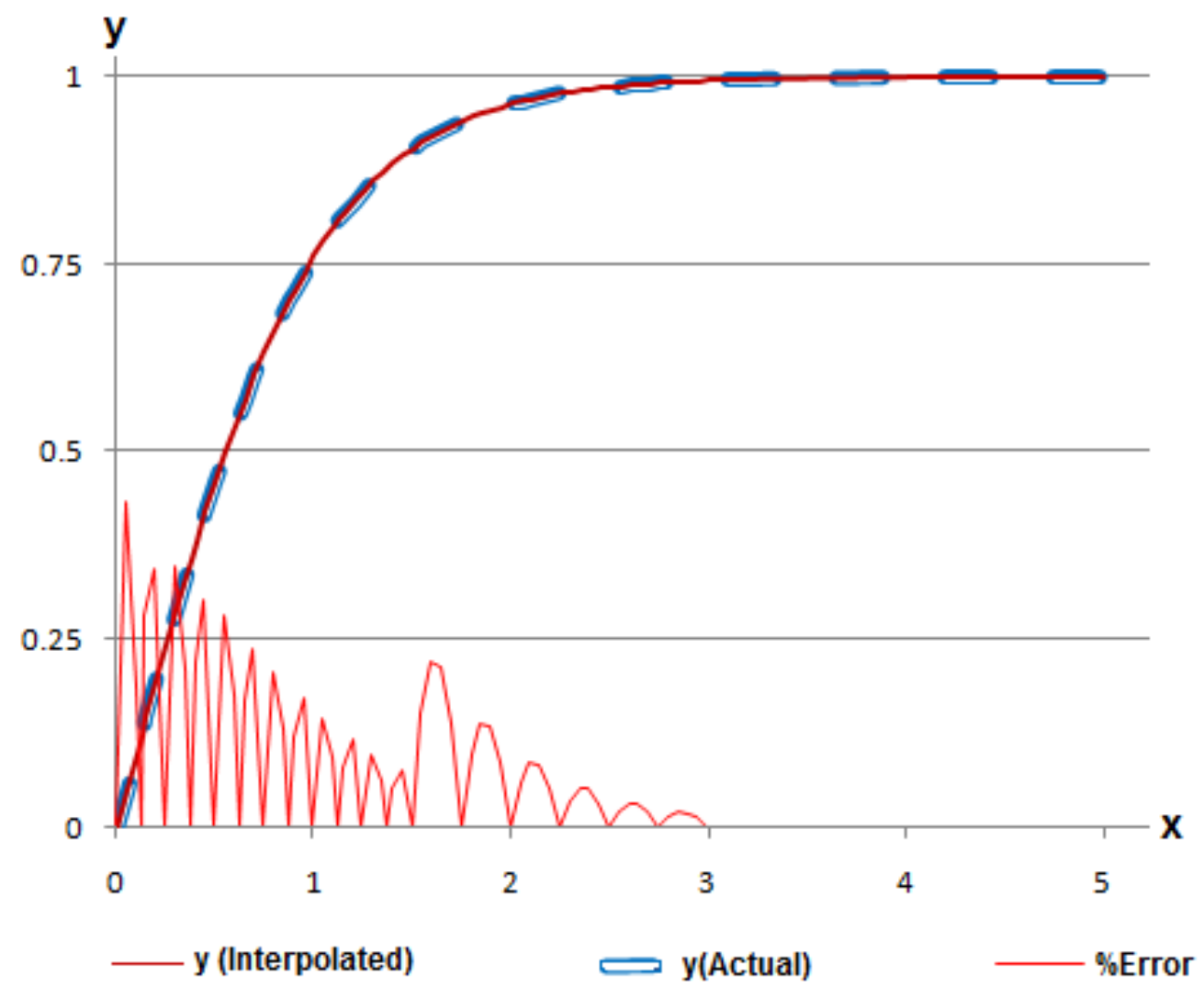

**Figure A2.16 PWL approximation of Tanh function at non-uniform spacing**

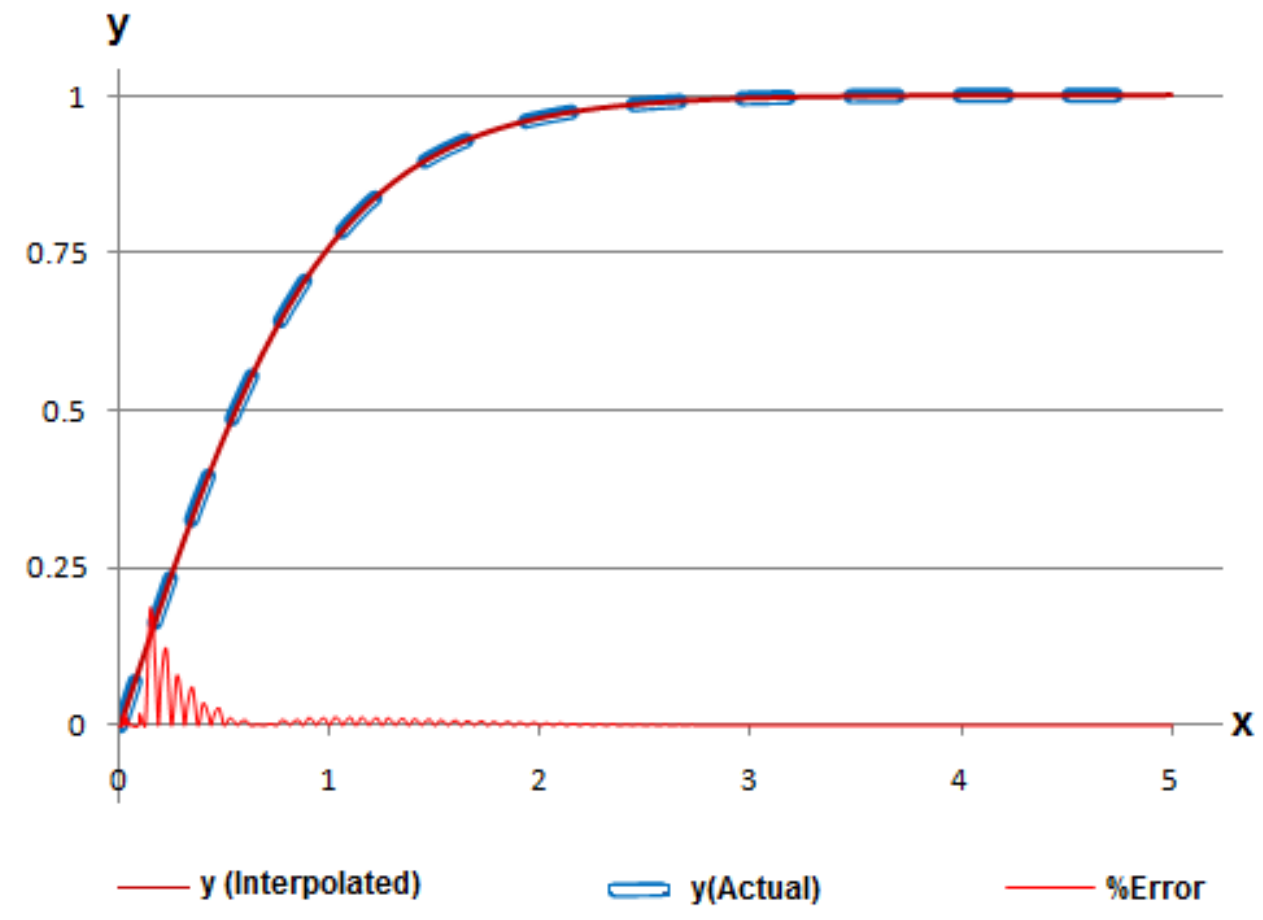

**Figure A2.17 SOI approximation of tanh function at non-uniform spacing**

## A2.11 ARN Resonator

The equation to calculate the output of ARN node is given by eqn. 4.8. The approximation methods are applied to this function and the graphs for the approximation (non-uniform spacing) are shown in Figure A2.18 and Figure A2.19. The resonator in an ARN has several curves to be approximated. Similar methods can be used to implement each of the resonance curves. For the current implementation, the values of $\rho$ are 1, 1.76, 2, 2.42, 3 and 5. However, it is also possible to reconfigure these values by changing the values of LUT.

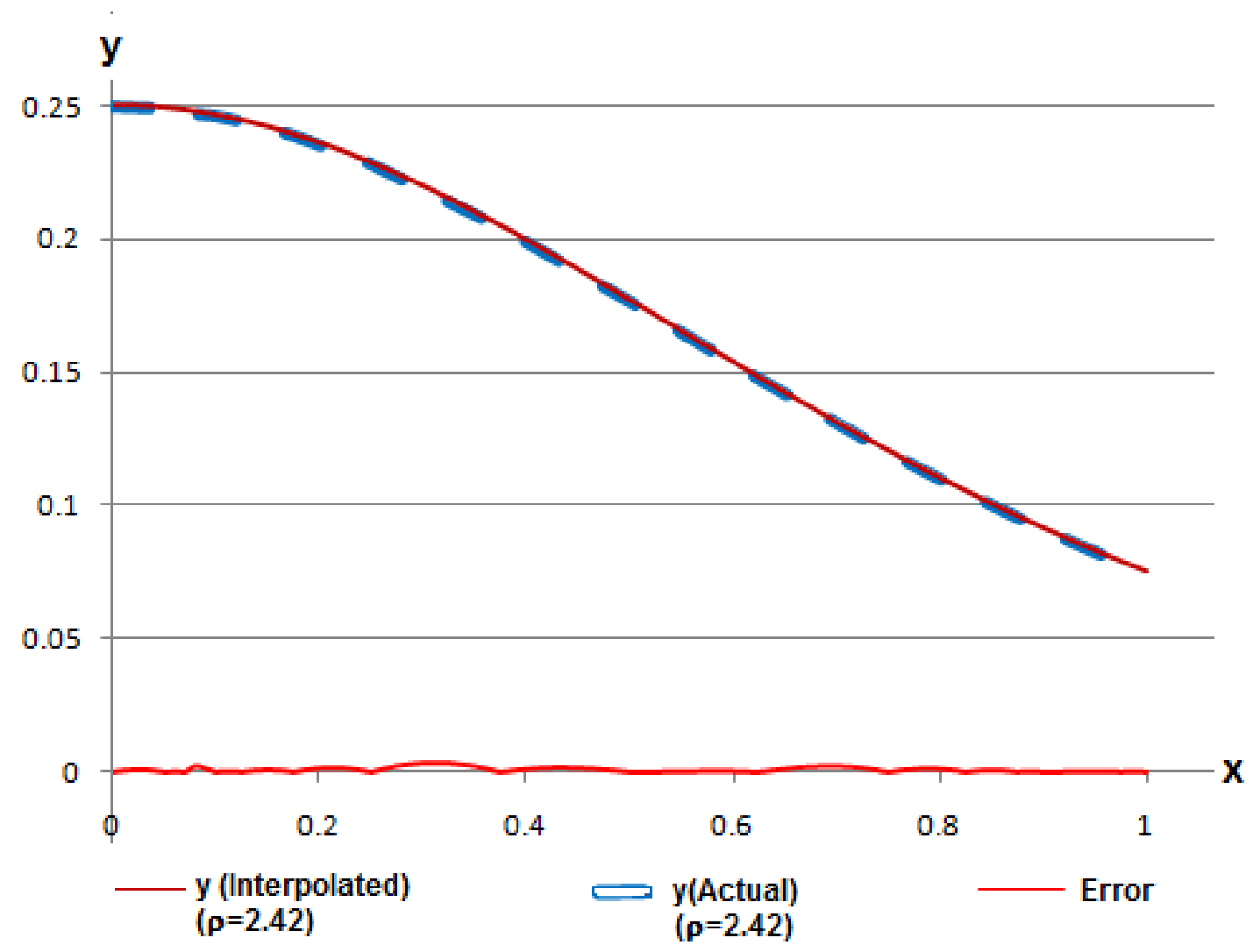

**Figure A2.18 PWL Approximation of a ARN resonator for $\rho$=2.42**

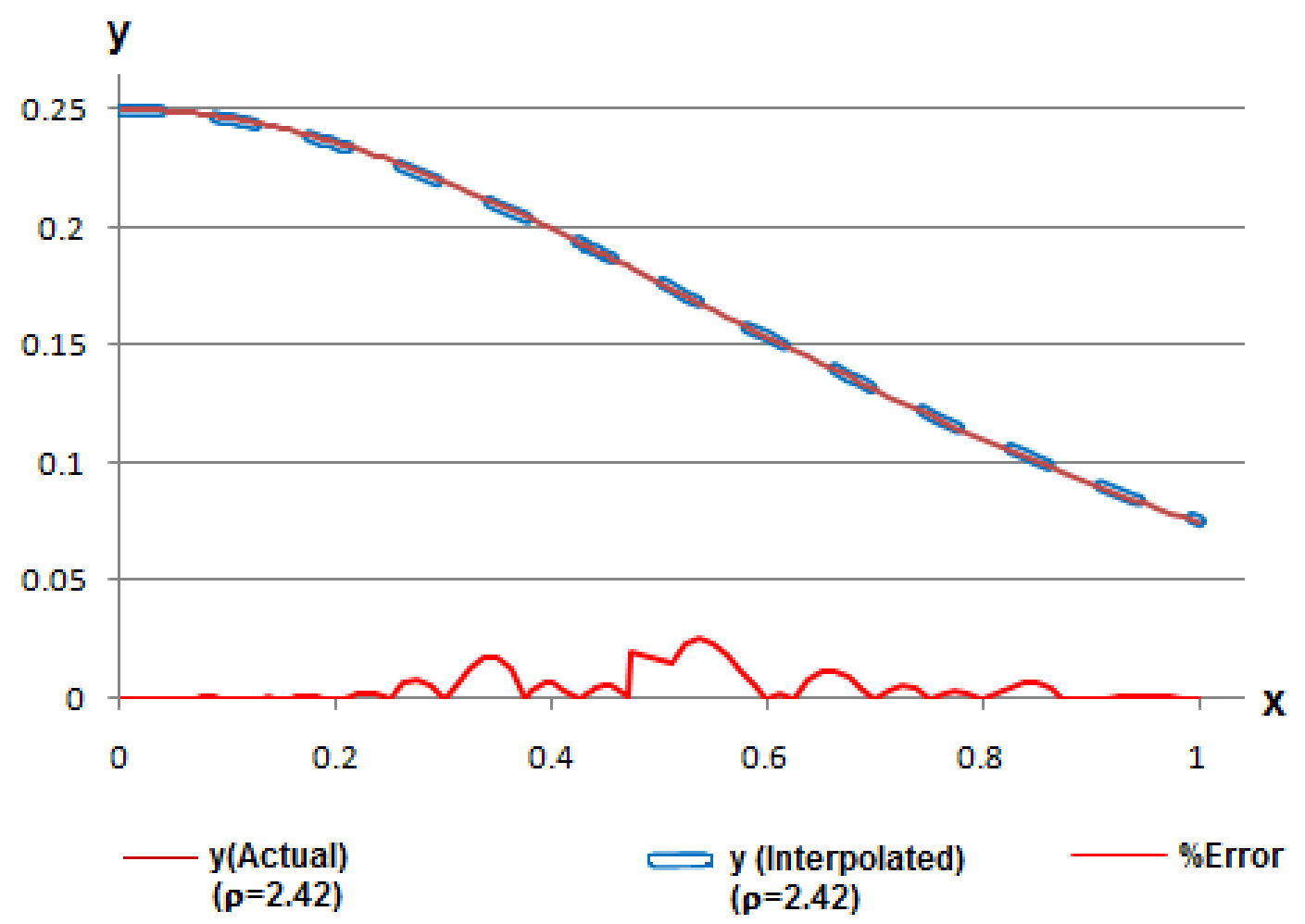

**Figure A2.19 SOI Approximation of a ARN resonator for $\rho$=2.42**

## A2.12 Implementation

As it is mentioned earlier, SOI method requires 3 fetch, 2 addition and 2 multiplication operations. This is independent of the function being implemented and therefore, a common procedure is used for the implementation of sigmoid and resonator.

The algorithm of implementing a function using PWL and SOI are as follows:

```
Prerequisites:
    a) Perform the approximation of a curve to be implemented using the
       eqn. 5.2 or eqn. 5.3.
    b) Store the pre-computed values in a look-up-table.
```

PWL Implementation:

a) Apply a test input.

b) Compare the input (x) with the stored values to identify the two stored points ($x_1$, $x_2$) such that, $x_1<x<x_2$. Return the values of m, $x_1$, $y_1$

c) Perform the subtraction of $x-x_1$

d) Perform the multiplication: the result of step c is multiplier with m, i.e., $(x-x_1)\times m$

e) Perform the addition: result of step d is added with $y_1$, i.e., $((x-x_1)\times m)+y_1$

f) Transfer the result to bus and indicate the completion of operation.

SOI Implementation:

a) Apply a test input.

b) Compare the input (x) with the stored values to identify the two stored points ($x_1$, $x_2$, $x_3$) such that, $x_1<x<x_3$. Return the values of a, b, c

c) Perform the first multiplication: multiply a with x, i.e., $a \times x$

d) Perform the first addition: the result of step c is added with b, i.e., $(a \times x) + b$

e) Perform the second multiplication: result of step d is multiplied with x, i.e., $((a \times x) + b) \times x$

f) Perform the second addition: the result of step e is added with c, i.e., $(((a \times x) + b) \times x) + c)$

g) Transfer the result to bus and indicate the completion of operation.

The sample structure of LUT for both PWL and SOI methods is shown in Figure A2.20.

| x | y | m |
|---|---|---|
| 0 | 0.5 | 0.2499 |
| 0.25 | 0.5621 | 0.2411 |
| ⋮ | ⋮ | ⋮ |
| 0.75 | 0.6791 | 0.2075 |

(a)

| x | a | b | c |
|---|---|---|---|
| 0 | -0.0077 | 0.2506 | 0.5 |
| 0.25 | -0.2231 | 0.2578 | 0.4991 |
| ⋮ | ⋮ | ⋮ | ⋮ |
| 0.75 | -0.0426 | 0.2821 | 0.4915 |

(b)

**Figure A2.20 Structure of a look-up-table for implementation using (a) PWL (b) SOI**

It is also important to note here that, the gradient of sigmoid curve is significant only for few values of x, i.e, for any values of $x < -5$, the output is 0 and for any values of $x > 5$, the output is 1. As the input values can be between -5 to 5 and the output value can be between 0 and 1, 16-bit number format is used for all the implementations.

Please note that the common procedure for implementing any non-linear functions like sigmoid, tanh and ARN resonators is given in Table A2.8. This procedure is based on the approximation methods discussed in chapter 5. However, the values of LUT for both PWL and SOI need to be changed according to the function being implemented.

**Table A2.8 Pseudo code for Sigmoid/tanh/ARN resonator**

```
module declaration
    I/O declaration
    Initialization
    LUT for SOI
    LUT for PWL
    #Choose the method of approximation
    For SOI{
            Compare the input with LUT values and get the corresponding values
of a,b &c
            Perform two multiplications and two additions as follows
            Multiply a and x
            Add the result with b
            Multiply the result with x
            Add the result with c
           }
    Transfer the result to output port
    Raise the endFlag

    For PWL{
            Compare the input with LUT values and get the corresponding values
of x1,y1&m
            Perform one subtraction, one multiplication and one addition as
follows
            Subtract x with x1
            Multiply the result with m
            Add the result with y1
           }
    Transfer the result to output port
    Raise the endFlag
Endmodule
```

## A2.13 Implementation of Multi-Operand adder

The basic modules required to implement multi-operand adder are single column adder and 4×M adder. The single column adder is used to calculate the sum of a column as described in chapter 6. Column adder can be implemented as a count one's logic: adding a single bit numbers is same as counting the number of ones in a binary logic. This single column adder is used as a look-up-table for the implementation of 4×M adder. Larger adders like 16×16 adders can be implemented using a series of 4×M adders.

**Table A2.9 Count one's logic/ single column adder**

```
/* Description: This module is used to count the number of one's in a 4-bit
input.  This logic is same as adding the four 1-bit number in binary.  When
adding four operands of m-bits each, this can be used as a column adder.  This
logic for this module is based on the structure given in Figure 6.5 */

//Module instance
module countOnes co(inValue, clk, reset, startCount, outLUT);
//Implementation
module countOnes(in1, in2, in3, in4, out);

//IO declaration
    input in2, in3, in4;
    input [3:0] in1;
    output wire [2:0] out;
    wire [2:0] tLUT;

//logic construction
    xor x1 (temp1, in1[2], in1[1]);//b xor c
    and a1 (temp2, ~in1[3], in1[0]);//~ad
    or o1 (temp3, in1[3], temp2);//~ad+a
    and a2 (temp4, temp3, temp1);//(~ad+a)(b xor c)
    and a3 (temp5, ~in1[3], in1[2], in1[1]);//~abc
    and a4 (temp6, in1[3], ~in1[2], ~in1[1], in1[0]);//a~b~cd
    and a5 (temp7, in1[3], in1[2], in1[1], ~in1[0]);//abc~d
    xor x2 (tLUT[0], in1[3],in1[2], in1[1],in1[0]);
    or o2 (tLUT[1], temp4, temp5, temp6, temp7);
    and a6 (tLUT[2], in1[3], in1[2], in1[1], in1[0]);
    and a7 [2:0] (out, tLUT, ~in3);
endmodule
```

**Table A2.10 Pseudo code for 4×M adder**

```
module declaration
    I/O declaration - #Allocate 20-bits (i.e., 16+log215) for output port
    Initialization
    Instantiate the single column adder module
    Take the inputs on load command
    Start the addition on start command
    Map and store the inputs column-wise
    column=0
    Repeat{
            Perform the column addition using single column adder
            Add it with the previous result
            Right shift the result and transfer LSB to temp Sum
            Increment the column
          } until all m columns are over
    Transfer the temp sum to bus
    Raise the endFlag
```

**Table A2.11 Pseudo code for 16×16 adder using 4×16 adder**

```
/* Description: This module is used to perform the addition of 16-operands of
16-bits each.  The 4x16 module is used as the basic module and is been
instantiated.  It is also possible to optimize the logic using 4x4 and 2x4
adders instead of 4x16 adders in 2nd and 3rd stage.  However, to maintain the
uniformity and easy understanding we have only used 4x16 module */

module declaration
    I/O declaration - #Allocate 20-bits (i.e., 16+log2 15) for output port
    Initialization
    Instantiate the 4x16 adder module
_________stage 1________________________________________________
          #Divide the 16-inputs into 4 groups #(U1, U2, U3, U4): each group
having 4-operands of 16-bits each
    Present each group to 4x16 adder
    Wait for the endFlag
_________stage 2________________________________________________
    Separate the result of 4x16 adder modules into sum and carry
    Group them into 4 sums and 4 carry
    Present the sum to 4x16 module #U5
    Present the carry to 4x16 module #U6
    Wait for the endFlag
_________stage 3________________________________________________
    Transfer the sum of U5 to lower bits of output port
    Add the carry of U5 with sum of U6 and transfer the sum to higher bits of
output port
    Raise the endFlag
Endmodule
```

# Appendix – 3

## Justification for ARN

Often, we encounter a question from the readers about what is the justification for experimenting with a new neural network when CNN and LSTM like networks already provide specific solutions. Biological systems evolve irrespective of the level of maturity or perfection reached by a species or an individual. There is always a chance that an existing but unused gene will prove useful in case of ecological catastrophes. Evolution in newer neural architectures should continue irrespective of the success rate of an existing model. Introduction of capsule networks to address the problem of spatial relations in CNN by none other than Prof. Geoffrey Hinton is an example of such a case. Use of spiking neural model in Intel's Loihi processor has a similar origin.

One more question often asked is that of accuracy of recognition. Is 94% recognition sufficient? Unlike the algorithmic methods where the results are 100% correct, inaccuracies in recognition/classification of input are a characteristic of all neural networks. Biological species survive without reaching optimal solution, even when one exists. Often, in the complex real world, individual with an optimal solution may not guarantee survival. Adaptability is a better survival trait than having a specific optimal solution. Exposure to the environment is more important than absolute result. Often, we observe that the performance of ANN improves with larger training set. We have not been able to use large data sets only because our image recognition tests were done on a low level laptop and not on GPU array assisted servers. Intelligent selection of training set may also contribute to increased recognition rate.

This raises an important question - what is the accuracy of recognition essential for AGI to work? Anything better than 50% accuracy may be acceptable; 70-80% should be very good.

# Appendix – 4

## Contents of CD

### A4.1 Directory structure of data on CD

| No. | Directory | Contents |
|---|---|---|
| | /2KL16PEJ14 | Base directory |
| 1 | /2KL16PEJ14/ Mandatory Documents | 1) Copy of Office Order of Ph.D Registration<br>2) Copy of Course Work Completion Certificate<br>3) Copy of the confirmation letter<br>4) Approval letter issued from VTU for change of Title<br>5) Approval letter issued from VTU for change of Research Centre<br>6) Adjudication format -1<br>7) No Due Certificate from Research Centre<br>8) Certificate from Guide<br>9) Letter of submission from Guide<br>10) Evidential proof from HOD and Head of Research Center for open seminar-1<br>11) Evidential proof from HOD and Head of Research Center for open seminar-2<br>12) Evidential proof from HOD and Head of Research Center for Colloquium<br>13) Online Fee payment challan<br>14) Residence certificate<br>15) Six months Progressive report |
| 2 | /2KL16PEJ14/ Thesis | 2KL16PEJ14.PDF |
| 3 | /2KL16PEJ14/ Publications | Soft copies of our papers |
| 4 | /2KL16PEJ14/ Python Code | Python codes for Image Recognition using ARN |
| 5 | /2KL16PEJ14/ verilog Code | Verilog code for Hardware modules |

# A4.2 Sample Codes

| No. | File name | Description |
|---|---|---|
| 1 | ARN.py | ARN library |
| 2 | trainARN.py | To train ARN for image recognition in MNIST dataset |
| 3 | testARN.py | To test ARN for image recognition in MNIST dataset |
| 4 | serialMultiplier.v | Module to perform multiplication of two binary numbers of maximum 16-bits each |
| 5 | sigmoid.v | Module to perform sigmoid operation on the given input |
| 6 | resonator.v | Module to implement ARN resonator |
| 7 | serialAdder4x4.v | Module to perform the serial addition of 4-binary operands of 4-bits each |
| 8 | parallelAdder4x4.v | Module to perform the parallel addition of 4-binary operands of 4-bits each |
| 9 | adder16x16.v | Module to perform the parallel addition of 16-binary operands of 16-bits each |
| 10 | neuronARN16.v | Module to calculate the output of 16-input ARN neuron |
| 11 | neuronMLP16.v | Module to calculate the output of 16-input MLP neuron |

**Visvesvaraya Technological University**
Jnana Sangama, Belagavi – 590 018.

**Prof.Satish Annigeri** Ph.D
**Registrar (Evaluation)**

**Phone: (0831) 2498136**
**Fax: (0831) 2405461**

Ref.No. VTU/BGM/Exam /2019-20/ 2339 Date: 10 DEC 2019

**Acceptance Letter**

Sir/Madam,

The soft copy of Ph.D./M.Sc. (Engineering by research) thesis of **Mr./Mrs. Shilpa Mayannavar** bearing **USN 2KL16PEJ14** has been submitted for Anti-plagiarism check at the office of the undersigned through "Turn-it-In" package. The scan has been carried out and the scanned output reveals a match percentage of **03% which is within the acceptable limit of 25%.** To obtain the comprehensive report of the plagiarism test, research scholar can send a mail to **apc@vtu.ac.in** along with the USN, Name, Name of the Guide/Co-guide, Research centre and title of the thesis.

Registrar (Evaluation)

To, **Shilpa Mayannavar**

Research Scholar

Electronics and Communication

KLE College of Engineering and Technology, Belagavi.

Copy to: **Dr.Uday U Wali**, Professor, Dept. of Electronics & Communication, KLES Dr. M. S. Sheshgiri CET, BELAGAVI.